\documentclass[b5paper,11pt,oldfontcommands]{memoir}

\usepackage[utf8]{inputenc}
\usepackage{amsmath}
\usepackage{amsfonts}
\usepackage[dutch,english]{babel}
\usepackage{bibentry}
\nobibliography*
\usepackage{graphicx}
\usepackage{natbib}
\usepackage{rotating}
\usepackage{tabularx}
\usepackage{url}
\usepackage[table]{xcolor}
\usepackage{enumerate}
\usepackage{subcaption}
\usepackage{pgffor}

\usepackage{booktabs}
\usepackage{covington}
\usepackage{rotating}

\usepackage{multirow}
\usepackage{multicol}
\usepackage{scalerel}
\usepackage{stackengine}

\usepackage{slashbox}

\usepackage{tikz}
\usetikzlibrary{arrows,calc,positioning,trees,fit,backgrounds,bayesnet,shapes.geometric}

\usepackage{tikz-dependency}

\newcommand{\RNum}[1]{\uppercase\expandafter{\romannumeral #1\relax}}

\usepackage{hyperref}
\hypersetup{
    colorlinks,
    citecolor=black,
    filecolor=black,
    linkcolor=black,
    urlcolor=black
}

\definecolor{myred}{RGB}{215,25,28}
\definecolor{myorange}{RGB}{253,174,97}
\definecolor{mygreen}{RGB}{0,136,55}
\definecolor{mybrown}{RGB}{166,97,26}
\definecolor{myblue}{RGB}{44,123,182}

\usepackage[default]{droidserif}
\usepackage[defaultsans]{lato}
\usepackage[T1]{fontenc}
\usepackage{enumitem}
\usepackage{microtype}
\usepackage{pdfpages}

\renewcommand{\normalsize}{\fontsize{10.5pt}{12.6pt}\selectfont}

\setstocksize{240mm}{170mm}
\settrimmedsize{240mm}{170mm}{*}

\settrims{0.5\trimtop}{0.5\trimedge}

\settypeblocksize{170mm}{120mm}{*}

\setlrmargins{30mm}{*}{*}
\setulmargins{35mm}{*}{*}

\checkandfixthelayout

\title{\sffamily\Huge\bfseries{
One Model to Rule them all
}
\\{\makebox[\textwidth]{\Large\bfseries{
Multitask and Multilingual Modelling for Lexical Analysis}
}}
}
\author{Johannes Bjerva}
\date{}

\makeatletter
\makechapterstyle{bjerva}{
        \chapterstyle{default}
        
        \renewcommand*{\chapnamefont}{\sffamily\Huge}
        \renewcommand*{\printchaptername}{
                \flushright\chapnamefont\MakeUppercase\@chapapp}

}
\chapterstyle{bjerva}
\makeatother

\addtopsmarks{ruled}{\nouppercaseheads}{}
\makeoddhead{ruled}{\sffamily\rightmark}{}{\sffamily\thepage}
\makeevenhead{ruled}{\sffamily\thepage}{}{\sffamily\leftmark}
\makeoddfoot{ruled}{}{}{}
\makeevenfoot{ruled}{}{}{}
\renewcommand*{\parttitlefont}{\Huge\bfseries\sffamily}

\renewcommand*{\printparttitle}[1]{\parttitlefont{#1}}

\usepackage[export]{adjustbox}
\newcommand{\partimage}[2][]{\gdef\@partimage{\vfill\includegraphics[#1,left]{#2}}}
\renewcommand{\printparttitle}[1]{\parttitlefont #1\vfil\@partimage\vfil}
\aliaspagestyle{part}{empty}
\aliaspagestyle{chapter}{empty}

\setsecheadstyle{\sffamily\noindent\bfseries}
\setsubsecheadstyle{\secheadstyle}
\setsubsubsecheadstyle{\secheadstyle}

\makeatletter
\renewcommand*{\l@part}[2]{\l@chapapp{{#1}}{#2}{\cftchaptername}}
\renewcommand*{\l@chapter}[2]{\l@chapapp{{#1}}{#2}{\cftchaptername}}
\renewcommand*{\l@appendix}[2]{\l@chapapp{{#1}}{#2}{\cftchaptername}}
\makeatother

\addto\captionsenglish{}

\newcommand{\absprelude}{\noindent\normalfont{\textbf{Abstract|}}\abstracttextfont}

\captionnamefont{\bfseries}
\captionstyle{\itshape}
\captiondelim{|}

\begin{document}

\selectlanguage{english}

\frontmatter

\maketitle
\thispagestyle{empty}


\pagebreak
\thispagestyle{empty}

\begin{figure}[ht]
\begin{minipage}[b]{0.5\linewidth}
\includegraphics[height=1.3cm]{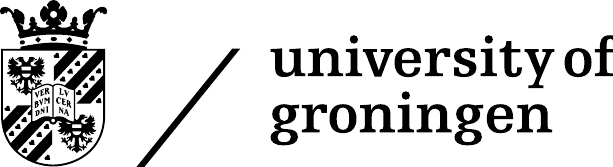}
\end{minipage}
\begin{minipage}[b]{0.5\linewidth}
\centering
\begin{flushright}
\vspace{0.15cm}
\includegraphics[height=1cm]{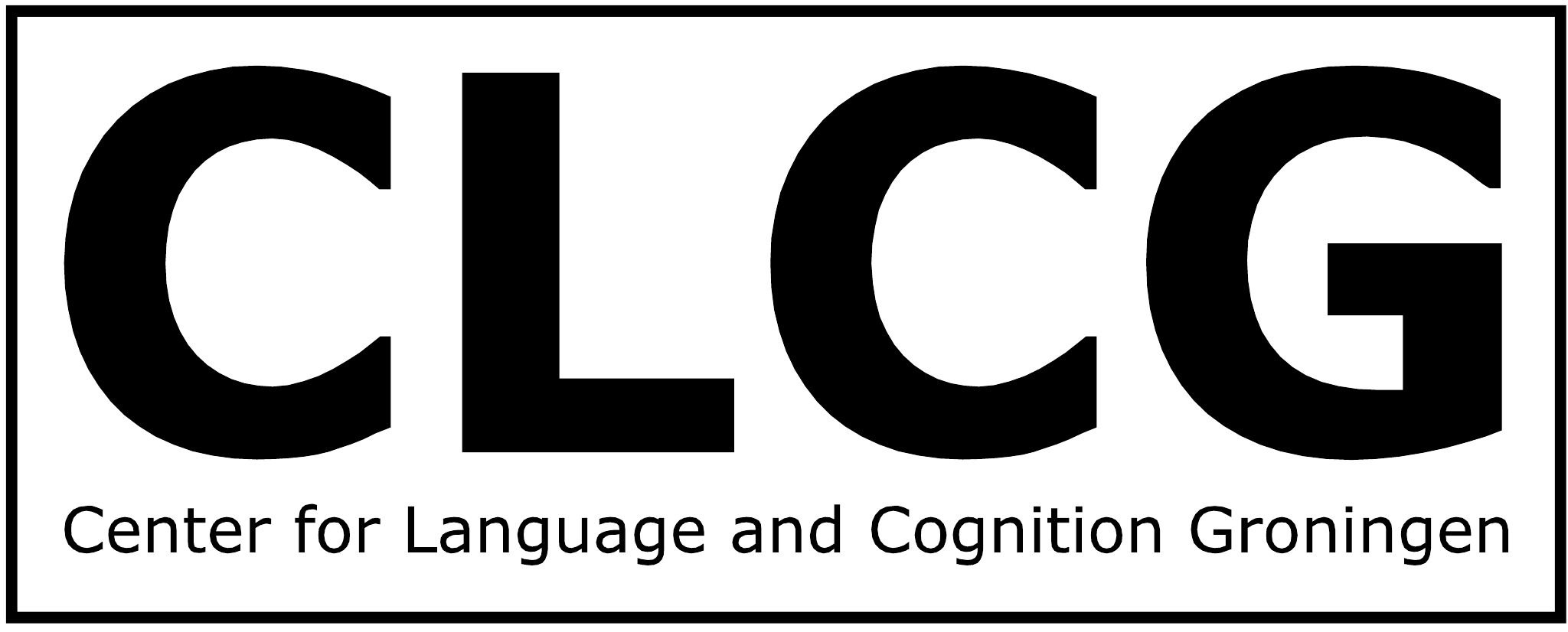}
\vspace{0.15cm}
\end{flushright}
\end{minipage}
\end{figure}

{\footnotesize
\noindent
The work in this thesis has been carried out under the auspices of the Center for Language
and Cognition Groningen (CLCG) of the Faculty of Arts of the University of Groningen.
\\
}

\vfill
\begin{figure}[h]
\includegraphics[width=2cm]{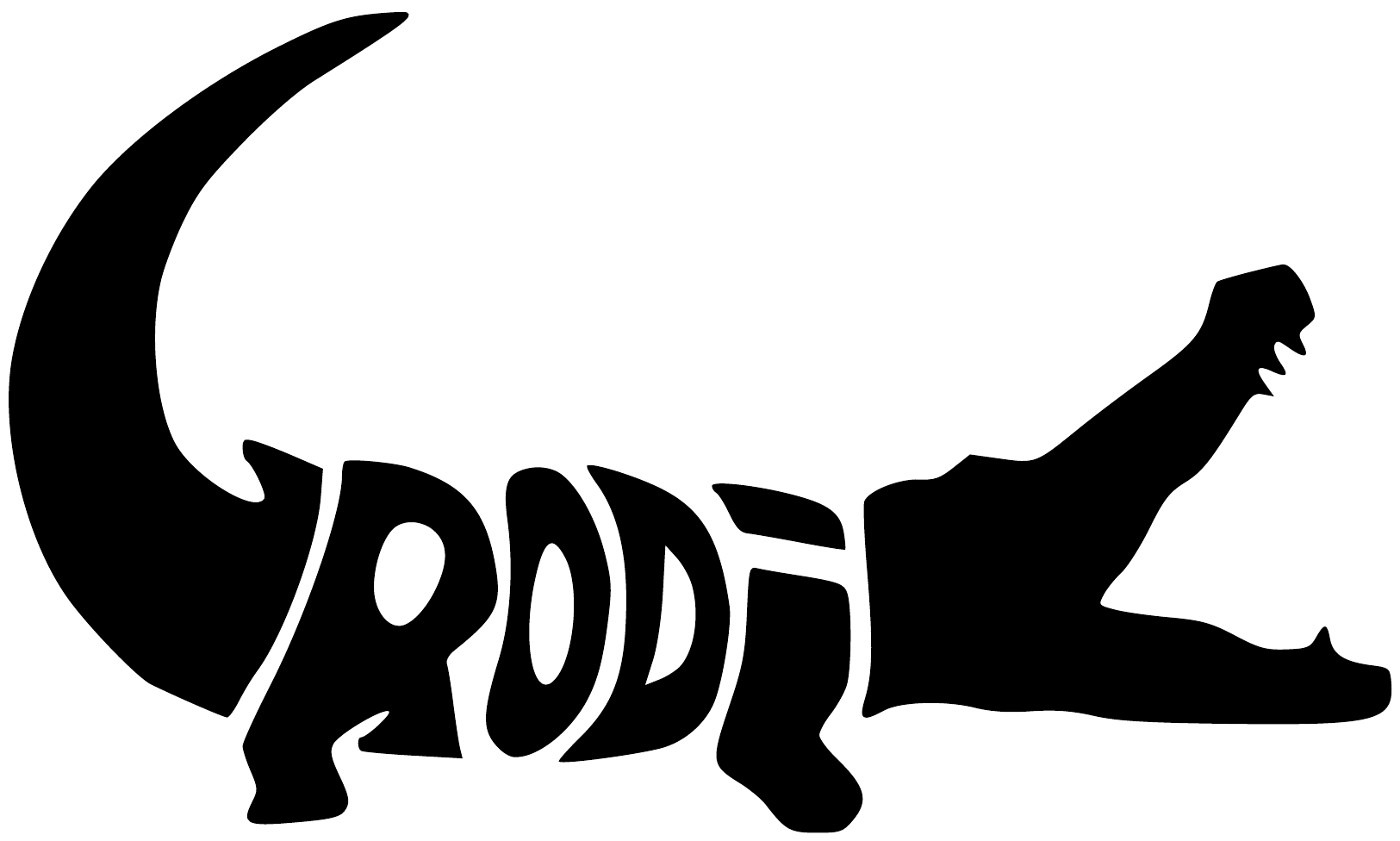}
\end{figure}
\vspace{0.1cm}
{\footnotesize
\noindent
Groningen Dissertations in Linguistics 164\\
ISSN: 0928-0030\\
ISBN: 978-94-034-0224-6 (printed version)\\
ISBN: 978-94-034-0223-9 (electronic version)\\
}
\vspace{0.05cm}\\
{\footnotesize
\noindent
\copyright~2017, Johannes Bjerva\\
}
\vspace{0.05cm}\\
{\footnotesize
\noindent
Document prepared with \LaTeXe\ and typeset by pdf\TeX\ \\ (Droid Serif and Lato fonts)\\
Cover art: \textit{Cortical Columns}. \copyright~2014, Greg Dunn \\
21K, 18K, and 12K gold, ink, and dye on aluminized panel.\\
Printed by Off Page (\url{www.offpage.nl}) on G-print 115g paper. 
}

\pagebreak
\thispagestyle{empty}

\begin{figure}[h]
\includegraphics[height=1.6cm]{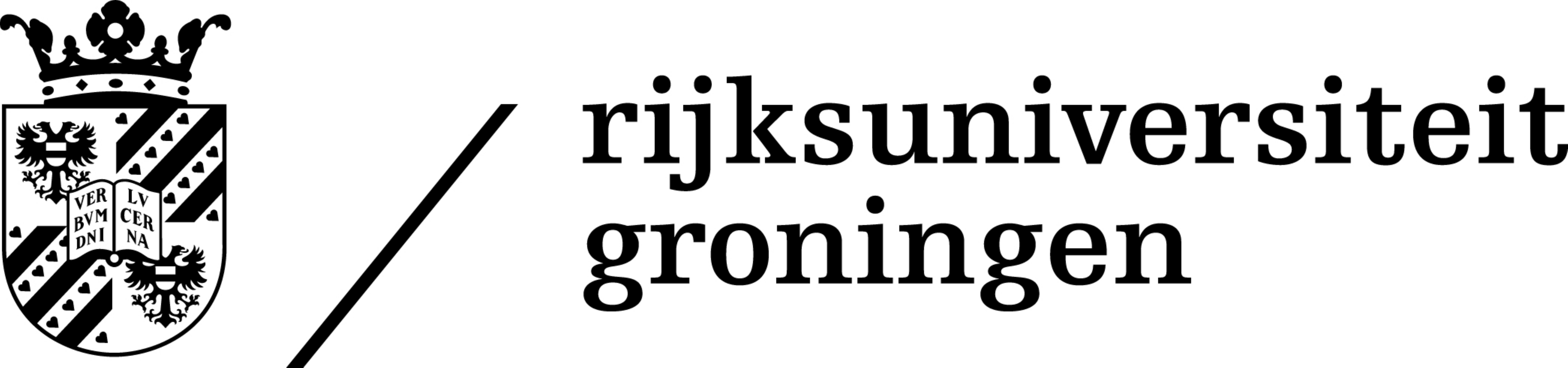}
\end{figure}
\vspace{0.5cm}
\begin{center}
        {\huge\bfseries{One Model to Rule them all}

        \smallskip

        {\Large\bfseries{Multitask and Multilingual Modelling\\ for Lexical Analysis}}}

        \bigskip

        \bigskip

        \smallskip

        {\large\bfseries{Proefschrift}}

        \bigskip

        \bigskip

        \smallskip

        ter verkrijging van de graad van doctor aan de\\
        Rijksuniversiteit Groningen\\
        op gezag van de\\
        rector magnificus prof. dr. E. Sterken\\
        en volgens besluit van het College voor Promoties.

        \bigskip
        De openbare verdediging zal plaatsvinden op\\

        \bigskip
        donderdag 7 december 2017 om 14.30 uur \\

        \bigskip

        \bigskip
        door

        \bigskip

        \bigskip
       {\large\bfseries{Johannes Bjerva}}

        \bigskip
       geboren op 21 maart 1990\\
       te Oslo, Noorwegen
\end{center}

\pagebreak
\thispagestyle{empty}
\noindent
\textbf{Promotor}\\
Prof. dr. ing. J. Bos\\

\noindent\textbf{Copromotor}\\
Dr. B. Plank\\

\noindent\textbf{Beoordelingscommissie}\\
Prof. dr. A. Søgaard\\
Prof. dr. J. Tiedemann\\
Prof. dr. L. R. B. Schomaker\\


\cleardoublepage
\chapter*{Acknowledgements}
\begin{abstract}\end{abstract}
\noindent
This has been a bumpy ride, to say the least, and having reached the end of this four-year journey, I owe a debt of gratitude to all of my friends, family, and colleagues.
Your support and guidance has certainly helped smooth out most of the ups and downs I've experienced.

First of all, I would like to thank my PhD supervisors.
Johan, the freedom you allowed me during the past four years has been one of the things I've appreciated the most in the whole experience.
This meant that I could pursue the track of research I felt was the most interesting, without which writing a whole book would have been much more arduous -- thank you!
Barbara, thank you for agreeing to join as co-supervisor so late in my project, and for putting in so much time during the last couple of months of my PhD.
The thesis would likely have looked quite different if you hadn't started in Groningen when you did.
I owe you a huge thanks, especially for the final weeks -- reading and commenting on the whole thesis in less than 24 hours during your vacation. Just, wow!

Next, I would like to thank Anders Søgaard, Jörg Tiedemann, and Lambert Schomaker for agreeing to form my thesis assessment committee.
I feel honoured that you took the time to read this thesis, and that you deemed it scientifically sound.
I also want to thank everyone who I have collaborated with throughout these years, both in Groningen and in Stockholm.
Thanks Calle for agreeing to work with a sign language novice such as myself.
Raf, it was an enlightening experience to work with you and see the world of `real' humanities research.
Robert, we should definitely continue with our one-week shared task submissions (with various degrees of success).

Most of the last few years were spent at the computational linguistics group in Groningen.
I would especially like to thank all of my fellow PhD students throughout the years.
Thanks Kilian, Noortje, Valerio, and Harm for welcoming me with open arms when I joined the group.
Special thanks to Kilian for being so helpful with answering all of my billions and billions of questions involved in finishing this thesis.
Also, a special thanks to both Noortje and Harm for the times we shared when I had just moved here (especially that first New Year's eve!).
Hessel, thanks for all the help with administrative matters while I was abroad, especially for sending a gazillion of travel declarations for me.
Rik and Anna, it was great getting to know you both better during the last few months -- hopefully you find the bookmark to be sufficiently sloth-y.
Dieke and Rob, thanks for being such great laid back drinking buddies and travel companions.
To all of you, and Pauline -- I hope we will continue the tradition of going all out whenever I come to visit Groningen!
A big thanks to the rest of the computational linguistics group, especially Gertjan, Malvina, Gosse, John, Leonie, and Martijn.
Also thanks to all other PhD students who started with me: Luis, Simon, Raf, Ruben, Aynur, Jonne, and everyone else whose name I've failed to mention.
A special thanks to Ben and Kaja - I hope we keep up our board-game centred visits to one another, no matter where we happen to pursue our careers.



I spent most of my final year in Stockholm, and it was great being in that relaxed atmosphere during one of the more intense periods of the past few years.
Most of all, I am sincerely grateful to Calle, to Johan (and Klara, Iris, and Vive), and to Bruno.
Your support is truly invaluable, and by my lights, there is not much more to say than \textit{<dank>}:(since a pal's always needed).
I'd also like to thank Johan and Calle especially for agreeing to observing weird Dutch traditions by being my paranymphs.
Josefina, El{\'i}sabet, and David, thank you all for being there for me during the past year.
Thanks to the computational linguistics group for hosting me during this period, especially Mats, Robert, Kina, and Gintar\.e.
Finally, thanks to everyone else at the Department of Linguistics at Stockholm University.


Having moved to Copenhagen this autumn has been an extremely pleasant experience.
I would like to thank the entire CoAStal NLP group for simply being the coolest research group there is.
Especially, I'd like to thank Isabelle for making the whole process of moving to Denmark so easy.
Thanks to Anders, Maria, Mareike, Joachim, Dirk, and Ana for being so welcoming.
I'd also like to thank everyone else at the image section at DIKU, especially the running buddies at the department, and most especially Kristoffer and Niels.


Finally, I would like to thank my family.
This bumpy ride would have been challenging to get through without their support through everything.
Kiitos, mamma! Takk/tack Aksel, Paulina, Julian, och Lucas!
Takk Olav, og takk Amanda!

\bigskip
\noindent
Let's see where the journey goes next!

\bigskip
\noindent Copenhagen, November 2017\\


\newpage
\tableofcontents



\mainmatter
\setcounter{secnumdepth}{4}

\cleardoublepage
\chapter{Introduction}
\begin{abstract}
\end{abstract}
\noindent
When learning a new skill, you take advantage of your preexisting skills and knowledge.
For instance, if you are a skilled violinist, you will likely have an easier time learning to play cello.
Similarly, when learning a new language you take advantage of the languages you already speak.
For instance, if your native language is Norwegian and you decide to learn Dutch, the lexical overlap between these two languages will likely benefit your rate of language acquisition.
This thesis deals with the intersection of learning multiple tasks and learning multiple languages in the context of Natural Language Processing (NLP), which can be defined as the study of computational processing of human language.
Although these two types of learning may seem different on the surface, we will see that they share many similarities.

Traditionally, NLP practitioners have looked at solving a single problem for a single task at a time.
For instance, considerable time and effort might be put into engineering a system for part-of-speech (PoS) tagging for English.
However, although the focus has been on considering a single task at a time, fact is that many NLP tasks are highly related.
For instance, different lexical tag sets will likely exhibit high correlations with each other.
As an example, consider the following sentence annotated with Universal Dependencies (UD) PoS tags \citep{nivre:2016}, and semantic tags \citep{bjerva:2016:semantic}.\footnote{PMB 01/3421. Original source: Tatoeba. UD tags obtained using UD-Pipe \citep{udpipe}}\footnote{The semantic tag set consists of 72 tags, and is developed for multilingual semantic parsing. The tag set is described further in Chapter~\ref{chp:semtag}.}

\begin{examples}
    \item{\gll We must draw attention to the distribution of this form in those dialects .
    \texttt{PRON} \texttt{AUX} \texttt{VERB} \texttt{NOUN} \texttt{\color{myblue}ADP} \texttt{\color{mygreen}DET} \texttt{NOUN} \texttt{\color{myblue}ADP} \texttt{\color{mygreen}DET} \texttt{NOUN} \texttt{\color{myblue}ADP} \texttt{\color{mygreen}DET} \texttt{NOUN} \texttt{PUNCT}
\gln\glend}
    \item{\gll We must draw attention to the distribution of this form in those dialects .
\texttt{PRO\ } \texttt{NEC} \texttt{EXS\ } \texttt{CON} \texttt{\color{myblue}REL} \texttt{\color{mygreen}DEF} \texttt{CON} \texttt{\color{myblue}AND} \texttt{\color{mygreen}PRX} \texttt{CON} \texttt{\color{myblue}REL} \texttt{\color{mygreen}DST} \texttt{CON} \texttt{NIL}
\gln\glend}
\end{examples}

\noindent While these tag sets are certainly different, the distinctions they make compared to one another in this example are few, as there are only two apparent systematic differences.
Firstly, the semantic tags offer a difference between definite (\texttt{\color{mygreen}DEF}), proximal (\texttt{\color{mygreen}PRX}), and distal determiners (\texttt{\color{mygreen}DST}), whereas UD lumps these together as \texttt{\color{mygreen}DET} (highlighted in green).
Secondly, the semantic tags also differentiate between relations (\texttt{\color{myblue}REL}) and conjunctions (\texttt{\color{myblue}AND}), which are both represented by the \texttt{\color{myblue}ADP} PoS tag, highlighted in blue.
Hence, although these tasks are undoubtedly different, there are considerable correlations between the two, as the rest of the tags exhibit a one-to-one mapping in this example.
This raises the question of how this fact can be exploited, as it seems like a colossal waste to not take advantage of such inter-task correlations.
In this thesis I approach this by exploring multitask learning (MTL, \citeauthor{caruana:1993}, \citeyear{caruana:1993,mtl}), which has been beneficial for many NLP tasks.
In spite of such successes, however, it is not clear \textit{when} or \textit{why} MTL is beneficial.  

Similarly to how different tag sets correlate with each other, languages also share many commonalities with one another.
These resemblances can occur on various levels, with languages sharing, for instance, syntactic, morphological, or lexical features.
Such similarities can have many different causes, such as common language ancestry, loan words, or being a result of universals and constraints in the properties of natural language itself (see, e.g., \citeauthor{chomsky:2005},\citeyear{chomsky:2005}, and \citeauthor{facultyoflanguage}, \citeyear{facultyoflanguage}).
Consider, for instance, the following German translation of the previous English example, annotated with semantic tags.\footnote{PMB 01/3421. Original source: Tatoeba.}

\begin{examples}
    \item{\gll Wir müssen die Verbreitung dieser Form in diesen Dialekten beachten .
    \texttt{PRO} \texttt{NEC} \texttt{DEF} \texttt{CON} \texttt{PRX} \texttt{CON} \texttt{REL} \texttt{PRX} \texttt{CON} \texttt{EXS} \texttt{NIL}
\glt\glend}
\end{examples}

\noindent Comparing the English and German annotations, there is a high overlap between the semantic tags used, and a high lexical overlap.
As in the case of related NLP tasks, this begs the question of how multilinguality can be exploited, as it seems like an equally colossal waste to not consider using, e.g., Norwegian PoS data when training a Swedish PoS tagger.
There are several approaches to exploiting multilingual data, such as annotation projection and model transfer, as detailed in Chapter~\ref{chp:mtl_bg}.
The approach in this thesis is a type of model transfer, in which such inter-language relations are exploited by exploring multilingual word representations, which have also been beneficial for many NLP tasks.
As with MTL, in spite of the fact that such approaches have been successful for many NLP tasks, it is not clear in which cases it is an advantage to \textit{go multilingual}.

Given the large amount of data available for many languages in different annotations, it is tempting to investigate possibilities of combining the paradigms of multitask learning and multilingual learning in order to take full advantage of this data.
Hence, as the title of the thesis suggests, the final effort in this thesis is to arrive at \textit{One Model to rule them all}.

This thesis approaches these two related aspects of NLP by experimenting with deep neural networks, which represent a family of learning architectures which are exceptionally well suited for the aforementioned purposes (described in Chapter~\ref{chp:nn}).
For one, it is fairly straightforward to implement the sharing of parameters between tasks, thus enabling multitask learning (discussed in Chapter~\ref{chp:mtl_bg}).
Additionally, providing such an architecture with multilingual input representations is also straightforward (discussed in Chapter~\ref{chp:mtl_bg}).
Experiments in this thesis are run on a large collection of tasks, both semantic and morphosyntactic in nature, and a total of 60 languages are considered, depending on the task at hand.

\section{Chapter guide}

The thesis is divided into five parts, totalling 9 chapters, aiming to provide answers to the following general research questions (\textbf{RQ}s):

\newcounter{rtaskno}
\newcommand{\rtask}[1]{\refstepcounter{rtaskno}\label{#1}}

\begin{enumerate}
    \item [{\bf RQ~\ref{rq:stag}}] To what extent can a semantic tagging task be informative for other NLP tasks? \rtask{rq:stag}
    \item [{\bf RQ~\ref{rq:mtl}}] How can multitask learning effectivity in NLP be quantified? \rtask{rq:mtl}
    \item [{\bf RQ~\ref{rq:sts}}] To what extent can multilingual word representations be used to enable zero-shot learning in semantic textual similarity? \rtask{rq:sts}
    \item [{\bf RQ~\ref{rq:multiling}}] In which way can language similarities be quantified, and what correlations can we find between multilingual model performance and language similarities?  \rtask{rq:multiling}
    \item [{\bf RQ~\ref{rq:mmmt}}] Can a multitask and multilingual approach be combined to generalise across languages and tasks simultaneously?  \rtask{rq:mmmt}
\end{enumerate}

\subsection*{Part I -- Background}
The goal of Part I is to provide the reader with sufficient background knowledge to understand the material in this thesis.
Chapter~\ref{chp:nn} contains a crash-course in neural networks, introducing the main concepts and architectures used in NLP.
Chapter~\ref{chp:mtl_bg} provides an introduction to multitask learning and multilingual learning, which are the two central topics of this work.

\subsection*{Part II -- Multitask Learning}

In Part II, the goal is to investigate multitask learning (MTL), in particular by looking at the effects of this paradigm in NLP sequence prediction tasks.
In Chapter~\ref{chp:semtag} we present a semantic tag set tailored for multilingual semantic parsing.
We attempt to use this tag set as an auxiliary task for PoS tagging, observing what effects this yields, answering {\bf RQ~\ref{rq:stag}}.
Chapter~\ref{chp:mtl} then delves deeper into MTL, attempting to find when MTL is effective, and how this effectiveness can be predicted by using information-theoretic measures ({\bf RQ~\ref{rq:mtl}}).

\subsection*{Part III -- Multilingual Learning}

Having looked at similarities between tasks, we turn to similarities between languages in Part III.
In Chapter~\ref{chp:multiling_sim}, we attempt to make a language-agnostic solution for semantic textual similarity, by exploiting multilingual word representations, thus answering {\bf RQ~\ref{rq:sts}}.
Having seen the results of combining related languages in this task, we try to quantify these effects in Chapter~\ref{chp:multilingual}, aiming to answer {\bf RQ~\ref{rq:multiling}}.

\subsection*{Part IV -- Combining Multitask and Multilingual Learning}
In Part IV we want to combine the paradigms of multitask learning and multilingual learning in order to make \textit{One Model to rule them all}.
Chapter~\ref{chp:gapfill} presents a pilot study taking a step in this direction, looking at predicting labels for an unseen task--language combination while exploiting other task--language combinations.

\subsection*{Part V -- Conclusions}

Finally, Chapter~\ref{chp:conclusions} contains an overview of the conclusions from this thesis.
In addition to this, we provide an outlook for future work in this direction, in particular focussing on the combined multitask--multilingual paradigm.

\section{Publications}

This thesis is based on the following publications:

\begin{enumerate}
    \item \bibentry{bjerva:2016:semantic}
    \item \bibentry{bjerva:2017:mtl}
    \item \bibentry{bjerva:2017:sts}
    \item \bibentry{bjerva:2017:sts:semeval}
    \item \bibentry{bjerva:2017:multilingual}
\end{enumerate}

\noindent Some parts of the thesis may also refer to the following peer-reviewed publications completed in the course of the PhD:
\begin{enumerate}
  \setcounter{enumi}{5}
    \item \bibentry{bjerva:14:eacl}
    \item \bibentry{bjerva:14:semeval}
    \item \bibentry{bjerva:2015}
    \item \bibentry{bjerva:2016:cl4lc}
    \item \bibentry{bjerva:2016:dsl}
    \item \bibentry{gronup}
    \item \bibentry{haagsma:metaphor}
    \item \bibentry{bjerva:2016:amr}
    \item \bibentry{bjerva:jdmdh}
    \item \bibentry{gmb:hla}
    \item \bibentry{pmb}
    \item \bibentry{bjerva:2017:sigmorphon}
    \item \bibentry{sjons:2017}
    \item \bibentry{rug:nli}
    \item \bibentry{bjerva:nli}
\end{enumerate}


\partimage[width=0.7\textwidth]{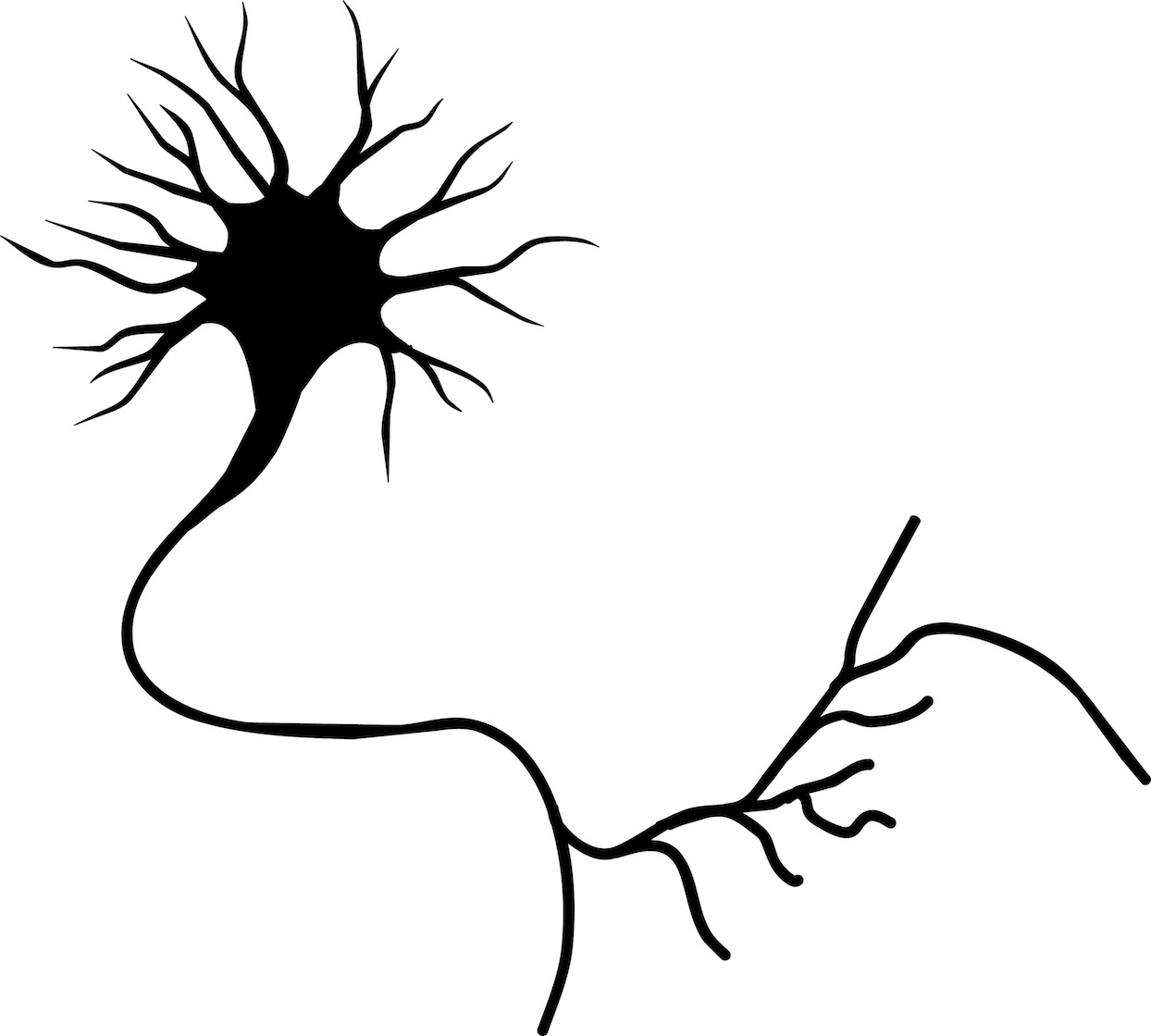}
\part{Background}


\renewcommand*{\thefootnote}{\fnsymbol{footnote}}
\chapter[An Introduction to Neural Networks]{
An Introduction \\to Neural Networks\hspace{-8pt}}
\renewcommand*{\thefootnote}{\arabic{footnote}}

\label{chp:nn}

\begin{abstract}
\absprelude
Deep Neural Networks are at the forefront of many state-of-the art approaches to Natural Language Processing (NLP).
The field of NLP is currently awash with papers building on this method, to the extent that it has quite aptly been described as a tsunami \citep{manning:2015}.
While a large part of the field is familiar with this family of learning architectures, it is the intention of this thesis to be available for a larger audience.
Hence, although the rest of this thesis assumes familiarity with neural networks, this chapter is meant to be a foundational introduction for those with limited experience in this area.
The reader is assumed to have some familiarity with machine learning and NLP, but not much beyond that.

We begin by exploring the basics of neural networks, and look at the three most commonly used general architectures for neural networks in NLP: Feed-forward Neural Networks, Recurrent Neural Networks, and Convolutional Neural Networks.
Some common NLP scenarios are then outlined together with suggestions for suitable architectures.
\end{abstract}
\clearpage



\section{Introduction}

The term \textit{deep learning} is used to refer to a family of learning models, which represent some of the most powerful learning models available today.
The power of this type of model lies in part in its intrinsic hierarchical processing of input features, which allows for learning representations at multiple levels of abstraction \citep{lecun:nature}.
This type of model is commonly referred to by several umbrella terms, such as \textit{deep learning}, and \textit{(deep) neural networks}.\footnote{While some make a distinction between \textit{deep} and non-deep neural networks (NNs) depending on the amount of layers used, there is no real consensus on where the line between these models should be. For the sake of consistency, I attempt to refer to this family of models as NNs, or some specification thereof, as consistently as possible.}
In this chapter, I aim to introduce the basic concepts of NNs, at a level sufficient to understand the work in this thesis.
The following sections are meant to cover the most basic workings in an intuitive, and theoretically supported, manner.
In addition to this background, I give an overview of the scenarios in which different recurrent NN architectures might be suitable in NLP (Section~\ref{sec:rnn_use_cases}).

\subsection*{History}

Neural networks have a long history, and have been popular in three main waves.\footnote{Only a very brief overview of the history is given here. For more details, the reader is referred to, e.g., \citeauthor{dl_history}, \citeyear{dl_history}, or \citeauthor{BengioBook}, \citeyear{BengioBook}.}
In the first wave, roughly between the $40s-60s$, they appeared under the moniker of \textit{cybernetics} (e.g., \citeauthor{cybernetics}, \citeyear{cybernetics}), inspired by the Hebbian learning rule \citep{hebb}.
In this wave, the \textit{perceptron} was first outlined \citep{perceptron}, which is still relatively popular today (see Section~\ref{sec:perceptron}).
Next, in the $80s$ and $90s$, \textit{connectionism} was on the rise.
In this wave, the algorithm for backwards propagation of errors \citep{backprop} was described, which is at the core of how neural networks are trained (see Section~\ref{sec:backprop}).
After an AI winter lasting roughly from $97$ to $06$, we finally arrive at the current wave (or tsunami), in which the term \textit{deep learning} is favoured.
This wave was initiated by works on deep belief networks \citep{hinton:06}, and has been the subject of much attention after successes in, e.g., reducing error rates in some tasks by more than 50\% \citep{lecun:nature}.
The recent advances made in the current wave further include breakthroughs in both recognition \citep{resnets:2016} and generation \citep{goodfellow:2014} of images, in NLP tasks such as machine translation \citep{bahdanau:2014,google:nmt} and parsing \citep{chen:2014}, as well as in the strategic board game Go \citep{alphago}, and the first-person shooter Doom \citep{doom_dl}.

%
%
%
%
%
%

%

\section{Representation of NNs, terminology, and notation}
\label{sec:nn_rep}

\begin{figure}[htbp]
    \centering
    \includegraphics[width=0.6\textwidth]{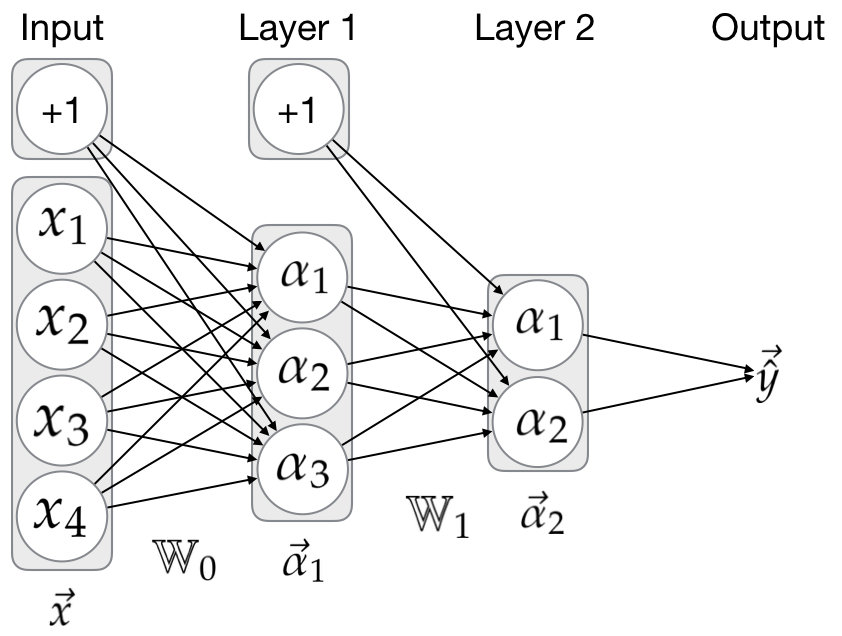}
    \caption{A basic Neural Network.}
    \label{fig:first_nn}
\end{figure}

Before embarking upon this journey and exploring the wondrous world of NNs, it is necessary to equip ourselves with some common ground in terms of terminology, notation, and how NNs are generally represented in this thesis.
Figure~\ref{fig:first_nn} contains a NN, which we will go through in detail.

\noindent First, note that the network is divided into three vertical slices.
Each such slice represents a \textit{layer}, marked by a light grey field.
Each layer contains one or more white circles, each representing a \textit{unit}, or \textit{neuron}.\footnote{In this thesis, the term \textit{unit} is preferred. While this is conventional in much of NLP, it is also debatable whether borrowing terminology for neural networks from neuroscience is motivated at all (this is discussed further in Section~\ref{sec:brain}).}
In this network, each unit has a connection to every unit in the following layer.
These connections are represented by arrows, which denote some weighting of the output of the unit at the start of the arrow, for the input of the unit at the end of the arrow.
Each layer can be described mathematically as a vector of \textit{activations}.
In the case of the first layer (the input layer), these activations are equal to the input (i.e.\ $\vec{\alpha}_0=\vec{x}$).
The final layer encodes the output of the network, which is denoted by $\hat{y}$.
Each layer up until the final layer also contains a special unit, marked by $+1$, which is called the \textit{bias} unit.\footnote{Although this can be discussed at length, suffice it to say that including bias units facilitates learning.}
The collection of all arrows between two layers, can be described mathematically as a matrix of weights (e.g.,\ $\mathbb{W}_0$).\footnote{We will cover ways in which to learn these weights later in this chapter (Section~\ref{sec:optimisation}).}
The application of the weight matrix to the input of the network can be described in linear algebraic notation as
\begin{equation}
    \begin{aligned}
        \vec{z}_1 &= \mathbb{W}_0\vec{x} = \mathbb{W}_0\vec{\alpha}_{0}, \\
    \end{aligned}
\end{equation}
\noindent where $\vec{z}_1$ is the vector (i.e.\ a series of numbers) resulting from this linear transformation.
The $\vec{z}$-vectors can be referred to as pre-activation vectors.
Each hidden layer thus first encodes the sum of the multiplications of each of the activations in the previous layer by some weight.
For instance, if we set $\mathbb{W}_0$ to be matrix of ones, then the pre-activation value of the first unit in the first hidden layer $z_0^1=\Sigma_{i=0}^{4}\vec{x}_i$.
The final piece of the puzzle is to calculate the output of each unit in the layer, by applying an activation function to the pre-activation vector,
\begin{equation}
    \begin{aligned}
        \vec{\alpha}_1 &= \sigma(\vec{z}_1), \\
    \end{aligned}
\end{equation}
where $\sigma$ is some non-linear activation function.\footnote{Traditionally the activation function ($\sigma$) used is some \textit{sigmoidal} function, such as the logistic function. However, many functions are suitable, given that they satisfy certain properties (see Section~\ref{sec:act_func}).}
Essentially, this is all that a basic FFNN is -- a series of matrix-vector multiplications, with non-linearities applied to it.
Now, how can this be used to solve problems, and how does the network learn to do this?
The answers to these questions will be made clear in the course of the following few pages.

\subsubsection*{Notation}
Before we continue, a brief note on the notation used in this thesis.
Scalars are represented with lower case letters ($x,y$), vectors are represented with lower case letters with arrows ($\vec{x}, \vec{y}$), and matrices are represented with blackboard upper case letters ($\mathbb{X}, \mathbb{Y}$).
Subscripts are used to denote the layer number, and where necessary, a superscript is used to denote indexation, to indicate the unit number in the case of deep networks.
For instance, $\alpha_i^j$ indicates the activation of unit $j$ in layer $i$, and $\mathbb{W}_i$ indicates the weight matrix for layer $i$.
Activation functions (Section~\ref{sec:act_func}) are denoted with $\sigma$, occasionally subscripted with the actual function used ($\sigma_{ReLU}$).

\section{Feed-forward Neural Networks}
\label{sec:ffnn}

A Feed-forward Neural Network (FFNN), also known as a multilayer perceptron, is perhaps the most basic variant of neural networks, and is the kind depicted in the previous figure.
As mentioned, a neural network can be seen as a collection of non-linear functions applied to a collection of matrices and vectors, thus mapping from one domain (e.g., words) to another (e.g., PoS tags).
Let us consider a concrete example, in which $\mathbb{X}$ contains information about the current weather, and $\mathbb{Y}=\{0,1\}$ denotes human-annotated labels denoting whether or not the weather is considered good.
In this case, $\mathbb{X}$ contains several variables, each representing a certain type of weather (e.g.\ $x_1$ = calm weather, and $x_2$ = sunny weather).\footnote{Note that we begin numbering of features with $1$, as the index $0$ is reserved for the bias terms.}
Table~\ref{tab:weather_and} represents the weather judgements of this example, where $y=1$ indicates good weather, and $y=0$ indicates not-so-good weather.

\begin{table}[h]
    \centering
    \caption{Weather appraisal (mimicking the logical \textsc{and} function).}
    \label{tab:weather_and}
    \begin{tabular}{rrr}
        \toprule
        \textbf{Calm ($x_1$)} & \textbf{Sunny ($x_2$)} & \textbf{Label ($y$)} \\
        \midrule
        $0$ & $0$ & $0$ \\
        $0$ & $1$ & $0$ \\
        $1$ & $0$ & $0$ \\
        $1$ & $1$ & $1$ \\
        \bottomrule
    \end{tabular}
\end{table}

\begin{figure}[h]
    \centering
    \includegraphics[width=0.5\textwidth]{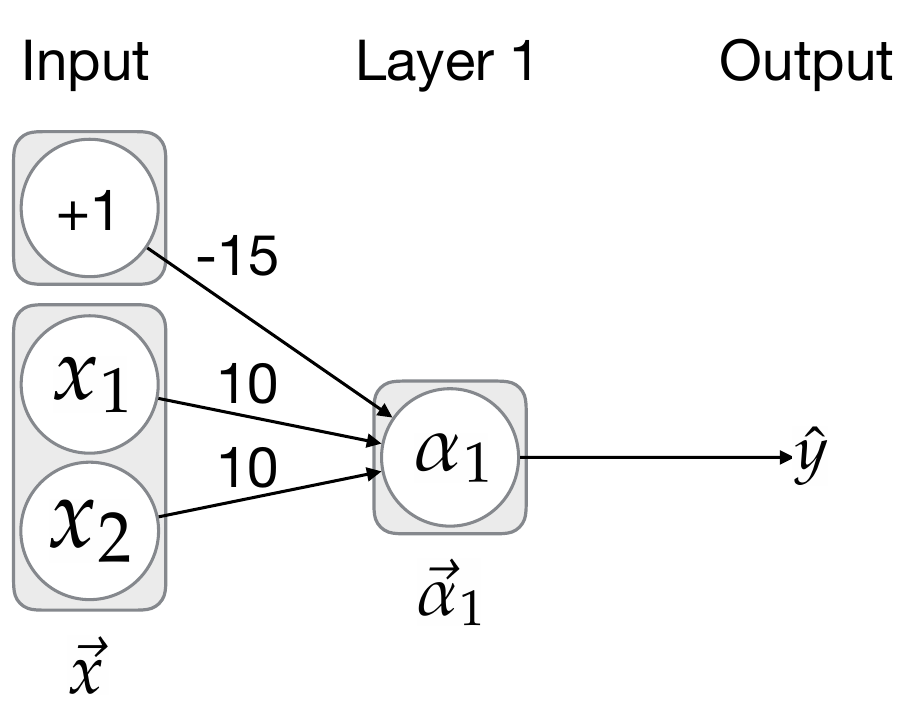}
    \caption{A Neural network coding the \textsc{and} logical function.}
    \label{fig:nn_and}
\end{figure}

\noindent The table shows the judgements of someone who considers weather to be good (i.e.\ $y=1$) only when it is both calm \textit{and} sunny (i.e.\ the logical \textsc{and} function).
Let us now consider our second neural network, which can solve the problem of determining whether the weather is good, based on this person's judgements, in Figure~\ref{fig:nn_and}.

\begin{table}[h]
    \centering
    \caption{Weather appraisal by the neural network in Figure~\ref{fig:nn_and}.}
    \label{tab:activations_and}
    \begin{tabular}{rrrrr}
        \toprule
        \textbf{Calm ($x_1$)} & \textbf{Sunny ($x_2$)} & $z_1$ & $\sigma(z_1) = \sigma(\alpha_1) = \hat{y}$ & \textbf{Label ($y$)} \\
        \midrule
        $0$ & $0$ & $-15$  & $\approx0$ & 0 \\
        $0$ & $1$ & $-5$   & $\approx0$ & 0 \\
        $1$ & $0$ & $-5$   & $\approx0$ & 0 \\
        $1$ & $1$ & $5$    & $\approx1$ & 1 \\
        \bottomrule
    \end{tabular}
\end{table}

\begin{figure}[h]
    \centering
    \includegraphics[width=0.5\textwidth]{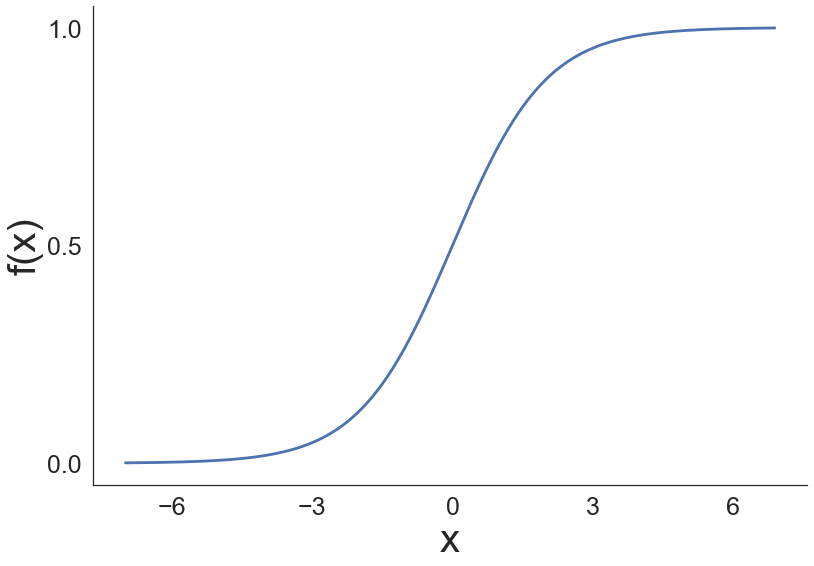}
    \caption{Plot of the logistic function.}
    \label{fig:logistic}
\end{figure}

\noindent Applying the calculations detailed in the previous section to this network yields the results shown in Table~\ref{tab:activations_and}.
If only one of $x_1,x_2$ is active, the activation $\alpha_1$ is approximately $0$, while the activation is $1$ if both $x_1$ and $x_2$ are active.
We get these values, by applying the perhaps most commonly used activation function to $z$.
This is the logistic function, defined as
\begin{equation}
    \begin{aligned}
        f(z)&=\frac{1}{1+e^{-z}},
    \end{aligned}
\end{equation}
where $e$ is Euler's number.
Plotting this function, yields the graph in Figure~\ref{fig:logistic}.
Hence, the value of $f(x)$ approaches 0 when $x<0$, and approaches 1 when $x>0$.

\label{sec:perceptron}
\noindent The simple neural network considered here, is what is also referred to as a \textit{perceptron}, although, technically, a perceptron uses the \textit{step} function as its activation function -- in other words, if $x<\lambda$ where $\lambda$ is some threshold, then $f(x)=0$, and if $x>\lambda$, then $f(x)=1$ \citep{perceptron}.
This is a very simple and useful architecture, but there are many problems which can not be easily solved by a perceptron, such as those in which the decision boundary to be learned is non-linear.
Take, for instance, the problem given in Table~\ref{tab:weather_xor}.

\begin{table}[h]
    \centering
    \caption{Weather appraisal (mimicking the logical \textsc{xor} function).}
    \label{tab:weather_xor}
    \begin{tabular}{ccc}
        \toprule
        \textbf{Snowy ($x_1$)} & \textbf{Sunny ($x_2$)} & \textbf{Label ($y$)} \\
        \midrule
        $0$ & $0$ & $0$ \\
        $1$ & $0$ & $1$ \\
        $0$ & $1$ & $1$ \\
        $1$ & $1$ & $0$ \\
        \bottomrule
    \end{tabular}
\end{table}

\begin{figure}[h]
    \centering
    \includegraphics[width=0.7\textwidth]{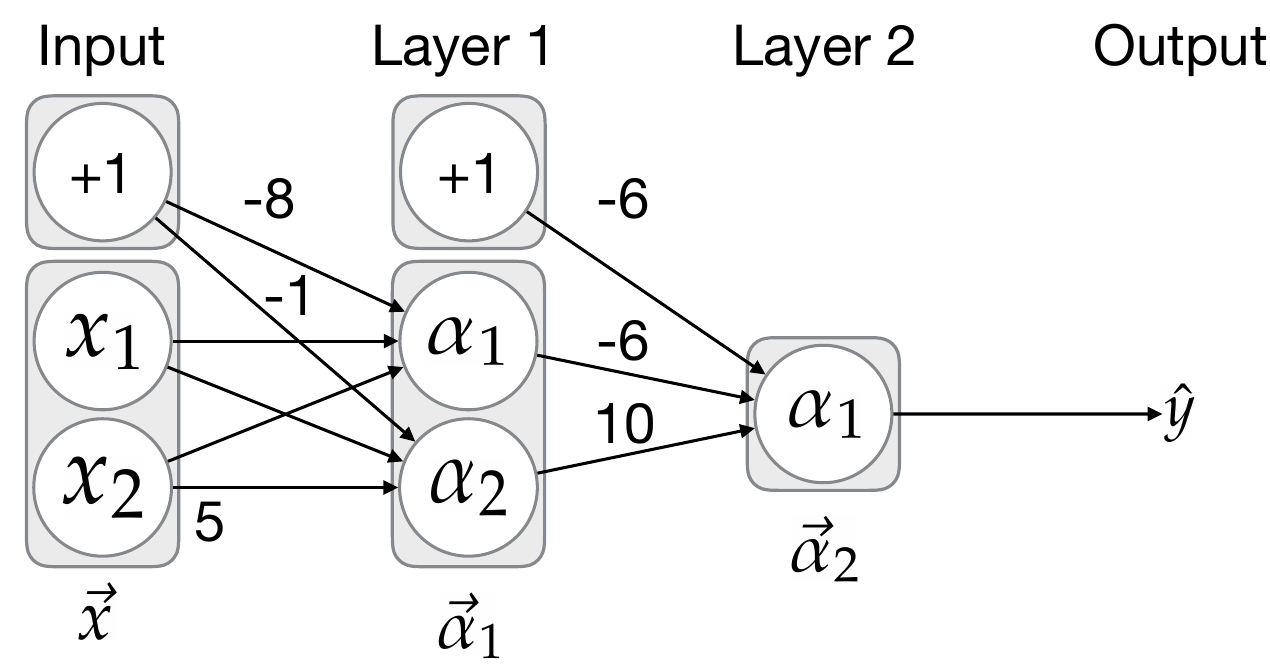}
    \caption{A Neural network coding the \textsc{xor} logical function.}
    \label{fig:nn_xor}
\end{figure}

\noindent This table shows the labels provided by some annotator who considers weather to be good (i.e.\ $y=1$) if it is \textit{either} snowy \textit{or} sunny -- but not both (i.e.\ the logical \textsc{xor} function).
As mentioned, a single unit (i.e.\ a perceptron) is not able to learn this decision boundary \citep{Minsky:1988}.\footnote{This is frequently cited as a potential catalyst for the \textit{the AI winter}, in which funding and interest in artificial intelligence was at a low. Nonetheless, it was known at the time that the \textsc{xor} problem could be solved by a neural network with more hidden units \citep{backprop}.}
Let us now have a look at a neural network which encodes this function.
This is depicted in Figure~\ref{fig:nn_xor}.
For the sake of clarity, all weights between $x_1,x_2$ and $a_1,a_2$ are set to $5$, but only one of these weights is shown.

Applying the calculations detailed in the previous section to this network yields the results shown in Table~\ref{tab:activations_xor}.
If both or none of $x_1,x_2$ are active, then the network will output $0$, whereas if one and only one of $x_1,x_2$ are active, the network will output $1$. Exactly what we want!

\begin{table}[h]
    \centering
    \caption{Weather appraisal by the neural network in Figure~\ref{fig:nn_xor}.}
    \label{tab:activations_xor}
    \begin{tabular}{rrrrrrr}
        \toprule
        \textbf{Calm ($x_1$)} & \textbf{Sunny ($x_2$)} & $z_1^1$ & $z_1^1$ & $z_2^1$ & $\sigma(z_2^1) = \alpha_1^2 = \hat{y}$ & $y$ \\
        \midrule
        $0$ & $0$ & $-8$ & $-1$ & $-6$  & $\approx0$ & 0 \\
        $1$ & $0$ & $-3$ & $4$  & $4$   & $\approx1$ & 1  \\
        $0$ & $1$ & $-3$ & $4$  & $4$   & $\approx1$ & 1  \\
        $1$ & $1$ & $3$  & $9$  & $-2$  & $\approx0$ & 0  \\
        \bottomrule
    \end{tabular}
\end{table}

\noindent The last two examples have shown how neural networks can encode certain simple functions.
It turns out that neural networks can do much more than this, and are in fact \textit{universal function approximators} \citep{cybenko:1989,hornik:1989}.
What this means, is that no matter the function, there is guaranteed to be a neural network with a single layer and a finite number of hidden units, such that for each potential input $x$, (a close approximation of) the value $f(x)$ is output from the network.

The class of networks discussed here is useful for many tasks in NLP, and can be used as a simple replacement for other classifiers.
Furthermore, they can be expanded by adding more units to each layer, or by adding more layers.
This allows such networks to learn to solve interesting NLP problems, like language modelling \citep{bengio:2003,vaswani:2013}, and sentiment classification \citep{iyyer:2015}.


Before going into other NN architectures, we will first consider some of the inner workings of NNs.
This includes how we represent our input, how weights are obtained, and finally some limitations which motivate the use of more complex architectures than FFNNs.

\subsection{Feature representations}
\label{sec:feature_rep}

In many NLP problems, we are interested in mapping from some textual language representation ($x$) to some label ($y$).
This textual representation can take many forms, both depending on the problem at hand, and on the choices made when approaching the problem.
As an example, say we are interested in doing sentiment analysis, i.e., given a text ($x$), predict whether the text is positive or negative in sentiment ($y$).
The perhaps simplest way of representing the text is to count the occurrences of each word in the text.
The intuition behind this is that if a text contains many negative words (\textit{horrible, bad, appalling}), it is more likely to be negative in sentiment than if it contains many positive words (\textit{wonderful, good, exquisite}).
Since these \textit{features} (i.e. counts of each word) need to be passed to an FFNN, they need to be represented as a single fixed-length vector $\vec{x}$.
What one might then do, is to assign an index to each unique word, and assign the count of each word to that index in the vector.
This can be referred to as a \textit{bag-of-words} model.

Although this type of feature representation is sufficient for some problems, and is traditionally used extensively, more recent developments include using other types of representations based on distributional semantics.
This is covered in more detail in the next chapter, in Section~\ref{sec:distrep}.

\subsection{Activation Functions}
\label{sec:act_func}

As stated in Section~\ref{sec:ffnn}, each hidden unit applies an activation function to the sum of its weighted inputs.
While many functions might be used, an activation function should have certain properties.
One such property is that the function needs to be non-linear.
It is for this kind of function that it has been proven that a two-layer neural network is a universal function approximator \citep{cybenko:1989}.
Additionally, the function should be monotonic, as the error surface associated with a single-layer model will then be convex \citep{wu:2009}.\footnote{A monotonic function is either non-increasing or non-decreasing in its entirety.}
There are several other important properties, which are not covered here.
Some of the more commonly used activation functions are listed in Table~\ref{tab:act_func}.

\begin{table}[htbp]
    \centering
    \caption{Commonly used activation functions in neural networks.}
    \label{tab:act_func}
    \begin{tabular}{ll}
        \toprule
        \textbf{Name} & \textbf{Function} \\
        \midrule
        Logistic (aka. sigmoid) & $f(x)=\frac{1}{1+e^{-x}}$ \\
        &  \\
        Hyperbolic Tangent (tanh) & $f(x)=\frac{2}{1+e^{-2x}}-1$ \\
        &  \\
        Rectified Linear Unit (ReLU)       & $f(x) = \left \{	\begin{array}{rcl}
	0 & \mbox{for} & x < 0\\
	x & \mbox{for} & x \ge 0\end{array} \right.$ \\
    &  \\
        Leaky ReLU       & $f(x) = \left \{	\begin{array}{rcl}
	0.01x & \mbox{for} & x < 0\\
	x     & \mbox{for} & x \ge 0\end{array} \right.$ \\
    &  \\
        Softmax & $f(\vec{x})_i = \frac{e^{x_i}}{\sum_{k=1}^K e^{x_k}}$ for $i = 1, …, K$ \\
        \bottomrule
    \end{tabular}
\end{table}

The traditionally popular logistic function was already described in Figure~\ref{fig:logistic}.
We will now consider some other commonly used activation functions.
Activation functions turn out to be one of the areas in which biological inspiration has been directly applicable to the development of neural networks.
The Rectified Linear Unit (ReLU) is in fact remarkably similar to what happens in a biological neuron \citep{relu:biological:a,relu:biological:b}.
That is to say, when the input is below a certain threshold, the neuron does not fire, and when the input is above this threshold, the neuron fires with a current proportional to its input.
ReLUs have been found to make it substantially easier to train deep networks \citep{relu}, and are currently very widely used.
One disadvantage of ReLUs is that they can wind up in a state in which they are inactive for almost all inputs, meaning that no gradients flow backward through the unit.
This, in turn, means that the unit is perpetually stuck in an inactive state, which at a large scale can decrease the network's overall capacity.
This is mitigated by using leaky ReLUs, for which even input $<0$ leads to some activity, allowing for error propagation given any input value.

The softmax function is generally only used at the final layer in classification problems, as it yields a probability distribution based on its input.

\subsection{Learning}
Learning in an FFNN happens in two phases.
First, in the forward propagation pass, the network sends a given input through the network, and produces some output.
Then, the \textit{error} of this output is calculated, as compared to some target label, and this error is sent back through the network, updating the weights of the network so as to make a more accurate prediction given the input in the next forward pass.\footnote{This is referred to as \textit{backward propagation of errors}, and is covered later in this section.}

We have already seen the largest part of the forward pass, as in the examples with the \textsc{and} and \textsc{xor} functions in Section~\ref{sec:ffnn}.
The only remaining part of the forward pass, is how the error of the network is calculated -- for this, a loss function is necessary.

\subsubsection*{Loss functions}
\label{sec:loss_function}

The loss functions used in NNs, generally fall into two classes -- those used for classification problems (i.e.\ when attempting to predict some discrete class label, out of a finite set of labels), and those used for regression problems (i.e.\ when attempting to predict some continuous score).
Classification is one of the most common cases in NLP (e.g.\ in POS tagging, NER, language identification, and so on).
In such cases, the activation function of the final layer is the softmax function, which allows for interpreting the layer's activations as a probability distribution over the labels under consideration.
Most often, the cross-entropy between this predicted probability distribution and the target probability distribution is used to calculate the error, or loss $L$, such that

\begin{equation}
    \begin{aligned}
    L_{cross-entropy}(\vec{\hat{y}}, \vec{y}) &=\ - \sum_{i} \vec{y}_i \log  \vec{\hat{y}}_i,
    \end{aligned}
\end{equation}
\noindent where $L$ denotes the loss function, $\vec{y}$ is the target probability distribution over labels, $\vec{\hat{y}}$ is the model's predicted model distribution given an input $x$.
A high error thus indicates that the predicted probability distribution is not consistent with the target probability distribution, and therefore changes should be made accordingly in the backward propagation pass.

Another loss function, common in regression, is the squared error function, defined as
\begin{equation}
    \begin{aligned}
    L_{squared} = (\hat{y} - y)^2,
    \end{aligned}
\end{equation}
where $\hat{y}$ is the predicted label, and $y$ is the true label.
This function is commonly used in regression, and is especially handy for explaining backpropagation, as in the next section.

\subsubsection*{Backpropagation}
\label{sec:backprop}

\begin{figure}[htbp]
    \centering
    \includegraphics[width=0.6\textwidth]{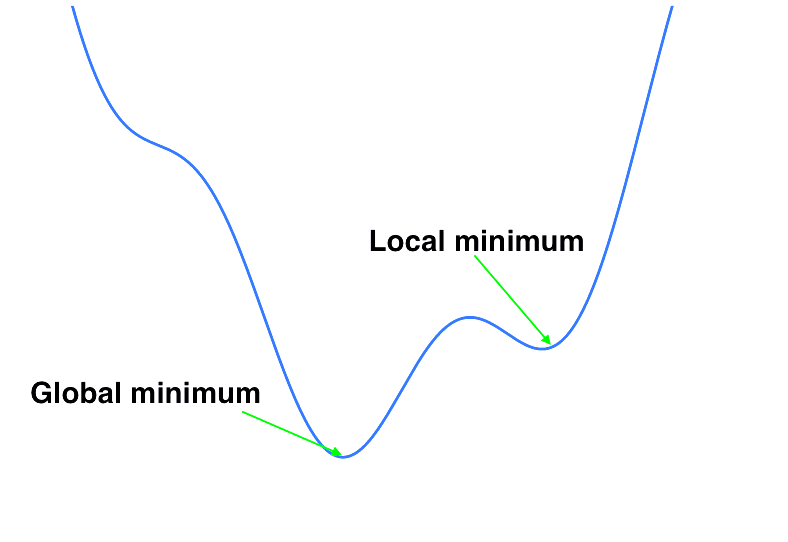}
    \caption{Non-convex error surface.}
    \label{fig:error_surface}
\end{figure}

Backward propagation of errors, or \textit{backprop} \citep{backprop,lecun:1998}, is an algorithm for calculating the gradient of the loss function, for each weight.
The gradient can, in turn, be used to update the weights by using an optimisation algorithm, such as gradient descent (discussed further in Section~\ref{sec:optimisation}).
Intuitively seen, gradient-based methods operate by viewing the errors as a geometric area, and use the slope of the area in which they are (i.e.,\ the gradient) in order to shift weights towards obtaining an error in a minimum of this area, as in Figure~\ref{fig:error_surface}.

\begin{figure}[htbp]
    \centering
    \includegraphics[width=0.8\textwidth]{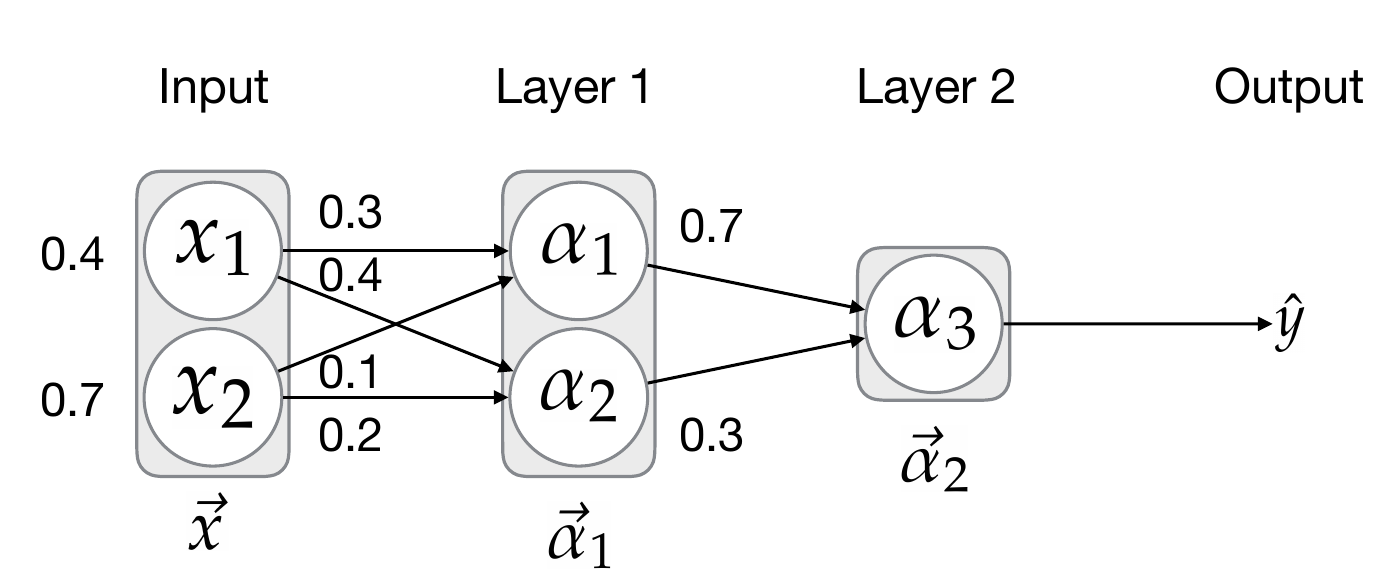}
    \caption{A neural network with weights for our backpropagation example.}
    \label{fig:backprop_ex1}
\end{figure}

Backprop relies on the fact that the partial derivative of the error of a certain weight $\mathbb{W}_i^j$, with respect to the loss function, can be easily calculated if we know the partial derivative of the outputs in the layer following that weight.
It turns out that this is, indeed, the case, as the derivative of the output layer is quite easily obtained.
The output error of a given unit, $\delta_i$, is calculated as
\begin{equation}
    \label{eq:backprop}
    \begin{aligned}
    \delta_i &= \begin{cases}
    \alpha_i(1-\alpha_i)(\alpha_i-y_i)& \text{for output units } i,\\
    \alpha_i(1-\alpha_i)(\alpha_i\sum_{\ell\in L} \delta_\ell \mathbb{W}_{i\ell})  & \text{for other units } i,
    \end{cases}
    \end{aligned}
\end{equation}
where $\alpha_i$ is the activation of the current unit, $y_i$ is the target output, $L$ is the collection of all units receiving input from the current unit, and $\mathbb{W}_{i\ell}$ is the weight from the current unit to unit $\ell$.
Let us consider a concrete example, and go through the forward pass, calculation of the error, and the backward pass.
The network in Figure~\ref{fig:backprop_ex1} shows a neural network with its weights.

%
%
%
%

Assuming that the activation function used is the logistic function, the following calculations hold:
\begin{equation}
    \begin{aligned}
    \alpha_1 &= \sigma(x_1\times0.3 + x_2\times0.1) = \sigma(0.4\times0.3 + 0.7\times0.1) = 0.547,\\
    \alpha_2 &= \sigma(x_1\times0.4 + x_2\times0.2) = \sigma(0.4\times0.4 + 0.7\times0.2) = 0.574,\\
    \alpha_3 &= \sigma(\alpha_1\times0.7 + \alpha_2\times0.3) = \sigma(0.19\times0.7 + 0.3\times0.3) = 0.635.
    \end{aligned}
\end{equation}
\noindent Assuming that the target output is $y=0.4$, we can now calculate the error.
Applying equation~\ref{eq:backprop}, we can obtain the error of the output, namely
\begin{equation}
    \begin{aligned}
    \delta_3 &= \alpha_3 (1-\alpha_3)(\alpha_3-y), \\
    &= 0.635(1-0.635)(0.635-0.4), \\
    &= 0.054.
    \end{aligned}
\end{equation}
The errors of the two hidden units can also be calculated, yielding
\begin{equation}
    \begin{aligned}
    \delta_2 &= \alpha_2 (1-\alpha_2)(\alpha_2 \delta_3\mathbb{W}_{2\ell}),\\
    &= 0.547(1-0.547)(0.547\times0.027\times0.3), \\
    &= 0.001,
    \end{aligned}
\end{equation}
and
\begin{equation}
    \begin{aligned}
    \delta_1 &= \alpha_1 (1-\alpha_1)(\alpha_1 \delta_3\mathbb{W}_{1\ell}),\\
    &= 0.547(1-0.547)(0.547*0.027*0.7), \\
    &= 0.002.
    \end{aligned}
\end{equation}

\noindent We now need to update the weights used, via gradient descent.
This can be done by shifting the weights with some constant with respect to the error obtained,
\begin{equation}
    \label{eq:grad_descent}
    \begin{aligned}
     \Delta \mathbb{W}_{ij} &= -\gamma\alpha_i\delta_j,
    \end{aligned}
\end{equation}
where $\Delta \mathbb{W}_{ij}$ is the amount with which to change $\mathbb{W}_{ij}$ (i.e.,\ the weight between the firing and receiving unit), $\gamma$ is some learning rate, $\alpha_i$ is the activation of the firing unit, and and $\delta_j$ is the error of the receiving unit.
Hence, if we set $\gamma=1$, the changes of the weights are calculated as
\begin{equation}
    \begin{aligned}
        \Delta\mathbb{W}_{x_1,\alpha_1}      &= -\gamma x_1\delta_1     = -1\times0.4\times0.002 = -0.0008,\\
        \Delta\mathbb{W}_{x_1,\alpha_2}      &= -\gamma x_1\delta_2     = -1\times0.4\times0.001 = -0.0004,\\
        \Delta\mathbb{W}_{x_2,\alpha_1}      &= -\gamma x_2\delta_1     = -1\times0.7\times0.002 = -0.0014,\\
        \Delta\mathbb{W}_{x_2,\alpha_2}      &= -\gamma x_2\delta_2     = -1\times0.7\times0.001 = -0.0007,\\
        \Delta\mathbb{W}_{\alpha_1,\alpha_3} &= -\gamma\alpha_1\delta_3 = -1\times0.547\times0.054 = -0.030,\\
        \Delta\mathbb{W}_{\alpha_2,\alpha_3} &= -\gamma\alpha_2\delta_3 = -1\times0.574\times0.054 = -0.031.\\
    \end{aligned}
\end{equation}
Using the new weights yields the following activations in the next forward pass, given the same input:
\begin{equation}
    \begin{aligned}
        \alpha_1 &= \sigma(0.4\times(0.3+\Delta\mathbb{W}_{x_1,\alpha_1}) + 0.7\times(0.1+\Delta\mathbb{W}_{x_2,\alpha_1}) = 0.547,\\
        \alpha_2 &= \sigma(0.4\times(0.4+\Delta\mathbb{W}_{x_2,\alpha_1}) + 0.7\times(0.2+\Delta\mathbb{W}_{x_2,\alpha_2}) = 0.574,\\
        \alpha_3 &= \sigma(0.547\times(0.7+\Delta\mathbb{W}_{\alpha_1,\alpha_3}) + 0.574\times(0.3+\Delta\mathbb{W}_{\alpha_2,\alpha_3}) = 0.627,
    \end{aligned}
\end{equation}
and the output error
\begin{equation}
    \begin{aligned}
    \delta_3 &= \alpha_3 (1-\alpha_3)(\alpha_3-y), \\
    &= 0.318(1-0.318)(0.318-0.4), \\
    &= 0.053,
    \end{aligned}
\end{equation}
which is smaller than the previous error where $\delta_3=0.054$.
This process is repeated with other training examples, until some criterion is reached, such as a sufficiently low average loss.

\subsubsection*{Optimisation Methods}
\label{sec:optimisation}

Backpropagation, as described in the previous section, can provide us with the derivatives of the error surface.
This can be used in a variety of ways to update the weights.
What we just saw, in Equation~\ref{eq:grad_descent}, is known as gradient descent.
One of the most commonly used optimisation methods is Stochastic Gradient Descent (SGD).
In SGD, a minibatch of $n$ samples is drawn from the training set, the gradient is calculated based on this batch, and the weights are then updated accordingly \citep{bottou:1998}.
Other algorithms, such as AdaGrad \citep{duchi:2011} and RMSProp \citep{hinton:2012}, learn and adapt the learning rate ($\gamma$) for each weight.
Modifying the learning rate in this manner can both increase the rate at which the error decreases, and lead to lower overall errors.
A recent and increasingly popular optimisation method is Adam, which is similar to RMSProp and yields better results on a many problems \citep{adam}.
The choice of optimisation method is not all that straightforward, and no real consensus exists for how this should be done \citep{schaul:2014}.
Hence, commonly, trial-and-error is applied in order to make this choice, by experimentally investigating performance on a development set.

\subsubsection*{Finding the global minimum}
The goal of an optimisation algorithm is to find the global minimum, as shown in Figure~\ref{fig:error_surface}.
What so-called \textit{gradient-based} optimisation algorithms do, is to calculate the derivative with respect to this error surface, and shift the weights so as to move towards the closest of all local minima.
Such local minima can, however, be the source of a host of problems, if the loss is high compared to the global minimum.
This is a frequently occurring issue, and it is possible to construct small neural networks in which this scenario appears \citep{sontag:1989,brady:1989,gori:1992}.
It turns out, however, that practically speaking, when considering larger neural networks, it is not particularly important to find the global minimum.
This has to do with the fact that, in the case of supervised learning with deep neural networks, most local minima appear to have a low loss function value, roughly equivalent to that of the true global minimum \citep{saxe:2013,dauphin:2014,goodfellow:2015,choromanska:2015}.

\subsubsection*{Parameter Initialisation}
There are several methods for initialising the weights in a neural network.
Naively, one might think to set the all weight matrices $\mathbb{W}=1$, however due to how backpropagation works, this will result in all hidden units representing the same function, and receiving the exact same weight updates.
Therefore, some random process is required.
Common methods include those introduced by \citet{glorot:init}, and \citet{saxe:2013}.
When employing the ReLU activation function, \citet{he:2015:init} show that weights should be initialised based on a Gaussian distribution with standard deviation $\sqrt{\frac{2}{d_{in}}}$, where $d_{in}$ is the input dimensionality.
In the case of recurrent neural networks, which are covered in Section~\ref{sec:rnn}, particular care needs to be taken, and weight initialisation is often done using orthogonal matrices (cf. \citeauthor{BengioBook}[p.404], \citeyear{BengioBook}).

\subsubsection*{Regularisation in Neural Networks}

One of the most common problems when training an ML system in general, is that of overfitting -- and neural networks are no exception.
Overfitting occurs when the network does not generalise to data outside of the training set, while having a low loss on the training set itself.
Generalisation is one of the most important parts of learning, as learning without generalisation is simply the memorisation of a training set.
A model which has only memorised the training set is of little practical value, as it will most likely fail miserably on unseen examples.
In order to avoid overfitting, regularisation techniques are typically employed.
The probably most common regularisation technique used today, is dropout \citep{dropout}.
In dropout, every activation has a probability $p$ of not being included in the forward and backward passes, during training.
This procedure leads to significantly lower generalisation error, as the network needs to be more robust, and less reliant on specific units.
In the case of recurrent neural networks, which are covered next, specific variants of dropout exist. such as recurrent dropout \citep{recurrent_do}, or variational dropout \citep{rnn:dropout} in which the same dropout mask is used for each time step.
Another commonly used manner of regularisation is \textit{weight decay}, in which the magnitudes of weights are decreased according to some criterion \citep{weightdecay}.
%

\section{Recurrent Neural Networks}
\label{sec:rnn}

Although FFNNs are suitable for many problems, they do not take the structure of the input into account.
Although it is possible to attempt to enforce this in such a network, this has several disadvantages, such as the fact that the amount of parameters which need to be tuned can become prohibitively large.
Luckily, there are architectures for dealing with structure, as this is not entirely unimportant when considering natural language.
Two such approaches are covered in the following sections.

\begin{figure}
    \includegraphics[width=\textwidth]{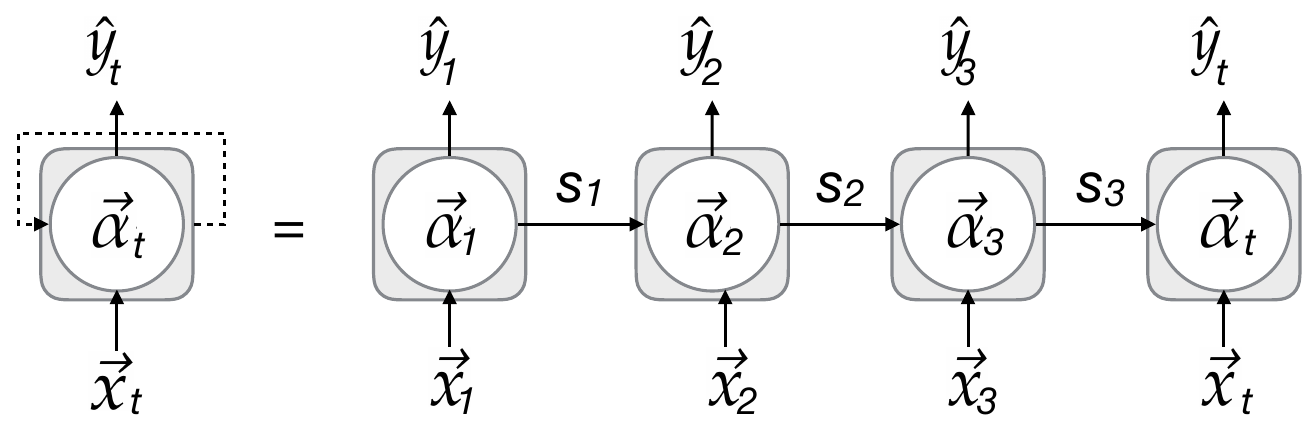}
    \caption{A simple RNN with a connection from the hidden state of the previous time step to the current side step. Left side shows the FFNN with the loop, whereas the right side shows the unrolled network.}
    \label{fig:rnn}
\end{figure}

Recurrent Neural Networks (RNNs) are an extension of feed-forward neural networks, which are designed for sequential data \citep{rnn}.
They can be thought of as a sequence of copies of the same FFNN, each with a connection to the following time step in the sequence, sharing parameters between time steps.
RNNs take a sequence of \textit{arbitrary length} as input ($x_1, x_2,\ldots,x_t$), and return another sequence ($\hat{y}_1, \hat{y}_2,\ldots,\hat{y}_t$).
Each $x_t$ in the input sequence is a vector representation of element $n_{t}$ in the sequence.
Each $\hat{y}_t$ in the output sequence can take advantage of information in the sequence up to step $t$ in the input sequence.
This is illustrated in Figure~\ref{fig:rnn}.
Each layer is shown as containing only one unit, which is here meant as an abstraction depicting the entire internal representation of the RNN.
The left side of the figure depicts an FFNN with a loop, whereas the right side shows the \textit{unrolled} version of the network.
The output of the hidden layer is passed as an input to the hidden layer in the next time step.

An RNN is essentially a group of FFNNs with connections to one another.
This connection is a sort of loop, going from the hidden layer of the network at time $x_t$ to the hidden layer at $x_{t+1}$.
In other words, $\vec{\hat{y_t}}$ is calculated as

\begin{equation}
    \begin{aligned}
        \vec{z}_t       &= \mathbb{W}_s\vec{x}_{t} + \mathbb{U}\vec{s}_{t-1}, \\
        \vec{s}_t       &= \sigma_s(\vec{z}_t), \\
        \vec{\hat{y}}_t &= \sigma_y(\mathbb{W}_y\vec{s}_t),
    \end{aligned}
\end{equation}

\noindent where $\mathbb{W}_s$ is the matrix of weights for the current time step's input ($\vec{x}_t$), $\mathbb{U}$ is a weight matrix for the connections from the previous time step, $\vec{s}_t$ is a state vector representing the history of the sequence, $t$ is the index of the current time step, $\mathbb{W}_y$ is the matrix of weights for the output, and the rest is defined as for FFNNs.
This is what is also referred to as an Elman net, or a Simple RNN \citep{rnn}.
The advantage of having access to $\vec{s}$, is that the network can take advantage of preceding information when outputting $\vec{\hat{y}}_t$.
For instance, in the case of POS tagging, if the current input is \textit{fly}, and the state vector shows that the previous word was \textit{to}, we most likely want to output the tag \textit{verb}.
Hence, in this way, the prediction at each time step is conditioned on the inputs in the entire preceding sequence.
There are also variants of this, in which the net's outputs are used to calculate the state vector, as in the case of Jordan nets \citep{jordan:1997}, which is defined such that
\begin{equation}
    \begin{aligned}
        \vec{z}_t       &= \mathbb{W}_s\vec{x}_{t} + \mathbb{U}\vec{\hat{y}}_{t-1}.
    \end{aligned}
\end{equation}

Although RNNs, in theory, can learn long dependencies (i.e.\ that an output at a certain time step is dependant on the state at a time step far back in the history), and can handle input sequences of arbitrary length, they are in practice heavily biased to the most recent items in the given sequence, and thus difficult to train on long sequences with long dependencies \citep{bengio:1994}.
For instance, in the case of language modelling, given a sentence such as \textit{My mother is from Finland, so I speak fluent $\ldots$}, it is quite likely that the omitted word should be \textit{Finnish}.
However, as the distance between such dependencies grows, it becomes increasingly difficult for an RNN to make use of such contextual information.
In general, this is because deep neural networks suffer from having \textit{unstable} gradients, as the gradients calculated by backprop (Section~\ref{sec:backprop}) are dependant on the output of the network, which can be quite far away from the first layers in the network.
One problem with this is that this can lead to \textit{vanishing} gradients (i.e.\ the gradient becomes very small).
This happens since the gradient in early layers of the network are the result of a large number of multiplication operators on numbers $<1$.
One might consider the fact that, since these multiplications involve the weights in the network, we might just set the weights to be really large.
Although this might seem like a good idea, this will likely lead to the converse of the issue one is trying to avoid, namely that of \textit{exploding} gradients.
Since most common optimisation methods are gradient-based, this is problematic (see Section~\ref{sec:optimisation} for optimisation details).

\subsection{Long Short-Term Memory}

\begin{sidewaysfigure}[p]
    \centering
    \includegraphics[width=0.9\textwidth]{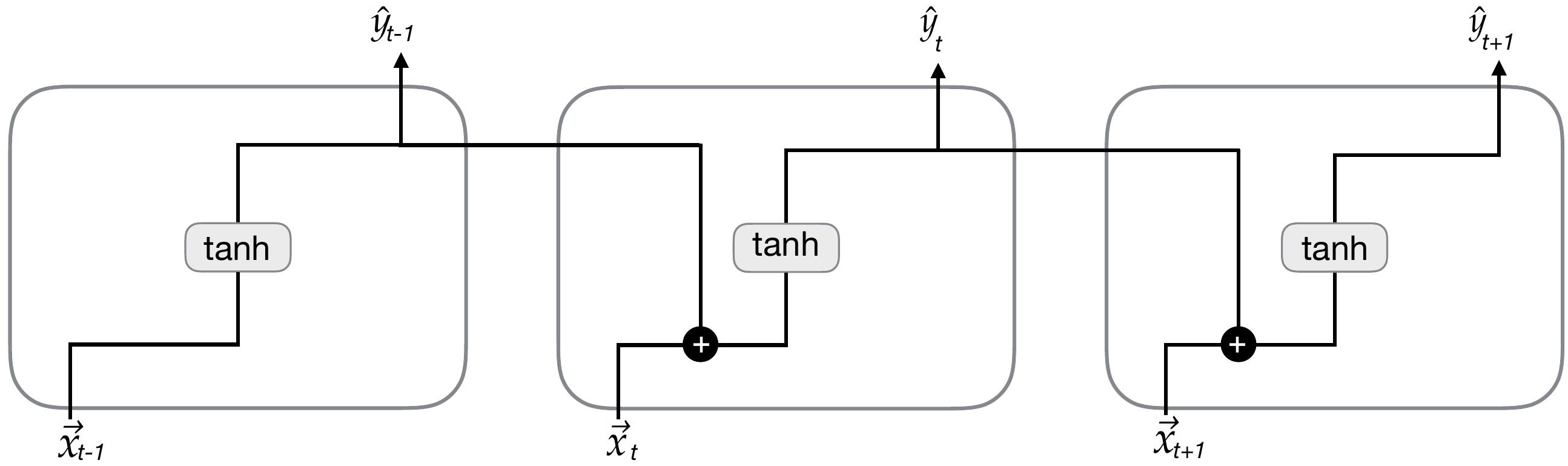}
    \caption{Internal view of an RNN in three time steps.\label{fig:rnn_internal}}
    \includegraphics[width=0.9\textwidth]{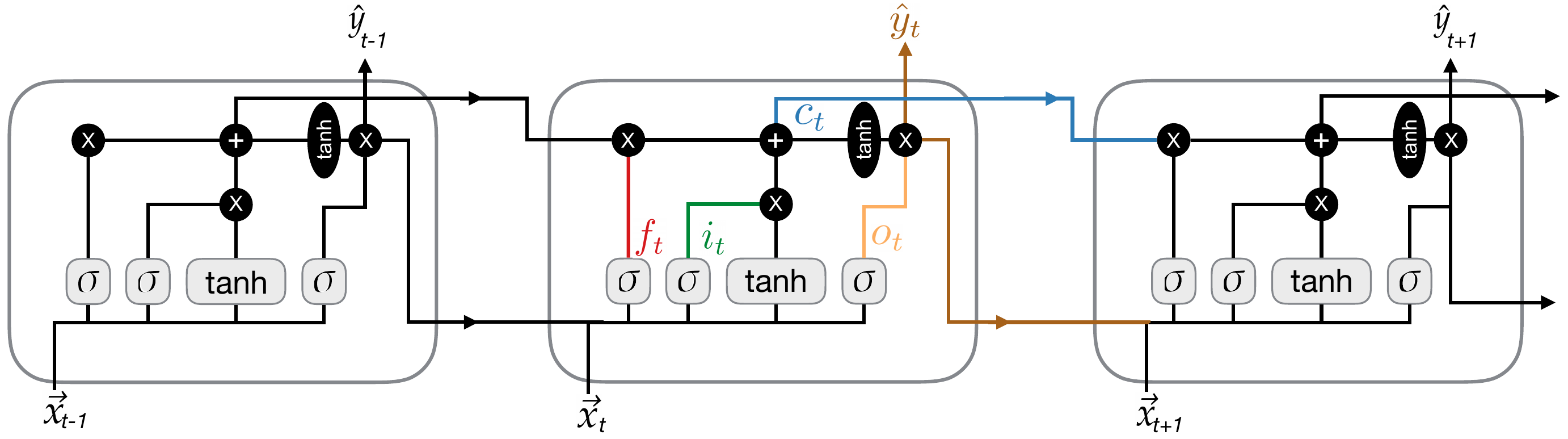}
    \caption{Internal view of an LSTM in three time steps.\label{fig:lstm_internal}}
\end{sidewaysfigure}

Previous work has attempted to solve this problem by adapting the optimisation method used \citep{bengio:2013,pascanu:2013,sutskever:2014}, however the more successful approach has been to modify the neural network architecture itself.
Because of such efforts, there are several types of RNNs specifically designed to cope with this issue, essentially by enforcing a type of protection of the memory of the history of the input sequence, storing and maintaining important features, while neglecting and forgetting unimportant features.
One such method, namely Long Short-Term Memory (LSTM), was described in \citet{lstm}, and saw an explosion in popularity around 2014, following several influential papers (e.g.~\citet{sundermeyer:lstm,sutskever:2014,dyer:2015}).
An LSTM is an extension of RNNs, with \textit{memory cells}, engineered to cope with the issue of unstable gradients, and have been shown to be able to capture long-range dependencies \citep{lstm,cho:2015}.
For an overview of the many variations of LSTMs which appear in the literature, see \citet{greff:2017}.

Whereas an RNN only has a single internal layer (Figure~\ref{fig:rnn_internal}), typically with a \textit{tanh} (hyperbolic tangent) activation, an LSTM is somewhat more complicated (Figure~\ref{fig:lstm_internal}).
An LSTM contains gates, denoted in the figure by the $\sigma$ layers, which are used to modify the extent to which old information is remembered or forgotten.
Part of the explanation for LSTMs involves the observation that they, on the surface, can be seen as a combination of Elman nets and Jordan nets, in that both the cell state and the hypothesis are passed between states.
In detail, an LSTM is implemented as follows
\begin{equation}
    \begin{aligned}
    \color{myred}{f_t} &= \sigma(\mathbb{W}_{f} x_t + \mathbb{U}_{f} {\color{mybrown}{\hat{y}_{t-1}}} + b_f), \\
    \color{mygreen}{i_t} &= \sigma(\mathbb{W}_{i} x_t + \mathbb{U}_{i} {\color{mybrown}{\hat{y}_{t-1}}} + b_i), \\
    \color{myorange}{o_t} &= \sigma(\mathbb{W}_{o} x_t + \mathbb{U}_{o}{ \color{mybrown}{\hat{y}_{t-1}}} + b_o), \\
    \color{myblue}{c_t} &= {\color{myred}{f_t}} \circ {\color{myblue}{c_{t-1}}} + {\color{mygreen}{i_t}} \circ \sigma_c(\mathbb{W}_{c} x_t + \mathbb{U}_{c} {\color{mybrown}{\hat{y}_{t-1}}} + b_c), \\
    \color{mybrown}{\hat{y}_t} &= {\color{myorange}{o_t}} \circ \sigma({\color{myblue}{c_t}}),
    \end{aligned}
\end{equation}
where $\color{myred}{f_t}$ represents the output of the \textit{forget gate}, $\color{mygreen}{i_t}$ represents the output of the \textit{input gate}, $\color{myorange}{o_t}$ represents the output of the \textit{output gate},
$\color{myblue}{c_t}$ represents the \textit{cell state}, ${\color{mybrown}{\hat{y}_{t}}}$ represents the output vector, $\mathbb{W}$ and $\mathbb{U}$ represent the weight matrices,
$x_t$ represents the current input, and $b$ represents bias units.
Each of these parts are covered in detail in the following sections.

\subsubsection*{Cell state}
The cell state, $\color{myblue}{c_t}$, is the line coded in blue in Figure~\ref{fig:lstm_internal}.
It is similar to the state vector, $\vec{s}$, in a simple RNN, but its content is maintained and protected by three gates.
The forget gate's output $\color{myred}{f_t}$ determines to which extent each feature in the cell state should be kept.
This is done by observing the current input $x_t$, and the previous time step's output $\color{mybrown}{\hat{y}_{t-1}}$.

For instance, say we have a POS tagger which at time $t$ observes a word which could be either a noun or a verb (e.g.\ \textit{fly}).
If the previous word was \textit{to}, this would be useful to keep in mind in order to better predict the next tag.
Following this time step, however, we might want to forget about this word, when predicting the next tag.

Adding information to the cell state happens in two steps.
We first decide on which values in the cell state to update again observing $x_t$ and $\color{mybrown}{\hat{y}_{t-1}}$, again deciding for each dimension the extent to which we will add information.
This is denoted by the input gate $\color{mygreen}{i_t}$.
The vector which is added to the cell state, is calculated by passing $x_t$ and $\color{mybrown}{\hat{y}_{t-1}}$ through a non-linearity, marked by tanh in the figure.
In the POS tagging example, we want to add the information regarding the preceding determiner.
These two vectors are then summed, resulting in our new cell state $\color{myblue}{c_t}$.

\subsubsection*{Output}
Finally, we need to output a value from the current time step.
This output is based on the cell state, $\color{myblue}{c_t}$, which is first run through a non-linearity (usually tanh), and filtered again by $\color{myorange}{o_t}$, which also observes $x_t$ and $\color{mybrown}{\hat{y}_{t-1}}$, when deciding on which dimensions to keep, and to what extent.

\subsubsection*{Gated Recurrent Units}
In addition to the many LSTM variants \citep{greff:2017}, Gated Recurrent Units (GRUs) represent a different variant of gated RNNs which was independently developed to LSTMs and similar in both purpose and implementation \citep{cho:ea:2014}.
The main difference between LSTMs and GRUs is the fact that GRUs do not have separate memory cells, and only include two gates -- an update gate, and a reset gate \citep{gru}.
This in turn means that GRUs are computationally somewhat more efficient than LSTMs.
In practice, both LSTMs and GRUs have been found to yield comparable results \citep{gru,jozefowicz:2015}.
On a general level, the performance of various gated RNN architectures is, at least in the case of large amounts of data, closely tied to the number of parameters \citep{collins:2017,lstm:eval}.

\subsubsection*{Bi-directionality}
Many properties of language depend on both preceding and proceeding contexts, so it is useful to have knowledge of both of these contexts simultaneously.
This can be done by using a bi-directional RNN variant, which makes both forward and backward passes over sequences, allowing it to use both contexts simultaneously for the task at hand \citep{birnn,graves:schmidhuber:2005,goldberg:primer}.
Bi-directional GRUs and LSTMs have been shown to yield high performance on several NLP tasks, such as POS tagging, named entity tagging, and chunking \citep{wang2015:unified,yang:2016,plank:2016}.

\subsection{Common use-cases of RNNs in NLP}
\label{sec:rnn_use_cases}

In NLP, there are four general scenarios for producing some sort of analysis for a given text.
Consider that we have the following sentence as input:

\begin{examples}
    \item \textit{I'm not fussy.}\footnote{PMB 76/2032, Original source: Tatoeba}
\end{examples}

\noindent We might want to analyse this unit as a whole, for instance in order to judge that the text is in English, and not in some other language, or to determine the native language of the person writing it, or the sentiment of the text itself.
This can be referred to as a \textbf{many-to-one} scenario, since we have several smaller units (e.g.\ words or characters), which we want to translate into a single score or class, depending on the task at hand.

On the other hand, we might want to analyse the sentence word by word, by assigning, e.g., a part-of-speech (POS) tag or a semantic tag to each word in the sentence.
This can be referred to as a \textbf{one-to-one} scenario, since every single unit in the text (e.g.\ each word token) has a direct correspondence to a single tag.\footnote{The term 'one-to-one' is also used for simple classification cases where there is no sentential context available. We see this as simply being a special case in which the sequence length $=1$. This is equivalent to the relation between FFNNs and RNNs, in which an FFNN can be seen as a special case of RNNs (or vice versa).}

We might also want to carry out some task in which the sentence should be translated to some other form, for instance translating the sentence to German, or some other language.
If the sentence was written in some non-standard form of English, we might want to produce a normalised version of the sentence, or in a different setting we might want to generate an inflected form of some word in the same language.
This can be referred to as a \textbf{many-to-many} scenario, as there is no structural one-to-one correspondence between the input $X$ and the output $Y$.

A final logically possible case, is the \textbf{one-to-many} scenario.
This is a highly uncommon scenario, as it is not generally the case that one tries to predict several things from an atomic unit.
Although one could argue that some tasks fit this scenario, such as caption generation, this is not really a one-to-many scenario, as the image is not an atomic unit, but is read by the NN as a matrix of pixels.

\begin{figure}[p!]
    \centering
    \includegraphics[width=0.55\textwidth]{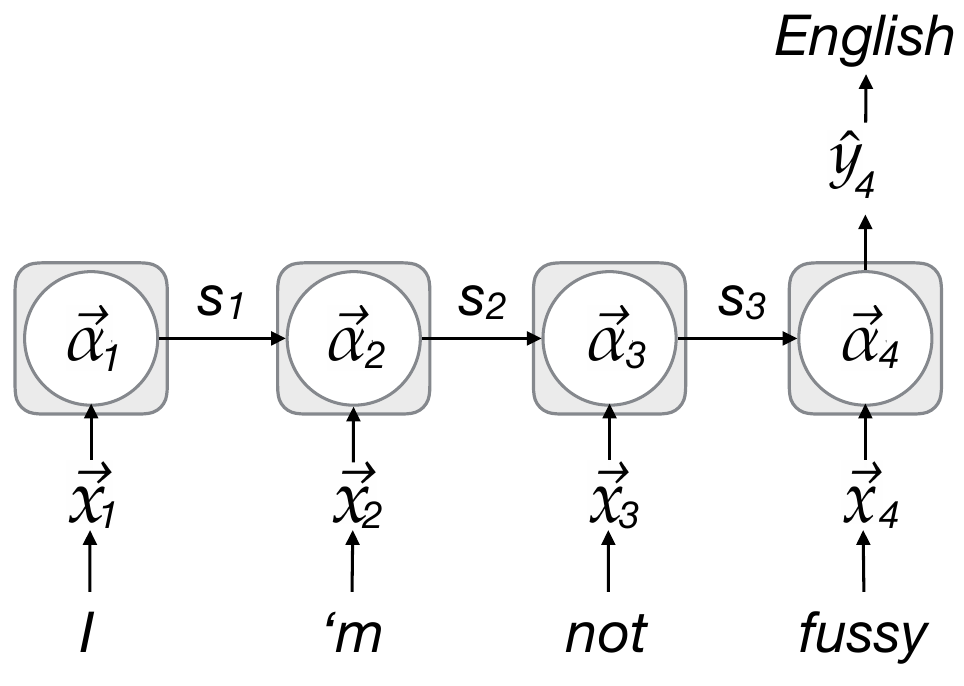}

    \vspace{0.7cm}
    \includegraphics[width=0.55\textwidth]{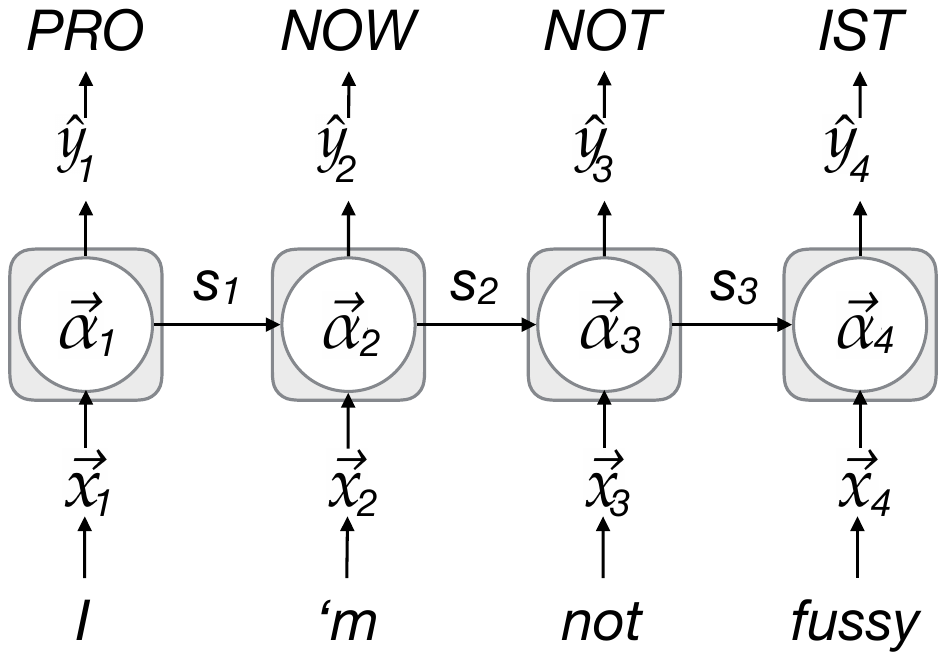}

    \vspace{0.7cm}
    \includegraphics[width=0.84\textwidth]{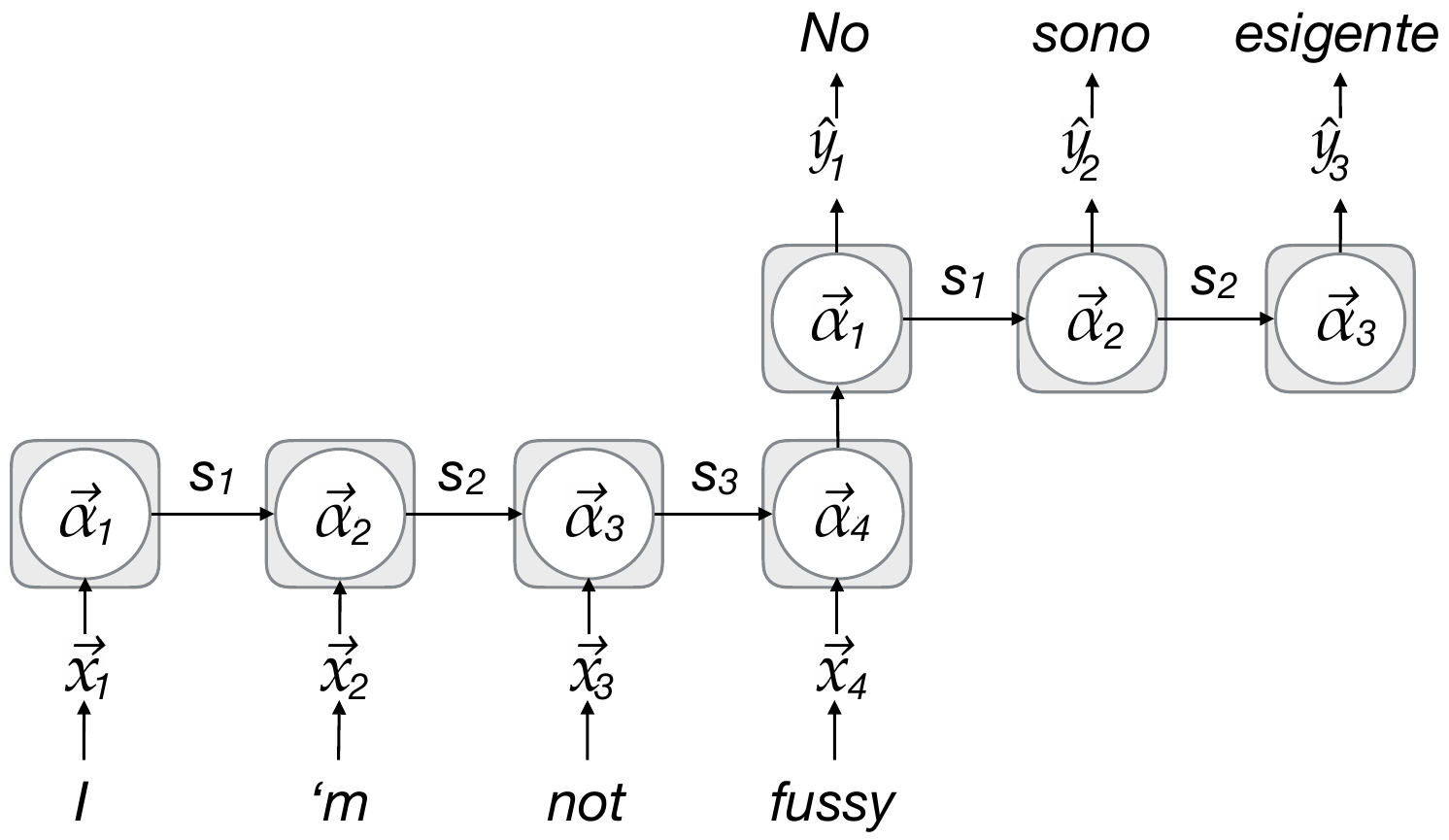}
    \caption{Common scenarios in which RNNs are applied in NLP. From top to bottom: many-to-one, one-to-one, and many-to-many.}
    \label{fig:nlp_scenarios}
\end{figure}

A schematic overview of the three relevant scenarios is given in Figure~\ref{fig:nlp_scenarios}.
The versatility of these three scenarios is evident when observing the current NLP scene, in which common practise is to cast a problem to fit one of these scenarios, and to then throw a Bi-LSTM at the problem.

\subsubsection*{Many-to-one}

Many NLP tasks deal with going from several smaller units to a single prediction.
This essentially means that these units need to be compressed into a single vector, onto which a softmax layer can be applied in order to arrive at a probability distribution over the classes at hand (e.g.\ a set of languages to identify).
For this type of problem, a number of possibilities exist, such as

\begin{enumerate}
    \item Averaging the vectors representing each unit in the sentence (i.e.\ average pooling);
    \item Applying an RNN and using the final state output vector as a representation of the sentence (depicted in Figure~\ref{fig:nlp_scenarios});
    \item Applying convolutions in order to arrive at a condensed representation.\footnote{Convolutional Neural Networks are described in Section~\ref{sec:cnn}.}
\end{enumerate}

Approach 1) is the most simplistic of these, and has been successfully applied in previous work (e.g.\ \citet{socher:2013:reasoning,zhang:2015:semrel}).
The main advantage of this approach is indeed its simplicity, as calculating the mean of the vectorial representations of a sentence is both a very cheap operation, as well as an operation which allows for the application of a simple FFNN on top of this vector.
With this approach, however, the structure inherent in the natural language signal is left unexploited.

Approaches 2) and 3) both offer more expressive power as compared to approach 1), as they both take advantage of the structure inherent in the input signal.
This is not the case when summing the vectors, which is in a way analogous to a bag-of-words approach.
Structure is naturally of paramount importance in natural language, so taking advantage of this is good.
Although structure in natural language is generally hierarchical, even using the sequential structure is better than assuming no structure at all.
There are, however, some recent architectures which do encode the hierarchical structure of language, such as tree LSTMs \citep{tree_lstm}, and RNN-Grammars \citep{rnng}.

These three approaches are meant to give an overview of some straight-forward manners of obtaining such representations. Other, more sophisticated approaches, include skip-thought vectors, in which sentence-level representations are learned with the objective of being able to predict surrounding sentences in a document \citep{skipthought}.
A systematic overview of such methods is given by \citet{hill:2016}, who conclude that the best suited approach depends on the intended application of such representations.

Another use case for this approach is when building hierarchical models, in the sense that one, e.g., may want to have word representations which are aware of what is going on on a sub-word level.
For this, one might apply a many-to-one RNN, and use the final state output vector as a word vector \citep{ballesteros:2015,plank:2016}.
Alternatively, one can use convolutions to arrive at this type of word vector, as in \citet{dossantos:14}.

\subsubsection*{One-to-one}
The one-to-one case is perhaps one of the most common scenarios in NLP.
This covers tagging task scenarios, as well as simple classification scenarios.
Many NLP tagging tasks have seen relatively large improvements when applying variants of RNNs similarly to what is depicted in Figure~\ref{fig:nlp_scenarios}.
Recently, gated variants such as LSTMs and GRUs, often in a bi-directional incarnation are applied \citep{wang2015:unified,huang:2015,yang:2016,plank:2016}.
Such RNNs are highly suited for this type of task, as it is highly informative for, e.g., POS tagging to know which words occur both before and after the word at hand.
Such dependencies might also have quite large spans, which both LSTMs and GRUs are able to capture well.
To contrast with older feature-based models, it would require a fair bit of feature engineering to decide on what types of spans to include in feature representations, lest one wishes to suffer from the sparsity of simply using $n$-gram features with large values of $n$.

\subsubsection*{Many-to-many}

This paradigm is also what is frequently referred to as an encoder-decoder architecture in the literature, or sequence-to-sequence learning problems \citep{bahdanau:2014,sutskever:2014}.
A frequent approach here, for instance in machine translation, is to apply an RNN from which one takes the final time step's output to be a representation of the entire sentence, as represented in Figure~\ref{fig:nlp_scenarios}, which may seem like a bold thing to do.\footnote{\textit{You can't cram the meaning of a whole sentence into a single vector!}\\--Ray Mooney, as communicated by Kyunghyun Cho in his NoDaLiDa 2017 keynote \mbox{(\url{https://play.gu.se/media/1_xt08m5je})}}
It turns out that this is in fact often not sufficient, although one can obtain surprisingly good translations this way.
However, results improve dramatically when going a step further by incorporating an attentional mechanisms.
This is not focussed upon in this thesis, and will not be explained in full detail.
Essentially, an attention mechanism can learn which parts of the source sentence to attend to, when producing the target sentence translation.
For instance, such a mechanism might learn an implicit weighted word-alignment between the source and target sentences, thus facilitating translation.

Many NLP tasks can be solved with a many-to-many approach.
Machine translation has already been mentioned, and has in no small degree been the driving force behind research in this direction.
Apart from this, the approach has been applied to
morphological inflection \citep{kann:2016,sigmorphon:2016,bjerva:2017:sigmorphon,sigmorphon:2017},
AMR parsing \citep{riga:semeval16,konstas:2017,rik:amr:semeval,rik:amr:clin},
language modelling (e.g.~\citeauthor{vinlays:2015}, \citeyear{vinlays:2015}),
generation of Chinese poetry \citep{yi:2016},
historical text normalisation \citep{korchagina:2017},
and a whole host of other tasks.

\section{Convolutional Neural Networks}
\label{sec:cnn}
Certain machine learning problems, such as image recognition, deal with input data in which spatial relationships are of utmost importance.
While simpler image recognition problems, such as handwritten digit recognition, can be carried out relatively successfully with simple FFNNs, this is often not sufficient.
Recall that the input for an FFNN is simply a single vector $\vec{x}$, meaning that the network has no notion of adjacency between, e.g., two pixels.
A Convolutional Neural Network (CNN) is a type of network explicitly designed to take advantage of the spatial structure of its input.
The origins of CNNs go back to the 1970s, but the seminal paper for modern CNNs is considered to be \citet{convnets:lecun}, although other work exists in the same direction (e.g., \citet{lecun:1989,convnets:waibel}).
CNNs have been used extensively in NLP, and can in many cases be used instead of an RNN (contrast, e.g., \citet{dossantos:14} who use CNNs for character-based word representations, and \citet{plank:2016} who use RNNs for the same purpose).

Although NLP is the focus of this thesis, we will approach CNNs from an image recognition perspective, as this is somewhat more intuitive.
This is in part due to the fact that image recognition was the intended application of CNNs upon their conception.
On a general level, convolutions can be carried out on input of arbitrary dimensionality.
As mentioned, two-dimensional input (e.g.\ images) were the original target for CNNs.
More recent work has extended this to three-dimensional input (e.g.\ videos).
In the case of NLP, it is often the case that one-dimensional input is used, for instance applying a CNN to a text string.
There are three basic notions which CNNs rely upon: local receptive fields, weight sharing, and pooling.

\subsection{Local receptive fields}
In the case of image recognition, an image of $n\times n$ pixels can be coded as an input layer of $n\times n$ units.\footnote{This is assuming greyscale, i.e., one value per pixel.}
In a CNN, this input is processed by sliding a window of size $m\times m$ across this image.
This window, or patch, is known as a local receptive field.
After passing this window over the input image, the following layer contains a representation based on $m\times m$ sized slices of the input image.
Intuitively, this can be seen as blurring the input image somewhat, as the spatial dimensions of the image are generally reduced through this process.

\begin{figure}[htbp]
    \includegraphics[width=\textwidth]{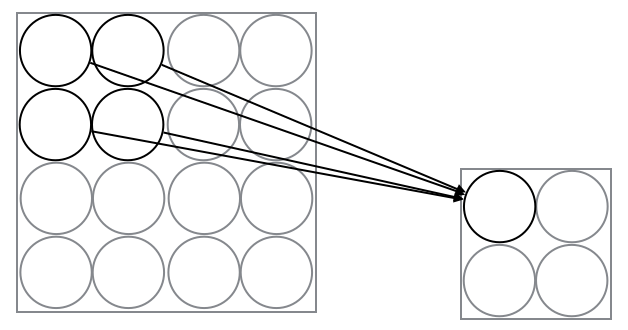}
    \caption{Illustration of a local receptive field of size $2x2$, which results in a new image of size $2x2$ due to the stride length being $2$.
    \label{fig:cnn_local}}
    \includegraphics[width=\textwidth]{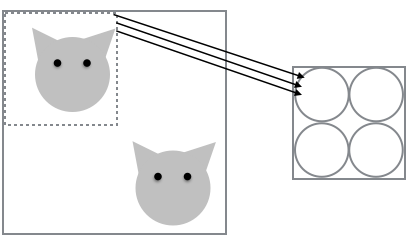}
    \caption{Weight sharing example with a cat. The dotted line represents the local receptive field used, the first square represents the entire input domain, and the second square represents the following convolutional layer.
    \label{fig:cnn_cat}}
\end{figure}

The length with which this window moves is referred to as its \textit{stride}, and is most often set to $1$, meaning that the window simply shifts by one pixel at a time.
Although stride lengths of 2 and 3 are encountered in the literature, it is fairly uncommon to see larger stride lengths than this.
Figure~\ref{fig:cnn_local} shows a convolution, where $n=4$, $m=2$, a stride of $2$ is used, which yields a new layer with size $2\times2$.

\subsection{Weight sharing}
A key notion of CNNs is the fact that weights are shared between each such local receptive field, and the units to which they are attached.
Hence, in our example, rather than having to learn $n\times i = 64$ weights (where $i=n\times n$ is the total number of units in the first hidden layer), as in an FFNN, only $m\times m = 4$ parameters need to be learned.
Therefore, all units in the first hidden layer capture the same type of features from the input image in various locations.
For instance, imagine you want to identify whether a picture contains cats (as in Figure~\ref{fig:cnn_cat}).
In this figure, units in the first hidden layer might encode some sort of \textit{cat detector}.
The fact that weights are shared in this manner, results in CNNs being robust to translation invariance.
This means that a feature is free to occur in different regions in the input image.
Intuitively, taking our cat detector as an example, it is naturally the case that a cat is a cat, regardless of where in the image it happens to hide.

Each such \textit{cat detector}, is referred to as a \textit{feature map}, or a \textit{channel}.\footnote{The \textit{channel} terminology makes sense when considering an input image in, e.g., RGB formatting, in which the intensities of each colour is represented in a separate colour channel.}
For a CNN to be useful, normally more than one feature map is learnt.
That is to say, while the figures shown so far only show a single feature map, an input image is normally mapped to several smaller images.
As an example, another feature map in Figure~\ref{fig:cnn_cat} might learn a \textit{dog detector}.
A more realistic example can be found in facial recognition, in which one feature map might learn to detect eyes, while another learns to detect ears, and yet another learns to detect mouths.
Local receptive fields generally speaking look at all feature maps of the previous layers, hence the combination of eyes, ears, and mouths might be used to learn a feature map representing an entire face.

\begin{figure}[htbp]
    \centering
    \includegraphics[width=0.7\textwidth]{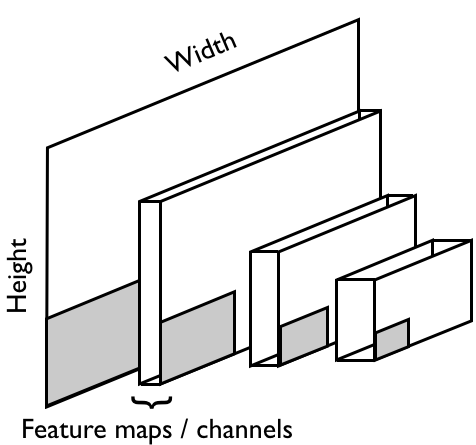}
    \caption{General CNN structure. Each layer shrinks the width and height of the input image, and increases the number of feature maps. The grey regions denote the sizes of the local receptive fields.}
    \label{fig:cnn_structure}
\end{figure}

The fact that we generally map to several feature maps means that, at each layer, the spatial size of the image shrinks (i.e.\ $m<n$), while the depth of the image increases, as shown in Figure~\ref{fig:cnn_structure}.
Finally, following a series of convolutional layers, it is common practise to attach an FFNN prior to outputting predictions.

In NLP, the situation is somewhat different, as we normally do not have an image as input, but rather some sort of textual representation.
Commonly, this will either be a string of words, characters, or bytes. 
An intuition for how this works, is that something resembling an n-gram feature detector is learnt given a window.
This type of approach has been applied successfully in various tasks, for instance to obtain word-level representations which take advantage of sub-word information \citep{dossantos:14,bjerva:2016:semantic}, and for sentence classification \citep{kim:2014}.


\subsection{Pooling}
A pooling layer takes a feature map and condenses this into a smaller feature map.
Each unit in a pooling layer summarises a region in the previous layer, generally using a simple arithmetic operation.
Frequently, operations like maximum pooling (max pooling) or average pooling are used \citep{maxpool}.
These operations take, e.g., the maximum of some region to be a representation of that entire region, thus reducing dimensionality by essentially applying simple non-linear downsampling.
Max pooling can thus be seen as a way for each max pooling unit to encode whether or not a feature from the previous layer was found anywhere in the region which the unit covers.
The intuition is that the downsampled version of the feature map, which yields feature locations which are rough, rather than precise, is sufficient in combination with the relative location to other such downsampled features.
Importantly, this operation reduces the dimensionality of feature maps, thus reducing the number of parameters needed in later layers.

\section{Residual Networks}
\label{sec:resnet}
\label{sec:resnets}

\begin{figure}
    \centering
    \includegraphics[width=0.9\textwidth]{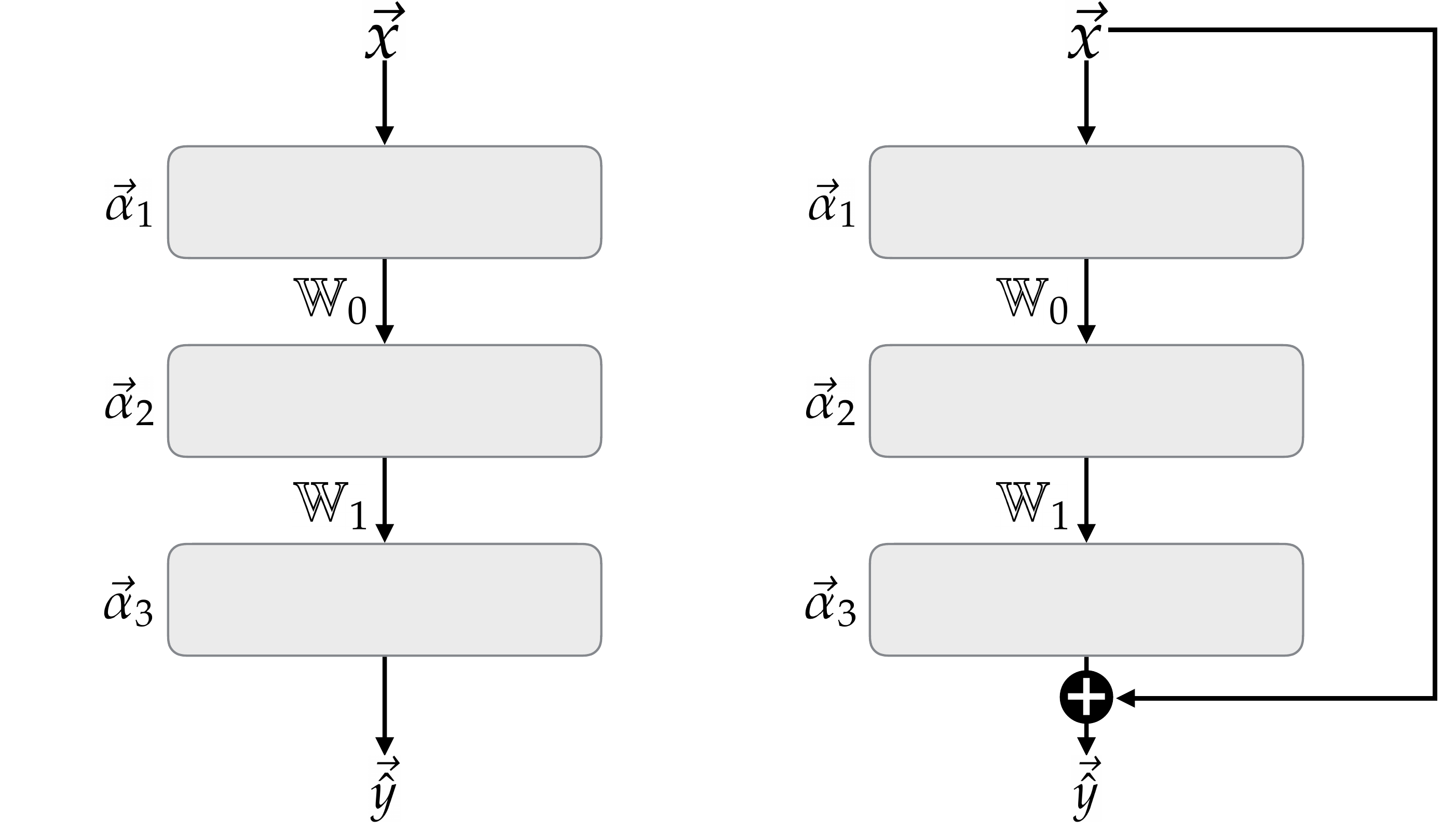}
    \caption{Illustration of a residual network (right) as compared to a standard network without skip connections (left). The skip connection here passes the input vector $\vec{x}$ and adds this to the activations $\vec{\alpha}_3$, thus providing the network with a shortcut. The squares represent an abstract block of weights, such as, e.g., a fully connected layer in an FFNN, a convolutional block in a CNN, or an LSTM cell.}
    \label{fig:resnet}
\end{figure}

Residual Networks (ResNets) define a special class of networks with skip connections between layers, as depicted in Figure~\ref{fig:resnet}.
This facilitates the training of deeper networks, as these skip connections ease the propagation of errors back to earlier layers in the network.

Such skip connections are referred to as residual connections, and can be expressed as
\begin{equation}
    \begin{aligned}
    &y_l = h(x_l) + \mathcal{F}(x_l,\mathcal{W}_l),\\
    &x_{l+1} = f(y_l),
    \end{aligned}
\end{equation}
\noindent where $x_l$ and $x_{l+1}$ are the input and output of the $l$-th layer, $\mathcal{W}_l$ is the weights for the $l$-th layer, and $\mathcal{F}$ is a residual function \citep{resnets:2016} such as the identity function \citep{resnets:2015}, which we also use in our experiments.
Although ResNets were developed for the use in CNNs, the skip connections are currently being used, e.g., in LSTMs \citep{google:nmt}.
ResNets can be intuitively understood by thinking of residual functions as paths through which information can propagate easily.
This means that, in every layer, a ResNet learns more complex feature combinations, which it combines with the shallower representation from the previous layer.
This architecture allows for the construction of much deeper networks.
ResNets have recently been found to yield impressive performance in image recognition tasks, with networks as deep as 1001 layers \citep{resnets:2015,resnets:2016}, and are thus an interesting and effective alternative to simply stacking layers.
Another useful feature of ResNets is that they act as ensembles of relatively shallow networks, which may help to explain why they are relatively robust to overfitting in spite of their large number of parameters \citep{veit:resnet}.

ResNets have recently been applied in NLP to morphological re-inflection \citep{robert:sigmorphon:2016}, language identification \citep{bjerva:2016:dsl}, sentiment analysis and text categorisation \citep{conneau:2016}, semantic tagging \citep{bjerva:2016:semantic}, as well as machine translation \citep{google:nmt}.
Recently proposed variants include wide residual networks, with relatively wide convolutional blocks, showing that resnets do not necessarily need to be as deep as the 1001 layers used in previous work \citep{resnets:wide}.

\section{Neural Networks and the Human Brain}
\label{sec:brain}
While there are numerous reasons to use neural networks, there are camps which might argue for applying NNs because they are \textit{'biologically motivated'}.
While it can be tempting to use the conceptual metaphor \citep{lakoff:metaphors} of \textsc{neural networks are the brain}, this can be rather misleading.
While usage of misleading metaphors can seem innocent, it has been shown that they do affect reasoning \citep{metaphors:reasoning}.
Therefore, a misleading metaphor should certainly not be used as an argument for the usage of neural networks -- there are plenty of other reasons for that.

Now, you might ask, is this metaphor really as misleading as it seems?
At best, neural networks (as used in NLP) are a mere caricature of the human brain.
On a physiological level, there is evidence that neurons do more than simply outputting some activation value based on a weighting of its inputs.
For instance, recent research has shown that a single neuron can encode temporal response patterns, without relying on temporal information in input signals.
Hence, the nature of how neurons work is quite different from what is encoded in a neural network, for instance in terms of information storage capacity \citep{jirenhed:2017}.
There is in fact compelling evidence that memory is not coded in (sets of) synapses, but rather internally in neurons (cf.~\citeauthor{gallistel:memory}, \citeyear{gallistel:memory}, and \citeauthor{gallistel:festschrift}, \citeyear{gallistel:festschrift} and references therein, notably \citet{purkinjecell}).
Additionally, backpropagation is not biologically plausible, although there is work on making biologically plausible neural networks \citep{bengio:2015}.

\subsection*{Convolutional Neural Networks and the Brain}
Similarly to the ReLUs discussed in Section~\ref{sec:act_func}, CNNs are biologically inspired.
When CNNs were invented \citep{convnets:lecun}, this was inspired by work which proposed an explanation for how the world is visually perceived by mammals \citep{hubelwiesel:1959,hubelwiesel:1962,hubelwiesel:1968}.
The attempts to reverse-engineer a similar mechanism, as in CNNs, have proven fruitful, as CNNs are indeed highly suitable for image recognition.
Furthermore, recent research has found some correlations between the representations used by a CNN, and those encoded in the brain in a study where the CNN could identify which image a human participant was looking at roughly 20\% of the time \citep{seeliger:2017}.
However, it remains to be seen whether anything similar to this is even plausible for natural language.


\section{Summary}

In this chapter, an intuitive and theoretically supported overview of neural networks was given, including a practical overview NLP.
We have seen that many common NLP problems can be classified into three categories: one-to-one, one-to-many, and many-to-many.
Appropriate deep learning architectures suitable for each of these categories were suggested.

While this chapter is meant to be a sufficient introduction to neural networks to understand this thesis, it is by no means a complete account of the topic.
For a more in-depth description of neural networks in general, I refer the readers to \citet{BengioBook}.
For a primer which is more geared towards NLP, see \citet{goldberg:primer}.

\renewcommand*{\thefootnote}{\fnsymbol{footnote}}
\chapter[Multitask Learning and Multilingual Learning]{
Multitask Learning\\ and Multilingual Learning\hspace{-8pt}}
\renewcommand*{\thefootnote}{\arabic{footnote}}
\label{chp:mtl_bg}

\begin{abstract}
\absprelude In this chapter, we build upon the background knowledge of neural networks presented in the previous chapter.
We will focus on different, but highly related, paradigms -- multitask learning, and multilingual model transfer.
Multitask learning is first presented in a general context, and then in the context of neural networks, which is the primary focus of this thesis.
We will then look at multilingual approaches in NLP, again first in a general context, and then in the context of model transfer with multilingual word representations, which is the secondary focus of this thesis.
In this thesis, we consider the first setting in Part II, the second setting in Part III, and include an outlook for a combined multilingual/multitask paradigm in Part IV.
\end{abstract}

\clearpage

\section{Multitask Learning}
\label{sec:mtl}

In Natural Language Processing (NLP), and machine learning (ML) in general, the focus is generally on solving a single task at a time.
For instance, one might invest significant amounts of time in making a Part-of-Speech tagger or a parser.
However, fact is that many tasks are related to one another.
The aim of multitask learning (MTL) is to take advantage of this fact, by attempting to solve several tasks simultaneously, while taking advantage of the overlapping information in the training signals of related tasks \citep{caruana:1993,mtl}.
When the tasks are related to each other, this approach can improve generalisation, partially since it provides a source of inductive bias, and since it allows for leveraging larger amounts of more diverse data.\footnote{Generally speaking, it is beneficial to have access to more data when training an ML model.}
Additionally, since related tasks can often make use of similar representations, this can lead to the tasks being learnt even better than when training on a single task in isolation.

The use of MTL is skyrocketing in NLP, and has been applied successfully to a wide range of tasks, for instance sequence labelling such as POS tagging \citep{collobert:2008,plank:2016}, semantic tagging \citep{bjerva:2016:semantic}, as well as chunking and supertagging \citep{sogaard2016deep}.
In addition to this, it is the primary focus of this thesis, and having some background knowledge on this will be useful for the following chapters.
The first part of this chapter is an attempt at providing an understanding of what MTL is and how it is applied.
While some general MTL scenarios are covered, the focus will be on MTL in the context of neural networks, and in the context of NLP.

\subsection{Non-neural Multitask Learning}

Before going into MTL in neural networks (NNs), we first take a look at the usage of this paradigm in other frameworks.
Generally speaking, we seek to exploit the fact that there are many tasks which are somehow related to one another \citep{caruana:1993,mtl,mtl:thrun}.
For instance, MTL can have the role of being a distant supervision signal, in the sense that the tasks used might be fairly distantly related.
Additionally, since MTL plays the role as a regulariser (see Chapter~\ref{chp:nn}), and lowers the risk of overfitting \citep{baxter:1997,baxter:2000}, MTL often improves generalisation.
This is in part because MTL reduces \textit{Rademacher complexity} \citep{baxter:2000,maurer:2006}.\footnote{A lower Rademacher complexity essentially indicates that a class of functions is easier to learn.}
Furthermore, MTL will push the weights of a model towards representations which are useful for more than one task.
Finally, MTL can be seen as a method of dataset augmentation, as it allows for using more data than when only considering a single task at a time.

A commonly made assumption in MTL is that only a handful of parameters or weights (see Chapter~\ref{chp:nn}) ought to be shared between tasks, and conversely that most parameters should not be shared \citep{argyriou:2007}.
This can intuitively be understood by considering that only a few features useful for a task $t_1$ might be useful for another task $t_2$.
For instance, imagine that we are building a joint POS tagger and language identification system.
A feature capturing capitalised words preceded by a determiner will both be a decent indicator of the language being, e.g., German, as well as that the capitalised word is a noun.
Other features, on the other hand, such as one indicating that the language is likely to be Norwegian or Danish if the letter \o \ is encountered, is not likely to be beneficial for POS tagging at all.
In other words, this type of \textit{parameter sparsity} can be phrased as that most parameters should not be shared, as many parameters are task specific.
In this type of approach, all shared parameters are generally considered by all tasks involved.
This puts the system at a relatively large risk of negative transfer, if one tries to combine this approach with tasks which are only slightly related.
In NLP we are often interested in exploiting even relatively weak training signals, which makes this particularly problematic.

Another approach is to learn clusters of tasks, which allows for letting related tasks share certain parameters, and relatively unrelated tasks perhaps only a few.
Such approaches have in common that they assume that the parameters which are beneficial for each other are geometrically close to one another in $n$-dimensional space \citep{evgeniou:2004,kang:2011}.
Other work has come up with other definitions of task similarities.
For instance, \citet{thrun:1995} consider two tasks to be similar simply if one improves performance on the other.
While other approaches to MTL have been used in the past, such as \citet{daume:2009} who approach MTL from a Bayesian perspective, and \citet{toutanova:2005} who train a joint classifier for semantic role labelling with automatically generated auxiliary tasks, the perhaps most popular approach in NLP is parameter sharing in NNs.

\subsection{Neural Multitask Learning}

We now turn to the main method used in this thesis, namely neural MTL.
There are two main approaches to this, differing in the manner in which parameters are shared -- \textit{hard} and \textit{soft} parameter sharing.
Currently, the less popular variant of the two in NLP is soft parameter sharing, and will not be covered in detail.
Briefly put, in this setting, parameters are constrained in a similar manner to the \textit{parameter sparsity} approach.
That is to say, the parameters between tasks are encouraged to be similar to one another, which allows for some transfer between tasks, or between languages \citep{duong:2015}.
However, as parameters are not explicitly shared between tasks, the risks of negative transfer are relatively low in this setting.
This approach is not explored in this thesis as hard parameter sharing offers several advantages, including ease of implementation, and computational effectivity, as the amount of parameters is kept almost constant as compared to having a single task.

Hard parameter sharing is currently more common, perhaps mainly due to the ease with which a neural MTL system with several tasks can be created.
This is the type of MTL discussed in the seminal works by \citet{caruana:1993,mtl}.
In this thesis we consider research questions tied to this type of MTL in the context of NLP, partially due to the versatility of the paradigm.
Apart from allowing for considering data from several tasks simultaneously, even corpora in different languages might be used in this approach, given some sort of unified input representations.\footnote{This is covered further in the second half of this chapter.}
Then, if the output labels between tasks correlate with one another to some extent, it seems quite intuitive that this approach should be beneficial.

In NLP, MTL is generally approached from the perspective that there is some \textit{main} task, i.e., the task in which we are interested, and some \textit{auxiliary} task, which should improve the main task.
It is important to note, however, that these labels are quite arbitrary.\footnote{The exception being cases in which the performance on the auxiliary task is disregarded in favour of the main task performance.}
There is not necessarily anything to distinguish a main task from an auxiliary task in an NN.
One might lower the weighting of the auxiliary task (i.e.\ multiply the loss for each batch by some $\lambda<1$), but this strategy appears to be relatively rare in the literature.

\begin{figure}[htbp]
    \centering
    \includegraphics[width=0.8\textwidth]{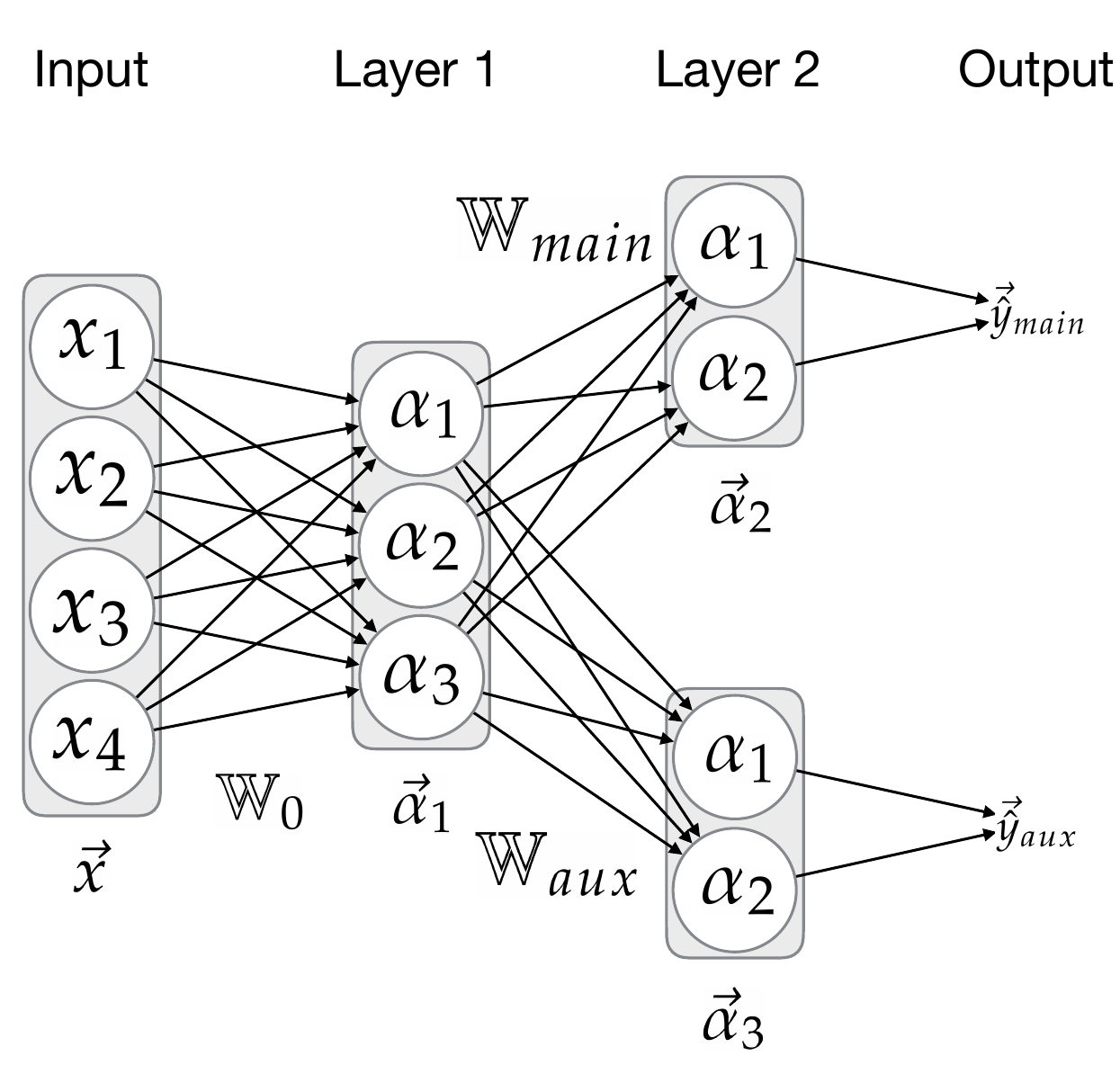}
    \caption{Common MTL architecture (bias units omitted for clarity).}
    \label{fig:mtl_nn}
\end{figure}

A common way of implementing hard parameter sharing, is to have a stack of layers for which weights are updated with respect to all tasks, with at least two output layers, each with task-specific weights (see Figure~\ref{fig:mtl_nn}).
Concretely, consider that we have $t$ corpora with different annotations, each containing pairs of input and output sequences ($\vec{x},\vec{y}^t$) for a single task.
While the inputs $x$ will largely be part of the same vocabulary, and can be shared across tasks, the tag sets used, and therefore the labels ($y^0\ldots y^t$) differ.
Note that the vocabularies in the tasks at hand do not necessarily need to overlap, but when considering a single natural language, this tends to be the case.
A common approach when training is to randomly sample such sequence pairs, predict a label distribution $\vec{\hat{y}}^t$, and update model parameters as calculated by the loss relative to the true label distribution $\vec{y}^t$ with backpropagation (see Chapter~\ref{chp:nn} for an overview of this).
Each task $t$ has a task-specific classifier ($f_{t=main}, f_{t=aux}$) with its own weight matrix ($\mathbb{W}_{main}, \mathbb{W}_{aux}$).
The output of the task-specific layer is then calculated using the softmax function (cf.\ Section~\ref{sec:act_func}), such that

\begin{equation}
    \begin{aligned}
    \vec{\hat{y}}_{t} &= \mbox{softmax}(\mathbb{W}_{t}\vec{\alpha}_n+b),
    \end{aligned}
\end{equation}
where $\vec{\alpha}_n$ denotes the activations of the layer before the output layer.

This architecture is common in NLP, with weights typically shared between the \textit{main} and \textit{auxiliary} task at all layers, up to the task-specific classification layer (i.e.\ the output layer).
A multitude of other possibilities do exist, as the task-specific output layers can be attached anywhere in the network.
This can be advantageous, as \citet{sogaard2016deep} found that including the lower-level task supervision at lower levels in the network was useful, in the case of using the low-level task of POS tagging in combination with CCG supertagging (i.e.\ assigning CCG lexical categories).
Most related work, including the experiments in this thesis, apply multitask learning akin to what is shown in Figure~\ref{fig:mtl_nn}.

\subsubsection*{Neural Multitask Learning in Natural Language Processing}
Hard parameter sharing in NNs is the target of considerable attention in the recent NLP literature.
Practically speaking, there appear to be two main approaches to MTL in the NLP literature.
Some work, such as \citet{ando:2005}, \citet{collobert:2008}, \citet{sogaard2016deep}, \citet{plank:2016}, \citet{bjerva:2016:semantic}, and \citet{augenstein:2017} take the approach of exploiting seemingly related NLP tasks, based on some linguistic annotation.
Other work, e.g.,~\citet{plank:keystroke}, and \citet{klerke:2016}, take the approach of exploiting data from non-linguistic sources (keystroke data and eye gaze data, respectively).
While these approaches are both useful and interesting, the focus of this thesis is the first approach, specifically in which an NLP sequence prediction task is used as an auxiliary task for some other NLP sequence prediction task.
This is partially motivated by the fact that using a word-level input for all tasks, allows for a one-to-one mapping between labels in different tag sets (given a specific token in context), which in turn opens up for the information-theoretic approach considered in Chapter~\ref{chp:mtl}.

\subsection{Effectivity of Multitask Learning}
Plenty of studies demonstrate the success of MTL, such as in computer vision \citep{torralba:2007,loeff:2008,quattoni:2008}, genomics \citep{obozinski:2010}, and the aforementioned NLP studies.
Apart from relatively straight-forward results showing that MTL is often beneficial, efforts have been put into experimentally investigating when and why MTL is advantageous in NLP.
\citet{alonso:mtl} look at a collection of semantic main tasks while using morphosyntactic and frequency-based tasks as auxiliary tasks.
They find that the success of an auxiliary tasks depends on the distribution of the auxiliary task labels, e.g., the distribution's entropy and kurtosis.\footnote{The kurtosis of a distribution is essentially a measure of its tailedness.}
\citet{bingel:2017} present a large systematic study of MTL in a collection of NLP tasks.
They find that certain dataset characteristics are predictors of auxiliary task effectivity, corroborating the findings of \citet{alonso:mtl}, and also show that MTL can help target tasks out of local minima in the optimisation process.
In \citet{bjerva:2017:mtl}, it is argued that entropy is not sufficient for explaining auxiliary task effectivity, and that measures which take the joint distribution between tasks into account offer more explanatory value (this is elaborated in Chapter~\ref{chp:mtl}).

In terms of data sizes \citet{benton:2017} suggest that MTL is effective given limited training data for the main task.
\citet{mtl:seq2seq}, however, highlight that the auxiliary task data should not outsize the main task data -- this is contradicted by \citet{augenstein:2017}, who highlight the usefulness of an auxiliary task when abundant data is available for such a task, and little for the main task.
Finally, \citet{mou:2016} investigate transferability of neural network parameters, by attempting to initialise a network for a main task with weights which are pre-trained on an auxiliary task, and highlight the importance of similarities between tasks in such a setting.
Finally, a promising recent innovation is that of sluice networks, in which a NN learns which parts of hidden layers to share between tasks, and to what extent \citep{sluice}.

\subsection{When MTL fails}
The cases in which MTL does not work are also deserving of attention.
While up until now we have assumed that applying MTL is a piece of cake, there are times when one adds an auxiliary task, causing the system to collapse like a house of cards.
This type of performance loss is referred to as \textit{negative transfer}, and can occur when two unrelated tasks share parameters.
This is generally something to avoid, as there are few, if any, advantages to worsening the generalisation ability of the network.
However, such results are rarely shared in the community, in part due to the \textit{file drawer problem} \citep{filedrawer}.
In short, the problem is that it is impossible to access, or even know of, studies which have been conducted and not published.
In the case of MTL, this issue might be alleviated by publishing results on all auxiliary tasks experimented with, even if only one or two such tasks improved performance.



\section{Multilingual Learning}

In the second half of this chapter, we turn to multilingual approaches.
Many languages are similar to each other in some respect, and similarly to related tasks, this fact can also be exploited in order to improve model performance with respect to, e.g., a given language.
While there are many approaches to multilingual NLP, with various use cases, the focus in this thesis is on model transfer, in which a single model is shared between languages.
We will nonetheless begin with an overview of the most common approaches.

As an example, consider NLP tagging tasks, which can be summed up as learning to assign a sequence of tags from a tag set $t$ to a sequence of tokens in language $l$.
In cross-lingual multitask NLP settings, there are many $l$/$t$ pairs which do not have any annotated data.
For instance, there is (at the time of writing) no annotated data for Welsh in the Universal Dependencies \citep{ud20}.
However, many NLP systems require input data from specific tag sets.
For instance, the Stanford Neural Network Dependency parser requires POS tags in its input \citep{chen:2014}, whereas the semantic parser Boxer requires semantic tags in its input \citep{step:boxer,pmb}.
Hence, for such tools to be applicable in multilingual settings, the tags they rely on need to be available for other languages as well, which highlights the importance of approaches which deal with this.
There are three frequently used approaches to solving this problem:
\begin{enumerate}[itemsep=-1ex]
    \item human annotation;
    \item annotation projection;
    \item model transfer.
\end{enumerate}

\noindent Although serious efforts have gone into furthering these approaches, they all have considerable drawbacks.
In brief, human annotation is time consuming and expensive, annotation projection is only applicable to texts which are both translated and aligned, and model transfer is generally only used in mono-lingual or mono-task settings.

\subsection{Human Annotation}
Generally speaking, annotating data manually is a very expensive and time-consuming manner of, e.g., producing some sort of linguistic labels for a sentence.
Although the process can be alleviated with gamification \citep{wordrobe,chamberlain:2014,jurgens:2014,Bos:2015}, considerable time and effort still needs to be invested into creating such crowd-sourcing systems.

\subsection{Annotation Projection}
\label{sec:annotation_proj}
Given an annotated sentence in a source language and a translation of that sentence in a target language, it is possible to transfer, or project, the annotation from the source language to the target language.
This approach is known as annotation projection, and relies on having access to parallel text for which at least one source language is annotated \citep{yarowsky:2001,hwa:2005}.
Usually, word alignments are used in order to project linguistic labels from source to target.
The resulting annotations can then be used to train a new monolingual system for the target language(s).
This approach has been applied successfully to various tasks, primarily syntactic parsing \citep{hwa:2005,tiedemann:2014,rasooli:15,agic:2016}, POS tagging \citep{yarowsky:2001}, and recently also semantic parsing \citep{evang:2016,evang:phd}.

Annotation projection has two main drawbacks.
Primarily, it is only applicable to texts which are both translated and aligned, whereas the majority of available texts are monolingual.
Furthermore, this approach relies heavily on the quality of the automatic word alignments.
Word-aligning parallel text is not always successful, for instance with very dissimilar languages, insufficient statistics, or bad translations \citep{ostling:2014,ostling:2015}.
Another approach for annotation projection relies on automatic translation.
This works by applying a machine translation (MT) system to generate a parallel text for which source language annotation exists \citep{tiedemannetal:2014}.
In other words, in addition to the difficulties of the annotation projection approach, this method places high requirements on availability of parallel texts for training an MT model.
In addition to these prerequisites, the involvement of a fully-fledged MT system in an annotation pipeline, will in itself increase its complexity severely.

\subsection{Model Transfer}

Model transfer deals with learning a single model which is shared between several languages \citep{zeman:08,mcdonald:11}.\footnote{Note the similarities to multitask learning with hard parameter sharing.}
This type of approach has been explored extensively in previous work.
Multilingual model transfer has been successfully applied to, e.g., POS tagging \citep{tackstrom:2013}, and syntactic parsing \citep{tackstrom:phd,ammar:2016}.
This is commonly done by using delexicalised input representations, as in the case of parsing \citep{zeman:08,mcdonald:11,tackstrom:2012,tackstrom:2013}.
A related situation, is the case of exploiting language similarities in order to train models for low-resource languages (see, e.g., \citet{georgi:2010}).

In this thesis, model transfer is framed as a special case of MTL.
That is to say, each language in the model can be seen analogously to a task.
This means that we are also free to choose whether we want to code tag predictions jointly as a single output layer, or have one separate output layer per language.
As with MTL with multiple tasks, we consider the same specific type of MTL across languages, namely hard parameter sharing in neural networks.

A common approach in parsing is to delexicalise the input representations in order to enforce uniformity across languages, by training a parser on sequences of PoS tags rather than sequences of words \citep{zeman:08,mcdonald:11}.
However, as we are looking at predicting such tags, we approach this by using input representations which are shared across languages.
This allows for training a neural network for several languages simultaneously in a language-agnostic manner, while still taking lexical semantics into account.
Apart from this advantage, implementing a system in this manner is straightforward.
Additionally, this approach offers the possibility of out-of-the-box zero-shot learning, as simply adding input representations for a different language is sufficient to enable this.

\textit{Zero-shot learning} is the problem of learning to predict labels $y$ which have not been seen during training.
This is especially relevant in cases such as MT, in which, e.g., many of the target forms which need to be produced for languages with rich morphology have not been seen.
In recent years, zero-shot learning has become increasingly popular, for instance in image recognition \citep{palatucci:2009,socher:2013}.
Recently, it has also been applied to MT, resulting in a model which even allows for translation into unseen languages \citep{google:zeroshot}.
One way of enabling zero-shot learning, is to use shared input representations.
For instance, in the case of character-based models, we can simply use the same alphabet in the inputs and outputs of each system, as in \citet{bjerva:2017:sigmorphon} and \citet{bjerva:2017:multilingual}.
In the case of word level input representations, one can employ word embeddings living in the same space, regardless of language.

\subsection{Model Transfer with Multilingual Input Representations}

Looking further at the problem of model transfer across languages, consider the following example, of an English sentence and its translation, as two separate input sequences to a neural network, with their corresponding annotated output sequences.\footnote{PMB 01/3421. Original source: Tatoeba.}

\begin{examples}
    \item{\gll We must draw attention to the distribution of this form in those dialects .
\texttt{PRON} \texttt{VERB} \texttt{VERB} \texttt{NOUN} \texttt{ADP} \texttt{DET} \texttt{NOUN} \texttt{ADP} \texttt{DET} \texttt{NOUN} \texttt{ADP} \texttt{DET} \texttt{NOUN} \texttt{PUNCT}
\glt\glend}
    \item{\gll Wir müssen die Verbreitung dieser Form in diesen Dialekten beachten .
\texttt{PRON} \texttt{VERB} \texttt{DET} \texttt{NOUN} \texttt{DET} \texttt{NOUN} \texttt{ADP} \texttt{DET} \texttt{NOUN} \texttt{VERB} \texttt{PUNCT}
    \glt\glend}
\end{examples}

\noindent Although the surface forms of these two sentences differ, as one is in English and one in German, multilingual word representations for the corresponding words in these two sentences ought to be close to one another.
Hence, if the NN only sees the English sentence in training, and the German sentence during test time, it ought to be fairly successful in tagging this 'unseen' sentence with suitable tags.\footnote{Considering that the semantic content of the two sentences ought to be highly similar, one could regard the translated sentence to be 'seen' if the original sentence was in the training data. This has the further implication that one, in this type of experiments, must take care not to allow corresponding sentences to occur in both training and evaluation data.}
However, one question is whether having access to the same sentence in a typologically more distance language such as Japanese also would be useful (this is approached in Chapter~\ref{chp:mtl}).

In order for such an approach to work, it is necessary that words with similar meanings in different languages are represented in a fairly similar way.
How do we arrive at word representations with such properties?
In the next few sections we will look at this, beginning at simple monolingual representations, and leading up to bilingual and multilingual representations. 


\subsection{Continuous Space Word Representations}
\label{sec:distrep}

In many NLP problems, we are concerned with processing some word-like unit, in order to arrive at some linguistically motivated and appropriate label.
In Section~\ref{sec:feature_rep}, we considered bag-of-words models for tasks such as this.
To recap, in this type of model we assign an index to each unique word.
Each word is then represented by a vector $\vec{x}$, with a dimensionality equal to the size of the vocabulary, since each word requires its own index.
As an example, consider a vocabulary size of five words, with three of those words being \textit{cat}, \textit{dog}, and \textit{coastal}, with their corresponding vector representations, such that
\begin{equation}
    \begin{aligned}
    \vec{x}_{cat} &= [0, 0, 1, 0, 0],\\
    \vec{x}_{dog} &= [0, 0, 0, 0, 1],\\
    \vec{x}_{coastal} &= [0, 1, 0, 0, 0].\\
    \end{aligned}
\end{equation}
In NLP, we are often interested in comparing words with one another, either simply in order to have some measure of their similarity, or because we are interested in the fact that similar words tend to have similar properties in down-stream tasks.
For instance, the words \textit{cat} and \textit{dog} are likely to have the same or similar linguistic analyses in many cases, such as both being tagged with the PoS tag \texttt{NOUN}.
A commonly used similarity measure between vectors is the cosine distance.
In this setting, a word representation as presented above is somewhat problematic, as the distances between the three words are equal, although we want a higher similarity between \textit{cat} and \textit{dog}.

The representation we have seen so far is known as a \textit{sparse} feature representation, as each word is represented by a vector of zeroes with one element set to one, also known as a \textit{one-hot vector}.
Apart from the drawback of similarity, this type of input representation can run into other problems, such as the dimensionality of the representations becoming too large to handle as vocabulary size grows.
This can be remedied in many ways, for instance by applying dimensionality reduction algorithms.
Commonly used algorithms include singular value decomposition (SVD), and random indexing \citep{kanerva:2000,sahlgren:2005}.
This does not help with the problem of similarities, however.

\begin{figure}[htbp]
    \centering
    \includegraphics[width=0.4\textwidth]{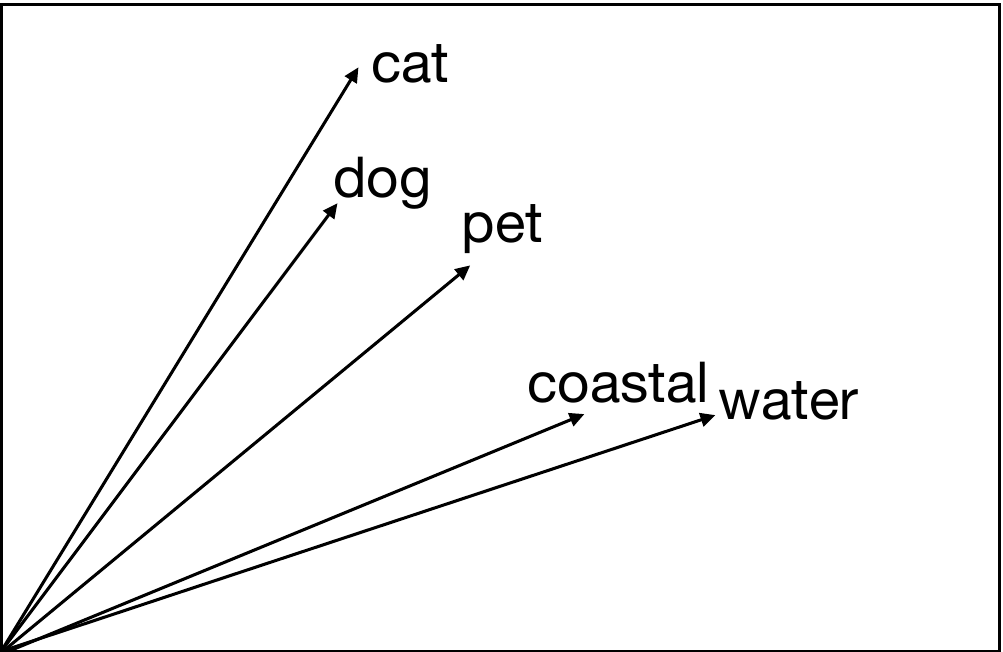}
    \caption{An example of a word space.}
    \label{fig:word_space}
\end{figure}

It turns out that one can arrive at word representations with nice properties of similarity by taking advantage of the \textit{distributional hypothesis}:
\begin{quotation}
    \noindent\textit{'Semantics is partly a function of the statistical distribution of words.'}\\
    --\citet{harris:1954}
\end{quotation}
\begin{quotation}
    \noindent\textit{'You shall know a word by the company it keeps.'}\\
    --\citet[p.11]{firth}
\end{quotation}
This means that the semantic content of a given word is related to other words occurring in similar contexts.
Furthermore, \citet{harris:1954} claims that the strength of this relation is proportional to the similarity between two words, such that if two words $w_1$ and $w_2$ are more similar in meaning than $w_1$ and $w_3$, then the relative distribution of the first pair will be more similar than that of the second pair.
One way of implementing this type of distributional semantics is to count word co-occurrences in a large corpus.
Let us now assign the two remaining indices in the five-dimensional representation used above to the words \textit{pet} and \textit{water}, such that

\begin{equation}
    \begin{aligned}
    \vec{x}_{pet} &= [1, 0, 0, 0, 0],\\
    \vec{x}_{water} &= [0, 0, 0, 1, 0].\\
    \end{aligned}
\end{equation}
These five vectors can then be used to generate new distributional vectors, $\vec{y}$, by representing each word by the sum of the vectors $\vec{x}$ of the words with which it co-occurs.
If \textit{cat} and \textit{dog} frequently co-occur with each other, and with \textit{pet}, whereas \textit{coastal} mainly co-occurs with \textit{water}, the resulting representations may be similar to

\begin{equation}
    \begin{aligned}
    \vec{y}_{cat} &= [25, 0, 20, 0, 10],\\
    \vec{y}_{dog} &= [30, 0, 10, 0, 20],\\
    \vec{y}_{coastal} &= [0, 10, 0, 100, 0].\\
    \end{aligned}
\end{equation}
In this representation, \textit{cat} and \textit{dog} are more similar to one another than they are to \textit{coastal}, which is exactly what we want.
A visualisation of such a word space is given in Figure~\ref{fig:word_space}.

\begin{figure}[htbp]
    \centering
    \includegraphics[width=0.4\textwidth]{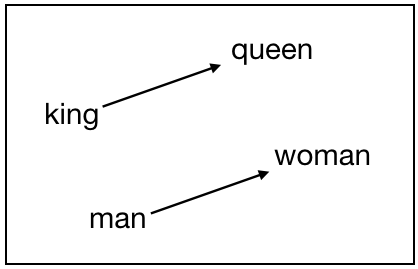}
    \caption{A word space in which adding the difference between \textit{woman} and \textit{man} to \textit{king} results in \textit{queen}.}
    \label{fig:word_similarities}
\end{figure}

The type of word representation discussed up until now is also known as a \textit{count}-based representation, as opposed to a \textit{prediction}-based representation \citep{baroni:predict}.\footnote{Whereas \citet{baroni:predict} suggest that prediction-based methods outperform count-based ones, \citet{levy:2014} show that the underlying differences between the approaches are small.}
Whereas a count-based representation can be seen as counting the words in a given context, a prediction-based representation can be made by attempting to predict that context.
Doing this with a neural network, the error obtained when attempting to make such predictions is used to update the representations, until a low error is obtained (see Chapter~\ref{chp:nn} for details on neural networks).
With the entry of the deep learning tsunami on the NLP scene \citep{manning:2015}, this type of dense word representations has become increasingly popular.
The availability of tools implementing such algorithms, such as \textit{word2vec}, undoubtedly helped push the popularity of this approach further.
This trend was introduced by \citet{collobert:2008}, \citet{turian:2010}, and \citet{collobert:2011}, and was further spearheaded by papers such as \citet{Mikolov:13:Regularities}, which showed that a simple neural model would encapsulate linguistic regularities in its embedded vector space.
The now infamous example of this property is shown in a figure in \citet{Mikolov:13:Regularities}, replicated in Figure~\ref{fig:word_similarities}, where the following relation holds
\begin{equation}
    \overrightarrow{\mathit{king}}+\overrightarrow{\mathit{woman}}-\overrightarrow{\mathit{man}}=\overrightarrow{\mathit{queen}}.
\end{equation}
In other words, the distance between \textit{man} and \textit{woman} is similar to that between \textit{king} and \textit{queen}, so adding this difference to the vector of \textit{king} results in a vector close to \textit{queen}.

In a prediction-based approach, the terminology used is that a word is \textit{embedded} into $n$-dimensional space.
Typically, this dimensionality is much lower (e.g.\ around $100$) than what is generally used in count-based approaches (e.g.\ around $1000$).
In the case of neural networks, these embeddings can be trained together with the rest of the network.
This results in a matrix of word vectors in which words with similar properties (under the task at hand) are close to one another.

\subsubsection*{Distributional vs. Distributed representations}
An useful distinction to make is that of \textit{distributed} vs.~\textit{distributional} representations.
A distributional word representation is based on a co-occurrence matrix, taking advantage of the distributional hypothesis.
Similarities between the resulting distributional word representations thus represent the extent to which they co-occur, and therefore also their semantic similarity.
A distributed representation, on the other hand, is simply one that is continuous.
That is to say, a word is represented by a dense, real-valued, and usually low-dimensional vector.
Such representations are generally known as word embeddings, with each dimension representing some latent feature of the word at hand.
One way to remember this is that such representations are \textit{distributed} across some $n$-dimensional space.
The first representations we saw were therefore distributional, but not distributed \citep{turian:2010}.
The word embeddings, on the other hand, can be said to be both.

\subsubsection*{Bilingual and Multilingual Word Representations}

Going from monolingual to bilingual word representations has been the subject of much attention in recent years.
One of the first approaches to bilingual word representations was shown by \citet{klementiev:2012}, followed by work such as \citet{wolf:2014}, and \citet{coulmance:2015}.
Parallel to approaches which aim at making good multilingual embeddings, are attempts at producing better monolingual embeddings by exploiting bilingual contexts, as in \citet{guo:2014}, \citet{suster:2016:naacl}, and \citet{suster:2016}.

In essence, the approaches to building such representations can be divided up into several categories.
Cross-lingual mapping can be done by first learning monolingual embeddings for separate languages, and then using a bilingual lexicon to map representations from one space to the other \citep{Mikolov:13:MT}.
Another approach is to mix contexts from different languages, and training pre-existing systems, such as word2vec, on this mixed data \citep{gouws:2015}.
The approach under consideration in this thesis is based on exploiting parallel texts, by jointly optimising a loss function when predicting multilingual contexts \citep{multisg}.

\begin{figure}[htbp]
    \centering
    \includegraphics[width=0.4\textwidth]{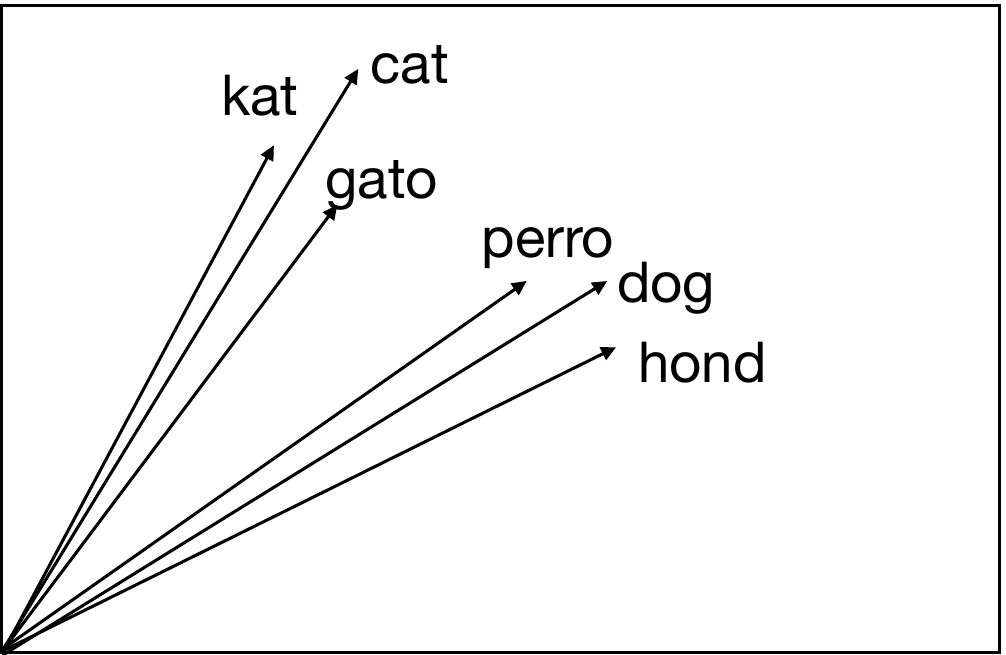}
    \caption{An example of a multilingual word space.}
    \label{fig:multiling_space}
\end{figure}

The true power of multilinguality is not unlocked until we can consider an arbitrary number of languages at a time.
Whereas bilingual word representations only encode two languages, a multilingual word space contains representations from several languages in the same space.
As before, we here also have the property that words with similar meanings are close to one another irrespective of the language (see Figure~\ref{fig:multiling_space}).

One such is the multilingual skip-gram model, as outlined by \citet{multisg}.\footnote{The skip-gram method to create word embeddings, in which a neural network attempts to predict the context of a word, is not to be confused with skip-grams in the sense of $n$-grams which are not necessarily consecutive. In the second sense,  we can define a k-skip-n-gram as a sequence of length n, in which words occur at distance k from each other.
In this thesis, only the first sense is of importance.}
As a variant of this model is used in Part III and Part IV, we will now cover this in more detail.

\subsubsection*{Multilingual Skip-gram}
\label{sec:multisg}
The skip-gram model has become one of the most popular manners of learning word representations in NLP \citep{Mikolov:13:Models}.
This is in part owed to its speed and simplicity, as well as the performance gains observed when incorporating the resulting word embeddings into almost any NLP system.
The model takes a word $w$ as its input, and predicts the surrounding context $c$.
Formally, the probability distribution of $c$ given $w$ is defined as
\begin{equation}
    p(c|w;\theta) = {\exp(\vec{c}^{T}\vec{w}) \over \Sigma_{c\in V} \exp(\vec{c}^{T}\vec{w})},
\end{equation}
where $V$ is the vocabulary, and $\theta$ the parameters of word embeddings ($\vec{w}$) and context embeddings ($\vec{c}$).
The parameters of this model can then be learned by maximising the log-likelihood over $(w,c)$ pairs in the corpus $C$,
\begin{equation}
    J(\theta) = \sum_{(w,c)\in D}\log p(c|w;\theta).
\end{equation}

\citet{multisg} provide a multilingual extension for the skip-gram model, by requiring the model to not only learn to predict English contexts, but also multilingual ones.
This can be seen as a simple adaptation of \citet[p.11]{firth}, i.e., you shall know a word by the \textit{multilingual} company it keeps.
Hence, the vectors for, e.g., \textit{dog} and \textit{perro} ought to be close to each other in such a model.
This assumes access to multilingual parallel data, as word alignments are used in order to determine which words comprise the multilingual context of a word.

\begin{figure}[htbp]
    \centering
    \includegraphics[width=0.8\textwidth]{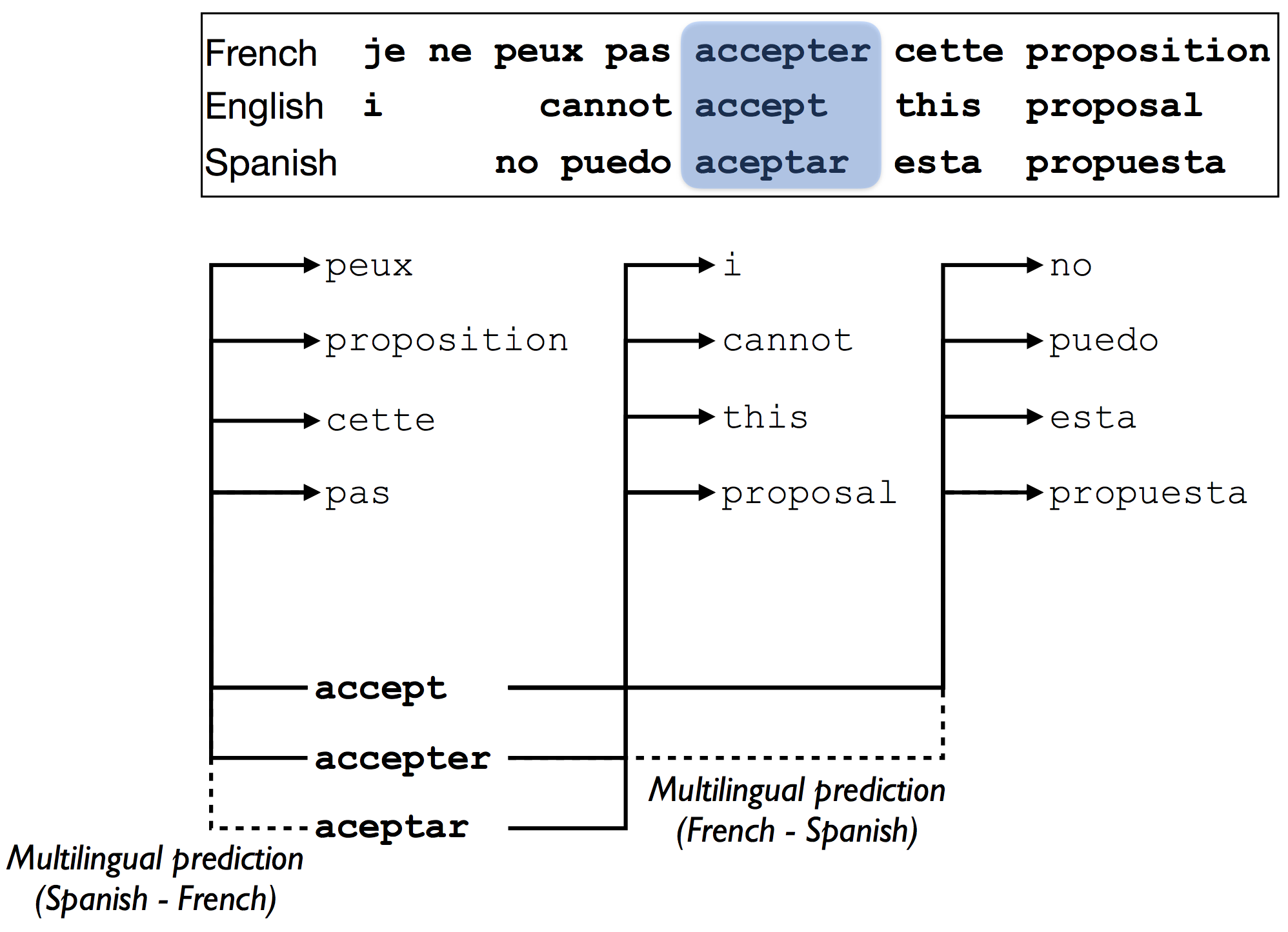}
    \caption{Multilingual skip-gram utilising multilingual contexts. Dashed lines indicate the additions of our loss function, i.e., predictions between every language pair.}
    \label{fig:multi_sg}
\end{figure}

Formally, the learning objective in multilingual skip-gram is defined in \citet{multisg} as
\begin{equation}
    \begin{aligned}
        J &= \alpha\sum_{l\in L}J_{mono_l} + \beta\sum_{l\in L, \{EN\}}J_{bi_{l},EN} \\
        J_{mono_l} &= \sum_{(w,c)\in D_{l\leftrightarrow l}} \log p(c|w;\theta) \\
        J_{bi_l,EN} &= \sum_{(w,c)\in D_{l\leftrightarrow EN}} \log p(c|w;\theta),
    \end{aligned}
\end{equation}
\noindent where L denotes the set of all languages, and $\alpha$ and $\beta$ are weight parameters for the monolingual and bilingual contexts, respectively.
In our work, however, we do not rely on always using English as a pivot, and rather use all bilingual pairings to generate contexts.
In other words, we also predict the French context based on the Spanish word, and vice versa, rather than only predicting from or to English.
This is visualised in Figure~\ref{fig:multi_sg}, in which the dashed lines indicate the additional predictions made using the loss described here, and used in \citet{bjerva:2017:sts}.

Formally, the joint objective function used here is defined as

\begin{equation}
    \begin{aligned}
        J &= \alpha\sum_{l\in L}J_{mono_l} + \beta\sum_{l_1\in L}\sum_{l_2\neq l_1\in L}J_{bi_{l_1},bi_{l_2}} \\
        J_{mono_l} &= \sum_{(w,c)\in D_{l\leftrightarrow l}} \log p(c|w;\theta) \\
        J_{bi_{l_1},bi_{l_2}} &= \sum_{(w,c)\in D_{l_1\leftrightarrow l_2}} \log p(c|w;\theta).
    \end{aligned}
\end{equation}

\section{Outlook}
In the first part of this chapter, we considered multitask learning, which is the focus of Part II of this thesis.
We will first see a case study, in which a MTL paradigm is shown to improve performance on two sequence labelling tasks.
Then we turn to a more theoretical investigation into why this is the case.

Following this, Part III also begins with a case study on multilinguality in a single NLP task.
The subsequent chapter then includes an empirical study of multilinguality in several tasks, and looks at change in performance when multilinguality is employed.

\partimage[width=1\textwidth]{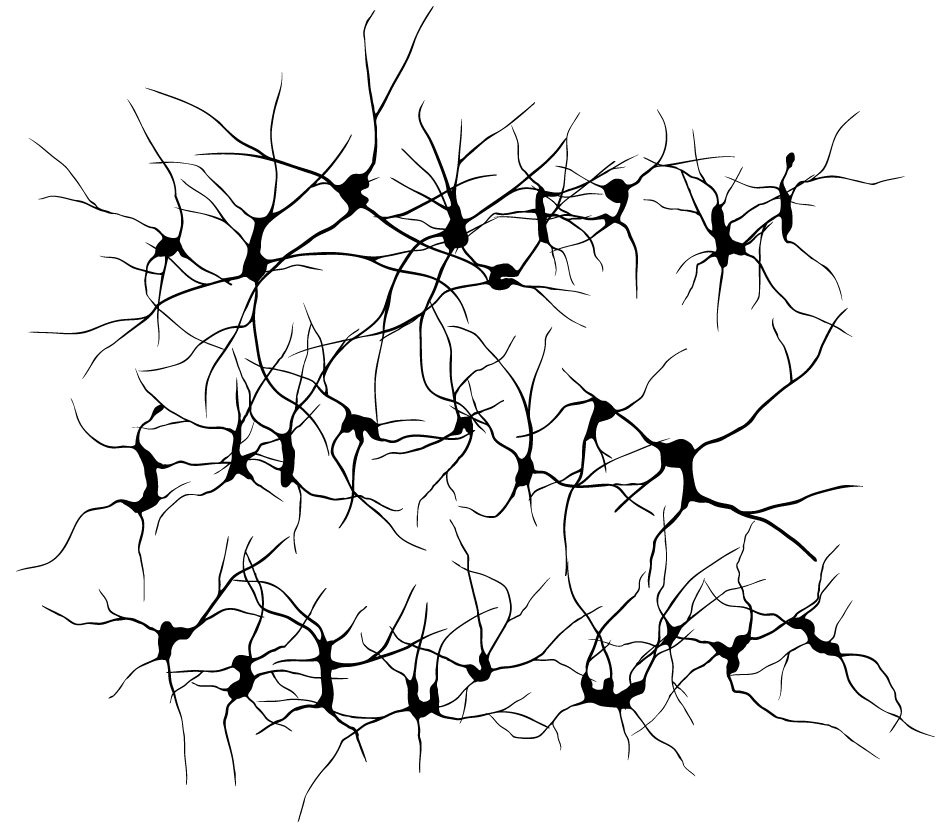}
\part{Multitask Learning}
\renewcommand*{\thefootnote}{\fnsymbol{footnote}}
\chapter[Multitask Semantic Tagging]{\footnote{Chapter adapted from:
\textbf{Bjerva, J.}, Plank, B., and Bos, J. (2016). Semantic tagging with deep residual networks. In Proceedings of COLING 2016, the 26th International Conference on Computational Linguistics: Technical Papers, pages 3531--3541. The COLING 2016 Organizing Committee}
\hspace{-6pt}Multitask Semantic Tagging\\with Residual Networks\hspace{-8pt}}

\renewcommand*{\thefootnote}{\arabic{footnote}}
\label{chp:semtag}
\label{chp:semtags}
\begin{abstract}
\absprelude
In this chapter, a semantic tag set is presented, which is tailored for multilingual semantic parsing.
As a first step towards exploring multitask learning, we will look at the effects of jointly learning this task, together with POS tagging, compared to learning the two tasks separately.
Furthermore, we will see a deep neural network tagger, which is the first tagger to use deep residual networks (ResNets).
The tagger uses both word and character representations, and includes a novel residual bypass architecture.
We evaluate the tagger separately on the semantic tagging task, on POS tagging, and in the multitask learning setting. 
In the multitask setting, the tagger significantly outperforms prior results on English Universal Dependencies POS tagging reaching 95.71\% accuracy on UD v1.2 and 95.67\% accuracy on UD v1.3.
\end{abstract}
\clearpage

\section{Introduction}
A key issue in computational semantics is the transferability of
semantic information across languages. Many semantic parsing systems
depend on sources of information such as POS tags
\citep{pradhan:2004,mrs,step:boxer,Butler:2010,berant:2014}. However,
these tags are often customised for the language at hand \citep{ptb} or
massively abstracted, such as the Universal Dependencies tagset
\citep{ud20,nivre:2016}. Furthermore, POS tags are syntactically oriented,
and therefore often contain both irrelevant and insufficient
information for semantic analysis and deeper semantic processing. This
means that, although POS tags are highly useful for many downstream
tasks, they are unsuitable both for semantic parsing in general, and
for tasks such as recognising textual entailment.


We present a novel set of semantic labels tailored for the purpose of
multilingual semantic parsing.  This tagset (i) abstracts over
POS and named entity types; (ii) fills gaps in semantic
modelling by adding new categories (for instance for phenomena like
negation, modality, and quantification); and (iii) generalises over
specific languages (see Section~\ref{sec:semtagging}).  We introduce and motivate this new task in this
chapter, and refer to it as \textit{semantic tagging}. Our experiments aim to answer the following
two research questions, in order to answer \textbf{RQ~\ref{rq:stag}}:

\begin{enumerate}
\setlength{\itemindent}{.29cm}
    \item [{\bf RQ~\ref{rq:stag}a}] Can we use recent neural network architectures to implement a state-of-the-art neural sequence tagger, which can easily be expanded for multitask learning?


    \item [{\bf RQ~\ref{rq:stag}b}] Semantic tagging is essential for deep semantic parsing, but can we find evidence that semtags are effective also for other NLP tasks, in a multitask learning setting?
\end{enumerate}
We will first look at the semantic tag set, before going into the details of the neural network architecture used, and exploring these research questions.

\section{Semantic Tagging}
\label{sec:semtagging}

\textit{Semantic tagging}, or \textit{semtagging}, is the
task of assigning semantic class categories to the smallest meaningful
units in a sentence. In the context of this chapter these units can be
morphemes, words, punctuation, or multi-word expressions.
These tags are designed so as to facilitate semantic analysis and parsing in cross-linguistic settings, and to be as language neutral as possible, so as to be applicable to several languages \citep{pmb}.
The tag set is motivated by the observation that linguistic information traditionally obtained for deep processing is insufficient
for fine-grained lexical semantic analysis. The widely used Penn
Treebank (PTB) Part-of-Speech tagset \citep{ptb} does not make the
necessary semantic distinctions, in addition to containing redundant
information for semantic processing.

In particular, there are significant differences in meaning between the determiners
\textit{every} (universal quantification), \textit{no} (negation), and
\textit{some} (existential quantification), but they all receive the
\texttt{DT} (determiner) POS label in PTB. Since determiners form a closed
class, 
one could enumerate all word forms for each
class. Indeed some recent implementations of semantic parsing
follow this strategy \citep{step:boxer,Butler:2010}. This might work
for a single language, but it falls short when considering a
multilingual setting. Furthermore, determiners like \textit{any} can
have several interpretations and need to be disambiguated in context.

In addition to this, consider the following examples of redundant information of some POS tagsets, when considering semantic analysis.
For instance, the tagset used in the PTB includes a distinction between \texttt{VBP} (present simple) and \texttt{VBZ} (present simple third person).
In the context of semantic analysis, this type of distinction is not necessary.

Semantic tagging does not only apply to determiners, but reaches all
parts of speech. Other examples where semantic classes disambiguate
are reflexive versus emphasising pronouns (both POS-tagged as
\texttt{PRP}, personal pronoun); the comma, that could be a
conjunction, disjunction, or apposition; intersective vs.\ subsective
and privative adjectives (all POS-tagged as \texttt{JJ}, adjective);
proximal vs.\ medial and distal demonstratives (see Example 2.1); subordinate vs.
coordinate discourse relations; role nouns vs. entity nouns.
\texttt{ROL} is used to separate roles from concepts, which is crucial in order to get accurate semantic behaviour \citep{pmb}.

The set of semantic tags that we use in this chapter is established in a
data-driven manner, considering four languages in a parallel corpus
(English, German, Dutch and Italian). This first inventory of classes
comprises 13 coarse-grained tags and 73 fine-grained tags (see
Table~\ref{table:classes}).\footnote{Experiments in this chapter are based on the tag inventory as detailed in this chapter. This is based on version 0.3 of the semantic tags used in the PMB in June 2016, and has since been revised.}
As can be seen from this table and the examples given below, the tagset
also includes named entity classes.

\begin{examples}
    \item{\gll We must draw attention to the distribution of this form in those dialects .
\texttt{PRO} \texttt{NEC} \texttt{EXS} \texttt{CON} \texttt{REL} \texttt{DEF} \texttt{CON} \texttt{AND} \texttt{PRX} \texttt{CON} \texttt{REL} \texttt{DST} \texttt{CON} \texttt{NIL}
\glt\glend}
    \item{\gll Ukraine 's glory has not yet perished , neither her freedom .
\texttt{GPE} \texttt{HAS} \texttt{CON} \texttt{ENT} \texttt{NOT} \texttt{IST} \texttt{EXT} \texttt{NIL} \texttt{NOT} \texttt{HAS} \texttt{CON} \texttt{NIL}
\glt\glend}
\end{examples}

\noindent In Example 2.1,\footnote{PMB 01/3421, Original source: Tatoeba} both \textit{this} and \textit{those} would be tagged as \texttt{DT}.
However, with our semantic tagset, they are disambiguated as \texttt{PRX} (proximal) and \texttt{DST} (distal).
In Example 2.2,\footnote{PMB 05/0936, Original source: Tatoeba} \textit{Ukraine} is tagged as \texttt{GPE} rather than \texttt{NNP}.

\begin{table}[htbp]
    \centering
    \caption{Semantic tags used in this chapter.\label{table:classes}}
    \begin{tabular}{lll|lll}
    \toprule
      \multirow{5}{*}{\textbf{ANA}} & PRO & pronoun & \multirow{3}{*}{\textbf{MOD}} & NOT & negation \\
      & DEF & definite & & NEC & necessity \\
      & HAS & possessive & & POS & possibility \\ \cline{4-6}
      & REF & reflexive & \multirow{2}{*}{\textbf{ENT}} & CON & concept \\
      & EMP & emphasizing & & ROL & role \\ \cline{1-3} \cline{4-6}
      \multirow{4}{*}{\textbf{ACT}} & GRE & greeting & \multirow{8}{*}{\textbf{NAM}} & GPE & geo-political ent. \\
      & ITJ & interjection & & PER & person \\
      & HES & hesitation & & LOC & location \\
      & QUE & interrogative & & ORG & organisation \\\cline{1-3}
      \multirow{8}{*}{\textbf{ATT}} & QUA & quantity & & ART & artifact \\
      & UOM & measurement & & NAT & natural obj./phen. \\
      & IST & intersective & & HAP & happening \\
      & REL & relation & & URL & url \\ \cline{4-6}
      & RLI & rel. inv. scope & \multirow{15}{*}{\textbf{EVE}} & EXS & untensed simple \\
      & SST & subsective & & ENS & present simple \\
      & INT & intensifier & & EPS & past simple \\
      & SCO & score & & EFS & future simple \\ \cline{1-3}
      \multirow{7}{*}{\textbf{LOG}} & ALT & alternative & & EXG & untensed prog. \\
      & EXC & exclusive & & ENG & present prog. \\
      & NIL & empty & & EPG & past prog. \\
      & DIS & disjunct./exist. & & EFG & future prog. \\
      & IMP & implication & & EXT & untensed perfect \\
      & AND & conjunct./univ. & & ENT & present perfect \\
      & BUT & contrast & & EPT & past perfect \\ \cline{1-3}
      \multirow{6}{*}{\textbf{COM}} & EQA & equative & & EFT & future perfect \\
      & MOR & comparative pos. & & ETG & perfect prog. \\
      & LES & comparative neg. & & ETV & perfect passive \\
      & TOP & pos. superlative & & EXV & passive \\ \cline{4-6}
      & BOT & neg. superlative & \multirow{3}{*}{\textbf{TNS}} & NOW & present tense \\
      & ORD & ordinal & & PST & past tense \\ \cline{1-3}
      \multirow{3}{*}{\textbf{DEM}} & PRX & proximal & & FUT & future tense \\ \cline{4-6}
      & MED & medial & \multirow{6}{*}{\textbf{TIM}} & DOM & day of month \\
      & DST & distal & & YOC & year of century \\ \cline{1-3}
      \multirow{3}{*}{\textbf{DIS}} & SUB & subordinate & & DOW & day of week \\
      & COO & coordinate & & MOY & month of year \\
      & APP & appositional & & DEC & decade \\
      & & & & CLO & clocktime \\
      \bottomrule
    \end{tabular}
\end{table}

\subsection*{Annotated data}

We use two semtag datasets. The Groningen Meaning Bank (GMB) corpus of
English texts (1.4 million words) containing silver standard semantic
tags obtained by running a simple rule-based semantic tagger
\citep{gmb:hla}. This tagger uses POS and named entity tags available
in the GMB (automatically obtained with the C\&C tools
\citep{candcboxer:2007} and then manually corrected), as well as a set
of manually crafted rules to output semantic tags. Some tags related
to specific phenomena were hand-corrected in a second stage.

Our second dataset, the PMB, is smaller but equipped with gold standard semantic
tags and used for testing \citep{pmb}.
It comprises a selection of 400 sentences of the English part of a
parallel corpus. It has no overlap with the GMB corpus.  For this
dataset, we used the Elephant tokeniser, which performs word,
multi-word and sentence segmentation \citep{elephant}. We then used the
simple rule-based semantic tagger described above to get an initial
set of tags. These tags were then corrected by a human annotator.

In order to enable the multitask learning setting, we look at the POS annotation in the English portion of the Universal Dependencies dataset, version 1.2 and 1.3~\citep{nivre:2016}.
An overview of the data used is shown in Table~\ref{tab:data}.
We use the official training, development, and test splits on the UD data.
For the semantic silver standard data set we split the data into 70\% for training, 10\% for development, and 20\% for testing.
We do not use any gold semantic tagging data for training or development, reserving the entire set for testing.

\setlength{\tabcolsep}{4pt}
\begin{table}[htbp]
  \caption{\label{tab:data} Overview of the semantic tagging data (ST Silver from the GMB, ST Gold from the PMB) and the Universal Dependencies data (UD), as of November 2016.}
    \footnotesize
\begin{center}
\begin{tabular}{lrrrr}
\toprule
\sc Corpus                               & \sc Train (sents/toks)  & \sc Dev (sents/toks) & \sc Test (sents/toks) & \sc tags \\
\midrule
ST Silver & 42,599 / 930,201 & 6,084 / 131,337 & 12,168 / 263,516 & 66 \\
ST Gold   & n/a              & n/a             & 356 / 1,718      & 66 \\
UD        & 12,543 / 204,586 & 2,002 / 25,148  & 2,077 / 25,096   & 17 \\
\bottomrule
\end{tabular}
\end{center}
\end{table}
\setlength{\tabcolsep}{6pt}

\section{Method}

To address \textbf{RQ 1a}, we will look at convolutional neural
networks (CNNs) and recurrent neural networks (RNNs), which are both
highly prominent approaches in the recent natural language processing
(NLP) literature (see Chapter~\ref{chp:nn}).  A recent development is the emergence of deep
residual networks (ResNets), a building block for CNNs (see Section~\ref{sec:resnet}).
In short, ResNets consist of several stacked residual units, which can be thought of as
a collection of convolutional layers coupled with a shortcut 
which aids the propagation of the signal in a neural network.  This
allows for the construction of much deeper networks, since keeping a
relatively clean information path in the network facilitates optimisation
\citep{resnets:2016}.  ResNets have recently shown state-of-the-art
performance for image classification tasks \citep{resnets:2015,resnets:2016}, and have also seen some recent use
in NLP \citep{robert:sigmorphon:2016,conneau:2016,bjerva:2016:dsl,google:nmt}.
However, no previous work has attempted to apply ResNets to NLP tagging tasks.

\begin{sidewaysfigure}[p]
    \centering
    \includegraphics[width=\textwidth]{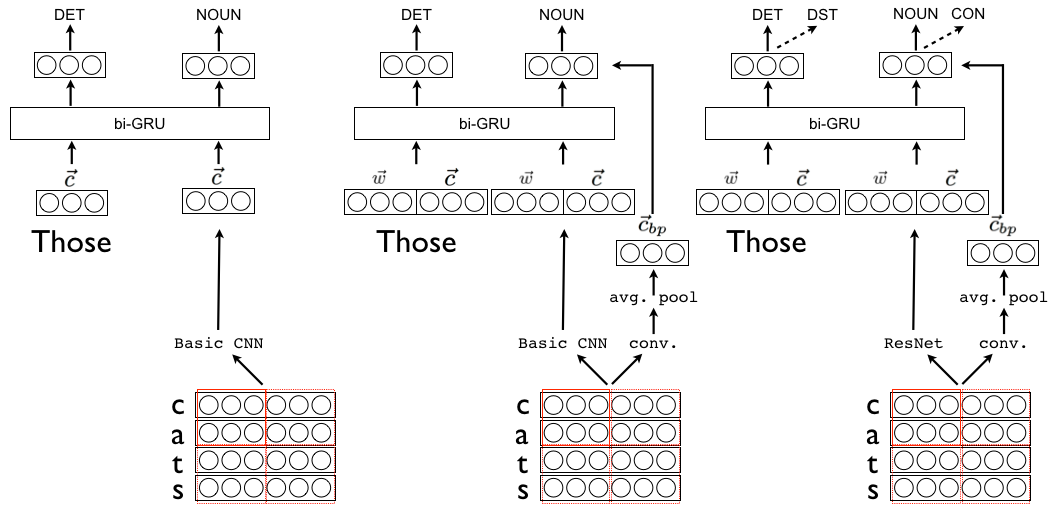}
    \caption{\label{fig:model_arch}Model architecture. Left: Architecture with basic CNN char representations ($\vec{c}$), Middle: basic CNN with char and word representations and bypass ($\vec{c}_{bp}\land\vec{w}$), Right: ResNet with an auxiliary task and residual bypass ($+${\sc aux$_{bp}$}).}
\end{sidewaysfigure}

Our tagger is a hierarchical deep neural network consisting of a bidirectional Gated Recurrent Unit (GRU) network at the upper level, and a CNN or a ResNet at the lower level, including an optional novel residual bypass function (cf.\ Figure~\ref{fig:model_arch}).
Bi-directional GRUs and LSTMs have been shown to yield high performance on several NLP tasks, such as POS tagging, named entity tagging, and chunking \citep{wang2015:unified,yang:2016,plank:2016}.
We build on previous approaches by combining bi-GRUs with character representations from a basic CNN and ResNets.

\subsection{Inception model}
Due to the success of GoogLeNet on the ImageNet 2014 challenge, we experiment with a variant of their model codenamed \textit{Inception} \citep{Inception:2015}.
The \textit{Inception} model applies convolutions of different sizes (e.g. $1\times1$, $3\times3$, $5\times5$) in parallel to the output of the previous layer.
In the system used by \citet{Inception:2015}, several such modules are stacked, forming a network of 22 layers.
The intuition behind building a deep network with modules using varying convolutional patch sizes, is that the features learned by each layer are expected to be both more abstract and less spatially concentrated.
Thus, in the lower layers of the model, the smaller patches are expected to capture most of the correlation statistics of the previous layer, whereas in deeper layers the larger patches are expected to do this (cf.~\citet{Arora:2014}).
Adding a large amount of convolutions na\"ively would explode the amount of parameters to be optimised.
Hence, $1\times1$ convolutions are added before the expensive larger convolutions.
This addition can be thought of as a dimensionality reduction and information compression step, analogous to the success of word embeddings \citep{Inception:2015}.

We apply \textit{Inception} modules directly following the layer containing embedded character representations, and pass this information through to the bi-GRU.
Our main modification on the original \textit{Inception} module is that we apply it in a sequence-to-sequence prediction task.

\subsection{Deep Residual Networks}
We use Deep Residual Networks (ResNets), as introduced in Section~\ref{sec:resnets}.
ResNets have recently been found to yield impressive performance in image recognition tasks, with networks as deep as 1001 layers \citep{resnets:2015,resnets:2016}, and are thus an interesting and effective alternative to simply stacking layers.
In this chapter we use the \textit{assymetric} variant of ResNets similarly to what is described in Equation 9 in \citet{resnets:2016}, namely

\begin{equation}
    \begin{aligned}
    &z_{l+1} = z_l + F(\sigma(z_l)),
    \end{aligned}
\end{equation}
where $z_l$ is the pre-activation of the $l$th layer, $\sigma(z_l)=\alpha_l$, $F$ is a convolutional function.
In other words, we add the pre-activation output of layer $l$ to the output of a convolutional block over the same output, and set this to be the pre-activation of the following layer.
This effectively gives the network a shortcut to the previous layer, which is useful when propagating errors backwards through the network.

In NLP, ResNets have been recently applied to morphological reinflection \citep{robert:sigmorphon:2016}, native language identification \citep{bjerva:nli,rug:nli}, sentiment analysis and text categorisation \citep{conneau:2016}, as well as machine translation \citep{google:nmt}.
Our work is the first to apply ResNets to NLP sequence tagging tasks.
We further contribute to the literature on ResNets by introducing a residual bypass function. The intuition is to combine both deep and shallow processing, which opens a path of easy signal propagation between lower and higher layers in the network.

\subsection{Modelling character information and residual bypass}
\label{sec:char}
Using sub-token representations instead of, or in combination with, word-level representations has recently obtained a lot of attention due to their effectiveness 
\citep{sutskever:2011,chrupala:2013,zhang:2015:char,chung:charactermt,gillick:2015}. 
There is much to be said for approaching NLP using sub-token information.
Doing so allows for much more compact models, since it is no longer necessary to learn weights associated with a large vocabulary, as when using word embeddings.  
Additionally, not relying on the linguistic artefact of white-space delimited text, opens up for learning whatever internal structure is the most appropriate for the task at hand \citep{gillick:2015}.
Sub-token information can be seen as both a potential replacement for token information, or simply as a supplement, which is the approach we take in this chapter.

The use of sub-token representations can be approached in several ways.
\citet{plank:2016} and \citet{yang:2016} use a hierarchical bi-directional RNN, first passing over characters in order to create word-level representations.
\citet{gillick:2015} similarly apply an LSTM-based model using byte-level information directly.
CNNs are used by \citet{dossantos:14}, who construct character-based word-level representations by running a CNN over the character representations of each word.
All of these approaches have in common that the character-based representation is passed through the entire remainder of the network.
Our work is the first to combine the use of character-level representations with both deep processing (i.e., passing this representation through the network) and shallow processing (i.e., bypassing the network in our residual bypass function).
We achieve this by applying our novel residual bypass function to our character representations, inspired by the success of ResNets (depicted in Figure~\ref{fig:model_arch}).
In particular, we first apply the bypass to a CNN-based model achieving large gains over a plain CNN, and later evaluate its effectiveness in a ResNet.
The bypass function allows both lower-level and higher-level features to be taken directly into account in the final layers of the network.
The intuition behind using such a global residual function in NLP is that character information primarily ought to be of importance for the prediction of the current word.
Hence, allowing these representations to bypass our bi-GRU might be beneficial.
This residual bypass function is not dependent on the usage of ResNets, and can be combined with other NN architectures as in our experiments.
We define the penultimate layer, $\alpha_{n-1}$, of a network with $n$ layers, using a residual bypass, as
\begin{equation}
    \begin{aligned}
        &\alpha_{n-1} = \sigma(z_{n-1}) + \alpha_{c},
    \end{aligned}
\end{equation}
where $\sigma$ is some activation function, $z_{n-1}$ is the pre-activation of layer $n-1$, and $\alpha_{c}$ is the activation of the character-level CNN, in the case of our experiments.

\subsection{System description}

The core of our architecture consists of a bi-GRU taking an input based on words and/or characters, with an optional residual bypass as defined in subsection~\ref{sec:char}.
We experiment with a basic CNN, ResNets, a variant of the \textit{Inception} model \citep{Inception:2015}, and our novel residual bypass function.
Our system is implemented in Keras using the Tensorflow backend \citep{keras,tensorflow}.\footnote{System code available at \mbox{\url{https://github.com/bjerva/semantic-tagging}}.}

We represent each sentence using both a character-based representation ($S_c$) and a word-based representation ($S_w$).
The character-based representation is a 3-dimensional matrix $\mathbb{S}_c^{s \times w \times d_c}$, where $s$ is the zero-padded sentence length, $w$ is the zero-padded word length, and $d_c$ is the dimensionality of the character embeddings.
The word-based representation is a 2-dimensional matrix $\mathbb{S}_w^{s \times d_w}$, where $s$ is the zero-padded sentence length and $d_w$ is the dimensionality of the word embeddings.
We use the English Polyglot embeddings \citep{polyglot} in order to initialise the word embedding layer, but also experiment with randomly initialised word embeddings.

Word embeddings are passed directly into a two-layer bi-GRU \citep{gru}.
We also experimented using a bi-LSTM.
However, we found GRUs to yield comparatively better validation data performance on semtags.
We also observe better validation data performance when running two consecutive forward and backward passes before concatenating the GRU layers, rather than concatenating after each forward/backward pass as is commonplace in NLP literature.

We use CNNs for character-level modelling.
Our basic CNN is inspired by \citet{dossantos:14}, who use character representations to produce local features around each character of a word, and combine these with a maximum pooling operation in order to create fixed-size character-level word embeddings.
The convolutions used in this manner cover a few neighbouring letters at a time, as well as the entire character vector dimension ($d_c$).
In contrast to \citet{dossantos:14}, we treat a word analogously to an image.
That is to say, we see a word of $n$ characters embedded in a space with dimensionality $d_c$ as an image of dimensionality $n\times d_c$.
This view gives us additional freedom in terms of sizes of convolutional patches used, which offers more computational flexibility than using only, e.g., $4\times d_c$ convolutions.
This view is applied to all CNN variations explored in this work.



To answer \textbf{RQ1b}, we investigate the effect of using semantic tags as an
auxiliary task for POS tagging, in a multitask learning paradigm (see Chapter~\ref{chp:mtl_bg}).
Since POS tags are useful for many NLP
tasks, it follows that semantic tags must be useful if they can improve POS tagging.
A neural network is trained with respect to a given loss function, such as the cross-entropy between the predicted tag probability distribution and the target probability distribution.
Recent work has shown that the addition of an auxiliary loss function can be beneficial to several tasks.
For instance, \citet{cheng:2015} use this paradigm for language modelling, by predicting the next token while also predicting whether
the sentence at hand contains a name.
\citet{plank:2016} use the log frequency of the current token as an auxiliary task, and find this to improve POS tagging accuracy.
Since our semantic tagging task is based on predicting fine semtags, which can be mapped to coarse semtags, we add the prediction of these coarse semtags as an auxiliary task for the semtagging experiments.
Similarly, we also experiment with POS tagging, where we use the fine semtags as an auxiliary information.

\subsubsection*{Hyperparameters}

All hyperparameters are tuned with respect to loss on the semtag validation set.
We use rectified linear units (ReLUs) for all activation functions \citep{relu}, and apply dropout with $p=0.1$ to both input weights and recurrent weights in the bi-GRU \citep{dropout}.
In the CNNs, we apply batch normalisation \citep{batchnorm} followed by dropout with $p=0.5$ after each layer.
In our basic CNN, we apply a $4\times8$ convolution, followed by $2\times2$ maximum pooling, followed by $4\times4$ convolution and another $2\times2$ maximum pooling.
Our ResNet has the same setup, with the addition of a residual connection.
We also experimented with using average pooling instead of maximum pooling, but this yielded lower validation data performance on the semantic tagging task.
We set both $d_c$ and $d_w$ to $64$.
All GRU layers have $100$ hidden units.
All experiments were run with early stopping monitoring validation set loss, using a maximum of 50 epochs.
We use a batch size of 500.
Optimisation is done using the ADAM algorithm \citep{adam}, with the categorical cross-entropy loss function as training objective.
The main and auxiliary loss functions have a weighting parameter, $\lambda$.
In our experiments, we weight the auxiliary task with $\lambda=0.1$, as set on the semtag auxiliary task, and a weighting of $\lambda=1.0$ for the main task. 

Multi-word expressions (MWEs) are especially prominent in the semtag data, where they are annotated as single tokens.
Pre-trained word embeddings are unlikely to include entries such as `International Organization for Migration', so we apply a simple heuristic in order to avoid treating most MWEs as unknown words. That is to say,
the representation of a MWE is set to the sum of the individual embeddings of each constituent word, such that
\begin{equation}
\begin{aligned}
    &\overrightarrow{\text{mwe}} = \sum_{w\in \text{mwe}}\vec{w},
\end{aligned}
\end{equation}
where $\text{mwe}$ is the MWE at hand, $w\in mwe$ is every word in this mwe, and $\vec{w}$ is the embedded vector representation of that word.


\section{Evaluation}

We evaluate our tagger on two tasks: semantic tagging and POS tagging.
Note that the tagger is developed solely on the semantic tagging task, using the GMB silver training and validation data.
Hence, no further fine-tuning of hyperparameters for the POS tagging task is performed.
We calculate significance using bootstrap resampling \citep{efron:bootstrap}.
The following independent variables are manipulated in our experiments:
\begin{enumerate}
    \item character and word representations ($\vec{w}, \vec{c}$);
    \item residual bypass for character representations ($\vec{c}_{bp}$);
    \item convolutional representations (Basic CNN and ResNets);
    \item auxiliary tasks (using coarse semtags on ST and fine semtags on UD).
\end{enumerate}

\noindent We compare our results to four baselines:
\begin{enumerate}
    \item the most frequent baseline per word (MFC), where we assign the most frequent tag for a word in the training data to that word in the test data, and unseen words get the global majority tag;
    \item the trigram statistic based TNT tagger which offers a slightly tougher baseline \citep{tnt};
    \item the {\sc Bi-lstm} baseline, running the off-the-shelf state-of-the-art POS tagger for the UD dataset \citep{plank:2016} (using default parameters with pre-trained Polyglot embeddings);
    \item we also use a baseline consisting of running our own system with only a {\sc Bi-gru} using word representations ($\vec{w}$), with pre-trained Polyglot embeddings.
\end{enumerate}

\subsection{Experiments on semantic tagging}
We evaluate our system on two semantic tagging (ST) datasets: our silver semtag dataset and our gold semtag dataset.
For the $+${\sc aux} condition we use coarse semtags as an auxiliary task.
Results from these experiments are shown in Table~\ref{tab:stag_results}.

\begin{table}[p]
    \small
\centering
\caption{\label{tab:stag_results}\label{tab:pos_results}
Experiment results on semtag (ST) and Universal Dependencies (UD) test sets (\% accuracy).
{\sc MFC} indicates the per-word most frequent class baseline,
{\sc TNT} indicates the TNT tagger, and
{\sc Bi-lstm} indicates the system by \citet{plank:2016}.
{\sc Bi-gru} indicates the $\vec{w}$ only baseline.
$\vec{w}$ indicates usage of word representations,
$\vec{c}$ indicates usage of character representations, and
$\vec{c}_{bp}$ indicates usage of residual bypass of character representations.
The $+${\sc aux} column indicates the usage of an auxiliary task.}
\begin{tabular}{llrrrr}
\toprule
& & ST Silver &  ST Gold  & UD v1.2     & UD v1.3  \\
\midrule
 \multirow{4}{*}{\textsc{Baselines}} & {\sc MFC} & 84.64	  &  77.39	  & 85.06	    & 85.07  \\
& {\sc TNT} & 92.09	  &  80.73	  & 92.66	    & 92.69  \\
& {\sc Bi-lstm} & 94.98	  &  82.96	  & 95.17	    & 95.04  \\
& {\sc Bi-gru} & 94.26	  &  80.26	  & 94.39	    & 94.32  \\
\midrule
 \multirow{4}{*}{\textsc{Basic CNN}} & $\vec{c}$ & 91.39	  &  69.21	  & 77.63	    & 77.51  \\
& $\vec{c}_{bp}$ & 90.18	  &  65.77	  & 83.53	    & 82.89  \\
& $\vec{c}_{bp}\land\vec{w}$ & 94.63	  &  76.83	  & 94.68	    & 94.89  \\
& $+${\sc aux}$_{bp}$ & 94.53	  &  80.73	  & 95.19	    & 95.34  \\
\midrule
 \multirow{6}{*}{\textsc{ResNet}} & $\vec{c}$ & 94.39	  &  76.89	  & 92.65	    & 92.63  \\
& $\vec{c}\land\vec{w}$ & 95.14	  &  {\bf 83.64}	  & 94.92	    & 94.88  \\
& $+${\sc aux} & 94.23	  &  74.84	  & {\bf 95.71}    & {\bf 95.67}  \\
& $\vec{c}_{bp}$& 94.23	  &  75.84	  & 92.45	    & 92.86  \\
& $\vec{c}_{bp}\land\vec{w}$ & {\bf 95.15}	  &  82.18	  & 94.73	    & 94.69  \\
& $+${\sc aux}$_{bp}$ & 94.58	  &  73.73	  & 95.51	    & 95.57  \\
 \bottomrule
\end{tabular}
\end{table}


%

\subsection{Experiments on Part-of-Speech tagging}
We evaluate our system on v1.2 and v1.3 of the English part of the Universal Dependencies (UD) data.
We report results for POS tagging alone, comparing to commonly used baselines and prior work using LSTMs, as well as using the fine-grained semantic tags as auxiliary information.
For the $+${\sc aux} condition, we train a single joint model using a multi-task objective, with POS and ST as our two tasks.
This model is trained on the concatenation of the ST silver data with the UD data, updating the loss of the respective task of an instance in each iteration.
Hence, the weights leading to the UD output layer are not updated on the ST silver portion of the data, and vice-versa for the ST output layer on the UD portion of the data.
Results from these experiments are shown in Table~\ref{tab:pos_results}.

\subsection{The Inception architecture}

For comparison with the ResNet, we evaluate the \textit{Inception} architecture on ST Silver data, and UD v1.2 and v1.3 (see Table~\ref{tab:Inception}).

\setlength{\tabcolsep}{6pt}
\begin{table}[htbp]
  \caption{\label{tab:Inception}
  Results when using the \textit{Inception} architecture on ST and UD data.
  $\vec{w}$ indicates usage of word representations,
  $\vec{c}$ indicates usage of character representations, and
  $\vec{c}_{bp}$ indicates usage of residual bypass of character representations.}
    \small
\begin{center}
\begin{tabular}{lcccc}
\toprule
                & $\vec{c}$ & $\vec{c}_{bp}$ & $\vec{c}_{bp}\land\vec{w}$ & ResNet\\
\midrule
ST Silver & 94.40 & 93.32 & 94.64  & \bf 95.15 \\
UD v1.2   & 90.82 & 89.78 & \bf 95.07  & 94.73 \\
UD v1.3   & 91.12 & 89.55 & \bf 94.90  & 94.69 \\
\bottomrule
\end{tabular}
\end{center}
\end{table}

\subsection{Effect of pre-trained embeddings}

In our main experiments, we initialise the word embedding layer with pre-trained polyglot embeddings.
We compare this with randomly initialising this layer from a uniform distribution over the interval $[-0.05,0.05)$, without any pre-training.
Results from these experiments are shown in Table~\ref{tab:no_emb}.

\setlength{\tabcolsep}{6pt}
\begin{table}[htbp]
  \caption{\label{tab:no_emb}
  Results under the $\vec{c}_{bp}\land\vec{w}$ and $\vec{c}_{bp}\land\vec{w}${\sc +aux} conditions, on ST and UD data, using randomly initialised word embeddings.
  Change in accuracy is indicated in brackets.}
    \small
\begin{center}
\begin{tabular}{lcc}
\toprule
                &  $\vec{c}_{bp}\land\vec{w}$ & {\sc +aux} \\
\midrule
ST Silver  & 95.11 (-0.04) & 94.57 (-0.01)  \\
UD v1.2    & 91.94 (-2.79) & 94.90 (-0.61) \\
UD v1.3    & 92.00 (-2.69) & 94.96 (-0.61)  \\
\bottomrule
\end{tabular}
\end{center}
\end{table}

\section{Discussion}
\subsection{Performance on semantic tagging}

The overall best system is the ResNet combining both word and character representations $\vec{c}_{}\land\vec{w}$.
It outperforms all baselines, including the recently proposed RNN-based bi-LSTM.
On the ST silver data, a significant difference ($p<0.01$) is found when comparing our best system to the strongest baseline ({\sc bi-lstm}).
On the ST gold data, we observe significant differences at the alpha values recommended by \citet{Sogaard:2014}, with $p<0.0025$.
The residual bypass effectively helps improve the performance of the basic CNN.
However, the tagging accuracy of the CNN falls below baselines.
In addition, the large gap between gold and silver data for the CNN shows that the CNN model is more prone to overfitting, thus favouring the use of the ResNet.
Adding the coarse-grained semtags as an auxiliary task only helps for the weaker CNN model.
The ResNet does not benefit from this additional information, which is already captured in the fine-grained labels.

It is especially noteworthy that the ResNet character-only system performs remarkably well, as it outperforms the {\sc Bi-gru} and TNT baselines, and is considerably better than the basic CNN.
Since performance increases further when adding in $\vec{w}$, it is clear that the character and word representations are complimentary in nature.
The high results for characters only are particularly promising for multilingual language processing, as such representations allow for much more compact models (see, e.g.,~\citet{gillick:2015}).
This further indicates that ResNet-based character representations can almost account for the same amount of compositionality as word representations.

\subsection{Performance on Part-of-Speech tagging}

Our system was tuned solely on semtag data.
This is reflected in, e.g., the fact that even though our $\vec{c}\land\vec{w}$ ResNet system outperforms the \citet{plank:2016} system on semtags, we are substantially outperformed on UD 1.2 and 1.3 in this setup.
However, adding an auxiliary task based on our semtags markedly increases performance on POS tagging.
In this setting, our tagger outperforms the {\sc Bi-lstm} system, and results in new state-of-the-art results on both UD 1.2 ($95.71\%$ accuracy) and 1.3 ($95.67\%$ accuracy).
The difference between the {\sc Bi-lstm} system and our best system is significant at $p<0.0025$.

The fact that the semantic tags improve POS tagging performance reflects two properties of semantic tags.
Firstly, it indicates that the semantic tags carry important information which aids the prediction of POS tags.
This should come as no surprise, considering the fact that the semtags abstract over and carry more information than POS tags.
Secondly, it indicates that the new semantic tagset and released dataset are useful for downstream NLP tasks.
This could be done indirectly, by running a POS tagger which has been trained in a multi-task setting with semtags.
Alternatively, the semtags would likely be useful features for downstream tasks.
In this chapter we show this by using semtags as an auxiliary task.
In future work we aim to investigate the effect of introducing the semtags directly as features into the embedded input representation.

\subsection{Inception}
The experiments using the \textit{Inception} architecture showed that the ResNet architecture we used performs better on semantic tagging.
The first of these results is in line with results in, e.g., image recognition, where ResNets are also superior to \textit{Inception} \citep{resnets:2015,resnets:2016}.
On UD PoS tagging, however, results using \textit{Inception} were marginally better.
This might be explained by the fact that the ResNet architecture was rather heavily tuned on semantic tagging, whereas \textit{Inception} was tuned to a lesser extent.
Nonetheless, both architectures outperform the use of a standard relatively shallow CNN, indicating that they may indeed be more suitable for similar tasks, given similar amounts of data.

\subsection{Residual bypass}
Our novel residual bypass function outperforms corresponding models without residual bypass in some cases.
Notably for POS tagging with a standard CNN, the increase in tagging accuracy is around 5\%.
Combining this with a ResNet does not have a large effect on tagging performance, resulting in slightly higher accuracy on UD 1.3, but lower on UD 1.2 and semtags.
This indicates that, although using a residual bypass allows for the character information to propagate more easily, this is not crucial for the model to capture subtoken information and use this effectively.
An interesting possibility for future research is to investigate the use of a residual bypass on word-level representations.

\subsection{Pre-trained embeddings}

For semantic tagging, the difference with random initialisation is negligible, with pre-trained embeddings yielding an increase in about 0.04\% accuracy.
For POS tagging, however, using pre-trained embeddings increased accuracy by almost 3 percentage points for the ResNet.

\section{Conclusions}
In this chapter, we first introduced a semantic tagset tailored for multilingual semantic parsing.
We compared tagging performance using standard CNNs and the recently emerged ResNets.
For semantic tagging, ResNets are more robust and result in our best model.
Combining word and ResNet-based character representations helps to outperform state-of-the-art taggers on semantic tagging, while allowing for straightforward extension to an MTL paradigm ({\bf RQ~\ref{rq:stag}a}).
Since we were interested in seeing whether the new tagset could be informative for other tasks, we used semantic tagging as an auxiliary task for PoS tagging.
This yielded state-of-the-art performance on the English UD 1.2 and 1.3 POS datasets, showing that semantic tags are informative for other NLP tasks ({\bf RQ~\ref{rq:stag}b}).
The fact that using semantic tags aided POS tagging raises the question of in which cases it is useful to have an auxiliary tagging task ({\bf RQ~\ref{rq:mtl}}), which is explored in the following chapter.


\renewcommand*{\thefootnote}{\fnsymbol{footnote}}
\chapter[Information-theoretic Perspectives on Multitask Learning]{\footnote{Chapter adapted from: \textbf{Bjerva, J.} (2017)
Will my auxiliary tagging task help? Estimating Auxiliary Tasks Effectivity in Multi-Task Learning, in Proceedings of the 21st Nordic Conference on Computational Linguistics, NoDaLiDa, 22-24 May 2017, Gothenburg, Sweden, number 131, pages 216–220. Link\"oping University Electronic Press, Link\"opings universitet. Best short-paper award.}
\hspace{-6pt}Information-theoretic Perspectives on Multitask Learning Effectivity\hspace{-8pt}}

\renewcommand*{\thefootnote}{\arabic{footnote}}
\label{chp:mtl}
\begin{abstract}
\absprelude
In the previous chapter, we saw that multitask learning improved the performance on POS tagging, when using semantic tagging as an auxiliary task.
In fact, multitask learning often improves system performance for various tasks in ML in general, and NLP in particular.
However, the question of \textit{when} and \textit{why} this is the case has yet to be answered satisfactorily.
Although previous work has hypothesised that this is linked to the label distributions of the auxiliary task, it can be argued that this is not sufficient.
In this chapter, we will see that information-theoretic measures which consider the joint label distributions of the main and auxiliary tasks offer far more explanatory value.
The findings in this chapter are empirically supported by experiments on morphosyntactic tasks on 39 languages, and by experiments on several semantic tasks for English.
\end{abstract}

\section{Introduction}
When attempting to solve a natural language processing (NLP) task, one can consider the fact that many such tasks are highly related to one another.
As discussed in Chapter~\ref{chp:mtl_bg}, a common way of taking advantage of this is to apply multitask learning (MTL, \citet{mtl}).
MTL has been successfully applied to many linguistic sequence prediction tasks, both syntactic and semantic in nature \citep{collobert:2008,cheng:2015,sogaard2016deep,bjerva:2016:semantic,ammar:2016,plank:2016,alonso:mtl,bingel:2017}.
This trend is in part owed to the fact that a specific type of MTL, namely hard parameter sharing in neural networks, is relatively easy to implement and often quite effective.
It is, however, unclear \textit{when} an auxiliary task is useful, although previous work has provided some insights \citep{mtl,alonso:mtl,bingel:2017}.
For a further overview of MTL, see Chapter~\ref{chp:mtl_bg}.

Currently, considerable time and effort need to be employed in order to experimentally investigate the usefulness of any given main task / auxiliary task combination.
In this chapter the aim is to alleviate this process by providing a means to empirically investigating the potential effectivity of an auxiliary task.
We aim to answer the following two research questions, in order to answer {\bf RQ~\ref{rq:mtl}}:
\begin{enumerate}
\setlength{\itemindent}{.29cm}
    \item [{\bf RQ~\ref{rq:mtl}a}] Which information-theoretic measures can be used to estimate auxiliary task effectivity?
    \item [{\bf RQ~\ref{rq:mtl}b}] To what extent do correlations between information-theoretic measures and auxiliary task effectivity generalise across languages and NLP tasks?
\end{enumerate}
\noindent Concretely, we apply information-theoretic measures to a collection of data- and tag sets, and investigate correlations between these measures and auxiliary task effectivity.
We investigate this both experimentally on a collection of syntactically oriented tasks on 39 languages, as well as on several semantically oriented tasks for English.
We take care to structure our experiments so as to generalise across many common real-world situations in which MTL is applied.
Concretely, we apply neural multitask learning (see Chapter~\ref{chp:mtl}), using a bi-directional GRU (bi-GRU, see Section~\ref{sec:rnn}), as introduced in Chapter~\ref{chp:semtag}, using hard parameter sharing.

\section{Information-theoretic Measures}
We wish to give an information-theoretic perspective on when an auxiliary task will be useful for a given main task.
For this purpose, we introduce some common information-theoretic measures which will be used throughout this work.\footnote{See \citet{cover:2012} for an in-depth overview.}

\subsection{Entropy}
\label{sec:entropy}
The \textbf{entropy} of a probability distribution, originally described in \citet{shannon:1949}, is a measure of its unpredictability.
That is to say, high entropy indicates a uniformly distributed tag set, while low entropy indicates a more skewed distribution.
Formally, the entropy of a tag set can be defined as
\begin{equation}
    H(X)=-\sum _{x\in X}p(x)\log p(x),
\end{equation}
where $x$ is a given tag in tag set $X$.

\subsection{Conditional Entropy}
\begin{figure}
    \centering
    \includegraphics[width=0.7\columnwidth]{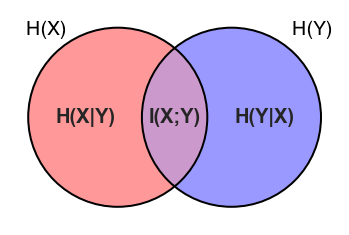}
    \vspace{-0.5cm}
    \caption{\label{fig:inf_theory}Information theory overview. The left circle denotes H(X), and the right circle H(Y). Blue ($H(X|Y)$) indicates the conditional entropy of $X$ given $Y$, red ($H(Y|X)$) indicates the opposite, and purple ($I(X;Y)$) indicates the mutual information of $X$ and $Y$.}
\end{figure}

It may be more informative to take the joint probabilities of the main and auxiliary tag sets in question into account, for instance using \textbf{conditional entropy}.
This is depicted in Figure~\ref{fig:inf_theory} as $H(X|Y)$ and $H(Y|X)$, with red and blue respectively.
Formally, the conditional entropy of a distribution $Y$ given the distribution $X$ is defined as
\begin{equation}
    H(Y|X) = \sum_{x\in X}\sum_{y\in Y} p(x,y)\log\frac{p(x)}{p(x,y)},
\end{equation}
where $x$ and $y$ are all variables in the given distributions, $p(x,y)$ is the joint probability of variable $x$ cooccurring with variable $y$, and $p(x)$ is the probability of variable $x$ occurring at all.
That is to say, if the auxiliary tag of a word is known, this is highly informative when deciding what the main tag should be.
In the case of a multitask setup, $Y$ and $X$ are the distributions of the main and auxiliary task tag sets respectively.
The variables $y$ and $x$ are specific tags in these tag sets.

\subsection{Mutual Information}
The \textbf{mutual information} (MI) of two tag sets is a measure of the amount of information that is obtained of one tag set, given the other tag set.
MI can be defined as
\begin{equation}
    I(X;Y)=\sum _{x\in X}\sum _{y\in Y}p(x,y)\log\frac{p(x,y)}{p(x)\,p(y)},
\end{equation}
where $x$ and $y$ are all variables in the given distributions, $p(x,y)$ is the joint probability of variable $x$ cooccurring with variable $y$, and $p(x)$ is the probability of variable $x$ occurring at all.
This is depicted in Figure~\ref{fig:inf_theory} as $I(X;Y)$, in purple, which also illustrates the alternative definition of MI, namely expressed in terms of entropy and conditional entropy as
\begin{equation}
    \begin{aligned}
    I(X;Y)&\equiv H(X) - H(X|Y)\equiv H(Y) - H(Y|X).
    \end{aligned}
\end{equation}
In the figure, this is depicted as that subtracting either $H(X)$ from $H(X|Y)$ or $H(Y)$ from $H(Y|X)$ will result in $I(X;Y)$.
MI describes how much information is shared between $X$ and $Y$, and can therefore be considered a measure of `correlation' between tag sets.
Should two tag sets be completely independent from each other, then knowing $Y$ would not give any information about $X$.

\subsection{Information Theory and MTL in NLP}
Entropy has in the literature been hypothesised to be related to the usefulness of an auxiliary task \citep{alonso:mtl}.
We argue that this explanation is not entirely sufficient.
Take, for instance, two tag sets $X$ and $X'$, applied to the same corpus and containing the same tags.
Consider the case where the annotations differ in that the labels in every sentence using $X'$ have been randomly reordered.
Such a situation is shown in the following examples:

{\small
\begin{examples}
    \item{\gll The quick brown fox jumps over the lazy dog  .
\texttt{DET} \texttt{ADJ} \texttt{ADJ} \texttt{NOUN} \texttt{VERB} \texttt{ADP} \texttt{DET} \texttt{ADJ} \texttt{NOUN} \texttt{PUNCT}
\glt\glend}
    \item{\gll The quick brown fox jumps over the lazy dog  .
\texttt{ADJ} \texttt{ADJ} \texttt{DET} \texttt{DET} \texttt{NOUN} \texttt{NOUN} \texttt{ADJ} \texttt{ADP} \texttt{VERB} \texttt{PUNCT}
\glt\glend}
\end{examples}
}
\noindent The tag distributions in $X$ and $X'$ do not change as a result of a reordering as in the examples, hence the tag set entropies will be the same.\footnote{Note that we look at the entropy of the marginal distribution of each tag set, as this is what has been hypothesised to be of importance in previous work.}
However, the tags in $X'$ are now likely to have a vanishingly low correspondence with any sort of natural language signal (as in the second sentence), hence $X'$ is highly unlikely to be a useful auxiliary task for $X$.
Measures taking joint probabilities into account will capture this lack of correlation between $X$ and $X'$.
In this work we show that measures such as conditional entropy and MI are much more informative for the effectivity of an auxiliary task than entropy.

\section{Data}
For our syntactic experiments, we use the Universal Dependencies (UD) treebanks on 39 out of the 40 languages found in version 1.3 \citep{nivre:2016}.\footnote{Japanese was excluded due to treebank unavailability.}
We experiment with POS tagging as a main task, and various dependency relation classification tasks (as defined in Section~\ref{sec:deprel}) as auxiliary tasks.
We also investigate whether our hypothesis fits with recent results in the literature, and train sequence taggers on the collection of semantically oriented tasks presented in \citet{alonso:mtl}, as well as on the semantic tagging task in \citet{bjerva:2016:semantic}.

Although calculation of joint probabilities requires jointly labelled data, this issue can be bypassed without losing much accuracy.
Assuming that (at least) one of the tasks under consideration can be completed automatically with high accuracy, we find that the estimates of joint probabilities are very close to actual joint probabilities on gold standard data.
In this work, we estimate joint probabilities by tagging the auxiliary task data sets with a state-of-the-art POS tagger.\footnote{Since the dependency relation auxiliary task data overlaps with POS tagging data, this allowed us to confirm that the differences between the measures obtained with estimated and gold data in this case are negligible (i.e. $\leq5\%$).}\footnote{The POS-tagger used is the deep bi-GRU ResNet-tagger described in Chapter~\ref{chp:semtag}.}

\subsection{Morphosyntactic Tasks}
\label{sec:deprel}
Dependency Relation Classification is the task of predicting the dependency tag (and its direction) for a given token. 
This is a task that has not received much attention, although it has been shown to be a useful feature for parsing \citep{ouchi:2014}.
We choose to look at several instantiations of this task, as it allows for a controlled setup under a number of conditions for MTL, and since data is available for a large number of typologically varied languages.

Previous work has suggested various possible instantiations of dependency relation classification labels, differing in the amount of information they encode \citep{ouchi:2014,ouchi:2016}.
In this work, we use labels designed to range from highly complex and informative, to relatively basic ones.\footnote{Labels are automatically derived from the UD dependency annotations.}
The labelling schemes used are shown in Table~\ref{tab:supertags}.
\begin{table}[h]
    \centering
    \caption{\label{tab:supertags} Dependency relation labels used in this work, with entropy in bits ($H$) measured on English. The labels differ in the granularity and/or inclusion of the category and/or directionality.}
    \begin{tabular}{lllr}
        \toprule
        {\bf Category} & {\bf Directionality} & {\bf Example} & {\bf $H$} \\
        \midrule
         Full   & Full   & nmod:poss/R\_L & 3.77 \\
         Full   & Simple & nmod:poss/R    & 3.35 \\
         Simple & Full   & nmod/R\_L      & 3.00 \\
         Simple & None   & nmod           & 2.03 \\
         None   & Full   & R\_L           & 1.54 \\
         None   & Simple & R              & 0.72 \\
        \bottomrule
    \end{tabular}
\end{table}
As an example, consider the following dependency graph:
\begin{center}
\begin{dependency}[theme = simple]
   \begin{deptext}[column sep=2em]
      No \& , \& it \& was \& n't \& Black \& Monday \& . \\
   \end{deptext}
   \deproot{4}{ROOT}
   \depedge{4}{1}{VMOD}
   \depedge{4}{2}{P}
   \depedge{4}{3}{SUB}
   \depedge{4}{5}{VMOD}
   \depedge{4}{7}{PRD}
   \depedge{7}{6}{NMOD}
   \depedge{4}{8}{P}
\end{dependency}
\end{center}
Table~\ref{tab:supertag_examples} shows examples of dependency relation labels based on this graph.
The labels encode the head of each word, as well as the relative position, or direction.
For instance, the word \textit{it} is the subject of a word on its right.
The more complex tags, for instance \textit{ROOT+SUB/L\_PRD/R}, include information regarding the dependents of each word.
In this case, the word \textit{was} has an obligatory \textit{sub} dependent to the left, and an obligatory \textit{prd} dependent on the right.

\setlength{\tabcolsep}{3pt}
\begin{table}[h]
    \centering
    \caption{\label{tab:supertag_examples} Examples of some dependency relation instantiations in context. The columns indicate the granularity used (category/directionality).}
    \resizebox{\columnwidth}{!}{
    \begin{tabular}{lllll}
        \toprule
        \textbf{Word} & \textbf{Full/Full} & \textbf{Simple/Simple} & \textbf{Simple/None} & \textbf{None/Simple} \\
        \midrule
        No     &  \sc vmod/r           & VMOD/R & VMOD & R \\
        ,      &  \sc P/R              & P/R & P & R \\
        it     &  SUB/R            & SUB/R & SUB & R \\
        was    & ROOT+SUB/L\_PRD/R  & ROOT & ROOT &  \\
        n't    & VMOD/L            & VMOD/L & VMOD & L \\
        Black  &  NMOD/R           & NMOD/R & NMOD & R \\
        Monday & PRD/L+L           & PRD/L & PRD & L \\
        .      &  P/L              & P/L & P & L \\
        \bottomrule
    \end{tabular}
    }
\end{table}
\setlength{\tabcolsep}{6pt}

The systems in the syntactic experiments are trained on main task data ($D_{main}$), and on auxiliary task data ($D_{aux}$).
Generally, the amount of overlap between such pairs of data sets differs, and can roughly be divided into three categories:
i) identity (identical data sets);
ii) overlap (some overlap between data sets);
and iii) disjoint (no overlap between data sets).
To ensure that we cover several possible experimental situations, we experiment using all three categories.
We generate ($D_{main}$, $D_{aux}$) pairs by splitting each UD training set into three portions.
The first and second portions always contain POS labels.
In the identity condition, the second portion contains dependency relations.
In the overlap condition, the second and final portions contain dependency relations.
In the disjoint condition, the final portion contains dependency relations.
Hence, the system always sees the exact same POS tagging data, whereas the amount of dependency relation data and overlap differs between conditions.
Each dependency relation instantiation is used in our experiments, paired with PoS tagging.
The data splitting scheme is shown in Table~\ref{tab:data_splits}.

\begin{table}
    \caption{Data splitting scheme. The training set is split into three equal parts. The annotations in each part differ per condition.    \label{tab:data_splits}}
    \centering
    \begin{tabular}{lccc}
        \toprule
        \textbf{Condition} & \textbf{Part I} & \textbf{Part II} & \textbf{Part III} \\
        \midrule
        \textbf{Identity}  & PoS & PoS $\land$ DepRel & n/a \\
        \textbf{Overlap}   & PoS & PoS $\land$ DepRel & DepRel \\
        \textbf{Disjoint}  & PoS & PoS & DepRel \\
        \bottomrule
    \end{tabular}
\end{table}

\subsection{Semantic Tasks}
\citet{alonso:mtl} experiment with using POS tagging, chunking, dependency relation tagging, and a frequency based measure as auxiliary tasks, with main tasks based on several semantically oriented tasks.
In this chapter, we limit ourselves to considering the PoS tagging auxiliary task, for the following semantic main tasks.

\subsubsection*{Named Entity Recognition}
For NER, we use the CONLL2003 shared-task data (e.g.~\ Person, Loc, etc., \citet{tjong:2003}).

\subsubsection*{Frames}
We use FrameNet 1.5 \citep{framenet} with the same data splits as \citet{das:2014} and \citet{hermann:2014}.
This data set is annotated for the joint task of frame detection and identification.
As in \citet{alonso:mtl}, we approach this task as a standard sequence prediction task.

\subsubsection*{Supersenses}
We experiment with the supersense version of SemCor \citep{miller:1993} from \citet{ciaramita:2006},
using course-grained semantic labels (e.g.~\ noun.person).

\subsubsection*{Semtraits}
We use the conversions of \citet{alonso:mtl} of supersenses into coarser semantic traits (e.g. Animate, UnboundedEvent, etc.), for which they used
the EuroWordNet list of ontological types for senses from \citet{vossen:1998}.

\subsubsection*{Multi-Perspective Question Answering}
We also consider the Multi-Perspective Question Answering (MPQA) corpus, using the coarse level of annotation \citep{deng:2015}.

\subsubsection*{Semantic Tags}
We also investigate the semantic tagging task of Chapter~\ref{chp:semtags}, using the same data splits \citep{bjerva:2016:semantic}.

We use the same setup as for our syntactic experiments, by using these semantic tasks as auxiliary tasks with POS tagging as the main task.

\section{Method}


\subsection{Architecture and Hyperparameters}
\begin{figure}
    \centering
    \includegraphics[width=0.6\textwidth]{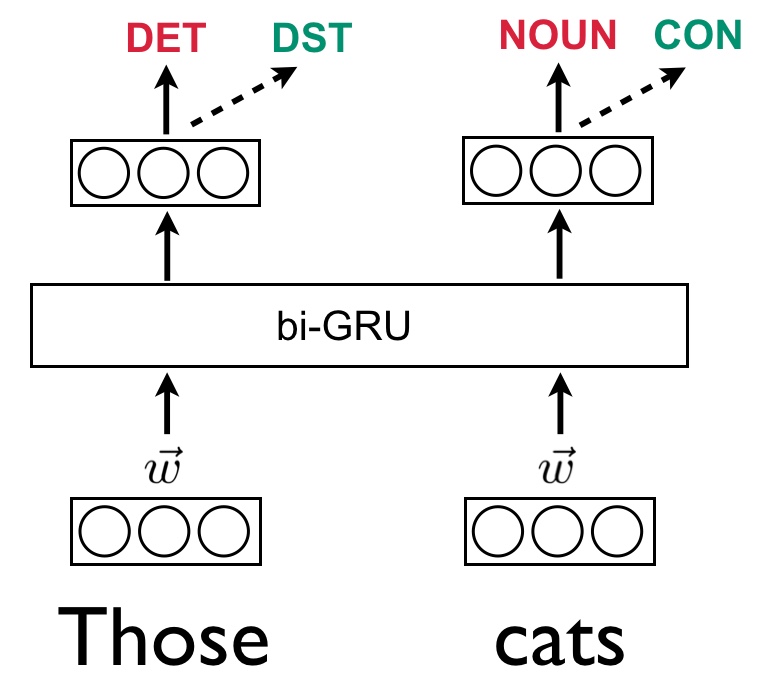}
    \caption{\label{fig:mtl_sys_arch}System architecture used in the multitask learning experiments.}
\end{figure}

We apply a deep neural network with the exact same hyperparameter settings in each syntactic experiment, with reasonably default parameter settings, similar to what was used in Chapter~\ref{chp:semtags}.
Our system consists of a two layer deep bi-GRU (100 dimensions per layer), taking an embedded word representation (64 dimensions) as input (see Figure~\ref{fig:mtl_sys_arch}).
We apply dropout ($p=0.4$) between each layer in our network \citep{dropout}.
The output of the final bi-GRU layer, is connected to two output layers -- one per task.
Both tasks are always weighted equally.
Optimisation is done using the Adam algorithm \citep{adam}, with the categorical cross-entropy loss function.
We use a batch size of 100 sentences, training over a maximum of 50 epochs, using early stopping and monitoring validation loss on the main task.

We do not use pre-trained embeddings.
We also do not use any task-specific features, similarly to \citet{collobert:2011}, and we do not optimise any hyper-parameters with regard to the task(s) at hand.
Although these choices are likely to affect the overall accuracy of our systems negatively, the goal of our experiments is to investigate the effect in \textit{change} in accuracy when adding an auxiliary task - not accuracy in itself.

\subsection{Experimental Overview}
In the syntactic experiments, we train one system per language, dependency label category, and split condition.
For sentences where only one tag set is available, we do not update weights based on the loss for the absent task.


\subsection{Replicability and Reproducibility}
In order to facilitate the replicability and reproducibility of our results, we take two methodological steps.
To ensure replicability, we run all experiments 10 times, in order to mitigate the effect of random processes on our results.\footnote{Approximately 10,000 runs using 400,000 CPU hours.}
To ensure reproducibility, we release a collection including:
i) A Docker file containing all code and dependencies required to obtain all data and run our experiments used in this work;
and ii) a notebook containing all code for the statistical analyses performed in this work.\footnote{\url{https://github.com/bjerva/mtl-cond-entropy}}

\section{Results and Analysis}

\setlength{\tabcolsep}{2.1pt}
\begin{table}[htbp]
    \centering
    \footnotesize
    \caption{\label{tab:mtl_results}Morphosyntactic tasks. Correlation scores and associated $p$-values, between change in accuracy ($\Delta_{acc}$) and entropy ($H(Y)$), conditional entropy ($H(X|Y)$, $H(Y|X)$), and mutual information ($I(X;Y)$), calculated with Spearman's $\rho$, across all languages and label instantiations.
    Bold indicates the strongest significant correlations.}
    \begin{tabular}{lrrrr}
        \toprule
        {\bf Condition} & $\rho(\Delta_{acc}, H(Y))$ & $\rho(\Delta_{acc}, H(X|Y))$ & $\rho(\Delta_{acc}, H(Y|X))$ & $\rho(\Delta_{acc}, I(X;Y))$ \\
        \midrule
        Identity     & $-$0.06 (p$=$0.214)      & 0.10 (p$=$0.020)  & 0.12 (p$=$0.013)             & 0.08 (p$=$0.114) \\
        Overlap      & 0.07  (p$=$0.127)        & 0.23 (p$<$0.001)  & 0.27 (p$<$0.001)             & \bf{0.43 (p$\ll$0.001)} \\
        Disjoint     & 0.08  (p$=$0.101)        & 0.26 (p$<$0.001)  & 0.25 (p$<$0.001)             & \bf{0.41 (p$\ll$0.001)} \\
        \bottomrule
    \end{tabular}
    \vspace{0.5cm}
    \centering
    \footnotesize
    \setlength{\tabcolsep}{6pt}
    \caption{\label{tab:semantic_results}Change in accuracy, and information theoretic measures, for the semantic tasks.}
    \begin{tabular}{lrrrrr}
        \toprule
        {\bf Auxiliary task} & $\Delta_{acc}$ & $H(Y)$ & $H(X|Y)$ & $H(Y|X)$ &  $I(X;Y)$ \\
        \midrule
        Frames               & -14.64  & 1.6 & 2.7 & 1.4 & 0.2 \\
        MPQA                 & -5.62   & 1.1 & 2.6 & 1.0 & 0.1 \\
        Supersenses          & -2.86   & 1.8 & 2.7 & 1.6 & 0.2 \\
        NER                  & -1.36   & 0.8 & 2.6 & 0.7 & 0.1 \\
        Semtraits            &  0.67   & 1.3 & 3.0 & 0.8 & 0.5 \\
        Semtagging           &  0.79   & 3.0 & 2.0 & 1.5 & 1.5 \\
        \bottomrule
    \end{tabular}
\end{table}
\setlength{\tabcolsep}{6pt}

\begin{figure}[htbp]
    \centering
    \includegraphics[width=0.6\textwidth]{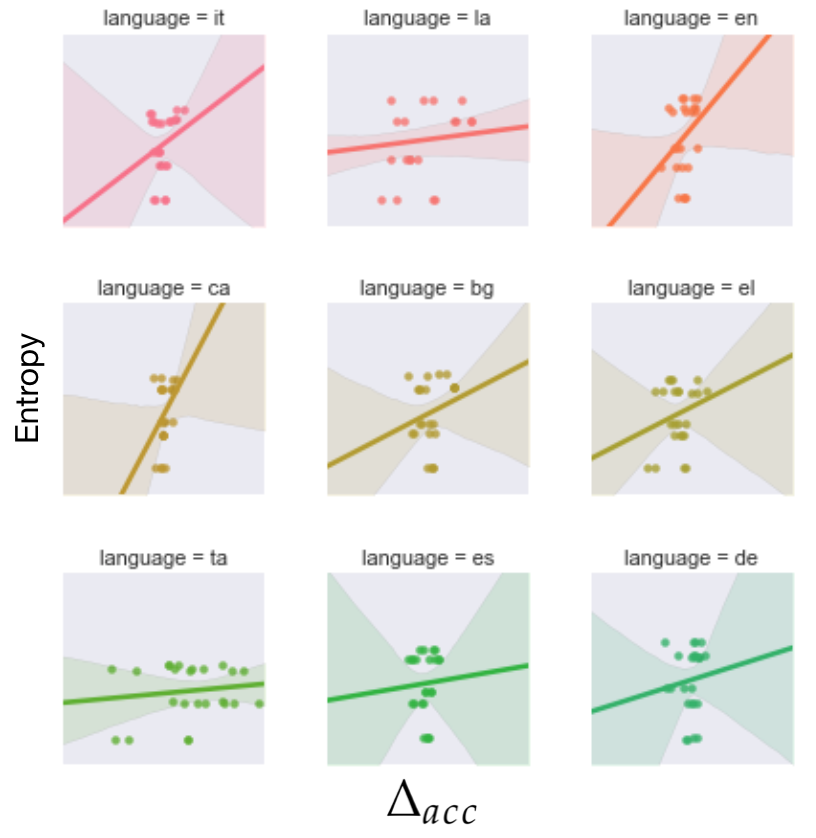}
    \caption{\label{fig:delta_entropy}Correlations between $\Delta_{acc}$ and entropy. Each data point represents a single experiment run.}
    \vspace{0.5cm}
    \includegraphics[width=0.6\textwidth]{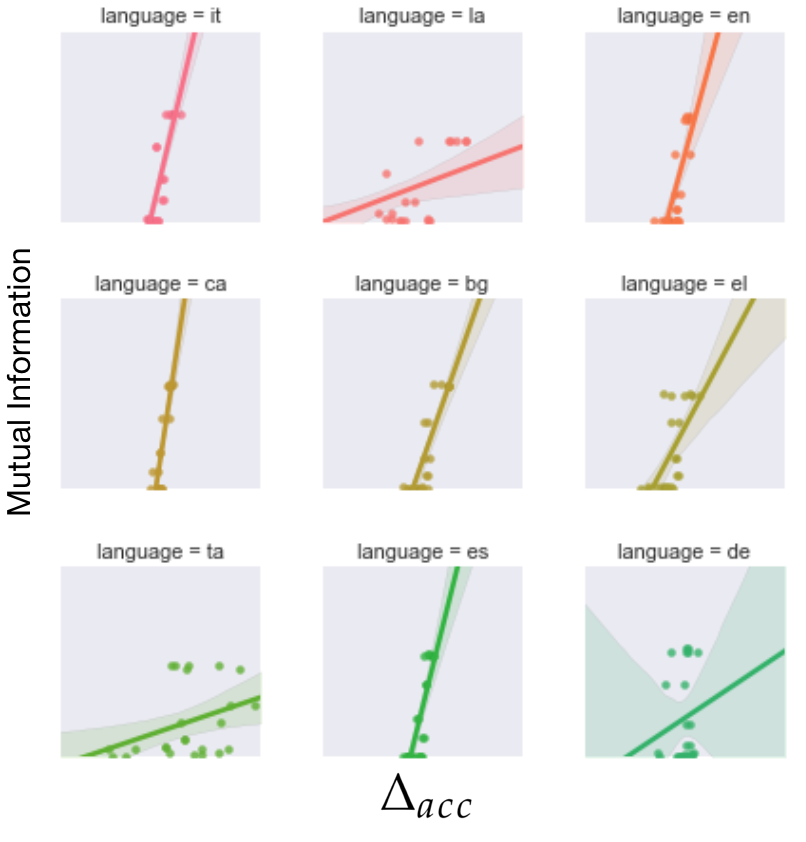}
    \caption{\label{fig:delta_mi}Correlations between $\Delta_{acc}$ and mutual information. Each data point represents a single experiment run.}
\end{figure}

\subsection{Morphosyntactic Tasks}
We use Spearman's $\rho$ in order to calculate correlation between auxiliary task effectivity (as measured using $\Delta_{acc}$) and the information-theoretic measures.
Following the recommendations in \citet{Sogaard:2014}, we set our $p$ cut-off value to $p<0.0025$.
Table~\ref{tab:mtl_results} shows that MI correlates significantly with auxiliary task effectivity in the most commonly used settings (overlap and disjoint).
The fact that no correlation is found in the identity condition between $\Delta_{acc}$ and any information-theoretic measure yields some interesting insights into the issue at hand.
Intuitively, this makes sense considering that the lack of any extra data in the identity setting does not offer much opportunity for the network to learn from the auxiliary data.
In other words, if the model is allowed to train on the same data with essentially the same label (i.e., if MI is high), this does not allow the model to learn anything new.
This is supported by the significant positive correlation between MI and $\Delta_{acc}$ in the overlap/disjoint conditions, in which the model does have access to more data.
This further suggests that in some cases, one of the most effective auxiliary tasks is simply more data for the same task (i.e.,\ the highest MI achievable).  
Additionally, this shows that high MI between tag sets in identical data is not necessarily helpful, and that in such a setting it may even be advantageous to have a less similar auxiliary task.

As hypothesised, entropy has no significant correlation with auxiliary task effectivity, whereas conditional entropy offers some explanation.
We further observe that these results hold for almost all languages, although the correlation is weaker for some languages, indicating that there are some other effects at play here.
For a sample of the languages, the correlations between $\Delta_{acc}$ and Entropy are shown in Figure~\ref{fig:delta_entropy}, and the correlations between $\Delta_{acc}$ and Mutual Information are shown in Figure~\ref{fig:delta_mi}.\footnote{These figures only show a subset of the languages under evaluation. See Appendix~\ref{app:langcorr} for a complete overview.}
We also analyse whether significant differences can be found with respect to whether or not we have a positive $\Delta_{acc}$, using a bootstrap sample test with 10,000 iterations \citep{efron:bootstrap}.
We observe a significant relationship ($p<0.001$) for MI.
We also observe a significant relationship for conditional entropy ($p<0.001$), and again find no significant difference for entropy ($p\geq0.07$).

\setlength{\tabcolsep}{4pt}
\begin{table}[htbp]
    \centering
    \footnotesize
    \caption{\label{tab:mtl_language_results}Correlation scores and associated $p$-values, between change in accuracy ($\Delta_{acc}$) and entropy ($H(Y)$), conditional entropy ($H(Y|X)$), and mutual information ($I(X;Y)$), calculated with Spearman's $\rho$.
    Bold indicates the strongest significant correlations per row.}
    \begin{tabular}{llrrl}
        \toprule
        {\bf Group} & {\bf Language} & $\rho(\Delta_{acc}, H(Y))$ & $\rho(\Delta_{acc}, H(Y|X))$ & $\rho(\Delta_{acc}, I(X;Y))$ \\
        \midrule
    \multirow{6}{*}{Germanic}
        & Danish         	& 0.27 (p$=$0.116) & 0.42 (p$=$0.011) & \bf{0.78 (p$\ll$0.001)} \\
        & Dutch          	& 0.31 (p$=$0.070) & 0.16 (p$=$0.337) & \bf{0.55 (p$<$0.001)} \\
        & English        	& 0.30 (p$=$0.076) & 0.19 (p$=$0.280) & \bf{0.58 (p$<$0.001)} \\
        & German         	& 0.03 (p$=$0.849) & 0.13 (p$=$0.448) & 0.18 (p$=$0.293) \\
        & Norwegian      	& -0.03 (p$=$0.858) & 0.23 (p$=$0.183) & 0.23 (p$=$0.177) \\
        & Swedish        	& -0.03 (p$=$0.843) & 0.29 (p$=$0.091) & 0.31 (p$=$0.068) \\
        \midrule

        \multirow{7}{*}{Romance}
        & Catalan        	& 0.34 (p$=$0.042) & 0.33 (p$=$0.047) & \bf{0.72 (p$\ll$0.001)} \\
        & French         	& 0.06 (p$=$0.734) & 0.38 (p$=$0.023) & 0.48 (p$=$0.003) \\
        & Galician       	& 0.10 (p$=$0.574) & 0.18 (p$=$0.304) & 0.28 (p$=$0.099) \\
        & Italian        	& 0.12 (p$=$0.503) & 0.52 (p$=$0.001) & \bf{0.67 (p$\ll$0.001)} \\
        & Portuguese     	& -0.02 (p$=$0.921) & 0.61 (p$<$0.001) & \bf{0.66 (p$<$0.001)} \\
        & Romanian       	& -0.31 (p$=$0.067) & 0.34 (p$=$0.040) & 0.04 (p$=$0.825) \\
        & Spanish        	& 0.02 (p$=$0.890) & 0.60 (p$<$0.001) & \bf{0.70 (p$\ll$0.001)} \\
        \midrule

        \multirow{7}{*}{Slavic}
        & Bulgarian      	& 0.20 (p$=$0.242) & 0.50 (p$=$0.002) & \bf{0.76 (p$\ll$0.001)} \\
        & Croatian       	& -0.24 (p$=$0.159) & 0.43 (p$=$0.009) & 0.22 (p$=$0.189) \\
        & Czech          	& -0.15 (p$=$0.376) & 0.49 (p$=$0.002) & 0.39 (p$=$0.017) \\
        & O.C. Slavonic	    & -0.08 (p$=$0.634) & 0.34 (p$=$0.044) & 0.35 (p$=$0.038) \\
        & Polish         	& 0.13 (p$=$0.437) & 0.40 (p$=$0.015) & \bf{0.59 (p$<$0.001)} \\
        & Russian        	& 0.29 (p$=$0.086) & 0.40 (p$=$0.015) & \bf{0.81 (p$\ll$0.001)} \\
        & Slovene        	& -0.24 (p$=$0.156) & 0.41 (p$=$0.014) & 0.19 (p$=$0.259) \\
        \midrule

        \multirow{2}{*}{Turkic}
        & Kazakh         	& 0.23 (p$=$0.172) & 0.04 (p$=$0.817) & 0.36 (p$=$0.030) \\
        & Turkish        	& 0.50 (p$=$0.002) & -0.17 (p$=$0.317) & 0.43 (p$=$0.008) \\
        \midrule

        \multirow{3}{*}{Uralic}
        & Estonian       	& 0.45 (p$=$0.006) & -0.14 (p$=$0.430) & 0.39 (p$=$0.017) \\
        & Finnish        	& 0.02 (p$=$0.924) & 0.37 (p$=$0.025) & \bf{0.50 (p$=$0.002)} \\
        & Hungarian      	& 0.14 (p$=$0.413) & 0.09 (p$=$0.594) & 0.27 (p$=$0.116) \\
        \midrule

        \multirow{12}{*}{Other}
        & Arabic         	& -0.16 (p$=$0.362) & 0.53 (p$<$0.001) & 0.47 (p$=$0.004) \\
        & Basque         	& 0.41 (p$=$0.014) & -0.01 (p$=$0.952) & \bf{0.49 (p$=$0.002)} \\
        & Chinese        	& -0.15 (p$=$0.399) & 0.46 (p$=$0.005) & 0.41 (p$=$0.012) \\
        & Farsi          	& 0.20 (p$=$0.244) & 0.41 (p$=$0.012) & \bf{0.75 (p$\ll$0.001)} \\
        & Greek          	& 0.20 (p$=$0.248) & 0.19 (p$=$0.264) & 0.44 (p$=$0.007) \\
        & Hebrew         	& 0.06 (p$=$0.724) & 0.37 (p$=$0.028) & \bf{0.52 (p$=$0.001)} \\
        & Hindi          	& -0.26 (p$=$0.121) & 0.24 (p$=$0.161) & 0.00 (p$=$0.979) \\
        & Irish          	& -0.24 (p$=$0.150) & 0.54 (p$<$0.001) & 0.35 (p$=$0.034) \\
        & Indonesian     	& -0.42 (p$=$0.011) & 0.51 (p$=$0.001) & 0.11 (p$=$0.510) \\
        & Latin          	& 0.19 (p$=$0.271) & 0.16 (p$=$0.362) & 0.47 (p$=$0.004) \\
        & Latvian        	& 0.64 (p$<$0.001) & -0.23 (p$=$0.171) & \bf{0.53 (p$<$0.001)} \\
        & Tamil          	& 0.16 (p$=$0.337) & 0.12 (p$=$0.482) & 0.31 (p$=$0.067) \\
        \bottomrule
    \end{tabular}
\end{table}
\setlength{\tabcolsep}{6pt}

\subsection{Language-dependent results}
Results per language are shown in Table~\ref{tab:mtl_language_results}.
While not all languages exhibit correlations below our selected $\alpha$-level, the non-significant languages still exhibit interesting trends in the same direction.
Note that there are two cases where Entropy is a fair predictor of $\Delta_{acc}$, namely for Latvian and Turkish.
However, in both of these cases the correlation is stronger still with MI.
Furthermore, the correlations between $\Delta_{acc}$ and entropy vary wildly between languages, sometimes exhibiting negative correlations.

\subsection{Semantic Tasks}
\begin{figure}[htbp]
    \centering
    \includegraphics[width=\textwidth]{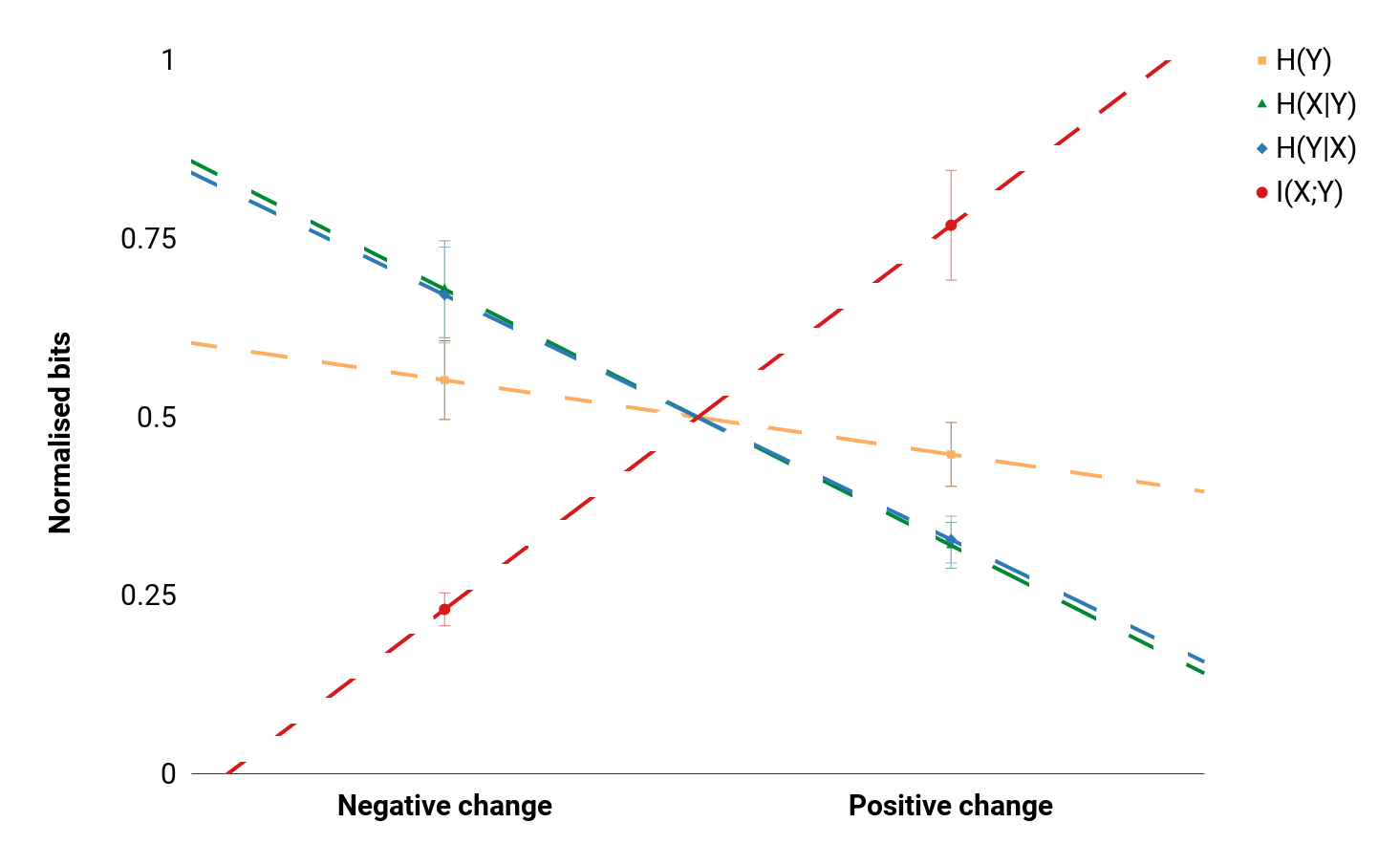}
    \caption{\label{fig:semantic_measures}Comparison of information theoretic measures and change in accuracy for the semantic tasks.
    Results are grouped by negative and positive change in $\Delta_{acc}$. Dashed lines are included for clarity, and are not meant to imply a linear correlation.}
\end{figure}

We do not have access to sufficient data points to run statistical analyses on the results obtained by \citet{alonso:mtl}, or by \citet{bjerva:2016:semantic}, and the results in Table~\ref{tab:semantic_results} do not reveal any obvious patterns.
A grouping of these results by whether or not $\Delta_{acc}$ was positive can be seen in Figure~\ref{fig:semantic_measures}, which offers some support to the results from the morphosyntactic tasks.
However, the lack of a clear pattern when looking at individual results per dataset serves to highlight the issue at hand, namely that even though MI offers some explanatory value, the interactions behind the workings of MTL are more complex than what can be explained purely by comparing joint distributions of tag sets.

\section{Conclusions}

We have examined the relation between auxiliary task effectivity and three information-theoretic measures.
The first research question which we aimed to answer in this chapter ({\bf RQ~\ref{rq:mtl}a}) was related to which information-theoretic measures are useful for estimation of auxiliary task effectivity.
While previous research hypothesises that entropy plays a central role, we show experimentally that this is not sufficient, and that conditional entropy is a somewhat better predictor, and that MI is the best predictor under consideration here.
This claim is corroborated when we correlate MI and change in accuracy with results found in the literature.
It is especially interesting that MI is a better predictor than conditional entropy, since MI is a symmetric measure, as it does not consider the order between main and auxiliary tasks.
For conditional entropy itself, the results for the two directionalities did not differ to a large extent.
Our findings should prove helpful for researchers when considering which auxiliary tasks might be helpful for a given main task.
Furthermore, it provides an explanation for the fact that there is no universally effective auxiliary task, as a purely entropy-based hypothesis assumes.

The fact that MI is informative when determining the effectivity of an auxiliary task can be explained by considering an auxiliary task to be similar to adding a feature.
That is to say, useful features are likely to be useful auxiliary tasks.
Interestingly, however, the gains of adding an auxiliary task are visible at test time for the main task, when no explicit auxiliary label information is available.

The second research question which we aimed to answer in this chapter ({\bf RQ~\ref{rq:mtl}b}), related to the generalisation ability of the information-theoretic measures as a measure of auxiliary task effectivity, across languages and NLP tasks.
We tested our hypothesis on 39 languages, representing a wide typological range, as well as a wide range of data sizes.
Our experiments were run on syntactically oriented tasks of various granularities.
We also corroborated our findings with results from semantically oriented tasks in the literature.

While the correlations between MI and $\Delta_{acc}$ were higher than for other information-theoretic measures, it is by no means a perfect predictor.
This highlights the fact that the interactions between tasks is more complex than simply the joint distribution between the tag sets at hand.
One possibility for future work is to take distributions over tags and words into account simultaneously.

Having considered interactions between tasks in MTL, and finding that task similarity is to some extent predictive of MTL effectivity, this raises the question of what the situation is in the case of multilingual learning.
This is explored further in the next part of this thesis.

\partimage[width=0.4\textwidth,center]{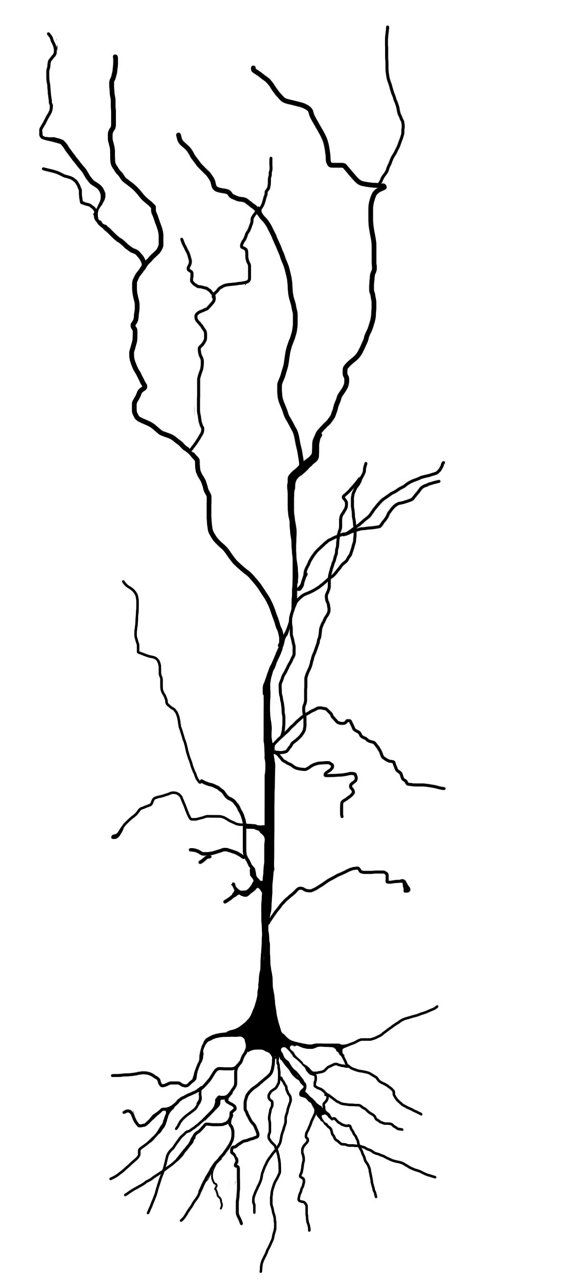}
\part{Multilingual Learning}
\renewcommand*{\thefootnote}{\fnsymbol{footnote}}
\chapter[Multilingual Semantic Textual Similarity]{\footnote{Chapter adapted from:
\textbf{Bjerva, J.} and \"Ostling, R. (2017a). Cross-lingual Learning of Semantic Textual Similarity with Multilingual Word Representations. In Proceedings of the 21st Nordic Conference on Computational Linguistics, NoDaLiDa, 22-24 May 2017, Gothenburg, Sweden, number 131, pages 211–215. Link\"oping University Electronic Press, Link\"opings universitet.}
\hspace{-6pt}Multilingual\\Semantic Textual Similarity\hspace{-8pt}}
\renewcommand*{\thefootnote}{\arabic{footnote}}
\label{chp:multiling_sim}
\begin{abstract}
\absprelude
Up until now, we have seen parameter sharing between tasks, in the context of multitask learning.
Another possibility is to share parameters between languages, in a sense casting model multilinguality as a type of multitask learning.
As a first example of multilingual learning, we look at cross-lingual semantic textual similarity.
This task can be approached by leveraging multilingual distributional word representations, in which similar words in different languages are close to each other in semantic space.
The availability of parallel data allows us to train such representations for a large number of languages.
Such representations have the added advantage of allowing for leveraging semantic similarity data for languages for which no such data exists.
In this chapter, the focus is on to what extent such an approach allows for enabling zero-shot learning for the task at hand.
We also investigate whether language relatedness has an effect on how successful this is.
We train and evaluate on six language pairs for semantic textual similarity, including English, Spanish, Arabic, and Turkish.
\end{abstract}

\clearpage
\section{Introduction}
In order to determine how similar a pair of sentences in two different languages are to one another, it is necessary to have a grasp of multilingual semantics.\footnote{The exception to this is methods based on machine translation, which are outlined in the following section.}
Continuous space word representations, which lend their strength from distributional semantics, are a clear candidate for this problem, as distances in such a space can be directly interpreted as semantic similarities (see Chapter~\ref{chp:mtl_bg}).
Given the constantly increasing amount of available parallel data (e.g., \citet{europarl},~\citet{opus},~\citet{uncorp}), it is possible to learn multilingual word representations for many languages.
In this chapter, we approach tasks on cross-lingual semantic textual similarity from SemEval-2016 \citep{semeval16} and SemEval-2017 \citep{semeval17} using such word representations.
This approach has the advantage that it allows for zero-shot learning while training on multiple languages, i.e., exploiting data from several source languages for an unseen target language.
We aim to answer the following specified research question, in order to answer {\bf RQ~\ref{rq:sts}}:
\begin{enumerate}
    \setlength{\itemindent}{.29cm}
    \item [{\bf RQ~\ref{rq:sts}a}] To what extent can multilingual word representations be used in a simple STS system so as to enable zero-shot learning for unseen languages?
    \item [{\bf RQ~\ref{rq:sts}b}] To what extent is the success of zero-shot learning dependent of language relatedness in this setting?
\end{enumerate}

\subsection*{Related work}
\label{sec:cross_sts}
Semantic Textual Similarity (STS) is the task of assessing the degree to which two sentences are similar in their meanings.
In the long-running SemEval STS shared task series, this is measured on a scale ranging from 0, indicating no semantic similarity, to 5, indicating complete semantic similarity (see \citeauthor{semeval17}, \citeyear{semeval12,semeval13,semeval14,semeval15,semeval16,semeval17}).
Monolingual STS is an important task, for instance for evaluation of machine translation (MT) systems, where estimating the semantic similarity between a system's translation and the target translation can aid both system evaluation and development.
STS is also a fundamental problem for natural language understanding, as being able to estimate the similarity between two sentences in their meaning content can be seen as a prerequisite for understanding.
The task is already a challenging one in a monolingual setting, such as when estimating the similarity between two English sentences.
In this chapter, we tackle the more difficult case of cross-lingual STS, e.g., estimating the similarity between an English and an Arabic sentence, in the context of shared tasks on cross-lingual STS at SemEval-2016 \citep{semeval16} and SemEval-2017 \citep{semeval17}.\footnote{SemEval-2016 Task 1: Semantic Textual Similarity: A Unified Framework for Semantic Processing and Evaluation -- Cross-lingual STS Subtask.}\footnote{SemEval-2017 Task 1: Semantic Textual Similarity.}

Previous approaches to the problem of cross-lingual STS have focussed on two main approaches.
The primary, and most successful approach, is to apply a MT system to non-English sentences, and translating these to English (e.g., \citeauthor{tian:2017}, \citeyear{tian:2017}, and \citeauthor{wu:2017}, \citeyear{wu:2017}).
The advantage of this approach is that the problem essentially boils down to estimating the similarity of two sentences in English.
There are at least two advantages to this.
First of all, the amount of resources available for English eclipse what is available for most, if not all, other languages.
Additionally, comparing two sentences in the same language allows for straight-forward application of features based on the surface forms of the sentences, such as word overlap, common substrings, and so on.
MT approaches tend to outperform purely multilingual approaches, with the winner of SemEval-2017, and many of the top systems of SemEval-2016 relying on this approach \citep{semeval16,semeval17}.
There are at least two drawbacks to this method, however.
Primarily, involving a fully-fledged MT system severely increases the complexity of a system.
Furthermore, such methods can be seen as bypassing the problem of cross-lingual STS, rather than tackling it directly, as no actual multilingual similarity assessments are necessarily carried out.

The amount of true multilingual approaches in the literature are somewhat more limited, with notable examples from SemEval-2016 such as \citet{cnrc:2016}, who make use of bilingual embedding space phrase similarities, in combination with cross-lingual machine translation metrics.
Another approach is represented by \citet{aldarmaki:2016}, who apply bilingual word representations in a matrix factorisation method, so as to assess STS without translation.
The method that bears the most resemblance to the approach taken in this chapter is \citet{ataman:2016}, who combine bilingual embeddings with machine translation quality estimation features \citep{quest}.
We expand upon this method by using \textit{multilingual} word embeddings as input, rather than bilingual ones.
One advantage of our approach, is that it allows for zero-shot learning while training on multiple languages (see Chapter~\ref{chp:mtl_bg}), as it does not depend on annotated STS training data for the target language, and only places requirements on the availability of parallel data.
This denotes the approach we take to {\bf RQ~\ref{rq:sts}a}, as well as {\bf RQ~\ref{rq:sts}b}.
Additionally, our method differs from \citet{ataman:2016} in that we choose a simpler architecture, using only such word representations as input.

\section{Cross-lingual Semantic Textual Similarity}

We will now look at the task of (cross-lingual) STS in more detail.
Given two sentences, $s_1$ and $s_2$, the task in STS is to assess how semantically similar these are to each other.
This is commonly measured using a scale ranging from 0--5, with 0 indicating no semantic overlap, and 5 indicating nearly identical content.
In the SemEval STS shared tasks, the following descriptions are used:
\begin{enumerate}
    \setcounter{enumi}{-1}
    \item The two sentences are completely dissimilar.
    \item The two sentences are not equivalent, but are on the same topic.
    \item The two sentences are not equivalent, but share some details.
    \item The two sentences are roughly equivalent, but some important information differs/missing.
    \item The two sentences are mostly equivalent, but some unimportant details differ.
    \item The two sentences are completely equivalent, as they mean the same thing.
\end{enumerate}

\setlength{\tabcolsep}{4pt}
\begin{table}[hptb]
    \centering
    \footnotesize
    \caption{\label{tab:sent_sim_examples}Examples of sentence similarities and corresponding entailment judgements.}
    \begin{tabular}{rlrl}
        \toprule
        \textbf{No.} & \textbf{Text / Hypothesis}      & \textbf{Score} & \textbf{Relation} \\
        \midrule
        \multirow{2}{*}{8678} & A skateboarder is jumping off a ramp	  & \multirow{2}{*}{4.8} & \multirow{2}{*}{entailment} \\
                              & A skateboarder is making a jump off a ramp	&  &  \\
        \midrule
        \multirow{2}{*}{2709} & There is no person boiling noodles        & \multirow{2}{*}{2.9} & \multirow{2}{*}{contradiction} \\
                              & A woman is boiling noodles in water          &  &  \\
        \midrule
        \multirow{2}{*}{219}  & There is no girl in white dancing         & \multirow{2}{*}{4.2} & \multirow{2}{*}{contradiction} \\
                              & A girl in white is dancing                   &  &  \\
        \bottomrule
    \end{tabular}
    \vspace{0.5cm}
\setlength{\tabcolsep}{2pt}
    \centering
    \footnotesize
    \caption{\label{tab:sent_sim_examples_multiling}Examples of cross-lingual sentence similarities.}
    \begin{tabular}{llr}
        \toprule
        \textbf{English / Spanish}  & \textbf{Score} \\
        \midrule
        The NATO mission officially ended Oct. 31.  &   \multirow{2}{*}{5} \\
        La misión de la OTAN terminó oficialmente oct. 31. & \\
        \midrule
        Mass Slaughter on a Personal Level  &   \multirow{2}{*}{3} \\
        El sacrificio masivo en un nivel personal & \\
        \midrule
        Support Workers' Union Will Sue City Over Layoffs    &  \multirow{2}{*}{1} \\
        Apoyo a los trabajadores "Unión va a demandar ciudad más despidos &  \\
        \bottomrule
\end{tabular}
\end{table}
\setlength{\tabcolsep}{6pt}

\noindent As an example of sentence similarities, consider the sentence pairs and their human-annotated similarity scores in Table~\ref{tab:sent_sim_examples}.
These examples are taken from the SemEval-2014 edition of the shared task on STS and Recognising Textual Entailment (RTE), giving us access to entailment information in addition to similarity scores for the purposes of the example \citep{semeval:2014}.\footnote{RTE is the task of assessing whether the meaning of one sentence (the \textit{hypothesis}) can be inferred from the other (the \textit{text}).}
Attempting to assess the semantic content of two sentences with a simple score notably does not take important semantic features such as negation into account, and STS can therefore be seen as complimentary to textual entailment.
For instance, in sentence No.\ 219 in Table~\ref{tab:sent_sim_examples}, the sentences have a high similarity score, even though their meanings are the opposite of one another.
It is also worth to note that STS is highly related to paraphrasing, as replacing an $n$-gram with a paraphrase thereof ought to alter the semantic similarity of two sentences to a very low degree.

Successful monolingual approaches in the past have taken advantage of both the relatedness with this task to paraphrasing, and to RTE.
\citet{bjerva:14:semeval} attempt to replace words in $s_1$ with paraphrases obtained from the Paraphrase Database (PPDB, \citeauthor{ppdb}, \citeyear{ppdb}), in order to increase the surface similarity with $s_2$.
Additionally, both \citet{bjerva:14:semeval} and \citet{beltagy:2016} make use of (features from) an RTE system to perform the task of STS.
Approaches similar to these can be applied in cross-lingual STS, if the sentence pair is translated to a language for which such resources exist.

As an example of sentence similarities in cross-lingual STS, consider the sentence pairs and their human-annotated similarity scores in Table~\ref{tab:sent_sim_examples_multiling}.
The first example contains a Spanish sentence which is a faithful translation of the English one, and has the highest similarity score (5).
Although the second example conveys a similar meaning, the translation expresses 'Mass Slaughter' in Spanish as 'A massive sacrifice', resulting in a lower similarity score (3), indicating a loose translation.\footnote{See \citet{bos:2014} for a further discussion of faithful, informative, and loose translations in the context of parallel corpora.}
In the third example sentence, the Spanish sentence conveys the opposite meaning of the English sentence, and has the lowest similarity score (1).

\section{Method}
As mentioned in Section~\ref{sec:cross_sts}, we approach the task of multilingual STS in a similar manner to \citet{ataman:2016}, with the addition that we use multilingual input representations, rather than bilingual ones.
We will now look at how our system is constructed, starting with the input representations.

\subsection{Multilingual word representations}
There are several methods available for obtaining multilingual word representations, as described in Chapter~\ref{chp:mtl_bg}.
In this chapter, we use a variant of the multilingual skip-gram method \citep{multisg}, as detailed in Chapter~\ref{chp:mtl_bg}.
This method was chosen as it is both relatively simple, and yields high-quality representations for down-stream tasks, as compared to other approaches \citep{multisg}.
The original method relies on using cross-lingual contexts, with English as a pivot language.
For instance, a Spanish word might be used to predict an English context, or the other way around.
Our approach differs in that we augment the learning objective so as to include multilingual contexts, such that we also use, for instance, a Spanish word to predict a French word (Figure~\ref{fig:multi_sg} in Chapter~\ref{chp:mtl_bg}).

We train 100-dimensional multilingual embeddings on the Europarl \citep{europarl} and UN corpora \citep{uncorp}, including data from bible translations.\footnote{Using the New Testament (approximately 140,000 tokens), available at \mbox{\url{http://homepages.inf.ed.ac.uk/s0787820/bible/}}.}\footnote{Training multilingual embeddings on this data yields a vocabulary coverage of over 85\% on the development sets of the languages at hand.}
This data was chosen partially since it allows us to learn such embeddings for a large number of languages, in addition to the availability of these corpora.
The dimensionality of the embeddings was chosen by balancing a sufficiently high number of dimensions with the computational resources necessary to compute these embeddings with the extended version of the multilingual skip-gram method.
Word alignment, which is required for the training of this type of multilingual embeddings, is performed using a tool based on the Efmaral word-alignment tool \citep{efmaral}.\footnote{We use the \textit{eflomal} tool, which uses less memory than \textit{efmaral}. Default parameters are used. Available at \url{https://github.com/robertostling/eflomal}.}
This allows us to extract a large amount of multilingual (word, context) pairs.
We then use these pairs in order to learn multilingual embeddings, by applying the \textit{word2vecf} tool \citep{word2vecf}.
In our experiments, we use the same parameter settings as \citet{multisg}, training using negative sampling \citep{Mikolov:13:Models}, and with equal weighting of monolingual and cross-lingual contexts.\footnote{Note that we do not use the same implementation as \citep{multisg}.}

\subsection{System architecture}

\begin{sidewaysfigure}
    \centering
    \includegraphics[width=\textwidth]{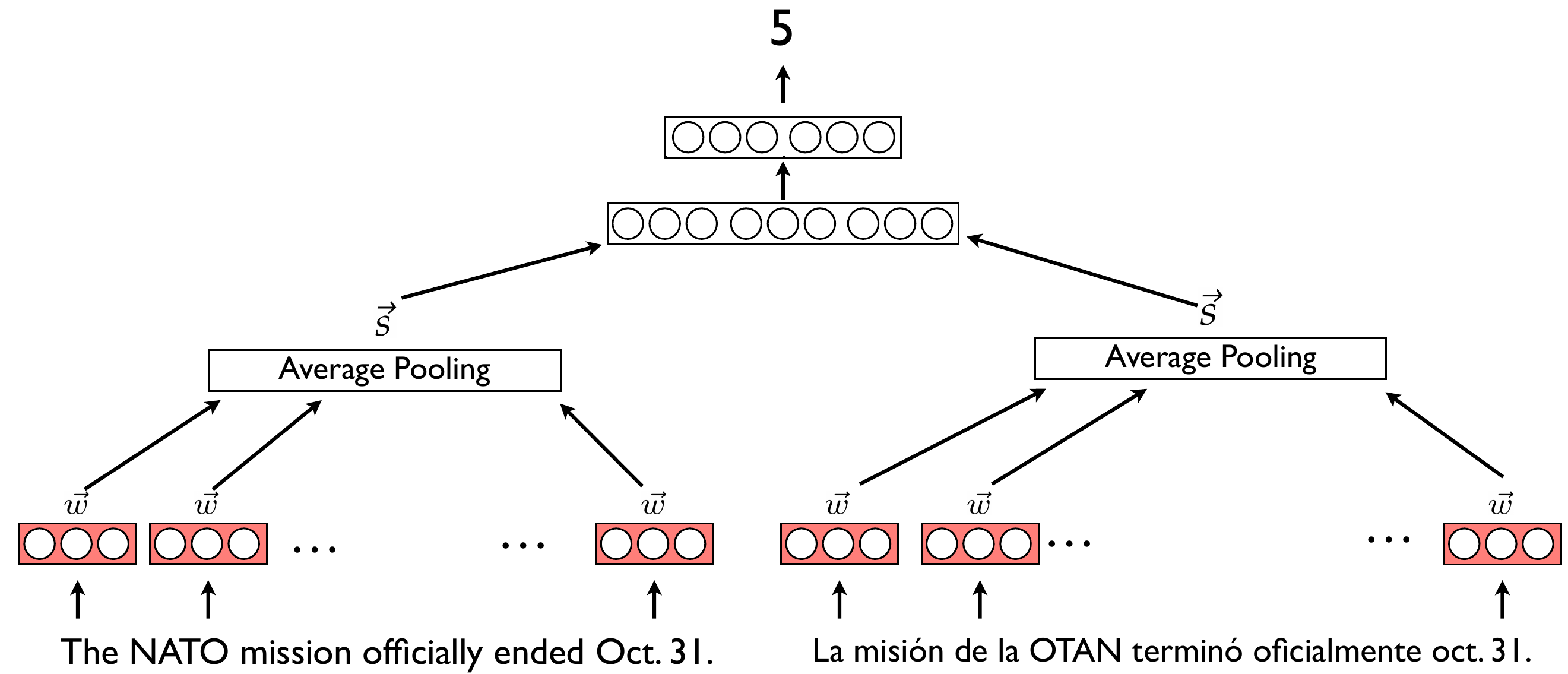}
    \caption{\label{fig:sts_sys_arch}System architecture used in the semantic textual similarity task.}
\end{sidewaysfigure}

We use a relatively simple neural network architecture, consisting of an input layer with pre-trained word embeddings and a network of fully connected layers.
This means that we need a sentence-level representation, based on the multilingual word representations, offering us a choice between methods such as those presented in Chapter~\ref{chp:nn}, Section~\ref{sec:rnn_use_cases}.
Given 100-dimensional word representations for each word in our sentence, we opt for the simplistic approach of averaging the vectors across each sentence, such that
\begin{equation}
    \begin{aligned}
        \vec{s} &= \frac{1}{|s|}\sum_{w\in s} \vec{w}, \\
    \end{aligned}
\end{equation}
where $w$ is a word in the sentence $s$, and $\vec{w}$ and $\vec{s}$ are their vectorial representations.
This is the same approach that is taken by \citet{ataman:2016}.
The resulting sentence-level representations are then concatenated and passed through two fully connected layers with ReLU activation functions ($200$ and $100$ units, respectively), prior to the output layer.
In order to prevent any shift from occurring in the embeddings, we do not update these during training.
The intuition here is that we do not want the representation for, e.g., \textit{dog} to be updated, which might push it further away from that of \textit{perro}.
We expect this to be especially important in cases where we train on a single language, and evaluate on another.
The system architecture is depicted in Figure~\ref{fig:sts_sys_arch}.

We apply dropout ($p=0.5$) between each layer \citep{dropout}.
All weights are initialised using the approach from \citet{glorot:init}.
We use the Adam optimisation algorithm \citep{adam}, monitoring the categorical cross-entropy of the sentence similarity score, while sanity-checking against the scores obtained as measured with Pearson correlation.
All systems are trained using a batch size of 40 sentence pairs, over a maximum of 50 epochs, using early stopping monitoring the loss on the validation set.
We report results using the model with the lowest validation loss.
Hyperparameters are kept constant in all conditions.

\subsection{Data for Semantic Textual Similarity}

As the SemEval STS task series has been running for several years, there is a substantial amount of data available.
We use data from previous editions of the tasks on (cross-lingual) STS.\footnote{This is what is generally recommended by the shared task organisers, and is followed by most participating systems.}
For English--English this includes data from SemEval 2012 through 2015 \citep{semeval12,semeval13,semeval14,semeval15}.
For English--Spanish this includes data from SemEval 2016 and 2017 \citep{semeval16,semeval17}.
For Spanish--Spanish this includes data from SemEval 2014 and 2015.
For English--Arabic this includes data from SemEval 2017.
Finally, for Arabic--Arabic, this includes data from SemEval 2017.
We use the concatenation of the training sets of previous editions for training, and validate and test on the most recent data for each language pair.
An overview of the available data is shown in Table~\ref{tab:data:sts}.

\begin{table}[htbp]
    \centering
    \caption{Training data used for (cross-lingual) STS from the SemEval shared task series.}
    \label{tab:data:sts}
    \begin{tabular}{lrr}
        \toprule
        \textbf{Language pair} & \textbf{N sentence pairs} & \textbf{SemEval edition(s)}\\
        \midrule
        English -- English   & 3,000 & 2012 -- 2015 \\
        English -- Spanish   & 3,900 & 2016 -- 2017 \\
        Spanish -- Spanish   & 1,500 & 2014 -- 2015 \\
        English -- Arabic    & 2,000 & 2017 \\
        Arabic \ \ -- Arabic & 900   & 2017 \\
        \bottomrule
    \end{tabular}
\end{table}

\section{Experiments and Results}
We investigate whether using a multilingual input representation and shared weights allow us to ignore languages in STS, after mapping words to their multilingual representations.
This, in turn, is one approach for enabling zero-shot learning for this task ({\bf RQ~\ref{rq:sts}a}).
We first train and evaluate single-source trained systems (i.e.\ on a single language pair), and evaluate this both using the same language pair as target, and on all other target language pairs.
In doing so, we investigate the extent to which the availability of parallel data allows us to train STS systems without access to STS training data for a given language.

Secondly, we investigate the effect of bundling training data together, in multi-source training, investigating which language pairings are helpful for each other.
Concretely, in single-source training, we only train on one out of the language pairs at a time, and evaluate the resulting single-source model on all language pairs.
In multi-source training, however, a model is trained on several language pairs at a time, and the resulting model is evaluated as in the single-source training setting.
This is done so as to offer insight into {\bf RQ~\ref{rq:sts}b}.

We measure performance between target similarities and system output using the Pearson correlation measure, as this is standard in the SemEval STS shared tasks.
We first present results on the development sets, and finally the official shared task evaluation results.

\subsection{Comparison with Monolingual Representations}
As a baseline, we compare multilingual embeddings with the performance obtained using the pre-trained monolingual Polyglot embeddings \citep{polyglot}.
Training and evaluating on the same language pair yields comparable results regardless of embeddings (Table~\ref{tab:monoling_comparison}).
This shows that our multilingual embeddings, at the very least, have comparable quality to purely monolingual embeddings in a monolingual setting.

\begin{table}[htbp]
    \centering
    \caption{Single-source training results (Pearson correlations) with monolingual embeddings (polyglot) as compared to multilingual embeddings (multilingual skipgram) on the SemEval-2017 development set. Rows indicate evaluation language, and rows indicate the embeddings used. Bold numbers indicate best results per row.}
    \label{tab:monoling_comparison}
    \begin{tabular}{lrr}
        \toprule
         & \textbf{Polyglot} & \textbf{Multilingual Skipgram}  \\
        \midrule
        \textbf{English}             & 0.68  & \textbf{0.69} \\
        \textbf{Spanish}             & 0.65  & 0.65          \\
        \textbf{Arabic}              & 0.70  & \textbf{0.71} \\
        \bottomrule
    \end{tabular}
\end{table}

\subsection{Single-source training}

\begin{table}[htbp]
    \centering
    \caption{Single-source training results with multilingual embeddings on the SemEval-2017 development set (Pearson correlations). Columns indicate training language pairs, and rows indicate testing language pairs. Bold numbers indicate best results per row.}
    \label{tab:single_source}
    \begin{tabular}{lrrrrr}
        \toprule
        \backslashbox{\small\textbf{Test}}{\small\textbf{Train}}  & \textbf{en--en} & \textbf{en--es} & \textbf{en--ar} & \textbf{es--es} & \textbf{ar--ar} \\
        \midrule
        \textbf{en--en}                                       & \textbf{0.69}  & 0.07           & -0.04           & 0.64           & 0.54           \\
        \textbf{en--es}                                       & 0.19           & \textbf{0.27}  & 0.00            & 0.18           & -0.04          \\
        \textbf{en--ar}                                       & -0.44          & 0.37           & \textbf{0.73}   & -0.10          & 0.62           \\
        \textbf{es--es}                                       & 0.61           & 0.07           & 0.12            & \textbf{0.65}  & 0.50           \\
        \textbf{ar--ar}                                       & 0.59           & 0.52           & \textbf{0.73}   & 0.59           & 0.71           \\
        \bottomrule
    \end{tabular}
\end{table}

\noindent Results when training on a single source corpus, using multilingual embeddings, are shown in Table~\ref{tab:single_source}.
Training on the target language pair generally yields the highest results, except for one case.
When evaluating on Arabic--Arabic sentence pairs, training on English--Arabic texts yields comparable, or slightly better, performance than when training on Arabic--Arabic.
Observing the results from a zero-shot learning perspective, it seems that certain language combinations can benefit from this approach.
Mainly, it seems to be the case that this approach is suitable when training on a monolingual source pair (such as English--English), and evaluating the model on a monolingual target pair (such as Spanish--Spanish).
The gap in performance between such cases, and a system where target and source languages are identical is relatively small.
One example of this is the results of evaluating on English--English when training on English--English ($0.69$) as compared to when training on Spanish--Spanish ($0.64$).
We can also observe that zero-shot learning in this setting is more successful between the Indo-European languages Spanish and English, than when involving the Semitic language Arabic.

\subsection{Multi-source training}

We combine training sets from two language pairs in order to investigate how this affects evaluation performance on target language pairs.
We copy the single-source setup, except for that we also add in the data belonging to the source-pair at hand, e.g., training on both English--Arabic and Arabic--Arabic when evaluating on Arabic--Arabic (see Table~\ref{tab:single_source_add}).

\begin{table}[htbp]
    \centering
    \caption{Results with one source language in addition to target-language data with multilingual embeddings on the SemEval-2017 development set (Pearson correlations). Columns indicate added source language pairs, and rows indicate target language pairs. Bold numbers indicate best results per row.}
    \label{tab:single_source_add}
    \begin{tabular}{lrrrrr}
        \toprule
        \backslashbox{\small\textbf{Test}}{\small\textbf{Train}}  & \textbf{en--en} & \textbf{en--es} & \textbf{en--ar} & \textbf{es--es} & \textbf{ar--ar} \\
        \midrule
        \textbf{en--en}                                       & 0.69          &  0.68          &  0.67           & 0.69           & \textbf{0.71}     \\
        \textbf{en--es}                                       & 0.22          &  0.27          & \textbf{0.30}   & 0.22           &  0.24             \\
        \textbf{en--ar}                                       & 0.72          &  0.72          & \textbf{0.73}   & 0.71           &  0.72             \\
        \textbf{es--es}                                       & 0.63          &  0.60          &  0.63           & 0.65           &  \textbf{0.66}    \\
        \textbf{ar--ar}                                       & 0.71          &  0.72          & \textbf{0.75}   &  0.70          &  0.71             \\
        \bottomrule
    \end{tabular}
\end{table}

We observe that the monolingual language pairings (English--English, Spanish--Spanish, Arabic--Arabic) appear to be beneficial for one another, also in this setting.
Interestingly, adding Arabic--Arabic training data seems to improve the performance on both English--English and Spanish--Spanish.
Although, this difference is small and likely not significant, it is interesting that the models performance does not worsen as an effect of adding this data.
This might have been the case, if model capacity is wasted on solving the task for Arabic--Arabic.

Finally, we run a multilingual ablation experiment, in which we train on two out of three of these language pairs, and evaluate on all three.
Notably, excluding all Spanish training data yields comparable performance to including it (Table~\ref{tab:ablation}).
Additionally, we see the largest drop in performance when evaluating on Arabic data without having trained on it.
This adds further support to the findings in Table~\ref{tab:single_source}, indicating that language relatedness is of importance for the success of zero-shot learning as applied here.

\begin{table}[htbp]
    \centering
    \caption{Multilingual ablation results with multilingual embeddings on the SemEval-2017 development set (Pearson correlations). Columns indicate \textit{ablated} language pairs, and rows indicate testing language pairs. The \textit{none} column indicates no ablation, i.e., training on all three monolingual pairs.
    The bold diagonal indicates results when the target language pair is not used as a source language pair.}
    \label{tab:ablation}
    \begin{tabular}{lrrrr}
        \toprule
        \backslashbox{\small\textbf{Test}}{\small\textbf{Ablated}}  & \textbf{en--en} & \textbf{es--es} & \textbf{ar--ar}   & \textbf{none} \\
        \midrule
        \textbf{en--en}                                       & \textbf{0.60}          &  0.69            & 0.69          & 0.65   \\
        \textbf{es--es}                                       & 0.64                   &  \textbf{0.64}   & 0.67          & 0.60   \\
        \textbf{ar--ar}                                       & 0.68                   &  0.66            & \textbf{0.58} & 0.72   \\
        \bottomrule
    \end{tabular}
\end{table}

\setlength{\tabcolsep}{2pt}
\begin{table}[htbp]
\centering
\caption{Results with multilingual embeddings on SemEval-2017 Shared Task Test sets. In the multi-source and ablation conditions, we use the systems with the best validation performance for the target language pair. The columns indicate the evaluation language, and the Primary column indicate the aggregated results over all languages, as used in SemEval-2017 \citep{semeval17}. The wmt column denotes the es--en test set drawn from WMT's quality estimation track.
Rows indicate our three systems, compared with the ECNU system \citep{tian:2017}, and the LIPN--IIMAS system \citep{arroyo:2017}.}
\label{tab:official_test}
\resizebox{\columnwidth}{!}{
\begin{tabular}{lrrrrrrrr}
    \toprule
   & \textbf{Primary} & \textbf{ar--ar} & \textbf{ar--en}  & \textbf{es--es}  & \textbf{es--en}  & \textbf{wmt} & \textbf{en--en} & \textbf{en--tr} \\
   \midrule
\textbf{Single-source}         & 0.315 & 0.289 & 0.105 & 0.661 & 0.239 & 0.030 & 0.691 & 0.188 \\
\textbf{Multi-source}          & 0.294 & 0.312 & 0.129 & 0.692 & 0.100 & 0.016 & 0.688 & 0.120 \\
\textbf{Ablation}              & 0.215 & 0.003 & 0.110 & 0.547 & 0.226 & 0.020 & 0.506 & 0.090 \\
\textbf{LIPN--IIMAS}           & 0.107 & 0.047 & 0.077 & 0.153 & 0.172 & 0.145 & 0.074 & 0.080 \\
\textbf{ECNU}                  & \bf 0.732 & \bf 0.744 & \bf 0.749 & \bf 0.856 & \bf 0.813 & \bf 0.336 & \bf 0.852 & \bf 0.771 \\
\bottomrule
\end{tabular}
}
\vspace{-1cm}
\end{table}
\setlength{\tabcolsep}{6pt}

\subsection{Results on SemEval-2017}

In order to compare our system's performance in itself with state-of-the-art systems, we participated in the official shared task results of SemEval-2017 \citep{bjerva:2017:sts:semeval}.
The results from the official evaluation are shown in Table~\ref{tab:official_test}.
Although our results for Spanish--Spanish and English--English are in line with our development results, the results for all other language pairs are far lower than expected, and worse than the best performing ECNU system \citep{tian:2017}.
The fact that the system was low in the ranking list for the shared task can be explained by several factors.
On the one hand, our approach was very simplistic, whereas other systems took more involved approaches.
For instance, the ECNU submission first translates all sentences to English, and then use an ensemble of four deep neural network models and three feature engineered models.
The features used included word alignments, summarisation and MT evaluation metrics, kernel similarities of bags of words, bags of dependencies, n-gram overlap, edit distances, length of common prefixes, suffixes, and substrings, tree kernels, and pooled word embeddings \citep{tian:2017}.
In contrast, our system only uses the latter of these features.
On the other hand, our system did outperform other systems in individual language tracks.
Additionally, in all tracks we outperform the LIPN--IIMAS system, which approaches the task using an attentional LSTM \citep{arroyo:2017}.

Underfitting might be an explanation for the low results obtained as compared to development, indicating that the approach we have taken is simply not sufficient to solve the task of (cross-lingual) STS well.
Comparing our approach with other systems, such as the ECNU system, does indeed reveal a staggering difference in complexity.



\subsection{Results on SemEval-2016}
In order to further evaluate our system, we compare with an approach which is relatively similar to ours, namely the FBK HLT-MT submission described by \citet{ataman:2016}.
We replicate the training, development, and test setting used in Shared task SemEval-2016, with the exception that we only evaluate on one of the domains \citep{semeval16}.
The results from the \textit{news} domain (English-Spanish) are shown in Table~\ref{tab:semeval16_comparison}.
On this dataset, the difference between our system and that of \citet{ataman:2016} is relatively small.

\begin{table}[htbp]
    \centering
    \caption{SemEval 2016 system comparison, comparing our single-source system with the FBK HLT-MT system \citep{ataman:2016}. The runs from FBK HLT-MT differ in the features used.}
    \label{tab:semeval16_comparison}
    \begin{tabular}{lr}
        \toprule
        \textbf{System} & \textbf{Score} \\
        \midrule
        FBK HLT-MT -- Run 1 & 0.243 \\
        FBK HLT-MT -- Run 2 & 0.244 \\
        FBK HLT-MT -- Run 3 & \bf 0.255 \\
        \midrule
        Our system  & 0.241 \\
        \bottomrule
    \end{tabular}
    \vspace{-1cm}
\end{table}

\section{Conclusions}

Although our system faired relatively poorly in the official results for SemEval-2017, the multilingual experiments presented in this chapter offer insights into research question ({\bf RQ~\ref{rq:sts}}).
Multilingual word representations allow us to leverage a large amount of data from parallel corpora, opening up for multilingual learning of semantic textual similarity.
This allows for zero-shot learning, which yielded relatively good performance on unseen target languages on the development sets under consideration.
This indicates that multilingual word representations are indeed suitable for enabling zero-shot learning for STS ({\bf RQ~\ref{rq:sts}a}).
As for language relatedness, we found that applying zero-shot learning, and sharing parameters, between the Indo-European languages Spanish and English was more beneficial in general than when involving the Semitic language Arabic ({\bf RQ~\ref{rq:sts}b}).
Having seen that language similarities to some extent are indicative of performance in zero-shot STS, this raises the question of whether this generalises to other tasks, to what extent language similarities are important for this, and how such language similarities can be quantified.
We approach this in the following chapter, where we will look at {\bf RQ~\ref{rq:multiling}}.

\renewcommand*{\thefootnote}{\fnsymbol{footnote}}
\chapter[Comparing Multilinguality and Monolinguality]{\footnote{Chapter adapted from: \textbf{Bjerva, J.} (in review)
Quantifying the Effects of Multilinguality in NLP Sequence Prediction Tasks.}
\hspace{-6pt}Quantifying the Effects of Multilinguality in NLP Sequence Prediction Tasks\hspace{-8pt}}
\renewcommand*{\thefootnote}{\arabic{footnote}}
\label{chp:multilingual}

\begin{abstract}
\absprelude The fact that languages tend to share certain properties can be exploited by, e.g., sharing parameters between languages.
This type of model multilinguality is relatively common, as taking advantage of language similarities can be beneficial.
However, the question of \textit{when} multilinguality is a useful addition in terms of monolingual model performance is left unanswered.
In this chapter, we explore this issue by experimenting with a sample of $60$ languages on a selection of tasks: semantic tagging, part-of-speech tagging, dependency relation tagging, and morphological inflection.
We compare results under various multilingual model transfer conditions, and finally observe correlations between model effectivity and two measures of language similarity.
\end{abstract}
\clearpage
\section{Introduction}

Languages tend to resemble each other on various levels, for instance by sharing syntactic, morphological, or lexical features.
Such similarities can have many different causes, such as common language ancestry, loan words, or being a result of universals and constraints in the properties of natural language itself (cf. \citet{chomsky:2005,facultyoflanguage}).
Several current approaches to problems in Natural Language Processing (NLP) take advantage of these similarities.
For instance, in model-transfer settings, parsers are frequently trained on (delexicalised) versions of entire treebanks, whereas in annotation projection, word alignments between translated sentences are used \citep{mcdonald:2011,tackstrom:2012,tiedemann:2015,ammar:2016,vilares:2016,agic:2016}.\footnote{These approaches are covered in Chapter~\ref{chp:mtl_bg}.}
In the case of language modelling, multilinguality can be taken advantage of, e.g., in order to model domain-specific or diachronically specific language variants \citep{ostling:langmod}.
In addition to these specific examples, multilinguality in NLP models has been used in a whole host of other tasks, such as part-of-speech (PoS) tagging and semantic textual similarity, and is especially useful for NLP for low resource languages  \citep{georgi:2010,tackstrom:2013,faruqui:2016,plank:2017}.
One concrete advantage of taking a multilingual approach, is that this allows for the exploitation of much larger amounts of data, as compared to using monolingual approaches.
This fact, together with the prevalence of multilingual approaches in modern NLP, highlight the importance of further research in this area.

Although the literature contains a large amount of successful multilingual approaches, it is not sufficiently clear in which cases multilinguality is likely to be helpful.
The previous chapter served as an example for this, with some tentative indications of which combinations were useful, based on typological relatedness.
Hence, it may be reasonable to assume that choosing typologically similar languages when building a multilingual model will be beneficial, however this is not always the case, and relying on intuition or personal language knowledge in such matters has its limitations.
Hence, the go-to approach when considering multilinguality as a means of performance improvement in an NLP setting, is the time-consuming and resource-exhausting process of trial and error.
In this chapter, the aim is to provide insight into how one might approach the selection of languages when considering model multilinguality.
We investigate the following research questions, in order to provide an answer to \textbf{RQ~\ref{rq:multiling}}:
\begin{enumerate}
\setlength{\itemindent}{.29cm}
    \item [{\bf RQ~\ref{rq:multiling}a}] Given a model trained on a language, $l_1$, does adding data for another language, $l_2$, increase the performance on $l_1$ if those languages are similar?
    \item [{\bf RQ~\ref{rq:multiling}b}] In which way can such similarities be quantified?
    \item [{\bf RQ~\ref{rq:multiling}c}] What correlations can we find between model performance and language similarities?
\end{enumerate}


\noindent

We experiment on four NLP tasks: semantic tagging, part-of-speech tagging, dependency relation tagging, and morphological inflection.
The language sample we use differs per task, and covers a total of $60$ languages from a typologically diverse sample.
After covering related work, we first present experiments in multilingual settings for each of these tasks (Sections~\ref{sec:semtag}, ~\ref{sec:ud}, and ~\ref{sec:sigmorphon}).
We then investigate the correlations between two different measures of language similarity, and the change in system performance observed in multilingual settings, in order to provide an answer to our research questions (Section~\ref{sec:quantification}).
An approach which can be considered as parallel to this effort, is works similar to \citet{rosa:2017}, in which similarities between languages are exploited when deciding on which features to use in a cross-lingual parser.

\section{Semantic Tagging}
\label{sec:semtag}

\subsection{Background}

The first task under consideration is semantic tagging, as introduced in \citet{bjerva:2016:semantic}, and described in more detail in Chapter~\ref{chp:semtag}.
In this chapter, we consider this task in a multilingual setting, which is possible since the Parallel Meaning Bank (PMB, \citeauthor{pmb}, \citeyear{pmb}) includes such data for four languages: English, Dutch, German, and Italian.

\subsection{Data}

We use semantic tagging data obtained from the PMB \citep{pmb}.
There is a relatively large amount of gold standard annotation for English, which we use in our experiments.
Note that, we only use data from the PMB in this setting, as opposed to the setting in Chapter~\ref{chp:semtag} in which we also use data from the Groningen Meaning Bank (GMB, \citeauthor{gmb:hla}, \citeyear{gmb:hla}).
For the languages other than English, i.e.,\ Dutch, German, and Italian, we rely on the projected tags based on this gold standard annotation.
These tags were projected as described by \citet{pmb}, using word alignments obtained with \textit{GIZA++} \citep{gizapp}.
As the amount of parallel text differs per language, this yields the data amounts listed in Table~\ref{tab:pmb_data}.

\begin{table}[htbp]
    \centering
    \caption{Overview of the semantic tagging data used in this work.}
    \label{tab:pmb_data}
    \begin{tabular}{lrrl}
        \toprule
        \textbf{Language} & \textbf{Tokens} & \textbf{Sentences} & \textbf{Status} \\
        \midrule
        English & 20,098 & 2,814 & Gold \\
        Dutch   & 3,446  & 506   & Projected \\
        German  & 13,702 & 1,960 & Projected \\
        Italian & 11,376 & 1,711 & Projected \\
        \bottomrule
    \end{tabular}
\end{table}

\subsection{Method}
\label{sec:lstm_tagger}
\begin{figure}[htbp]
    \centering
    \includegraphics[width=0.5\textwidth]{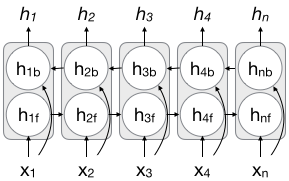}
    \caption{Sketch of the Bi-LSTM architecture used for our tagging tasks.}
    \label{fig:lstm}
\end{figure}

We employ a relatively simple neural network tagger for all of the tagging tasks in this study.
The tagger used is a bi-directional Long Short-Term Memory model (Bi-LSTM, \citet{lstm,graves:schmidhuber:2005}).
We use a single hidden layer for each direction, as shown in Figure~\ref{fig:lstm}.
The input-representations used are $100$-dimensional multilingual word embeddings trained on UN \citep{uncorp}, Europarl \citep{europarl}, and Bible data, using multilingual skip-gram \citep{multisg}, based on word alignments obtained with a variant of \textsc{Efmaral} \citep{efmaral}.\footnote{Using the default parameter settings for \textit{eflomal}: \\\url{https://github.com/robertostling/eflomal}.}\footnote{These are the same embeddings used in the previous chapter.}

The neural architecture used in this experiment is simpler than, e.g., the deep residual network Bi-GRU presented in Chapter~\ref{chp:semtag}.
This choice was made mainly for three reasons.
The primary motivation is that we wanted to rely only on multilingual word representations, seeing how far this will get us, without dealing with morphological dissimilarities, or exploiting morphological similarities directly.
Additionally, using only word representations is one way of using fewer parameters, meaning that fewer computational resources are needed.
Finally, although systems using character-based representations generally perform better than ones using only word-based representations, we are not interested in absolute performance \textit{per se}, but rather relative changes in performance when building multilingual models.

\subsubsection*{Hyperparameters}
We use the same hyperparameter settings for each of the experimental settings, so as to ensure comparability.
These are detailed in Table~\ref{tab:hyperparams_semtag}.
We always train for a maximum of 50 epochs, using the epoch at which validation loss was the best for evaluation.

\setlength{\tabcolsep}{5pt}
\begin{table}
    \small
    \caption{Hyperparameters used for semantic tagging.}
    \label{tab:hyperparams_semtag}
    \begin{tabular}{lll}
        \toprule
        \textbf{Hyperparameter}  & \textbf{Setting} & \textbf{Notes} \\
        \midrule
        Library             & Chainer \citep{chainer}          &   \\
        Loss function       & Categorical Cross-Entropy        &   \\
        Optimiser           & Adam \citep{adam}                &  \\
        Training iterations & Early stopping, \textit{best val loss}  &   \\
        Batch size          & 4 sentences                      &   \\
        Regularisation      & Dropout \citep{dropout}          & $p=0.5$   \\
        Regularisation      & Weight decay \citep{weightdecay} & $\epsilon=10^{-4}$   \\
        \bottomrule
    \end{tabular}
\end{table}
\setlength{\tabcolsep}{6pt}

\subsection{Experiments and Analysis}
We first look at a high-resource to low-resource scenario, using the languages for which we have a large amount of data (English, German, and Italian) as source languages, and all four languages as target languages.
We then run further experiments using all four languages as source languages.
Since only relatively few tokens are available for Dutch we train on $1000$ tokens, as this allows us to compare all four languages equally.
We reserve $500$ tokens for each language as test data, and split the remaining data into 80\% for training and 20\% for validation.
Since we deal with parallel texts, we make sure that the training, development, and test sets used for $l_1$ and any $l_2$ do not overlap in any way during training.

\subsubsection*{Transfer from high-resource to low-resource languages}
The goal of this experiment is to investigate zero-shot learning between languages for semantic tagging, in a scenario in which we use all training data available, giving us access to between 10,000 and 20,000 tokens of annotated data for the source languages.
Given the tentative results of the previous chapter, we expect that the Germanic languages will be more beneficial for one another, as compared to Italian.

\begin{table}[htbp]
    \centering
    \caption{Results on high-resource to low-resource transfer. Bold indicates the best source language for each target language, not considering the cases in which the source and target languages are identical, which are denoted by italics.}
    \label{tab:semtagging_hilo}
    \begin{tabular}{lrrr}
        \toprule
        \backslashbox{\small\textbf{Test}}{\small\textbf{Train}} & \textbf{English} & \textbf{German}   & \textbf{Italian} \\
        \midrule
        \textbf{English} & \it 75.03\% & \bf 49.20\% & 35.45\% \\
        \textbf{Dutch}   & 49.30\% & \bf 56.90\% & 36.31\% \\
        \textbf{German}  & \bf 41.99\% & \it 67.41\% & 41.17\% \\
        \textbf{Italian} & 35.74\% & \bf 39.11\% & \it  70.89\% \\
        \bottomrule
    \end{tabular}
\end{table}

The results from these experiments are shown in Table~\ref{tab:semtagging_hilo}, in which bold represent the best source languages for each target language, and italics denote the cases in which source and target languages are the same.
We can observe that the results when the source and target languages are both Germanic are higher than when the Romance language Italian is involved.
For instance, using German as a source language for Dutch results in relatively high accuracies.
This can be explained by two factors.
For one, it is likely that the quality of the multilingual word embeddings is higher when comparing German and Dutch.
Additionally, the extensive similarities between these two languages is likely to make zero-shot learning relatively easy.

Note that these results are not strictly comparable to those in Chapter~\ref{chp:semtag}, since we both use different and less training and evaluation data.
In absolute terms, performance on similar data is around 5\% worse in this experimental setting, although we train on approximately one order of magnitude less data than in Chapter~\ref{chp:semtag}.\footnote{In order to compare the architectures used, we also train and evaluate our tagger on the same data as in Chapter~\ref{chp:semtag}, in which case we obtain approximately the same performance as the baseline using only word representations.}

\subsubsection*{Transfer from low-resource source languages}

In the low-resource source scenario, with $1000$ training tokens per language, we compare some linguistic and input representation conditions.
We run experiments with two \textit{input representation} settings:
i) with \textit{frozen} pre-trained word representations;
ii) with \textit{updated} pre-trained word representations.
By \textit{frozen} word representations, we refer to representations which are kept at their initial states during learning.
That is to say, errors are not back-propagated into the embeddings.
In the \textit{updated} condition, however, embeddings are updated during training.
This comparison is done so as to investigate whether there is a difference when enforcing the multilingual embedding space to remain in its initial state, thus preserving multilingual distances.

In combination with the two representation settings, we also use two \textit{linguistic} settings:
a) monolingual training;
and b) multilingual training.
The monolingual training serves as a baseline, indicating how well we can transfer semantic tags from a source language to a target language, by only training on the target language.
In the multilingual setting, we add training data for the target language, comparing transfer between languages in a multilingual setting.
This comparison is done so as to investigate to what extent we can take advantage of data from two low resource languages, with the goal of benefitting both languages.

We first present results from the monolingual training in both input-representation settings, in Tables~\ref{tab:semtagging_pret_1k}, and~\ref{tab:semtagging_pretupd_1k}.

\begin{table}[htbp]
    \centering
    \caption{Results on monolingual semantic tagging.}
    \begin{subtable}[b]{\textwidth}
        \centering
    \caption{Training on 1k monolingual tokens, frozen embeddings.}
    \label{tab:semtagging_pret_1k}
    \begin{tabular}{lrrrr}
        \toprule
        \backslashbox{\small\textbf{Test}}{\small\textbf{Train}} & \textbf{English} & \textbf{Dutch} & \textbf{German}   & \textbf{Italian} \\
        \midrule
        \textbf{English} &  \bf 64.54\% & 37.24\%  & 36.40\%    &  28.54\% \\
        \textbf{Dutch} &  36.32\% & \bf 53.20\%  & 44.80\%    &  28.75\% \\
        \textbf{German} &  35.10\% & 39.12\%  & \bf 54.36\%    &  37.31\% \\
        \textbf{Italian}  &  30.29\% & 30.43\%  & 29.40\%    & \bf  61.20\% \\
        \bottomrule
    \end{tabular}
    \end{subtable}

\vspace{0.5cm}
\begin{subtable}[b]{\textwidth}
    \centering
    \caption{Training on 1k monolingual tokens, updated embeddings.}
    \label{tab:semtagging_pretupd_1k}
    \begin{tabular}{lrrrr}
        \toprule
        \backslashbox{\small\textbf{Test}}{\small\textbf{Train}}   & \textbf{English} & \textbf{Dutch} & \textbf{German}   & \textbf{Italian} \\
        \midrule
        \textbf{English}                                      &  \bf 64.90\% & 38.82\%  & 37.43\%     &  31.51\% \\
        \textbf{Dutch}                                        &  39.86\% & \bf  56.78\%  & 40.93\%     &  30.81\% \\
        \textbf{German}                                       &  39.14\% & 40.23\%  & \bf 55.46\%     &  39.07\% \\
        \textbf{Italian}                                      &  25.09\% & 34.02\%  & 31.21\%     &  \bf 49.63\%  \\
        \bottomrule
    \end{tabular}
\end{subtable}
\end{table}

Not surprising, monolingual models trained on the target language consistently perform better than when the source language is different from the target language, as in zero-shot learning.
Nonetheless, all results when source and target languages differ outperform a most frequent class baseline (17.35\%) by far.
This is expected, as the pre-trained models have been trained on a relatively large amount of parallel data, and have formed word spaces which are unified across languages, which allows them to generalise somewhat across languages, and confirms the quality of the embeddings themselves.
Comparing these results to those of the high-resource to low-resource setting, we can observe a steep drop in performance.
This is due to the fact that we have approximately an order of magnitude less data in the current setting.

Surprisingly, updating the pre-trained word embeddings during training increases results for most source/target combinations (English/Italian being the exception).
This was unexpected, as it is usually beneficial to update such embeddings during training, as this allows them to learn representations which are tuned for the task at hand.
It was, however, not expected that this would be the case when applying model transfer, as we expected that tuning, say, the English representation for \textit{dog} to be more task-specific, would skew it away from that of the Dutch equivalent \textit{hond}.
It would be interesting to explore this further, observing the resulting word-space after updating only one language in a multilingual language space.
Italian, interestingly, sees a severe drop in performance when updating embeddings in a monolingual setting.
This might be explained by the updated embeddings being overfit, and not generalising to the test set.

%

\subsubsection*{Transfer between low-resource source languages}
We now turn to transfer between low-resource source languages.
In this setting, we are mainly interested in seeing whether more related languages, i.e.\ English, Dutch, and German, are more beneficial to combine with one another, than with the typologically more distant Italian language.\footnote{We will consider how these similarities can be quantified in Section~\ref{sec:quantification}.}

\begin{table}[h]
    \caption{Results on multilingual semantic tagging.}
    \centering
    \begin{subtable}[b]{\textwidth}
        \centering
    \caption{Training on 1k+1k multilingual tokens, frozen embeddings.}
    \label{tab:semtagging_pret_1k1k}
    \begin{tabular}{lrrrr}
        \toprule
        \backslashbox{\small\textbf{Test}}{\small\textbf{Train}}   & \textbf{English} & \textbf{Dutch} & \textbf{German}   & \textbf{Italian} \\
        \midrule
        \textbf{English}                                      &  64.54\% & \bf 65.38\%  &  64.98\%      &  65.08\% \\
        \textbf{Dutch}                                        &  56.07\% & 53.20\%      &  \bf 60.20\%  &  54.96\% \\
        \textbf{German}                                       &  54.96\% & \bf 55.30\%  &  54.36\%      &  54.49\% \\
        \textbf{Italian}                                      &  47.69\% & 47.25\%      &  47.29\%      & \bf  61.20\% \\
        \bottomrule
    \end{tabular}
\end{subtable}

\vspace{0.5cm}
\begin{subtable}[b]{\textwidth}
    \centering
    \caption{Training on 1k+1k multilingual tokens, updated embeddings.}
    \label{tab:semtagging_pretupd_1k1k}
    \begin{tabular}{lrrrr}
        \toprule
        \backslashbox{\small\textbf{Test}}{\small\textbf{Train}}   & \textbf{English} & \textbf{Dutch} & \textbf{German}   & \textbf{Italian} \\
        \midrule
        \textbf{English}                                      &  64.90\% & 39.53\%  &  \bf  64.98\%    &  22.85\% \\
        \textbf{Dutch}                                        &  37.56\% & 56.78\%  &  \bf  60.20\%    &  26.69\% \\
        \textbf{German}                                       &  39.21\% & 40.12\%  &  \bf  55.46\%    &  21.87\% \\
        \textbf{Italian}                                      &  26.27\% & 34.47\%  &   47.29\%        &  \bf 49.63\% \\
        \bottomrule
    \end{tabular}
\end{subtable}
\end{table}

Tables~\ref{tab:semtagging_pret_1k1k}, and~\ref{tab:semtagging_pretupd_1k1k} contain the results from semantic tagging with multilingual training.
Considering the rows in each table, it is generally the case that training a model with a combination of English, Dutch and German, improves results more than combining one of these with Italian.
This seems to hold in both conditions, with and without updating the pre-trained vectors in training.

In contrast to the monolingual training case, we here do observe that freezing the vectors during training is beneficial for model performance.
It may be the case that, since the weights of the embeddings in both $l_1$ and $l_2$ are optimised, they are pushed even further apart than in the monolingual training case in which only one language's embeddings are affected.
Hence, in the frozen case, the integrity of the multilingual language space is maintained, allowing the model to learn cross-lingually.
A potential explanation for the drop in results on Italian when updating the embeddings might be the extent to which the languages are similar.
Since English, German, and Dutch are all Germanic languages, it is possible that this relatedness suffices to preserve the multilingual quality of the word space.

\subsection{Summary of Results on Semantic Tagging}
We have observed that training on similar languages is helpful for semantic tagging, in the sense that combining Germanic languages tended to be beneficial.
Additionally, training in a high-resource scenario on, e.g., German and using Dutch as a source language yielded better results than when training on low-resource Dutch (see Tables~\ref{tab:semtagging_hilo} and~\ref{tab:semtagging_pret_1k}).
This leads us to ask whether similar patterns can be found when observing a larger sample of languages, and on other tasks.

\section{Tagging Tasks in the Universal Dependencies}
\label{sec:ud}
The Universal Dependencies treebank offers an excellent testing ground for experiments on NLP model multilinguality.
The corpus collection contains many languages, with several layers of uniform annotation across languages \citep{ud:2016}.
We use version 2.0 of the UD treebanks for experiments in two tasks: PoS tagging and dependency relation tagging \citep{ud20}.
We evaluate on the $48$ languages for which training data is available.

\subsection{Data}
In order to balance our experiments for differing data sizes, we balance all training sets so as to have an equal number of tokens.
We set this amount to 20,000 tokens, in order to allow for inclusion of the smallest language in the UD (Vietnamese, $n=20285$).
Hence, the overall results obtained will be relatively low, but should make the effects of multilingual modelling clearer.
A part of the evaluation will deal with grouping languages per language group.
An overview of which languages are included in the Germanic, Romance, and Slavic families in these evaluations is given in Table~\ref{tab:lang_families}.

\begin{table}[htbp]
    \centering
    \caption{Language grouping of the Germanic, Romance, and Slavic languages used in our experiments.}
    \label{tab:lang_families}
    \begin{tabular}{ll}
        \toprule
        \textbf{Language family} & \textbf{Language} \\
        \midrule

        \multirow{8}{*}{\textbf{Germanic}} & Afrikaans \\
        & Danish  \\
        & Dutch  \\
        & English   \\
        & German   \\
        & Norwegian Bokmål  \\
        & Norwegian Nynorsk  \\
        & Swedish  \\
        \midrule
        \multirow{11}{*}{\textbf{Slavic}} & Belarusian    \\
        & Bulgarian  \\
        & Croatian  \\
        & Czech  \\
        & Old Church Slavonic   \\
        & Polish   \\
        & Russian  \\
        & Serbian  \\
        & Slovak   \\
        & Slovenian  \\
        & Ukrainian  \\
        \midrule
        \multirow{9}{*}{\textbf{Romance}} & Catalan \\
        & French   \\
        & Galician   \\
        & Italian  \\
        & Latin  \\
        & Portuguese  \\
        & Brazilian Portuguese \\
        & Romanian  \\
        & Spanish  \\
        \bottomrule
    \end{tabular}
\end{table}

\subsection{Method}
We employ the same tagger as described in the semantic tagging experiments of this chapter, detailed in Section~\ref{sec:lstm_tagger}.
We also use the same hyperparameter settings, as shown in Table~\ref{tab:hyperparams_semtag}, and the same input representations.
The main difference with the semantic tagging experiments is therefore simply the tasks at hand, and the amount of languages under consideration.

\subsubsection*{Experimental Setup}
We only use the setting with \textit{frozen} word representations, as established in the semantic tagging experiments.
We focus on results from the multilingual training settings, as we are interested in how these results differ between language pairs.
Additionally, we consider two tasks in this setup: PoS tagging, and dependency relation tagging.
Dependency relation tagging is the task of predicting the dependency tag (and its direction) for a given token.
This is a task that has not received much attention, although it has been shown to be a useful feature for parsing \citep{ouchi:2014,ouchi:2016}.\footnote{Dependency relation labels are discussed in more detail in Chapter~\ref{chp:mtl}.}
These \textit{deprel} tags can be derived directly from UD dependency parse trees, making it straight-forward to evaluate on this task for the same sample of languages in the same settings.
In this setting, we use the dependency relation instantiations with simple granularity and simple directionality (i.e., encoding the head and its relative position, for each word), described further in Chapter~\ref{chp:mtl} (Table~\ref{tab:supertags}).

\subsection{Results and Analysis}
Due to the large number of language pairs, we discuss the results from the mean accuracy on a language group when trained in combination with a sample of languages.\footnote{Results covering all languages are presented later in this chapter.}

\subsubsection*{PoS Tagging}
Table~\ref{tab:pos_results_pretfreeze} contains PoS tagging results with frozen embeddings.
Some noteworthy findings include the highest accuracies per language group, marked in bold.
These are generally obtained by languages which are in the same language group, although there are exceptions to this pattern.
Note that we do not develop on the language which we use for evaluation.
That is to say, e.g., when evaluating on Danish, the \textit{Germanic} column is calculated as the mean accuracy over German, Norwegian Bokm\aa l and Nynorsk, and not including Danish.

There are examples in which training on a language from the same language group worsens performance overall.
Some notable cases include training on Dutch for the Germanic languages.
The relatively poor performance as compared to Danish might be explained by the fact that this group includes four Scandinavian languages, meaning that Danish has three such languages which it is likely helpful for.
Dutch, on the other hand, has only two languages to which it is highly similar in the Germanic group, namely Afrikaans and German.

It is nonetheless somewhat puzzling that some non-Germanic languages yield better performance in the Germanic group than Dutch.
Considering the baseline column in the table, however, it is clear that the model only sees increases in performance in a few cases, with a loss in performance in nearly all cases.
Therefore a potential explanation to the overall drop in results might be the fact that, although all languages within a single group are related to one another, this relatedness might still be too distant to be exploited in the current setup.
For instance, the languages in the Slavic group represent a relatively large variety.

\begin{table}[htbp]
    \centering
    \caption{PoS results -- Training on 20k+20k multilingual tokens, frozen embeddings. Columns indicate average results over languages in that language group.}
    \label{tab:pos_results_pretfreeze}
    \begin{tabular}{lrrr}
        \toprule
        \textbf{Language} & \textbf{Germanic}  & \textbf{Romance} & \textbf{Slavic} \\
        \midrule
        Baseline  &  83.97\% & 84.35\% & 81.12\% \\
        \midrule
        Bulgarian &  83.89\%	& 82.93\%	& 76.53\% \\

        Czech &  83.79\%	& 82.34\%	& 76.28\% \\
        Danish &  \textbf{84.14\%}	& 84.20\%	& 79.73\% \\
        Finnish &  83.87\%	& 83.58\%	& 78.41\% \\
        French &  82.59\%	& \textbf{84.54\%}	& 79.73\% \\
        Italian	& 83.83\%	& 83.77\%	& 79.67\% \\
        Dutch &  82.46\%	& 84.32\%	& 79.40\% \\
        Polish &  81.93\%	& 84.44\%	& 78.58\% \\
        Portuguese &  83.35\%	& 81.14\%	& 78.42\% \\
        Russian &  83.85\%	& 82.77\%	& \textbf{82.26\%} \\
        \bottomrule
    \end{tabular}
\end{table}

\subsubsection*{Dependency Relation Tagging}

Table~\ref{tab:deprel_results_pretfreeze} contains results from dependency relation tagging in the frozen embeddings setting.
Interestingly, although the top results for Germanic and Slavic are from in-group languages, we observe the best results here for out-of-group languages for the Romance group.
The results of these experiments show a similar trend to those in the PoS experiments, with almost all results being worse than the baseline.

\begin{table}[htbp]
    \centering
    \caption{DepRel results -- Training on 20k+20k multilingual tokens, frozen embeddings. Columns indicate average results over languages in that language group.}
    \label{tab:deprel_results_pretfreeze}
    \begin{tabular}{lrrr}
        \toprule
        \textbf{Language} & \textbf{Germanic}  & \textbf{Romance} & \textbf{Slavic} \\
        \midrule
        Baseline  & 65.10\% & 70.31\% & 65.25\% \\
        \midrule
        Bulgarian &  59.01\%	& \textbf{69.52\%}	& 60.31\% \\
        Czech &  64.05\%	& 67.91\%	& 61.95\% \\
        Danish &  \textbf{65.56\%}	& 68.20\%	& 61.74\% \\
        Finnish &  64.87\%	& 63.34\%	& 61.78\% \\
        French &  64.99\%	& 67.63\%	& 62.23\% \\
        Italian	& 63.94\%	& 67.97\%	& 61.91\% \\
        Dutch &  64.71\%	& 66.33\%	& 62.29\% \\
        Polish &  64.15\%	& 64.68\%	& 60.98\% \\
        Portuguese &  64.92\%	& 67.36\%	& 62.31\% \\
        Russian &  64.17\%	& 68.32\%	& \textbf{65.30\%} \\
        \bottomrule
    \end{tabular}
\end{table}

\subsection{Summary of Results on the Universal Dependencies}
For semantic tagging, we saw increases in performance when combining the Germanic languages with one another.
The results from tagging tasks on the UD languages reveal that this granularity of language similarity is not sufficient to determine whether this type of model multilinguality will be successful, under the experimental conditions used here.
In fact, observing results aggregated by the language families Germanic, Romance, and Slavic, revealed a decrease in performance when transferring from almost all languages in these families, with some exceptions.
These results thus shed some light on two potential issues.
On the one hand, describing language similarities in terms of typological families is perhaps not sufficient for the purposes of this chapter.
On the other hand, the multilingual model architecture used in the tagging experiments of this chapter might not be sufficient to fully take advantage of language similarities.

\section{Morphological Inflection}
\label{sec:sigmorphon}
Having investigated two sequence labelling tasks, we now turn to a sequence-to-sequence prediction task, namely morphological inflection.
The 2017 shared task on morphological inflection offers a large amount of data for 52 languages \citep{sigmorphon:2017}.
Whereas the shared task has two sub-tasks, namely inflection and paradigm cell filling, we only evaluate on the inflection task.
Furthermore, we use the high-resource setting, in which we have access to 10,000 training examples per language.
The inflection subtask is to generate a target inflected form, given a lemma with its part-of-speech, as in the following example:

\begin{verbatim}
Source form and features: release V;NFIN
Target tag:               V;V.PTCP;PRS
Target form:              releasing
\end{verbatim}

\subsection{Method}

\begin{figure}[htbp]
    \centering
    \includegraphics[width=0.9\textwidth]{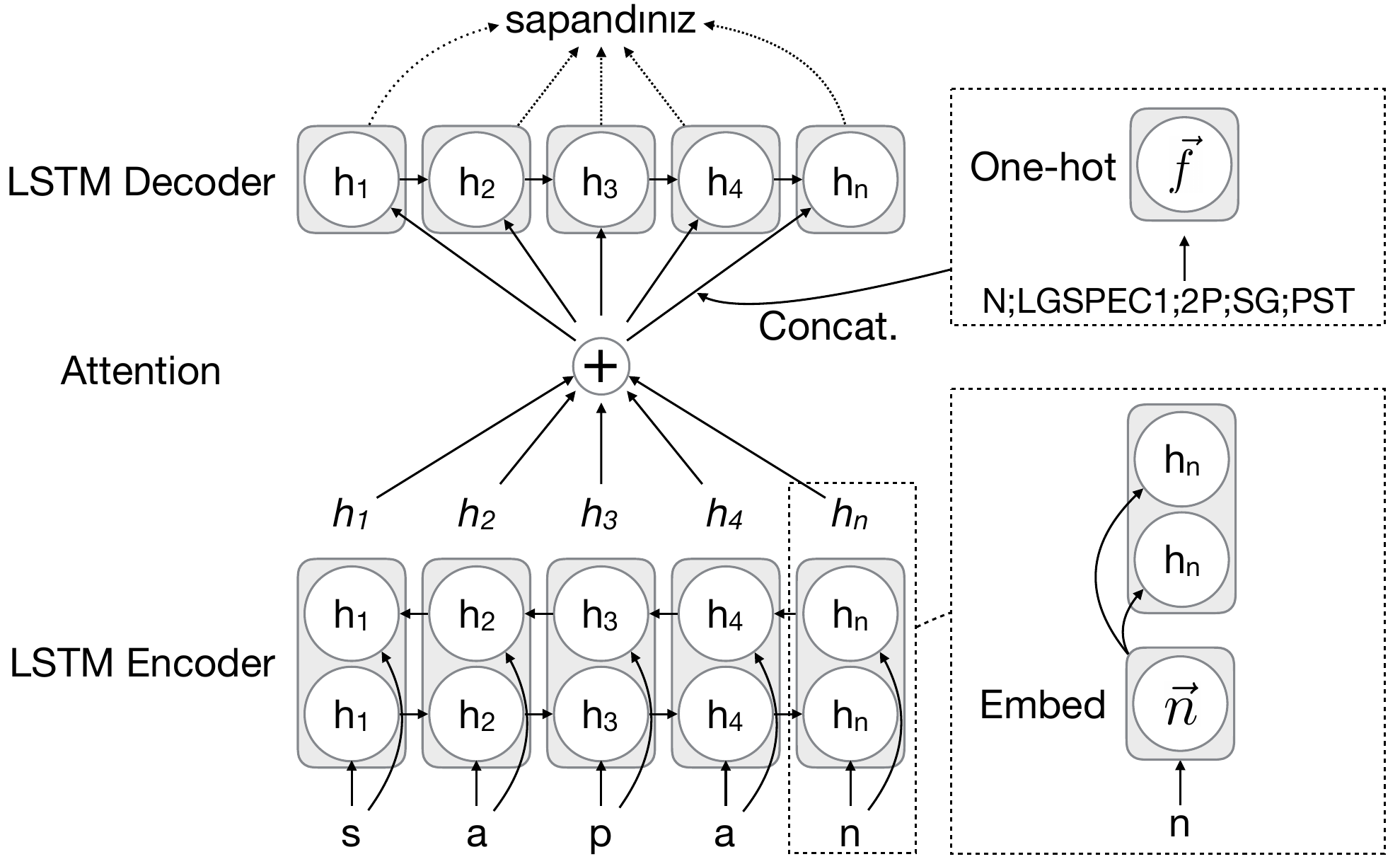}
    \caption{Architecture used for morphological inflection, consisting of an encoder-decoder with attention. The example depicts the production of the Turkish inflected form \textit{sapandınız}, based on the input \textit{sapan} and the tags \textit{N;LGSPEC1;2P;SG;PST}.}
    \label{fig:sigmorphon_arch}
\end{figure}

We employ a deep neural network for the experiments in morphological inflection.
This consists of an attentional sequence-to-sequence model, as described in \citet{bjerva:2017:sigmorphon}.\footnote{Available at \url{https://github.com/bjerva/sigmorphon2017}.}\footnote{In the SIGMORPHON shared task, this team placed as the 4th best \citep{sigmorphon:2017}.}
The system takes embedded character representations as input to a Bi-LSTM encoder.
The output of the encoder is passed through an attention mechanism, to an LSTM decoder which also takes the target form's morphological tags as features.
All layers in the network has 128 hidden units.
Optimisation is done using Adam \citep{adam} with default parameters.
Whereas \citet{bjerva:2017:sigmorphon} explore learning a single model per language, in this chapter we experiment with learning joint models across languages.
Additionally, we do not use an ensemble for the results presented in this chapter.
The system architecture is visualised in Figure~\ref{fig:sigmorphon_arch}

\subsubsection*{Experimental Setup}
We train our system using joint input and output representations.
In order to examine the effect of adding a language to the mix, we train each model as follows.
Given each language in the set of languages $l\in L$, we sample all language combinations $l_1, l_2$.
We then train on the entire \textit{high} dataset of $l_1$ (i.e.,~10,000 examples), combined with $n$ training examples from $l_2$, with $n=[0, 2^0, 2^1, \ldots, 2^{13}]$.\footnote{The SIGMORPHON-2017 shared task dataset contains three resource settings: low (100 examples), medium (1000 examples), and high (10000 examples). }
In other words, $l_1$ is our source language, and $l_2$ our target language.
This yields a total of $|L|\times|L|\times|n|=24,000$ experiments.
Note that the model is language agnostic, and apart from orthographic similarities between languages, has no way of knowing whether a certain string belongs to, e.g., Norwegian Nynorsk or Bokmål.
We train each model for a total of $36$ hours on a single CPU core, and report results using the model with the best validation loss.

\subsection{Results and Analysis}

\begin{figure}[p]
    \begin{subfigure}[b]{0.5\textwidth}
    \includegraphics[width=\textwidth]{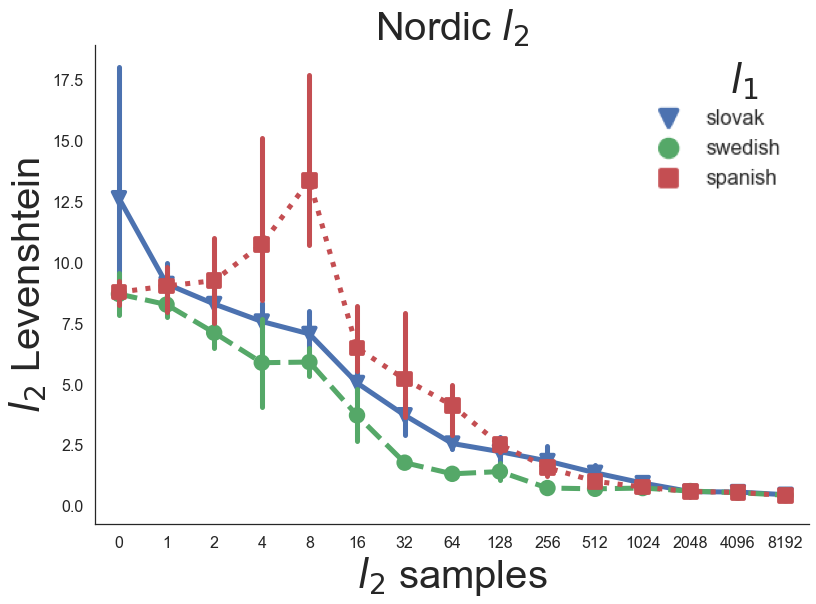}
    \caption{Mean Nordic $l_2$ Levenshtein.}
    \end{subfigure}
    \begin{subfigure}[b]{0.5\textwidth}
    \includegraphics[width=\textwidth]{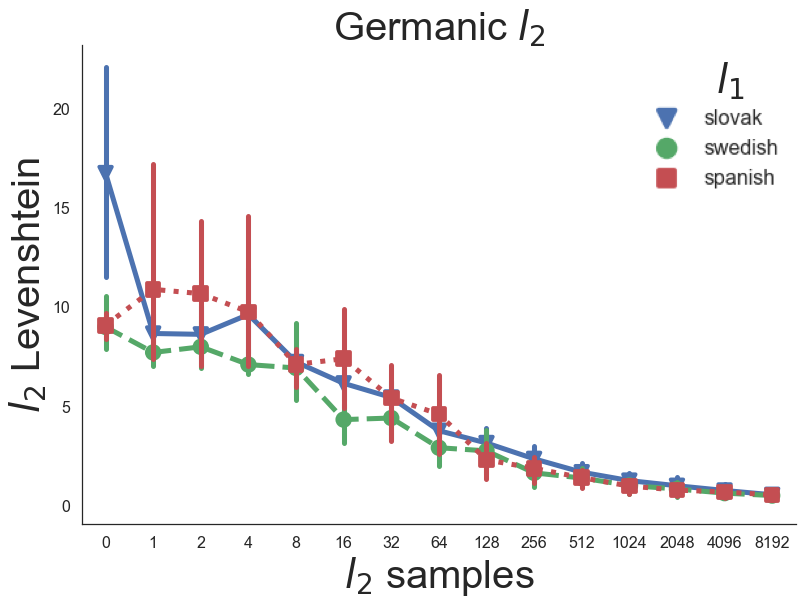}
    \caption{Mean Germanic $l_2$ Levenshtein.}
    \end{subfigure}

    \begin{subfigure}[b]{0.5\textwidth}
    \includegraphics[width=\textwidth]{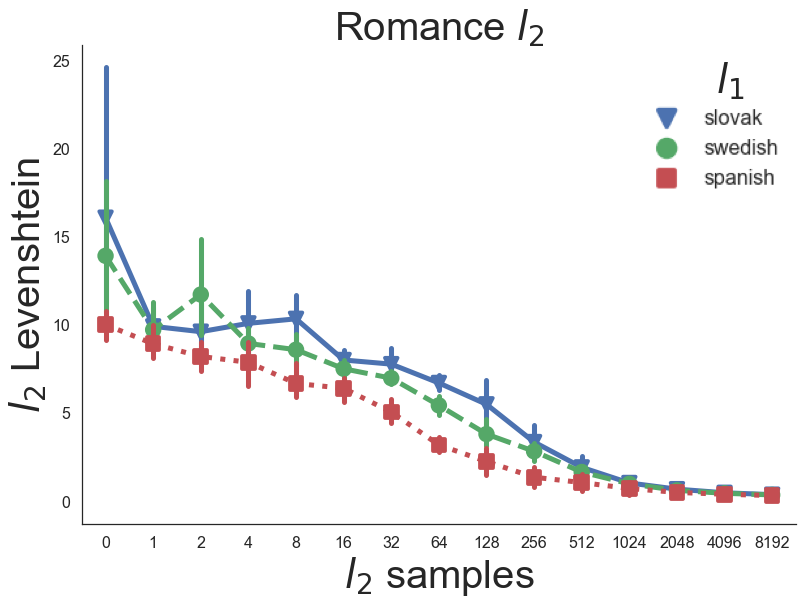}
    \caption{Mean Romance $l_2$ Levenshtein.}
    \end{subfigure}
    \begin{subfigure}[b]{0.5\textwidth}
    \includegraphics[width=\textwidth]{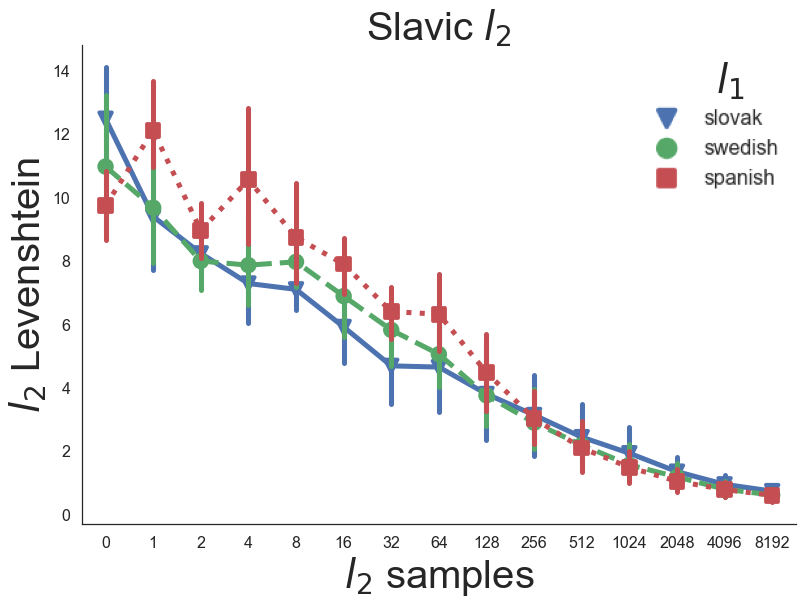}
    \caption{Mean Slavic $l_2$ Levenshtein.}
    \end{subfigure}

    \begin{subfigure}[b]{0.5\textwidth}
    \includegraphics[width=\textwidth]{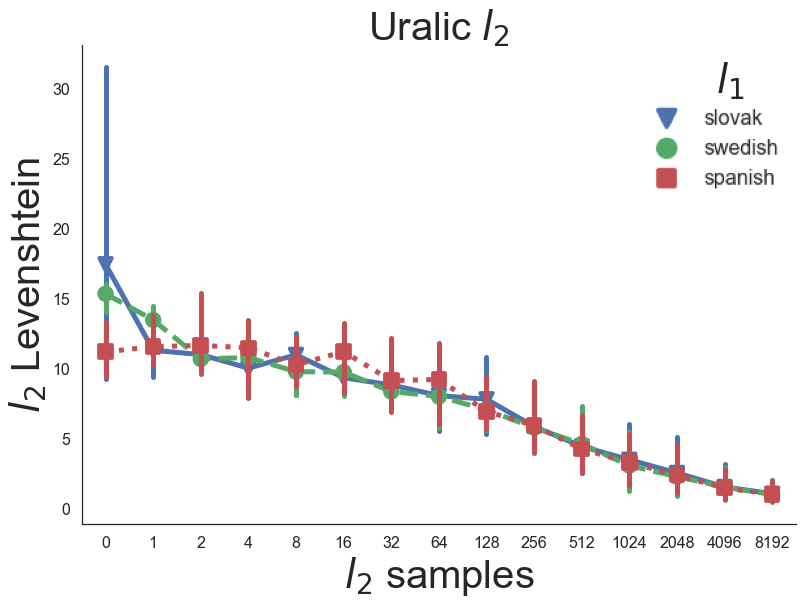}
    \caption{Mean Uralic $l_2$ Levenshtein.}
    \end{subfigure}
    \begin{subfigure}[b]{0.5\textwidth}
    \includegraphics[width=\textwidth]{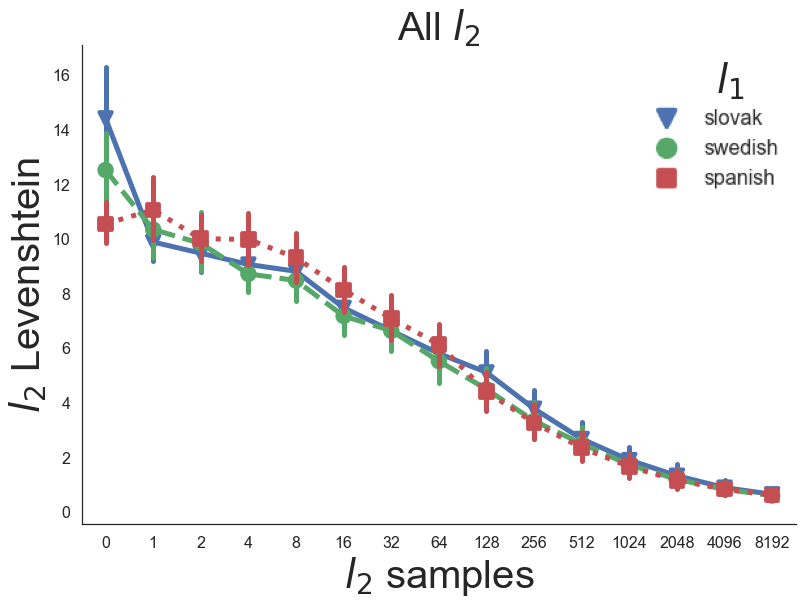}
    \caption{Mean $l_2$ Levenshtein for all $l_2$s.}
    \end{subfigure}
    \caption{Results on morphological inflection, with Slovak (blue, full, triangles), Swedish (green, dashed, circles), and Spanish (red, dotted, squares) as $l_1$.
    \label{fig:sigmorphon}}
\end{figure}

Evaluation is done using the standard metric for this task, namely the Levenshtein distance between the predicted form and the target form (i.e.\ lower is better).
Figure~\ref{fig:sigmorphon} shows results group-mean results on $l_2$ accuracy for training size $n$.
The green lines show results when the $l_1=\text{Swedish}$, the red lines when the $l_1=\text{Spanish}$, and the blue when $l_1=\text{Slovak}$.
Note that for performance is always better when training is combined with a language from the same language group.
Notable is the performance with $l_1=\text{Spanish}$ in the Nordic language group, where performance in fact drops when adding more $l_2$ samples at first.
This indicates that transfer from languages which are more similar is beneficial, as compared to transfer from less related languages.
This should come as no surprise, as the morphological similarities between, e.g., Norwegian and Danish are very pronounced, whereas similarities between Norwegian and Spanish are limited, if any exist.
As a transfer baseline, the bottom right shows an average across all language families, showing that none of the languages are inherently better as source languages, as confidence intervals overlap for almost all amounts of $l_2$ samples.

The results when using fewer than $256-512$ $l_2$ samples are, across the board, below baseline levels.
This can be explained by the fact that the system setup used in these experiments was not sufficient for cross-lingual transfer to be particularly successful.
Changes in the architecture, such as including language vectors, as described by \citet{ostling:langmod} and \citet{malaviya:2017} is one possibility of improving this.\footnote{Language vectors are described further in Section~\ref{sec:langvec}.}

In all cases, when sufficient $l_2$ data has been observed, the differences in performance with different $l_1$ is close to zero.
This indicates that the model is not relying on information from the $l_1$ in such cases.
The results obtained by the multilingual system when observing $2^{13}$ $l_2$ samples are similar to those obtained by the monolingual systems in \citet{bjerva:2017:sigmorphon}.
There are slight drops in performance across the board, which can be explained by two factors.
On the one hand, some net capacity is wasted (from the perspective of $l_2$ performance), as we encode several languages in the same model.
Additionally, we only observe $2^{13}$ $l_2$ samples, whereas the systems in \citet{bjerva:2017:sigmorphon} use all 10,000 samples available for training.

\subsection{Summary of Results on Morphological Inflection}
Similarly to the semantic tagging results, we observe that typologically related languages do tend to fare better in this transfer setting.
Perhaps most convincing are the results when using Swedish as the source language and evaluating on Nordic target languages.
This might be caused by the fact that the languages included in the Nordic (Danish, and Norwegian Bokmål and Nynorsk) group are highly similar to Swedish, whereas the other languages and language groups under consideration are more distinct.

\section{Estimating Language Similarities}
\label{sec:quantification}
So far, we have considered the research questions dealing with the effects of hard parameter sharing between languages.
The results have differed per task and per language combination, with the general trend that languages which \textit{seem} similar, tend to be beneficial in combination with one another.
This brings us to the final research question addressed in this chapter, namely, with what type of similarity measures does multilingual effectivity correlate?

As grouping by typological language families yielded a relatively large spread in results, one possibility is that language similarities should be quantified in a different manner.
We investigate two different measures to estimate the similarity between languages.
These measures have in common that the requirements to produce them are vanishingly small, meaning that they are not restricted to a few languages.
In fact, the measures are readily available for a significant portion of the languages in the world.
Furthermore, the two measures are quite different from one another -- one directly obtained in a data-driven manner, and one based on edit distances on a lexical level.

\subsection{Data-driven Similarity}
\label{sec:langvec}

The data-driven similarity measure which we employ is based on training language embeddings together with a Long Short-Term Memory language model \citep{lstm}.
The vectors are learned by conditioning the LSTM's prediction on an embedded language representation, when training the language model on a large collection of languages.
This leads to the model learning representations which encapsulate some type of language similarity, which means that the vectors can be used to calculate similarities between languages, and is presented by \citet{ostling:langmod}.\footnote{Thanks to Robert Östling for providing us with access to this resource.}
Furthermore, the method is applicable to languages with very limited data, such as all languages with, for instance, a translation of the New Testament (i.e.\ $\approx1000$ languages).
This approach to obtaining distributed can be compared to \citet{malaviya:2017}, in which similar representations are learned in a neural machine translation system.
We use the cosine distance between the vectors of two languages as a measure of their similarity.

\subsection{Lexical Similarity}

We calculate lexical similarity as in \citet{rama:2015}, by using normalised Levenshtein distance (LDN) between aligned word lists.
LDN is calculated by summing length normalised Levenshtein distances for pairs of words using, e.g., Swadesh lists.\footnote{Swadesh lists are standardised word lists, covering semantic concepts which are normally found in a given language, developed for the purposes of historical-comparative linguistics.}
While effects such as similarity between phoneme inventories could cause unrelated languages to seem related, LDN has the advantage that it compensates for such effects \citep{rama:2015}.

Lexically aligned lists, similar to the Swadesh lists, are obtained from the Automated Similarity Judgement Program (ASJP) database \citep{asjp}.\footnote{\url{http://asjp.clld.org/}}
The ASJP aims to offer 40-word lists for all of the world's languages, and currently offers such lists for 4664 languages.\footnote{As of 11-05-2017.}
These lists are linked on the meaning level, which allows for comparison of words across languages (see Table~\ref{tab:asjp} for an example).
The lists do not contain the orthographic representations of these words, but rather a phonemic representation.
Such a representation is beneficial for our purposes, as differences in orthography resulting from historical artefacts might otherwise skew the results.
For instance, while the orthographic representations of the 1st person singular pronoun in English and Norwegian have the maximum possible Levenshtein distance for the word pair (\textit{I} vs. \textit{jeg}), their phonemic representations reveal the commonalities (\textit{Ei} vs \textit{yEi}).

\begin{table}
    \caption{Examples from ASJP for English, Dutch, Norwegian, Finnish, and Estonian.}
    \label{tab:asjp}
    \centering
    \resizebox{\columnwidth}{!}{
    \begin{tabular}{llllll}
        \toprule
        \textbf{Word/Meaning} & \textbf{English} & \textbf{Dutch} & \textbf{Norwegian} & \textbf{Finnish} & \textbf{Estonian} \\
        \midrule
        I   & Ei  & ik  & yEi & minE  & mina \\
        you & yu  & yEi & d3  & sinE  & sina \\
        we  & wi  & vEi & vi  & me    & me   \\
        one & w3n & en  & En  & iksi  & uks  \\
        two & tu  & tve & tu  & kaksi & kaks \\
        \bottomrule
    \end{tabular}
    }
\end{table}

\begin{figure}[p]
    \centering
    \includegraphics[width=\textwidth]{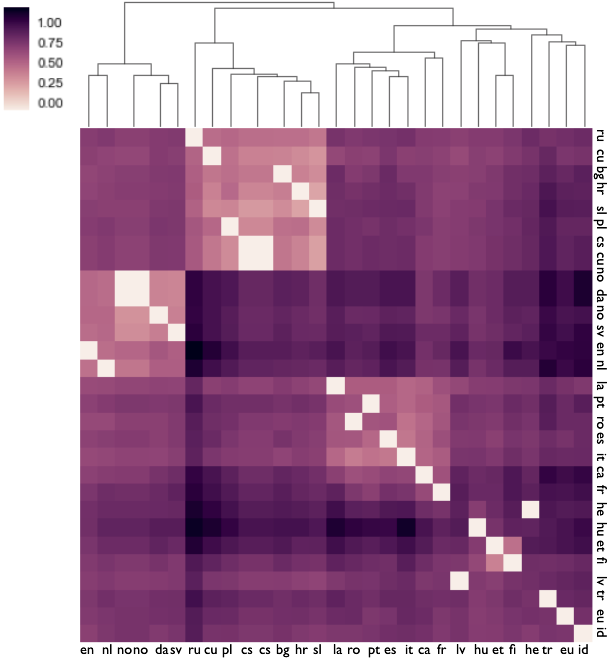}
    \caption{Distances calculated using LDN between ASJP lists, clustered with UPGMA.}
    \label{fig:lev-dist}
\end{figure}

Figure~\ref{fig:lev-dist} further illustrates the lexical distance measure.
Languages which are typologically similar to each other are automatically grouped together, using the hierarchical clustering algorithm \textit{UPGMA} (Unweighted Pair Grouping Method with Arithmetic-mean, cf. \citeauthor{upgma}, \citeyear{upgma}).

\subsection{Results and Analysis}
We will now consider the correlations observed between these language measures, and the results obtained from the multilingual experiments outlined in this chapter.
The results from the semantic tagging are not included in this analysis, as we have too few data points available for the semantic tagging task to allow for reliably quantitative analysis.
However, it is worth noting that the Germanic languages tend to help each other out, whereas Italian is generally less beneficial to performance.

\subsubsection*{Tagging Tasks in the Universal Dependencies}
\label{sec:ud_lang_corr}

\begin{figure}[htbp]
    \centering
    \begin{subfigure}[b]{0.45\textwidth}
        \includegraphics[width=\textwidth]{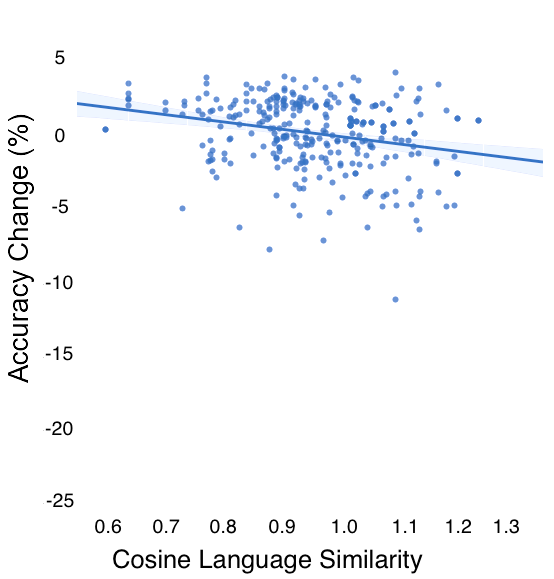}
        \caption{Language vector distances compared to change in accuracy on PoS tagging.}
        \label{fig:lang_corr_pos_vecs}
    \end{subfigure}
    \hfill
    \begin{subfigure}[b]{0.45\textwidth}
        \includegraphics[width=\textwidth]{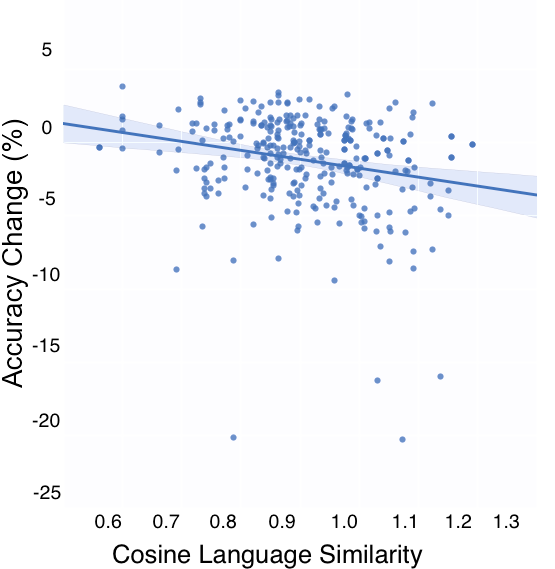}
        \caption{Language vector distances compared to change in accuracy on dependency relation tagging.}
        \label{fig:lang_corr_pos_edit}
    \end{subfigure}
    \caption{Language vector distances: Correlations between accuracy and language similarities.}
\end{figure}

Figure~\ref{fig:lang_corr_pos_vecs} shows language correlations with language vector similarities, across languages and conditions in the PoS tagging task (Spearman $\rho=-0.14$ ($p=0.001$)).
Figure~\ref{fig:lang_corr_pos_edit} contains the corresponding plot for the dependency relation task
(Spearman $\rho=-0.19$ ($p\ll0.001$)).
Although these correlations are statistically significant, it is debatable whether or not they are practically significant.
Given this amount of data points, statistically significant results are relatively likely, as the p-value indicates the risk of the correlation coefficient being equal to zero, given the data.

Although the correlations themselves are rather weak, it is interesting to observe that the patterns for both language similarities are rather similar.
This is likely due to the fact that both of these measures offer some explanatory value for the problem at hand, and might also be a side-effect of the fact that these two measures correlate rather well with one another ($\rho=0.7, p\ll0.001$).

%
%

\subsubsection*{Morphological Inflection}

\begin{figure}[htbp]
    \centering
    \begin{subfigure}[b]{0.45\textwidth}
        \includegraphics[width=\textwidth]{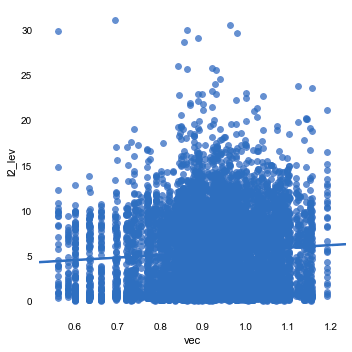}
        \caption{Language vector distances.}
        \label{fig:lang_corr_sigmorphon_vecs}
    \end{subfigure}
    \hfill
    \begin{subfigure}[b]{0.45\textwidth}
        \includegraphics[width=\textwidth]{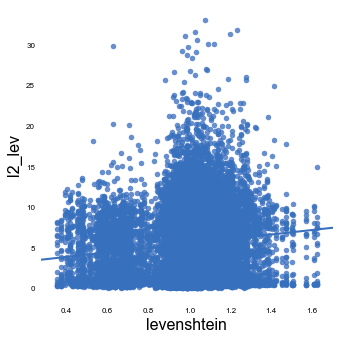}
        \caption{Levenshtein distances.}
        \label{fig:lang_corr_sigmorphon_edit}
    \end{subfigure}
    \caption{Morphological inflection: correlations between Levenshtein distance and language similarities.}
\end{figure}

The correlations between vector distances and performance in morphological inflection are weak, as seen in Figure~\ref{fig:lang_corr_sigmorphon_vecs} (Spearman $\rho=0.075$, $p<0.001$).
The correlation coefficient is somewhat higher when comparing with Levenshtein distance, as seen in Figure~\ref{fig:lang_corr_sigmorphon_edit} (Spearman $\rho=0.16$, $p\ll0.001$).


\subsection{When is Multilinguality Useful?}
As we used relatively little data for training the tagging models, so as to allow for inclusion of a large number of languages, the absolute performance obtained is quite low.
However, there appears to be some relation between the usefulness of adding in one more language to a model, and how similar those languages are.
Although this seems is quite intuitive, the effects observed in our training setting were more subtle than expected.
For instance, in many cases languages which are not particularly related appear to also increase system performance.
This might be explained by two factors.
On the one hand, it is possible that the quality of the word embeddings used is high enough so as to make the model fairly language agnostic.
An alternative explanation, is that the network simply uses the information from a second language to further adjust its prior.

In the case of morphological inflection, we saw that the edit-distance based measure of language similarity was more informative than the language vectors.
The fact that a measure based on edit distances is more successful here, is not altogether surprising as the task deals with minimising the Levenshtein distance between the predicted inflected form and the target form.

The effects seen in this work were weaker than expected, indicating that additional factors to language similarity as defined in this work govern the usefulness of multilinguality.
However, the weak correlations still hold, indicating that choosing languages which are similar either in terms of lexical distance, or in terms of language vectors, might be a good place to start.
An interesting prospect for future work, is to incorporate, e.g., the language vectors as a feature.
This might make it easier for the model to learn between which languages it is the most beneficial to share certain parameters (e.g.\ between Nordic languages), and between which languages such sharing would likely lead to negative transfer.

\vspace{-0.35cm}
\section{Conclusions}
\vspace{-0.35cm}
We investigated multilinguality in four NLP tasks, and observed correlations between performance in multilingual models with two measures of language similarity, in addition to a preliminary comparison based on typological language families.
On a general level, we found some cases in which using a source language related to the target language was beneficial, mainly in the case of semantic tagging ({\bf RQ~\ref{rq:multiling}a}).
We then looked at two measures of language similarities ({\bf RQ~\ref{rq:multiling}b}), which showed some correlations with multilingual model effectivity.
The correlations found were, however, rather weak, indicating that language similarities as defined in this work are not sufficient for explaining such improvements to a large degree ({\bf RQ~\ref{rq:multiling}c}).

In the next chapter, we will nonetheless continue on this path, attempting to both exploit language similarities, as well as similarities between tasks ({\bf RQ~\ref{rq:mmmt}}).

\partimage[width=1\textwidth]{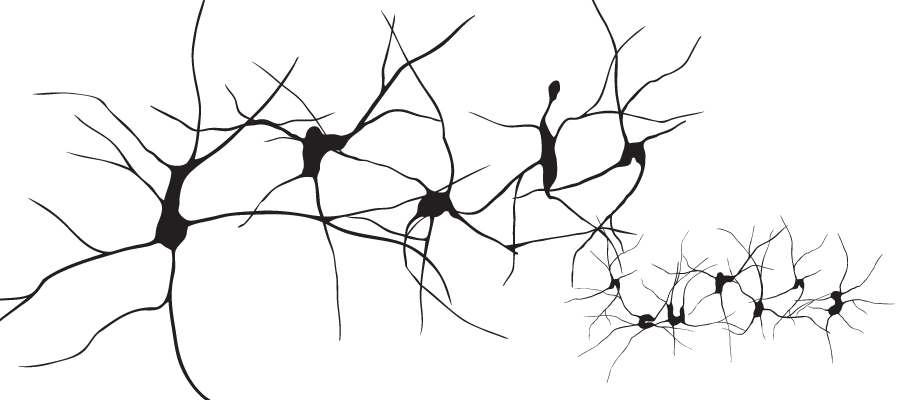}
\part{Combining Multitask and Multilingual Learning}
\renewcommand*{\thefootnote}{\fnsymbol{footnote}}
\chapter[Multitask Multilingual Learning]{One Model to rule them all: \mbox{\hspace{-6pt}Multitask Multilingual Learning}\hspace{-8pt}}

\renewcommand*{\thefootnote}{\arabic{footnote}}
\label{chp:gapfill}
\begin{abstract}
\absprelude
Multitask learning and multilingual learning share many similarities, and partially build on the same assumptions.
One such assumption is that similarities between tasks, or between languages, can be exploited with beneficial effects.
A natural extension of these two separate paradigms is to combine them, in order to take advantage of such similarities across both modalities simultaneously.
In this chapter, such a combined paradigm is explored, with the goal of building \textit{One Model to rule them all}.
The pilot experiments presented here take a first step in this direction, by looking at Part-of-Speech tagging and dependency relation labelling for a large selection of source languages.
We restrict ourselves to looking at three target languages representing some typological variety: Finnish, Italian, and Slovene.
Furthermore, we run experiments with a relatively simple model, using simple hard parameter sharing, and multilingual input representations.
In spite of this simplicity, promising results are obtained for these three languages.
For instance, a model which has not seen a single target language PoS tag performs almost equally to a model trained on target language PoS tagging only.
\end{abstract}

\clearpage
\section{Combining Multitask Learning and Multilinguality}
\vspace{-0.2cm}
While Part I of this thesis focussed on multitask learning (MTL), and Part II focussed on multilingual learning (MLL), we now turn to a combined paradigm.
In other words, in this final chapter of the thesis, we both consider several tasks and languages at the same time.
Let us first consider why such an approach might be useful.
For one, joint multilingual and multitask learning allows for taking advantage of the increasing amount of multilingual corpora with overlapping annotation layers, such as the Universal Dependencies \citep{ud20}, and the Parallel Meaning Bank \citep{pmb}.\footnote{In particular, we are interested in the fact that several languages have annotations within the same theoretical framework, following the same annotation guidelines, rather than a single language having several layers of annotation.}
Hence, this approach might significantly reduce \textit{data waste}, as one traditionally only considers a single task--language combination at a time.
An additional advantage of this paradigm is that it opens up for simultaneous model transfer between languages and tasks.
This essentially allows for applying zero-shot learning, in the sense of predicting labels for an unseen task--language combination while taking advantage of other task--language combinations.\footnote{I.e., zero-shot learning in a similar sense to \citet{google:zeroshot}.}
This approach has not been the subject of much attention in the field, perhaps due to its reliance on the combination of both MTL and MLL, which have only recently become popular.
Another issue is the fact that such combined systems put rather large demands on both access to data (alleviated by the UD project), and access to sufficient computing resources.
Although a full exploration of the possibilities of this paradigm is not carried out in this chapter, we do take a first step in this direction.
The main aim of this chapter is to provide an answer to the following research question, in order to answer \textbf{RQ~\ref{rq:mmmt}}.
\begin{enumerate}
\setlength{\itemindent}{.29cm}
    \item [{\bf RQ~\ref{rq:mmmt}a}] To what extent can a combined MTL/MLL system generate sensible predictions for an unseen task--language combination?
\end{enumerate}
Answering this question is considered as a step in the direction of \textit{One Model to rule them all}.
If successful, this will allow for bootstrapping off of more-or-less related languages and tasks,
which in turn will be highly useful for both low-resource languages and low-resource tasks.

\subsection*{Related work}
\label{sec:relwork}

For related work on the separate paradigms of multilingual and multitask learning, the reader is referred to Chapter~\ref{chp:mtl_bg}.
In this chapter, we are concerned with a combined multilingual and multitask learning paradigm.
Although little work has been done in multilingual multitask NLP, \citet{yang:2016} make some preliminary experiments in this direction by contrasting the two approaches, experimenting with NER, PoS tagging, and chunking on English, Dutch, and Spanish.
Their approach uses hard parameter sharing for certain layers, either between languages, or between tasks.
In the case of monotask MLL, they share character embeddings and weights of a character-based RNN, whereas in their monolingual MTL setup, they attach a task-specific conditional random field for each task.
Since their MLL setup depends on sharing character-based features, the approach is restricted to relatively related languages, and is not likely to work well for less related ones.
Indeed, \citet{yang:2016} apply their method only to the relatively closely-related languages English, Dutch and Spanish.
Recent work by \citet{fang:2017} exploits bilingual dictionaries in order to obtain cross-lingual embeddings, which are used to train a PoS tagger for a source language, which is then applied to a target language with embeddings in the same space.

This chapter expands on previous work by unifying MTL and MLL in a single system, using hard parameter sharing.
This allows for taking advantage of similarities between tasks and between languages simultaneously.
Rather than sharing character-level features, we focus on using multilingual word-level input representations.
One advantage of avoiding character-level features in a setup using hard parameter sharing, is that reliance on morphological similarities is reduced, which might otherwise lead to negative transfer when considering distantly related languages.
The work presented here therefore differs from \citet{yang:2016} in two main ways:
i) our method is not restricted to morphologically similar languages, and is applicable to a large portion of (combinations of) the languages in the world;
ii) we aim to combine a vast amount of data sources to generate reasonable predictions for a given unobserved task--language pair.
Additionally, our motivations are quite different.
Where \citet{yang:2016} aim to improve performance on a task--language pair for which annotated data exists, by adding a distant supervision signal from a different language for the same task, or a different task for the same language, we aim to induce tags for task--language combinations for which no annotated data exists.
This is important, since a multitask multilingual setting will usually resemble the scenario depicted in Table~\ref{tab:gapfill_illustration}.\footnote{We will refer to such tables as \textit{gap tables}, as they contain some filled (black) cells, and several white \textit{gaps} without data.}
Let us consider language $l_6$ and task $t_3$, as highlighted in red in the table.
In order to fill this \textit{gap}, we can choose from a few approaches:
i) Spend an enormous effort in finding, hiring and training annotators;
ii) Apply annotation projection to the text snippets which happen to be parallel text with languages for which annotation exists for $t_3$, or first translate the data from $l_6$ to such a language (cf. the translation approach described in Chapter~\ref{chp:mtl_bg});\footnote{Note that the requirements for annotated/parallel data are quite high in this case (cf. \citet{tiedemannetal:2014}).}
iii) Train a multilingual system with supervision from only languages with annotation for $t_3$ (cross-lingual model transfer);
or iv) Train a system on several task--language pairs in the matrix, including $l_7$ for other tasks.\footnote{The first three approaches can be considered traditional approaches, and are detailed in Chapter~\ref{chp:mtl_bg}.}
Our approach is essentially this final approach (iv).

\begin{table}[htbp]
  \caption{\label{tab:gapfill_illustration}Black cells indicate the availability of annotated data for a given task--language pair. Some languages have (almost) all cells filled, whereas some have a large amount of gaps.
  The red cell indicates a potential target task--language combination, for which no annotated data exists.}
    \centering
    \small
    \begin{tabular}{llllllllllllllll}
            & $l_1$ & $l_2$ & $l_3$ & $l_4$ & $l_5$ & $l_6$ &  $l_7$ & $l_8$ & $l_9$ & $l_{10}$ & $l_{11}$ & $l_{12}$ & $\cdots$ & $l_n$ \\
   $t_{1}$  &	 &	 &	 &	\cellcolor{black!80} &	\cellcolor{black!80} &	\cellcolor{black!80} &	\cellcolor{black!80} &	\cellcolor{black!80} &	\cellcolor{black!80} &\cellcolor{black!80} &	 &	\cellcolor{black!80} &	 &	\cellcolor{black!80}  \\
   $t_{2}$  &	 &	\cellcolor{black!80} &	\cellcolor{black!80} &	\cellcolor{black!80} &	\cellcolor{black!80} &	\cellcolor{black!80} &	\cellcolor{black!80} &\cellcolor{black!80} &	\cellcolor{black!80} &	\cellcolor{black!80} &	\cellcolor{black!80} &	\cellcolor{black!80} &	\cellcolor{black!80} &	\cellcolor{black!80} \\
   $t_{3}$  &	 &	 &	\cellcolor{black!80} &	\cellcolor{black!80} &	\cellcolor{black!80} &	\cellcolor{red!80} &	\cellcolor{black!80} &	\cellcolor{black!80} &	\cellcolor{black!80} &	 &	\cellcolor{black!80} &	\cellcolor{black!80} &	\cellcolor{black!80} &	 \\
   $t_{4}$  &	\cellcolor{black!80} &	\cellcolor{black!80} &	\cellcolor{black!80} &	\cellcolor{black!80} &	\cellcolor{black!80} &	\cellcolor{black!80} &	\cellcolor{black!80} &	\cellcolor{black!80} &	 &	\cellcolor{black!80} &	\cellcolor{black!80} &	 &	\cellcolor{black!80} &	\cellcolor{black!80}  \\
   $t_{5}$  &	\cellcolor{black!80} &	\cellcolor{black!80} &	\cellcolor{black!80} &	\cellcolor{black!80} &	\cellcolor{black!80} &	\cellcolor{black!80} &	\cellcolor{black!80} &	\cellcolor{black!80} &	 &	 &	\cellcolor{black!80} &	 &	\cellcolor{black!80} &	\cellcolor{black!80}  \\
   $t_{6}$  &	\cellcolor{black!80} &	\cellcolor{black!80} &	\cellcolor{black!80} &	\cellcolor{black!80} &	 &	 &	\cellcolor{black!80} &	\cellcolor{black!80} &	\cellcolor{black!80} &	\cellcolor{black!80} &	\cellcolor{black!80} &	\cellcolor{black!80} &	\cellcolor{black!80} &	\cellcolor{black!80} \\
   $t_{7}$  &	\cellcolor{black!80} &	\cellcolor{black!80} &	\cellcolor{black!80} &	\cellcolor{black!80} &	\cellcolor{black!80} &	\cellcolor{black!80} &	\cellcolor{black!80} &	\cellcolor{black!80} &	\cellcolor{black!80} &	\cellcolor{black!80} &	\cellcolor{black!80} &	\cellcolor{black!80} &	\cellcolor{black!80} &	\cellcolor{black!80} \\
   $t_{8}$  &	\cellcolor{black!80} &	\cellcolor{black!80} &	 &	\cellcolor{black!80} &	\cellcolor{black!80} &	\cellcolor{black!80} &	\cellcolor{black!80} &\cellcolor{black!80} &	\cellcolor{black!80} &	\cellcolor{black!80} &	\cellcolor{black!80} &	\cellcolor{black!80} &	\cellcolor{black!80} &	\cellcolor{black!80} \\
   $t_{9}$  &	\cellcolor{black!80} &	\cellcolor{black!80} &	 &	\cellcolor{black!80} &	\cellcolor{black!80} &	\cellcolor{black!80} &	\cellcolor{black!80} &	 &	\cellcolor{black!80} &	\cellcolor{black!80} &	\cellcolor{black!80} &	\cellcolor{black!80} &	\cellcolor{black!80} &	\cellcolor{black!80}  \\
   $t_{10}$ &	 &	 &	 &	\cellcolor{black!80} &	 &	\cellcolor{black!80} &	\cellcolor{black!80} &	 &	 &	 &	\cellcolor{black!80} &	\cellcolor{black!80} &\cellcolor{black!80} &	\cellcolor{black!80}\\
   $t_{11}$ &	 &	 &	 &	 &	 &	 &	 &	\cellcolor{black!80} &	 &	 &	\cellcolor{black!80} &	 &	\cellcolor{black!80} &	\cellcolor{black!80}  \\
   $t_{12}$ &	 &	\cellcolor{black!80} &	\cellcolor{black!80} &	 &	 &	 &	\cellcolor{black!80} &	 &	 &	 &	\cellcolor{black!80} &	 &	 &	\\
   $\cdots$ &	 &	 &	 &	\cellcolor{black!80} &	\cellcolor{black!80} &	 &	 &	 &	 &	\cellcolor{black!80} &	 &	 &	 &	 \\
   $t_{n}$  &	 &	 &	 &	 &	\cellcolor{black!80} &	 &	 &	 &	 &	 &	 &	 &	 &	 \\

    \end{tabular}
\end{table}

\section{Data}

\subsection{Labelled data}
While the goal is to extend this approach to a matrix such as in Table~\ref{tab:gapfill_illustration}, we will only look at two tasks in this chapter.
Additionally, since the goal in this pilot experiment is to see whether the proposed MTL/MLL paradigm is at all feasible, a setting in which very high correlations between main and auxiliary tasks can be found was chosen.
We therefore focus on PoS tagging and dependency relation (DepRel) labelling, as data for both of these tasks is available for a relatively large amount of languages through the Universal Dependencies (UD) project.
There are many possible ways of defining dependency relation labels, and in this chapter we use the \textit{simple/simple} paradigm described in Chapter~\ref{chp:mtl} (Table~\ref{tab:supertags}).
Furthermore, positive results have been obtained for this particular task combination, e.g., in Chapter~\ref{chp:mtl}.
In the experiments of this chapter, UD v1.2 is used \citep{ud12}.

\subsection{Unlabelled data}
We take a similar approach to enabling model multilinguality as in Part III.
That is to say, we use unified input representations, in the sense that we use multilingual word embeddings.
Although many options exist, we use multilingual skip-gram \citep{multisg}, for the same reasons as in Part III.

We train the embeddings in two resource settings.
The first is a high resource setting, in which we have access to a large amount of parallel text.
In this setting, we train embeddings on the Europarl corpus \citep{europarl}.
The second is a low resource setting, in which we only require very limited amounts of parallel data, and train embeddings on a collection of Bible corpora.
The low resource setting is, indeed, truly low-resource, as we only require approximately 140,000 tokens of training data.\footnote{This is the approximate token count for the English New Testament, and is bound to differ for other languages.}
Additionally, although the use of Bible corpora for multilingual representations can be criticised for many reasons, including the specificity of the domain, and the archaicness of the language, this data has the advantage that it is available for more than 1,000 languages.
While this does leave a long tail of approximately 5,000 languages for which such resources are not available, it nonetheless constitutes a leap forward from requiring Europarl-levels of data.

\section{Method}

\subsection{Architecture}

The system used is the same bi-GRU as described in Chapter~\ref{chp:multilingual}.
To recap, the bi-GRU is two layers deep, and uses only word-level multilingual embeddings as input.
The network has two output layers -- one for PoS tags, and one for dependency relation tags.
In the settings in which no dependency relations are observed, the weights of the corresponding layer are left unaltered.
This includes the baseline setting, and the settings with transfer solely from PoS tags.

Although using character-level representations would likely yield higher performance for some cases, there are two main reasons why this is not done.
Primarily, it is likely that using such representations would lead to negative transfer between less related languages.
This would need to be dealt with in a more sophisticated way than simple hard parameter sharing, for instance by using sluice networks, in which the parameter sharing itself is learnt \citep{sluice}.
Additionally, the goal of the experiments in this chapter is not to obtain the highest possible results, which is the trend in much of current NLP, but rather to investigate the differences between different transfer settings.

\subsection{Hyperparameters}
Hyperparameters were tuned to a small extent on the English development set when training on only English PoS tags, with the goal of asserting that the system performs on-par with the word-based Bi-GRU in Chapter~\ref{chp:semtag}.
The aim was to perform a relatively low amount of tuning, keeping parameters at fairly standard values.
These hyperparameters were used for all experiments in this chapter.
We use rectified linear units (ReLUs) for all activation functions \citep{relu}.
We apply dropout ($p=0.2$) at the input level, and recurrent dropout \citep{recurrent_do} between the layers in the network.
We use the Adam optimisation algorithm \citep{adam} with a batch-size of 10 sentences (randomly sampled from all source languages under consideration).
Training is done over a maximum of 50 epochs, using early stopping monitoring the loss on development sets of all source languages in the given experimental condition.
The weighting parameter $\lambda$, defining the weight of the auxiliary task is set to $\lambda=1.0$, i.e., weighting the main and auxiliary tasks equally.

\section{Experiments and Analysis}

For the purposes of evaluating whether the proposed approach is feasible, we consider several potential scenarios.
In all experiments, we look at filling the \textit{gap} of Finnish, Italian, and Slovene PoS tags.
These languages were chosen so as to represent some level of typological diversity, with one language from outside the Indo-European family (Finnish), and two fairly dissimilar Indo-European languages, of which one is a Romance language (Italian), and one is a Slavic language (Slovene).
The evaluation metric used in the experiments is the accuracy of PoS tagging on each of these languages, as evaluated on their UD development sets.\footnote{No tuning is performed on this set.}
We will successively increase the amount of, and variety of, data which the models are trained on.
We will also investigate the effect of adding more or less related languages to the training data.
Language relatedness is displayed in Table~\ref{tab:gap_languages}, and is defined heuristically, based on typological relatedness.

Every system is trained on the concatenation of the entire training set of all source languages involved in the setup at hand.
Validation is done on the concatenation of the development sets of all source languages in the setup at hand.

\begin{table}[p]
    \centering
    \caption{Source language overview table. Columns indicate whether the languages are considered to be related to the header of that column.}
    \label{tab:gap_languages}
    \begin{tabular}{rrrr}
        \toprule
        \textbf{Language} & \textbf{Finnish} & \textbf{Italian} & \textbf{Slovene} \\
        \midrule
        Basque             & & & \\
        Bulgarian             & & & x \\
        Croatian             & & & x \\
        Czech             & & & x \\
        Danish             & & & \\
        Dutch             & & & \\
        English             & & & \\
        Estonian             & x & & \\
        French             & & x & \\
        German             & & & \\
        Hebrew             & & & \\
        Hindi             & & & \\
        Hungarian             & x & & \\
        Indonesian             & & & \\
        Irish             & & & \\
        Kazakh             & & & \\
        Latin             & & x & \\
        Greek             & & & \\
        Norwegian             & & & \\
        Persian             & & & \\
        Portuguese             & & x & \\
        Romanian             & & x & \\
        Spanish             & & x & \\
        Swedish             & & & \\
        Tamil             & & & \\
        Turkish             & & & \\
        \bottomrule
    \end{tabular}
\end{table}

\subsubsection*{Training on English PoS}
Throughout the experimental overview, we will consider a gapped table as shown in Table~\ref{tab:gap_a}.
The black cells denote the source task--language combinations, and the red cells denote the target task--language combinations.
Note that, although we could train a joint system for all target languages, separate systems are trained for each target language.
This is because that, in following experiments, we look at languages which are related to the target languages to a smaller or larger extent.
As our target languages are typologically quite different, this requires us to train separate systems, as it would otherwise be impossible to add a language which is equally related to, e.g., both Finnish and Italian.
In the first experiment, we train on English PoS tags only, and evaluate on Finnish, Italian, and Slovene (Table~\ref{tab:gap_a}).
We also train a monolingual baseline system for each of the target languages which is used throughout this chapter, with the same general setup as the other systems, using the high-resource multilingual embeddings.

\begin{table}[htbp]
    \centering\small
    \caption{Gap table -- Training on English, PoS only. Evaluation is on the target languages Finnish, Italian, and Slovene.}
    \label{tab:gap_a}
    \begin{tabular}{rrr}
        \toprule
        \textbf{Language} & \textbf{PoS} & \textbf{DepRel} \\
        \midrule
        English           & \cellcolor{black!80} & \\
        Target languages  & \cellcolor{red!80} & \\
        \bottomrule
    \end{tabular}
\end{table}

\begin{figure}[h!]
    \centering
    \includegraphics[width=1.0\textwidth]{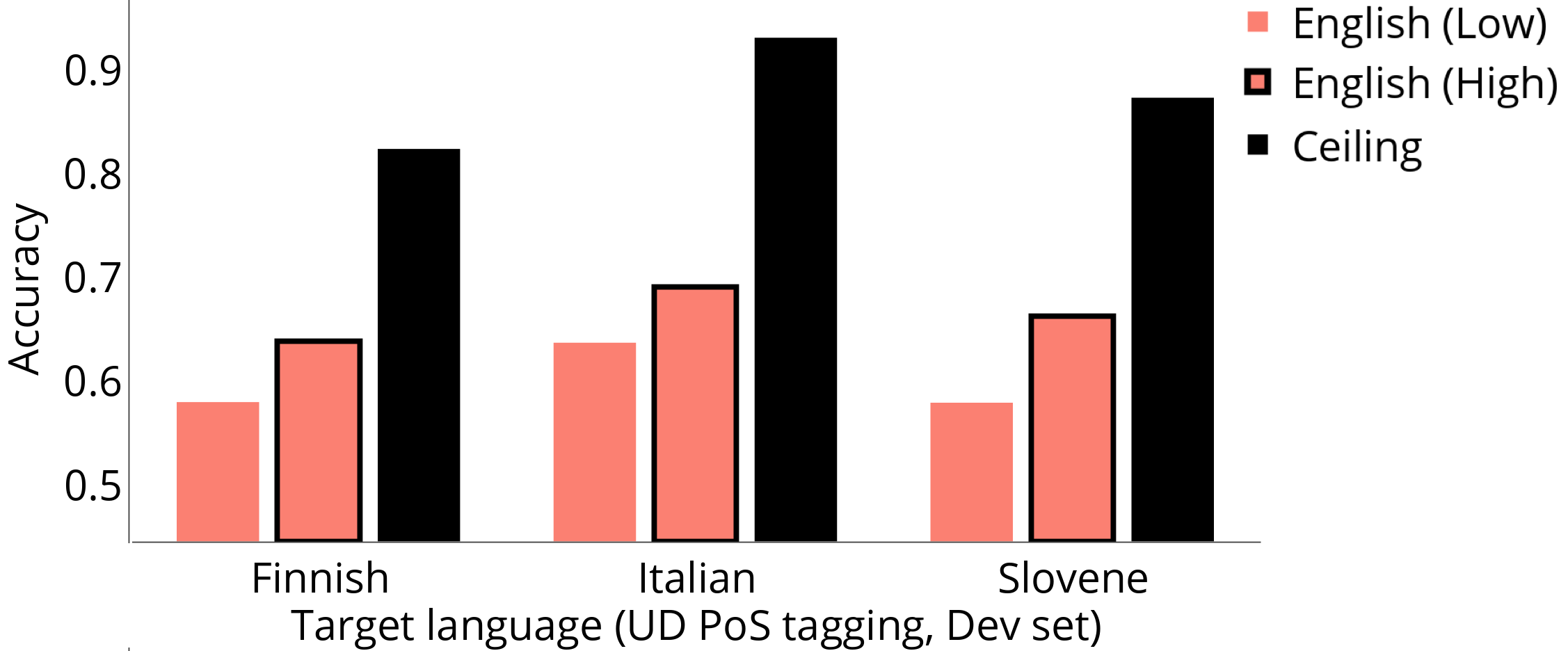}
    \caption{Red bars indicate training on English PoS. No border indicates training with low resource embeddings, and a black border indicates training with high resource embeddings. The black bars denote the monolingual baselines.}
    \label{fig:mmmt_a}
\end{figure}

The results from this setting can be observed in Figure~\ref{fig:mmmt_a}.
The red bars indicate the systems trained on English PoS, with a black border around the system using high resource embeddings, and no border for the system using low resource embeddings.
The black bars indicate the monolingual baseline systems.
Not surprisingly, transfer from English is not particularly successful, with performance far below baseline.
This shows that training on a single relatively unrelated language is not sufficient in this setting.
As expected, results when using embeddings trained on Europarl are somewhat higher than when using low resource embeddings.

\subsubsection*{Training on PoS from several languages}
Next, we add PoS training data from several languages, all relatively unrelated to the target languages.
In this setting, the system for each target language is trained on all languages labelled as unrelated to the source language in Table~\ref{tab:gap_languages}, as depicted in Table~\ref{tab:gap_b}.

\begin{table}[htbp]
    \centering\small
    \caption{Gap table -- Training on unrelated languages, PoS only.}
    \label{tab:gap_b}
    \begin{tabular}{rrr}
        \toprule
        \textbf{Language} & \textbf{PoS} & \textbf{DepRel} \\
        \midrule
        Unrelated languages   & \cellcolor{black!80} & \\
        Target languages           & \cellcolor{red!80} & \\
        \bottomrule
    \end{tabular}
\end{table}

\begin{figure}[h!]
    \centering
    \includegraphics[width=1.0\textwidth]{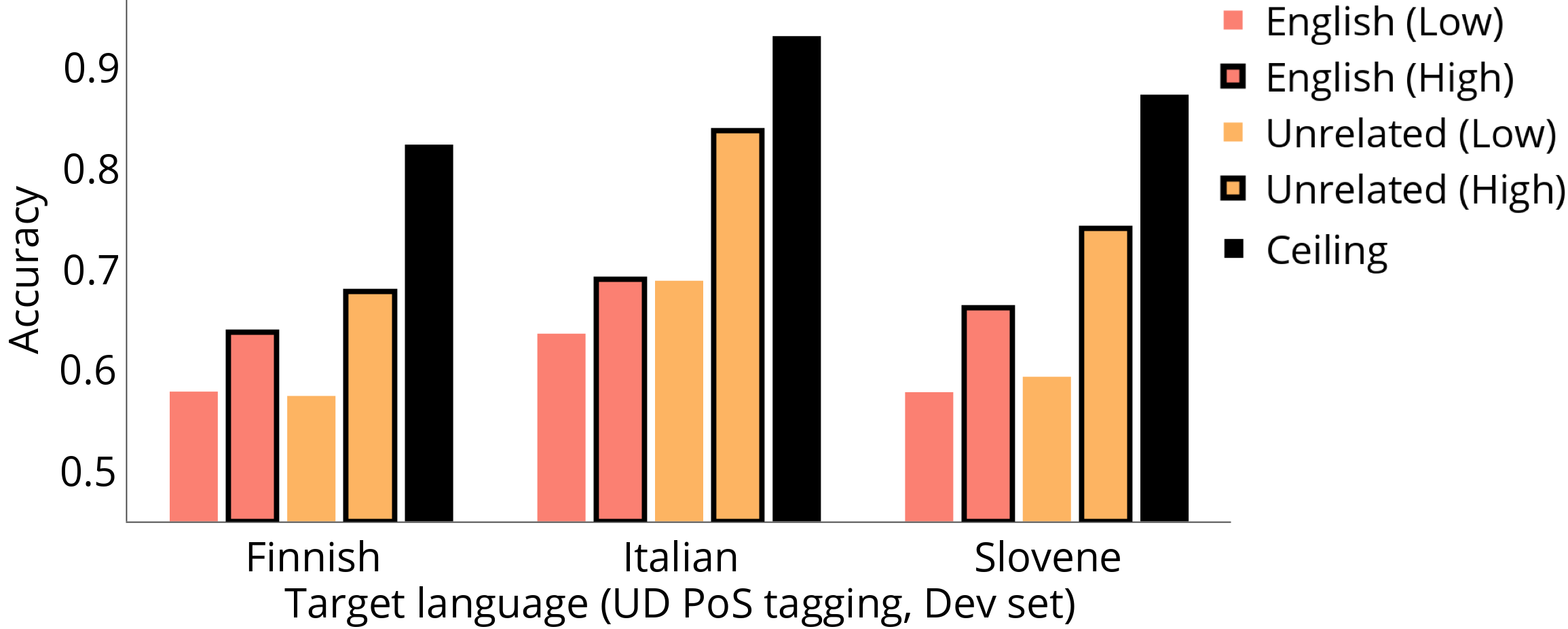}
    \caption{Red bars indicate training on English PoS. Orange bars indicate adding unrelated language PoS. No border indicates training with low resource embeddings, and a black border indicates training with high resource embeddings. The black bars denote the monolingual baselines.}
    \label{fig:mmmt_b}
\end{figure}

The results can be seen in Figure~\ref{fig:mmmt_b}.
Adding more languages to the training material does not affect results noticeably for Finnish.
For Italian and Slovene, however, the results improve somewhat, most notably when using high resource embeddings.
This might be due to the fact that the UD dataset contains an Indo-European bias, meaning that the so-called unrelated languages which we have added still share fairly distant ancestry.
We also see a rather large increase in the performance on Italian as compared to Slovene.
This can be explained by the fact that many of the unrelated languages added in this setting are Germanic.
Morphological complexity of Germanic languages is arguably relatively similar to Romance languages, such as Italian.
On the other hand, Slavic languages such as Slovene are much more morphologically complex.
This might have an effect on the quality of the multilingual word embeddings, leading to training on Germanic languages being more beneficial for Italian than it is for Slovene.
Finnish, being from the Finno-Ugric language branch, does not benefit from this setting, perhaps due to its typological distance from the added languages being larger.

\subsubsection*{Adding source language dependency relations}

We now add dependency relation data for the same collection of languages as in the previous setting (see Table~\ref{tab:gap_c}).
This is the first setting in which MTL is combined with the MLL experiments, as the network is now trained on both PoS tagging and dependency relation labelling.
The idea is that the correlations between PoS tags and DepRel labels can be learnt by the network in an implicit manner, which might be beneficial for system performance.
However, we do not expect positive results in this particular setting, considering that the mutual information between PoS tags and dependency relations is relatively high, and we are not adding any extra data (see Chapter~\ref{chp:mtl}).

\begin{table}[htbp]
    \centering\small
    \caption{Gap table -- Training on unrelated languages, PoS and dependency relations.}
    \label{tab:gap_c}
    \begin{tabular}{rrr}
        \toprule
        \textbf{Language} & \textbf{PoS} & \textbf{DepRel} \\
        \midrule
        Unrelated languages   & \cellcolor{black!80} & \cellcolor{black!80} \\
        Target languages   & \cellcolor{red!80} & \\
        \bottomrule
    \end{tabular}
\end{table}

\begin{figure}[h!]
    \centering
    \includegraphics[width=1.0\textwidth]{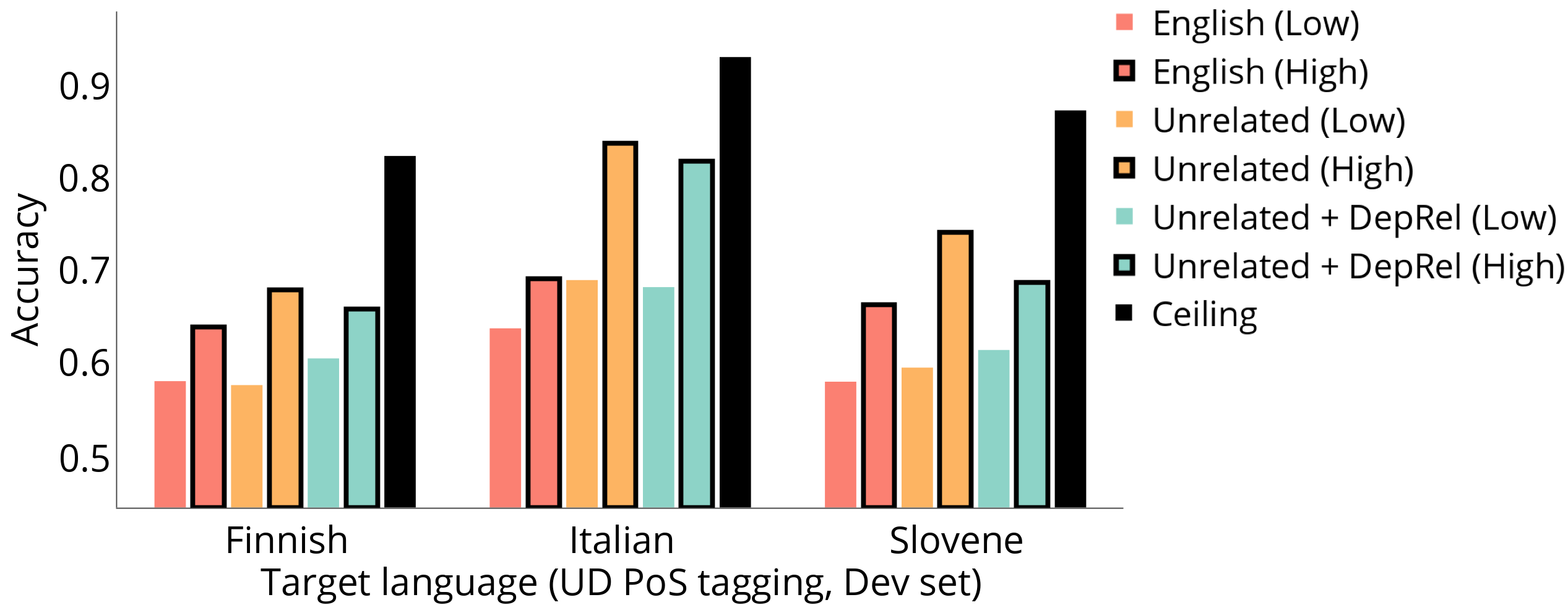}
    \caption{
    Green bars indicate adding source language dependency relations.
    No border indicates training with low resource embeddings, and a black border indicates training with high resource embeddings.
    }
    \label{fig:mmmt_c}
\end{figure}

Figure~\ref{fig:mmmt_c} shows that, indeed, this addition does not affect results to a large extent.
In fact, results drop somewhat in most settings, which may be owed to the fact that some of the net capacity is wasted, since two tasks need to be learned.
As expected, since the system has not seen any data for either task for the target languages, adding this data does not improve much, which can be explained by the findings in Chapter~\ref{chp:mtl}.

\subsubsection*{Adding target language dependency relations}

In this experiment, we add dependency relation data for the target languages in training (Table~\ref{tab:gap_d}).
The intuition behind this, is that the neural network ought to be able to make use of the implicitly learn correlations between tasks, thus learning to produce sensible PoS tags for the target languages, in spite of never having actually observed such tags.
In a sense, this is the first real combined MTL/MLL experiment in this chapter.

\begin{table}[htbp]
    \centering\small
    \caption{Gap table -- Training on English and unrelated languages, PoS and dependency relations.}
    \label{tab:gap_d}
    \begin{tabular}{rrr}
        \toprule
        \textbf{Language} & \textbf{PoS} & \textbf{DepRel} \\
        \midrule
        Unrelated languages   & \cellcolor{black!80} & \cellcolor{black!80} \\
        Target languages  & \cellcolor{red!80} & \cellcolor{black!80} \\
        \bottomrule
    \end{tabular}
\end{table}

\begin{figure}[htbp]
    \centering
    \includegraphics[width=1.0\textwidth]{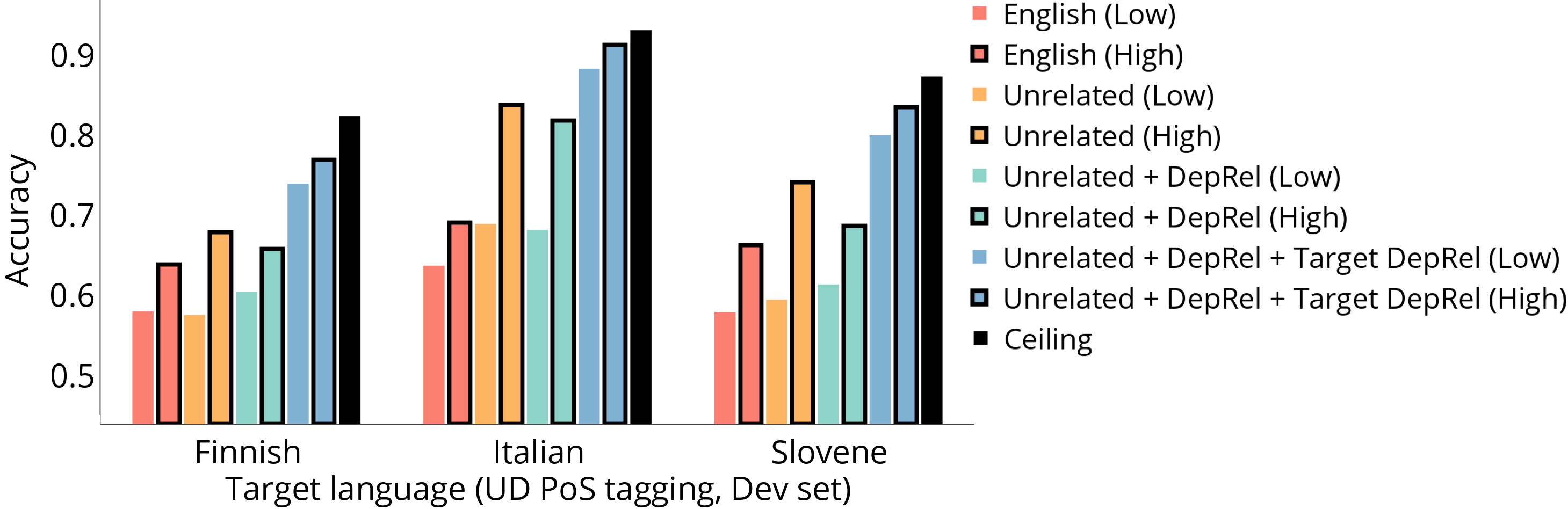}
    \caption{
    Blue bars indicate adding target language dependency relations.
    No border indicates training with low resource embeddings, and a black border indicates training with high resource embeddings.
    }
    \label{fig:mmmt_d}
\end{figure}

Results in Figure~\ref{fig:mmmt_d} show high resource embeddings almost reaching ceiling performance.
This can be interpreted as showing that the network has learned the correlations between the two tasks, allowing for generating sensible PoS tags for the target languages.
Another potential explanation is that adding extra data with high mutual information with PoS tagging ought to be useful, based on the findings in Chapter~\ref{chp:mtl}.

\subsubsection*{Adding related languages, no target dependency relations}

We here add data for related languages, as defined in Table~\ref{tab:gap_languages}.
This is the same as the third experimental setting (Table~\ref{tab:gap_c}), except we also look at related languages (depicted in Table~\ref{tab:gap_e}).
That is to say, we do not see any dependency relation tags for the target languages in this setting.

\begin{table}[htbp]
    \centering\small
    \caption{Gap table -- Training on unrelated and related languages, PoS and dependency relations.}
    \label{tab:gap_e}
    \begin{tabular}{rrr}
        \toprule
        \textbf{Language} & \textbf{PoS} & \textbf{DepRel} \\
        \midrule
        Unrelated languages   & \cellcolor{black!80} & \cellcolor{black!80} \\
        Related languages & \cellcolor{black!80} & \cellcolor{black!80} \\
        Target languages  & \cellcolor{red!80} & \\
        \bottomrule
    \end{tabular}
\end{table}

\begin{figure}[h!]
    \centering
    \includegraphics[width=1.0\textwidth]{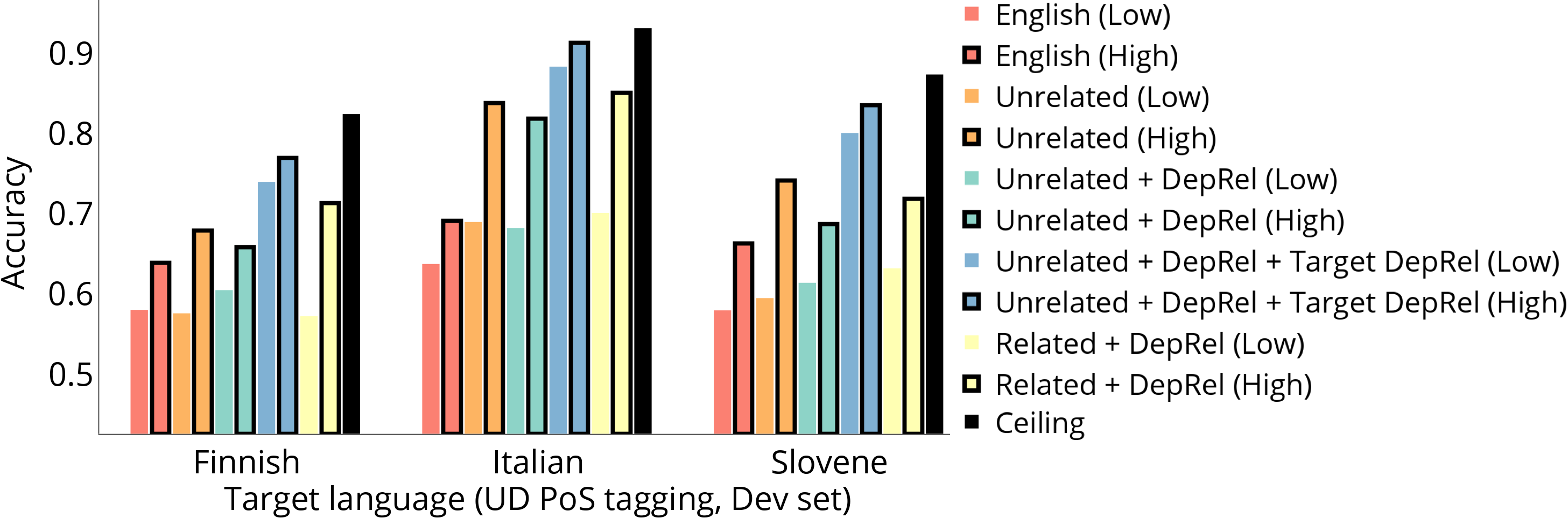}
    \caption{
    Green bars indicate adding source language dependency relations.
    Yellow bars indicate training on related languages.
    No border indicates training with low resource embeddings, and a black border indicates training with high resource embeddings.
    }
    \label{fig:mmmt_e}
\end{figure}

Comparing the yellow and green bars in Figure~\ref{fig:mmmt_e}, the change in results is not as large as what might have been anticipated.
This is somewhat surprising, as one could expect adding related languages to the mix to improve results significantly.
Nonetheless, especially in the high resource setting, some gains can be observed.
These results support the findings of Chapter~\ref{chp:multilingual}, in that training on similar languages can be beneficial in a multilingual scenario.
The relative gain for Finnish is especially high, which can be explained by the fact that the model finally has access to source data which to some extent resembles the target data.
As for Slovene and Italian, the gains in getting access to Slavic and Romance data, respectively, does not provide a very large benefit as compared to having access to Indo-European data.
As for the low resource settings in this experiment, very small differences can be observed.
This hints at the possibility that, without access to any target language data, the Bible-based embeddings are close to a performance ceiling.

\subsubsection*{Adding related languages, with target dependency relations}

Finally, we also add in the dependency relation data for the target languages (Table~\ref{tab:gap_f}).
This denotes the most complete experimental setting, as we only have a single gap to fill in the table, and have access to the largest possible amount of data.

\begin{table}[htbp]
    \centering\small
    \caption{Gap table -- Training on unrelated and related languages, PoS and dependency relations, as well as target language dependency relations.}
    \label{tab:gap_f}
    \begin{tabular}{rrr}
        \toprule
        \textbf{Language} & \textbf{PoS} & \textbf{DepRel} \\
        \midrule
        Unrelated languages   & \cellcolor{black!80} & \cellcolor{black!80} \\
        Related languages & \cellcolor{black!80} & \cellcolor{black!80} \\
        Target languages           & \cellcolor{red!80} & \cellcolor{black!80} \\
        \bottomrule
    \end{tabular}
\end{table}

\begin{sidewaysfigure}[p]
    \centering
    \includegraphics[width=\textwidth]{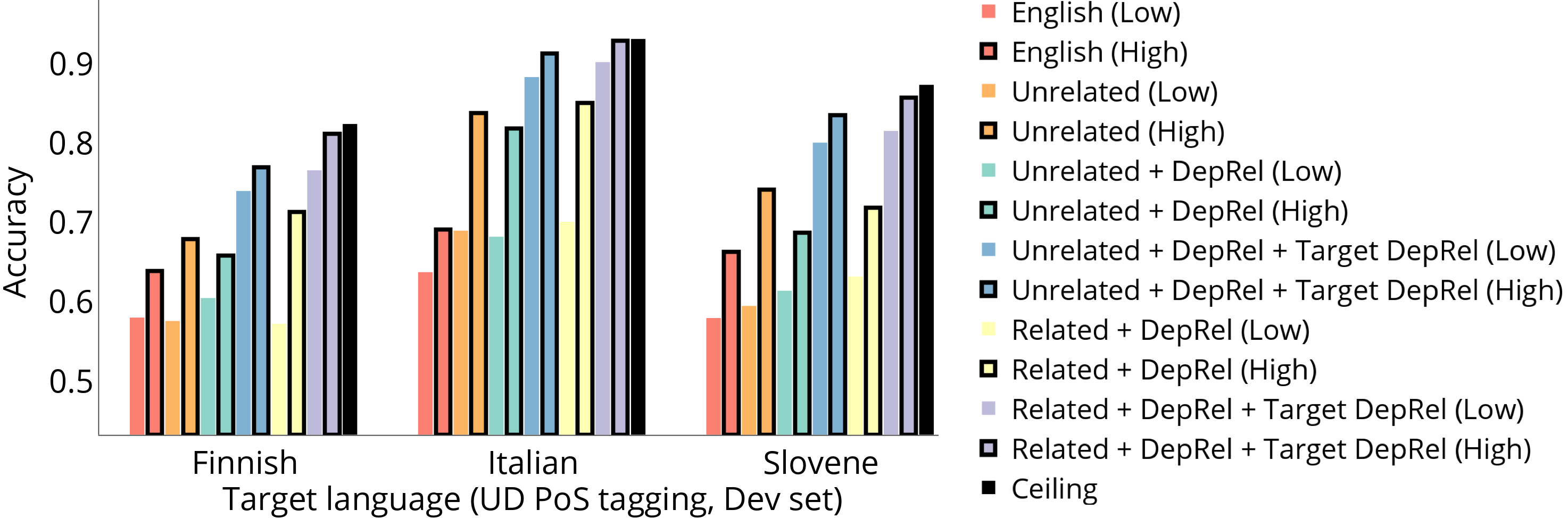}
    \caption{Red bars indicate training on English PoS. Orange bars indicate adding unrelated language PoS. Green bars indicate adding source language dependency relations. Blue bars indicate adding target language dependency relations. Yellow bars indicate training on related languages. Purple bars indicate adding target language dependency relations. No border indicates training with low resource embeddings, and a black border indicates training with high resource embeddings. The black bars denote the monolingual baselines.}
    \label{fig:mmmt_f}
\end{sidewaysfigure}

The results for this experiment are positive for all three languages, showing that it is possible to output PoS tags of decent quality, without having seen a single target-language PoS tag (Figure~\ref{fig:mmmt_f}).
Performance on Italian is especially promising, reaching the same level as the ceiling baseline, while Finnish and Slovene also show positive results.\footnote{An important caveat, however, is the fact that the setup is rather artificial, as one rarely will have dependency relation annotation for a language, without access to PoS tags. This is discussed further in Section~\ref{sec:gapfill_discussion}.}
Notably, although we use training data from related languages, the change in performance between this setting (purple bars) and the corresponding setting with unrelated languages (blue bars) is quite small.
Also noteworthy is the small distance between the two embedding types in this setting.
While ceiling performance is observed when using high resource embeddings, the low resource scenario also yields positive results.
Whereas most of the experiments did not provide much difference in the results with low resource embeddings, the two settings in which we have access to target-language data show that it may be sufficient with this resource scenario.
Should these results generalise to more exotic languages than the ones used in this study, then this type of multitask multilingual learning might indeed be a useful step towards improving NLP for low-resource languages.

\begin{figure}[htbp]
    \includegraphics[width=\textwidth]{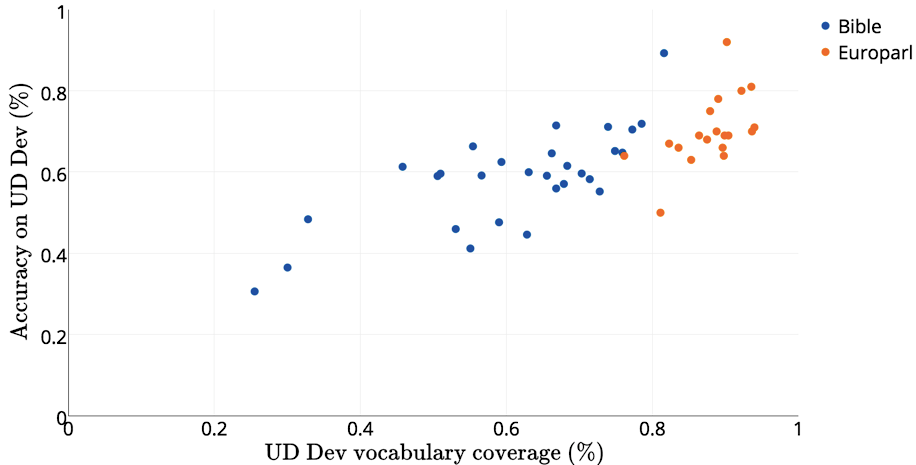}
    \caption{Accuracy of monolingual models on UD Dev compared to the vocabulary coverage of the high and low resource embeddings.}
    \label{fig:low_high_correlation}
\end{figure}

\section{Discussion}
\label{sec:gapfill_discussion}

While the general results in the high-resource scenario were positive, and indicate that the approach taken here is, at the very least, methodologically sound, the results were more varied in the low-resource scenario.
In general, the Bible-based embeddings did not yield any positive results, save for the experiments in which we also had access to some target-language training data.
This might be explained further by taking a look at model performance as compared to vocabulary coverage.
Figure~\ref{fig:low_high_correlation} shows the accuracy of monolingual models on UD Dev compared to the vocabulary coverage of the high resource and low resource embeddings on the same UD Dev set.
These models are trained in the same way as the ceiling baseline from the experimental setup in this chapter.
There is a relatively strong correlation between accuracy and vocabulary coverage, with most of the low-coverage region naturally occupied by the low-resource embeddings.
The fact that the link with vocabulary coverage is as pronounced as what we observe explains the large increases we observe when adding target language training data.
In doing so, we effectively increase the portion of the target language vocabulary which the model has observed, thus increasing performance on target language PoS tags.

While the results of the experiments in this chapter seem promising, there are some points which can be criticised.
One such matter, is the fact that it is hard to imagine a situation which is exactly as what was described here.
The assumption of the experiments was that we did have access to dependency relation annotations for the target languages, but did not have PoS tagged data.
As dependency relations constitute a more detailed level of description, this scenario is most likely not a very common one.
An interesting direction would therefore be to invert this setting, by trying to fill a dependency relation gap.
This is likely much more challenging than the current setup, as the mapping from PoS tags to dependency relations is more heterogenous than the inverse.
However, even though filling a gap for more intricate annotations than one has for a language is an interesting problem, filling a gap with annotations at a similar level as what already exists is also a useful application.
For instance, some languages have their own PoS tagged corpora, while they do not have any UD annotations.
This might be solved by mapping from one PoS tag set to another with the approach described in this chapter.

In spite of the aforementioned issues, the aim of the pilot study is to investigate whether or not the proposed combination of MTL and MLL is at all feasible.
The fact that we have seen positive results in such a simplistic setting, using hard parameter sharing and multilingual input representations, certainly indicates that this is the case.
A potential approach for dealing with languages for which parallel text does not exist in sufficient quantities, is to rely on bilingual dictionaries instead, as done by \citet{fang:2017}.

Refining this approach in the future therefore constitutes a highly interesting research direction, for which some approaches are detailed in the next and final chapter of this thesis, in Section~\ref{sec:final}.

\section{Conclusion}

We attempted to combine the paradigms of multilingual and multitask learning.
Providing the model with data for the target task for source languages, as well as auxiliary task data for the target and source languages, yielded promising results.
In fact, the results are almost on par with training a system directly on the target/source language, indicating that combining the paradigms of MTL and MLL has potential ({\bf RQ~\ref{rq:mmmt}a}).
Although the experimental setup was somewhat artificial, as we assumed access to a more complex level of annotation than what we aimed at producing, this approach constitutes a research direction which is worthwhile pursuing in the future.

\partimage[width=0.7\textwidth]{neuron}
\part{Conclusions}
\chapter[Conclusions]{
\hspace{-6pt}Conclusions\hspace{-8pt}}
\label{chp:conclusions}
\begin{abstract}
\end{abstract}
\noindent
While traditional NLP approaches consider a single task or language at a time, the aim of this thesis was to answer several research questions dealing with pushing past this boundary.
In doing so, the hope is that in the long term, low-resource languages can benefit from the advances made in NLP which are currently to a large extent reserved for high-resource languages.
This, in turn, may then have positive consequences for, e.g., language preservation, as speakers of minority languages will have a lower degree of pressure to using high-resource languages.
In the short term, answering the specific research questions posed should be of use to NLP researchers working towards the same goal.
We will now see the conclusions which can be drawn from each research part of this thesis.

\section{Part II - Multitask Learning}
In the first research part of the thesis, we began by exploring the following research question in Chapter~\ref{chp:semtag}.

\begin{quotation}
    \noindent\textit{'To what extent can a semantic tagging task be informative for other NLP tasks?'}\\
    --{\bf RQ~\ref{rq:stag}}
\end{quotation}

We found that semantic tags are informative for the task of PoS tagging.
Furthermore, the results obtained when exploiting this were state-of-the-art results at the time.
Additionally, we found that using coarse-grained semantic tags was not informative for semantic tagging.
This then raised more questions.
Why were the semantic tags useful for PoS tags, while coarse-grained semantic tags were not useful for semantic tagging?
A look at correlations between the tag sets showed that these were high in both cases.
Coarse-grained semantic tags have a one-to-one mapping with semantic tags.
Semantic tags do not have a one-to-one mapping with PoS tags, but still exhibit large correlations.
The idea was, then, that such high correlations between tag sets might correlate with auxiliary task effectivity, given differing data sets.
Thus, we aimed at answering the next research question in Chapter~\ref{chp:mtl}:

\begin{quotation}
    \noindent\textit{'How can multitask learning effectivity in NLP be quantified?'}\\
    --{\bf RQ~\ref{rq:mtl}}
\end{quotation}

Taking an information-theoretic perspective, we found that these correlations could be quantified fairly well by using mutual information.
Running experiments in various data overlap settings, on a large selection of languages and tasks, showed that the hypothesis was supported.
That is to say, providing the model with different data including annotations which correlate highly with the main task yields gains in performance.
However, providing the model with the same data with such highly correlated auxiliary annotations, does not yield any increase at all.
Intuitively, this makes sense if one thinks about it as follows.
The model has already seen sentence $x$ with some annotation.
Giving it the same sentence $x$ with highly correlated annotation does not give the model anything more to learn from -- after all, this example has practically already been observed!
However, giving the model a different sentence $y$ with highly correlated annotation essentially entails giving the model an extra training example.

\section{Part III - Multilingual Learning}
In the second content part of the thesis, the aim was to investigate similar research questions to Part I, focussing on similarities between \textit{languages} rather than between tasks.
We began by asking the following research question.

\begin{quotation}
    \noindent\textit{'To what extent can multilingual word representations be used to enable zero-shot learning in semantic textual similarity?'}\\
    --{\bf RQ~\ref{rq:sts}}
\end{quotation}

In Chapter~\ref{chp:multiling_sim} we found that a simple language-agnostic feed-forward neural network using multilingual word representations was able to solve the task of semantic textual similarity assessment to some extent.
Although results were below the current state-of-the-art for this task, some useful insights were gained.
Mainly, we found that languages which are more similar to one another are more suited for this approach, indicating that language similarity is important for the effectivity of model multilinguality.
This is similar to the case in MTL, where task relatedness is an important factor, and raised the following research question which we approached in Chapter~\ref{chp:multilingual}.

\begin{quotation}
    \noindent\textit{'In which way can language similarities be quantified, and what correlations can we find between multilingual model performance and language similarities?'}\\
    --{\bf RQ~\ref{rq:multiling}}
\end{quotation}

We looked at correlations between language similarity and multilingual model effectivity in two sequence prediction tasks, namely semantic tagging and PoS tagging, as well as in a sequence-to-sequence task, namely morphological inflection.
The overall results indicate that both measures of language similarity under consideration offer some explanatory value.
One interesting finding in the case of semantic tagging, was the fact that English, Dutch, and German benefitted from having their input representations updated during joint training. Combining these languages with Italian and updated embeddings, however, resulted in a serious drop in performance.
A potential reason for this is that language relatedness plays a large role in maintaining the quality of the multilingual embedding space in such a context.
In future work, it would therefore be interesting to observe the resulting word-space after updating word representations in such a setting.

\section{Part IV - Combining Multitask Learning and Multilinguality}
In the final research part of the thesis, the aim was to probe the possibilities of combining the paradigms of multitask learning and multilingual learning.
Chapter~\ref{chp:multilingual} aimed at providing an answer to the following research question.

\begin{quotation}
    \noindent\textit{'Can a multitask and multilingual approach be combined to generalise across languages and tasks simultaneously?'}\\
    --{\bf RQ~\ref{rq:mmmt}}
\end{quotation}

We looked at predicting labels for an unseen task--language combination, by taking advantage of other task--language combinations.
In the admittedly somewhat artificial setup, the target task was PoS tagging for three languages offering some typological diversity, namely Finnish, Italian, and Slovene.
In a high-resource scenario, assuming access to parallel text similar to Europarl, sensible tags could be produced for the target languages without seeing any annotated data for that target language.
In the low-resource scenario, assuming access to parallel text similar to the New Testament, similar results have the additional requirement of also having access to target-language annotations of some sort.
Finally, access to the high-resource scenario as well as target-language annotations yielded results on par with a monolingual monotask PoS tagger for the target language -- and that without seeing a single PoS tag for the target language.

\section{Final words}
\label{sec:final}

A large part of this thesis was motivated by the intuition that similarities between tasks and languages is one of the most important factors when considering a multitask or a multilingual approach.
Even though some correlations were found in experiments, attempting to correlate measures of task and language similarities with change in model performance, much of the change that is observed is left unaccounted for.
This highlights the case that even if such similarities are important, the situation is more complex than what can be explained purely by measures of correlation.

The successful experiments dealing with the combination of multitask and multilingual learning show the most potential for future research based on this thesis.
A plethora of new studies based on this idea can be imagined.
One could take advantage of morphological similarities by looking at character-level representations, investigating to what extent an architecture such as sluice networks can learn to share parameters for similar task--language combinations.
Another option is to probe into how much annotation is needed in order to bootstrap off of other languages than the target language at hand in order to predict reasonable labels for the target language.

A concrete proposal toward \textit{One Model to rule them all} at a larger scale, involving more languages and tasks, is to model this in a sluice network \citep{sluice}.
In this recently proposed architecture, the sharing of layers itself is learned by the network.
Combining a large amount of tasks in such a network should therefore allow for taking advantage of relevant similarities between tasks, while not sharing parameters in the case of dissimilarities which may lead to negative transfer.
Taking this one step further, by also involving multilingual learning as in this thesis, could also allow for learning between which languages to share parameters.
For instance, this architecture might be able to learn which parts of a character-RNN to share between which languages, for instance learning to only share these parameters between closely related languages, thus avoiding any negative transfer in this setting.
Further combining this approach with language vectors \citep{ostling:langmod,malaviya:2017} might facilitate exploitation of language similarities.
This approach might therefore alleviate many of the problems with hard parameter sharing, by allowing the model to only utilise parameter sharing for similarities between languages, while learning separate parameters for language-specific features.
As the amount of both unannotated parallel data, and annotated data with various universal annotation schemes increases, it is only a matter of choosing the right approach in order to arrive at \textit{One Model to rule them all}.

\partimage[width=0.0\textwidth]{neuron}
\appendix
\appendixpage

\chapter{Correlation figures for all languages in Chapter~\ref{chp:mtl}}
\label{app:langcorr}
\begin{figure}[htbp]
    \includegraphics[width=\textwidth]{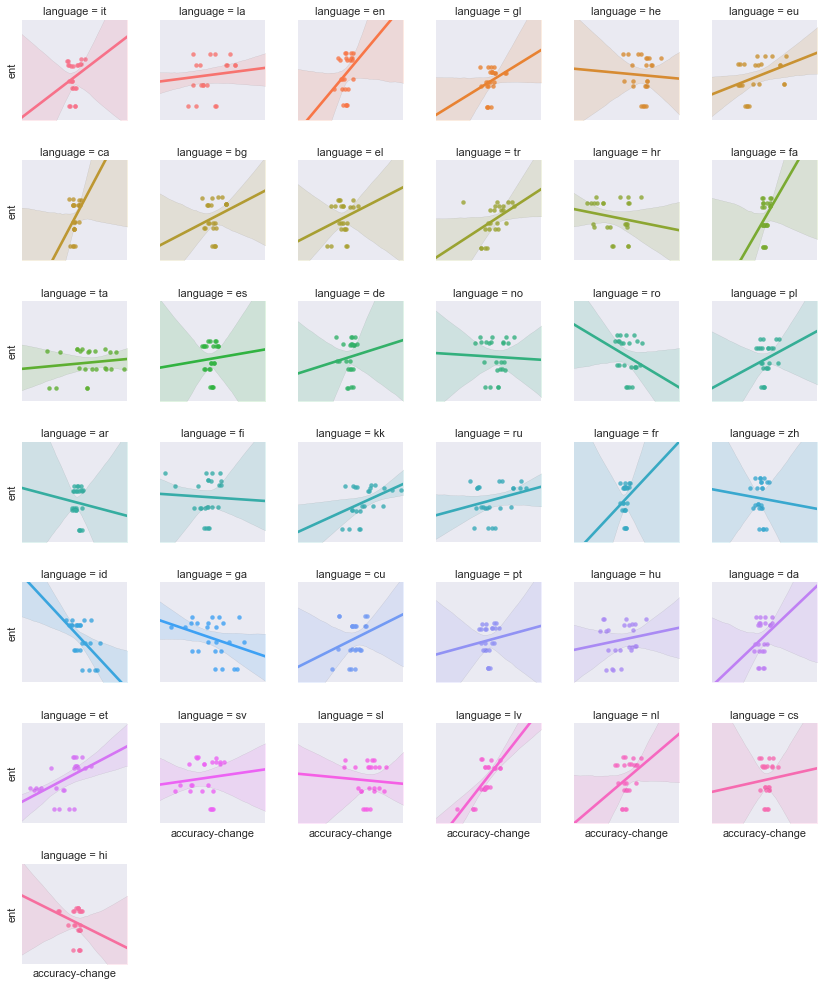}
    \caption{\label{fig:delta_entropy_fully}Correlations between $\Delta_{acc}$ and entropy. Each data point represents a single experiment run.}
\end{figure}

\begin{figure}[htbp]
    \includegraphics[width=\textwidth]{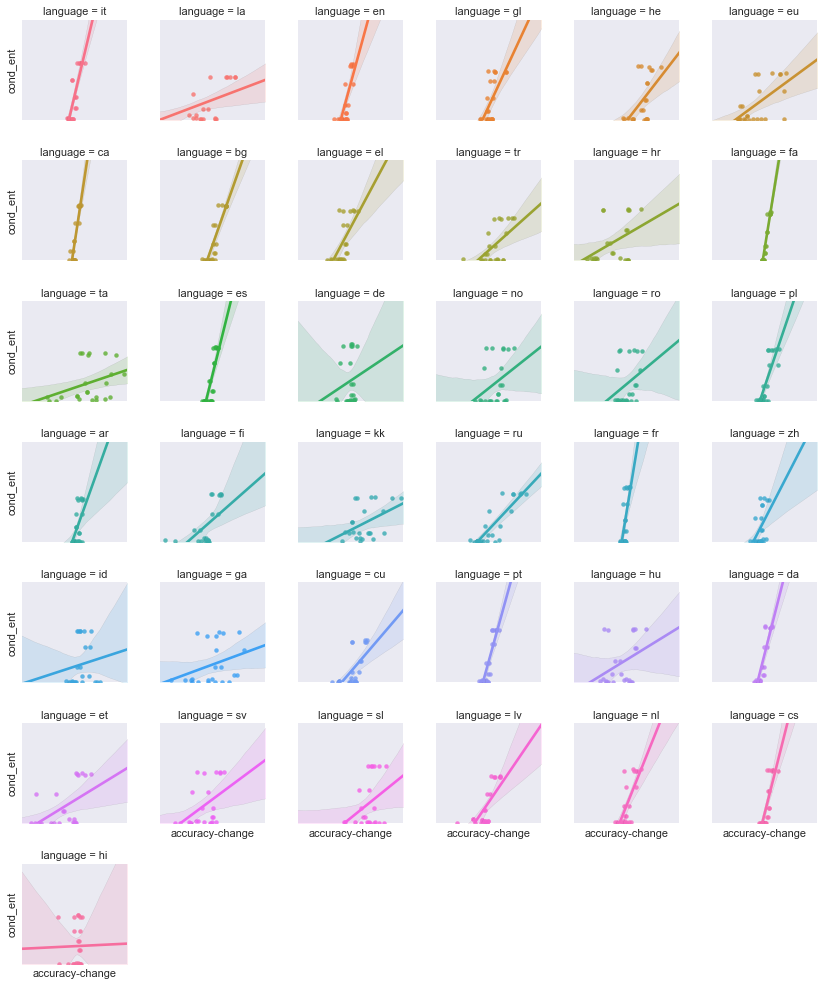} 
    \caption{\label{fig:delta_mi_full}Correlations between $\Delta_{acc}$ and mutual information. Each data point represents a single experiment run.}
\end{figure}

\clearpage
\vspace{-5cm}
\chapter{Bibliographical abbreviations}
\begin{abstract}\end{abstract}
  \vspace{-1cm}

\begin{itemize}
    \setlength\itemsep{-5pt}
    \item AAAI $\rightarrow$ Conference on Artificial Intelligence
    \item ACL $\rightarrow$ Annual Meeting of the Association for Computational Linguistics
    \item COLING $\rightarrow$ International Conference on Computational Linguistics
    \item CoNLL $\rightarrow$ Conference on Computational Natural Language Learning
    \item EACL $\rightarrow$ Conference of the European Chapter of the Association for Computational Linguistics
    \item EMNLP $\rightarrow$ Conference on Empirical Methods in Natural Language Processing
    \item HLT $\rightarrow$ Conference on Human Language Technology
    \item ICLR $\rightarrow$ International Conference on Learning Representations
    \item ICML $\rightarrow$ International Conference on Machine learning
    \item IJCNLP $\rightarrow$ International Joint Conference on Natural Language Processing
    \item LREC $\rightarrow$ Language Resources and Evaluation Conference
    \item NAACL $\rightarrow$ Conference of the North American Chapter of the Association for Computational Linguistics
    \item NIPS $\rightarrow$ Neural Information Processing Systems Conference
    \item NoDaLiDa $\rightarrow$ Nordic Conference on Computational Linguistics
    \item *SEM $\rightarrow$ Joint Conference on Lexical and Computational Semantics
    \item SemEval $\rightarrow$ International Workshop on Semantic Evaluation
\end{itemize}

\backmatter

\chapterstyle{default}
\bibliographystyle{apalike}
\footnotesize{
\bibliography{thesis.bib}
}

\chapterstyle{bjerva}

\clearpage
\normalsize{
\chapter{Summary}
\begin{abstract}
\end{abstract}
\noindent
When learning a new skill, you take advantage of your preexisting skills and knowledge.
For instance, if you are a skilled violinist, you will likely have an easier time learning to play cello.
Similarly, when learning a new language you take advantage of the languages you already speak.
For instance, if your native language is Norwegian and you decide to learn Dutch, the lexical overlap between these two languages will likely benefit your rate of language acquisition.
This thesis deals with the intersection of learning multiple tasks and learning multiple languages in the context of Natural Language Processing (NLP), which can be defined as the study of computational processing of human language.
Although these two types of learning may seem different on the surface, we will see that they share many similarities.

The traditional approach in NLP is to consider a single task for a single language at a time.
However, recent advances allow for broadening this approach, by considering data for multiple tasks and languages simultaneously.
This is an important approach to explore further as the key to improving the reliability of NLP, especially for low-resource languages, is to take advantage of all relevant data whenever possible.
In \textbf{Part I} of this thesis, we begin with an introduction to neural networks with a focus on NLP (Chapter 2), since such architectures are particularly well suited to combined learning of multiple tasks and languages.
We will then look at some ways in which neural networks can consider multiple tasks and languages at the same time (Chapter 3).
Specifically, we will consider multitask learning (MTL), and several common multilingual approaches.

In \textbf{Part II} of this thesis, I look at exploiting the fact that many NLP tasks are highly related to one another.
This is done by experimenting with MTL using hard parameter sharing, which has proven beneficial for a variety of NLP tasks.
In spite of such successes, however, it is not clear \textit{when} or \textit{why} MTL is beneficial in NLP.
Chapter 4 contains a case study in which semantic tagging is shown to be beneficial for POS tagging.
This further highlights the question of when MTL is beneficial in NLP tagging tasks, which is explored using information-theoretic measures in Chapter 5.

Multilingual models can leverage the fact that many languages share commonalities with one another.
These resemblances can occur on various levels, with languages sharing, for instance, syntactic, morphological, or lexical features.
While there are many possibilities for exploiting these commonalities, the focus in this thesis is on using multilingual word representations, as they allow for straight-forward integration in a neural network.
As with MTL, it is not clear in which cases it is an advantage to \textit{go multilingual}.
In \textbf{Part III}, I begin with presenting a case study on multilingual semantic textual similarity (Chapter 6).
Following this, I explore how similar languages need to be, and in which way, in order to for it to be useful to go multilingual (Chapter 7).

In \textbf{Part IV} of this thesis, I experiment with a combined paradigm, in which a neural network is trained on several languages and tasks simultaneously (Chapter 8).
Finally, the thesis is concluded in \textbf{Part V} (Chapter 9).
The experiments in this thesis are run on a large collection of mainly lexically oriented tasks, both semantic and morphosyntactic in nature, and on a total of 60 languages, representing a relatively wide typological range.

While traditional NLP approaches consider a single task or language at a time, the aim of this thesis was to answer several research questions dealing with pushing past this boundary.
In doing so, the hope is that in the long term, low-resource languages can benefit from the advances made in NLP which are currently to a large extent reserved for high-resource languages.
This, in turn, may then have positive consequences for, e.g., language preservation, as speakers of minority languages will have a lower degree of pressure to using high-resource languages.
In the short term, answering the specific research questions posed should be of use to NLP researchers working towards the same goal.

\chapter{Samenvatting}
\begin{abstract}
\end{abstract}
\selectlanguage{dutch}
\noindent
Wanneer je een nieuwe vaardigheid leert maak je gebruik van de vaardigheden en kennis die je al bezit.
Als je bijvoorbeeld een ervaren violist bent, is het waarschijnlijk makkelijker om ook cello te leren spelen.
Ook bij het leren van een nieuwe taal maak je gebruik van de talen die je al beheerst.
Bijvoorbeeld, als je moedertaal Noors is en je besluit dat je Nederlands wil gaan leren, dan maakt het grote aantal woorden dat op elkaar lijkt in die twee talen het waarschijnlijk makkelijker om de nieuwe taal te leren. 
In dit proefschrift kijk ik naar het leren van meerdere taken, het leren van meerdere talen en de combinatie daarvan, in het kader van \textit{natural language processing} (NLP): de computationele analyse van menselijke taal.
Hoewel deze twee soorten van leren aan de oppervlakte misschien anders lijken, zullen we zien dat ze meerdere overeenkomsten hebben.

De traditionele aanpak in NLP is om op één taak voor één taal te focussen.
Nieuwe ontwikkelingen maken het echter mogelijk om deze aanpak uit te breiden, door tegelijkertijd meerdere taken én meerdere talen te bekijken.
Dit is een veelbelovende richting voor verder onderzoek, aangezien het gebruiken van zoveel mogelijk relevante data de sleutel is tot het verbeteren van de betrouwbaarheid van NLP, met name voor kleinere talen.
\textbf{Deel I} van dit proefschrift begint met een inleiding over neurale netwerken, met een focus op NLP (Hoofdstuk~\ref{chp:nn}). Deze architecturen zijn bijzonder geschikt voor het gecombineerd leren van meerdere taken en talen.
Daarna zien we hoe gecombineerd leren werkt, door een aantal manieren waarop neurale netwerken tegelijkertijd meerdere taken en talen kunnen leren te beschrijven (Hoofdstuk~\ref{chp:mtl_bg}).
In het bijzonder kijken we naar \textit{multitask learning} (MTL) en enkele meertalige benaderingen van NLP.

In \textbf{Deel II} van dit proefschrift wordt onderzocht hoe we het feit dat veel taken in NLP sterk met elkaar zijn verbonden kunnen benutten.
Dit wordt gedaan door te experimenteren met MTL met \textit{hard parameter sharing}, dat voor veel taken in NLP succesvol is gebleken.
Ondanks deze successen is het echter niet helemaal duidelijk \textit{wanneer} of \textit{waarom} het gebruik van MTL voordelig is voor NLP.
Hoofdstuk 4 bevat een casestudy van MTL waarin ik laat zien dat het toekennen van semantische labels aan woorden een verbetering oplevert bij het herkennen van woordsoorten.
Deze vinding benadrukt het belang van het beantwoorden van de vraag `Wanneer is het gebruik van MTL voordelig voor NLP?'. In Hoofdstuk 5 wordt dit onderzocht aan de hand van begrippen uit de informatietheorie.

Meertalige modellen kunnen gebruik maken van het feit dat talen vaak overeenkomsten met elkaar hebben.
Deze overeenkomsten kunnen op verschillende niveaus optreden, bijvoorbeeld op syntactisch, morfologisch of lexicaal vlak.
Hoewel er veel mogelijkheden zijn om gebruik te maken van deze overeenkomsten, kijken we in dit proefschrift naar het gebruik van meertalige woordrepresentaties, omdat deze erg goed te combineren zijn met neurale netwerken.
Net als bij MTL is het hier niet duidelijk in welke gevallen het een voordeel is om meertalige modellen te bouwen. \textbf{Deel III} begint met een casestudy over meertalige semantische tekstuele gelijkenis (Hoofdstuk 6).
Daarna onderzoek ik in welke mate talen op elkaar moeten lijken, en op welke manier, om voordeel te kunnen halen uit het bouwen van een meertalig model (Hoofdstuk 7).

In \textbf{Deel IV} van dit proefschrift experimenteer ik met een gecombineerd paradigma waarin een neuraal netwerk verschillende talen én taken tegelijkertijd leert (Hoofdstuk 8).
Tenslotte worden de conclusies uit dit proefschrift gepresenteerd in \textbf{Deel V} (Hoofdstuk 9).

De experimenten in dit proefschrift worden uitgevoerd op een grote verzameling voornamelijk lexicaal georiënteerde taken, zowel semantisch en morfo\-syntactisch van aard, en op in totaal 60 talen, die een relatief grote typologische diversiteit vertegenwoordigen.

Terwijl de traditionele NLP-benaderingen slechts een enkele taak of taal tegelijk behandelen, is het doel van dit proefschrift juist om onderzoeksvragen die deze grenzen overschrijden te beantwoorden.
Dit heeft als doel dat kleinere talen op de lange termijn kunnen profiteren van de vooruitgang die in NLP wordt geboekt, aangezien deze momenteel grotendeels ten goede komt aan grotere talen.
Dit kan positieve gevolgen hebben voor bijvoorbeeld taalbehoud, omdat, onder andere, sprekers van minderheidstalen een lagere druk zullen voelen om grotere talen te gebruiken.
Op de korte termijn zullen de antwoorden op de specifieke onderzoeksvragen van dit proefschrift nuttig zijn voor NLP-onderzoekers die naar hetzelfde doel streven.

\includepdf[pages=-]{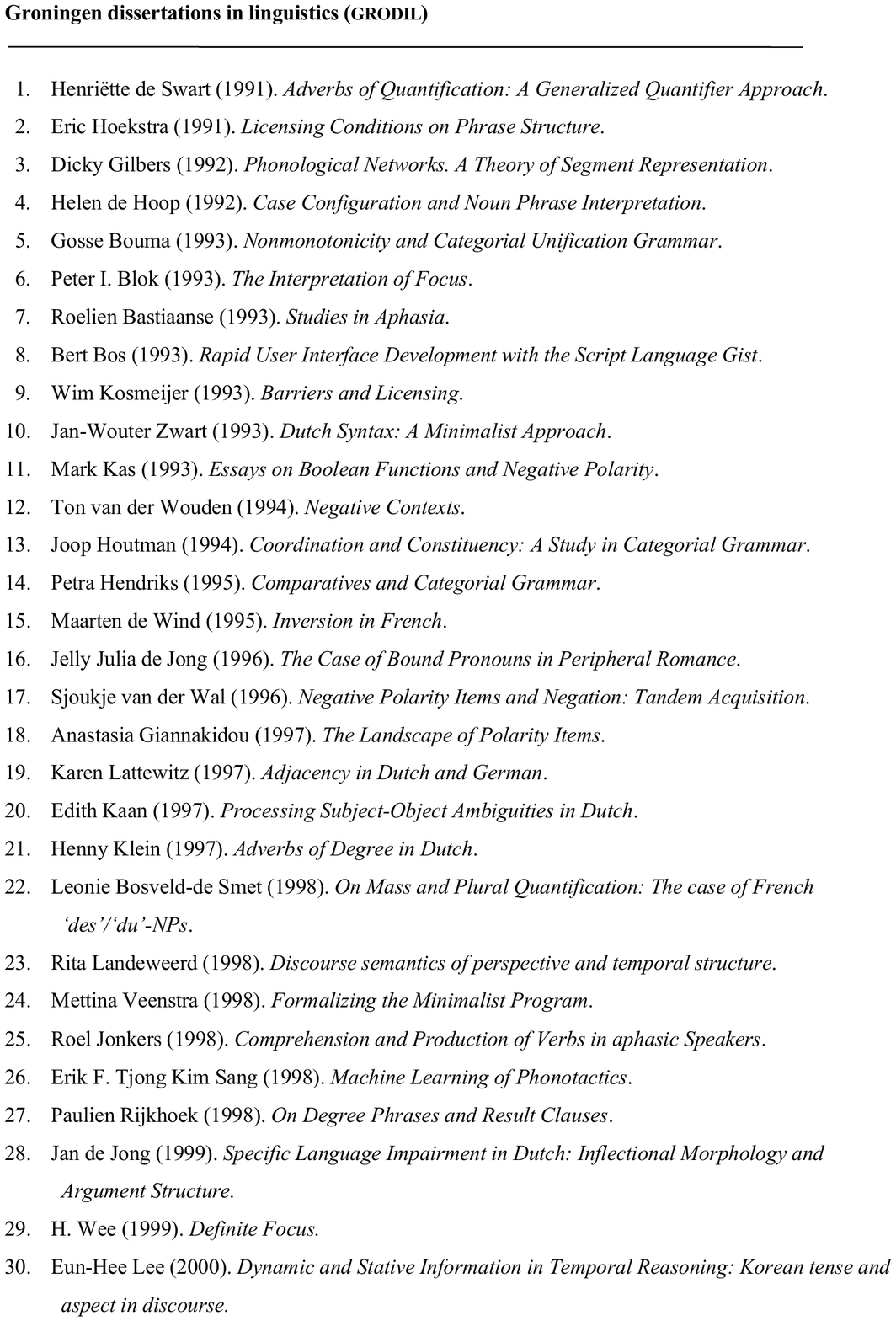}
}
\end{document}